\documentclass[11pt]{article}

\usepackage[truedimen,margin=25truemm]{geometry}

\usepackage{graphicx} 
\usepackage{epsfig} 
\usepackage{subfigure} 

\usepackage{amsmath}
\usepackage{amsthm}
\usepackage{amssymb}

\usepackage{algpseudocode}
\usepackage{algorithm}

\usepackage{color}

\usepackage{ulem}

\usepackage{booktabs}

\usepackage{hyperref}

\usepackage{graphicx} 
\usepackage{epsfig}
\usepackage{subfigure}

\usepackage{amssymb}
\usepackage{amsmath}
\usepackage{amsthm}
\usepackage{mathtools}
\usepackage[raggedright]{titlesec}

\def\vec#1{\mbox{\boldmath $#1$}}
\def\mat#1{\mbox{\bf #1}}

\usepackage{kasailab_def}

\title{Fast block-coordinate Frank-Wolfe algorithm for semi-relaxed optimal transport}

\author{Takumi Fukunaga \thanks{Department of Communications and Computer Engineering, School of Fundamental Science and Engineering, WASEDA University, 3-4-1 Okubo, Shinjuku-ku, Tokyo 169-8555, Japan (e-mail: f\_takumi1997@suou.waseda.jp) } \and Hiroyuki Kasai \thanks{Department of Communications and Computer Engineering, School of Fundamental Science and Engineering, WASEDA University, 3-4-1 Okubo, Shinjuku-ku, Tokyo 169-8555, Japan (e-mail: hiroyuki.kasai@waseda.jp)}}

\begin{document}

\maketitle

\newcommand{\changeHK}[1]{\textcolor{red}{#1}}
\newcommand{\changeTF}[1]{\textcolor{blue}{#1}}
\newcommand{\note}[1]{\textcolor{magenta}{#1}}

\begin{abstract}
Optimal transport (OT), which provides a distance between two probability distributions by considering their spatial locations, has been applied to widely diverse applications. Computing an OT problem requires solution of linear programming with tight mass-conservation constraints. This requirement hinders its application to large-scale problems. To alleviate this issue, the recently proposed relaxed-OT approach uses a faster algorithm by relaxing such constraints. Its effectiveness for practical applications has been demonstrated. Nevertheless, it still exhibits slow convergence. To this end, addressing a convex semi-relaxed OT, we propose a fast block-coordinate Frank-Wolfe (BCFW) algorithm, which gives sparse solutions. Specifically, we provide their upper bounds of the worst convergence iterations, and equivalence between the linearization duality gap and the Lagrangian duality gap. Three fast variants of the proposed BCFW are also proposed. Numerical evaluations in color transfer problem demonstrate that the proposed algorithms outperform state-of-the-art algorithms across different settings.
\end{abstract}

\section{Introduction}
\label{Seq:Intro}
The Optimal transport (OT) problem seeks an optimal {\it transport plan} or {\it transport matrix} by solving the total minimum transport cost from sources to destinations. This calculation requires {\it source mass conservation} from one source to targets, and versa, which are represented in formulation as a transport polytope. The OT problem can express the distance between two probability distributions, which is known as Wasserstein distance \cite{Peyre_2019_OTBook}. Consequently, this problem has been applied to widely diverse machine learning problems such as adversarial risk \cite{pmlr-v119-pydi20a}, inference with aggregate data \cite{singh2020inference}, graph optimal transport \cite{pmlr-v119-chen20e,Huang_arXiv_LCS_2020}, domain adaptation \cite{pmlr-v89-redko19a}, multi-view learning \cite{Kasai_ICASSP_2020}, and clustering \cite{fukunaga_ICPR_2020}. Among the OT problem formulations, the Kantorovich formulation is represented as convex linear programming (LP) \cite{Kantorovich_1942}. Thereby, many dedicated solvers such as an interior-point method and a network-flow method can obtain the solutions. It is, nevertheless challenging to solve large-scale problems efficiently because its computational cost increases cubically in terms of the data size.

To alleviate this difficulty, the Sinkhorn algorithm \cite{cuturi2013sinkhorn}, an {\it entropy-regularized} approach, works effectively on the OT problem, which is faster and which enables a parallel implementation. This computation includes a differentiable and unconstrained convex optimization. For that reason, it is easier to solve. In addition, the resultant OT distance is applicable to many machine learning problems by virtue of its differentiability. Furthermore, addressing its numerical unsuitability and non-robustness against small values of the regularizer, stabler variants have also been developed, but they are adversely affected by their slow convergences \cite{chizat2018scaling}. To reduce the runtime, a greedy algorithm of the Sinkhorn algorithm, the Greenkhorn algorithm \cite{altschuler2017nearlinear} and its accelerated variant \cite{lin2020efficient} have been proposed. It should be noted that these approaches produce a {\it dense} transport matrix because the entropy term is always positive. Along another avenue of development, a {\it smooth-regularized} approach exploits strong convexity and Lipschitz continuity \cite{blondel2018smooth}, where adding smooth terms onto the objective function enables harnessing of gradient-based approaches and dual formulations. One distinguishing feature is that regularization with the squared Euclidean norm obtains sparser solutions than the entropy-regularized approaches. Recent studies have exploited smoothness to the curse of dimensionality. Specifically, methods such as the smooth and strongly convex Brenier potentials \cite{pmlr-v108-paty20a_s} and the Gaussian-smoothed OT \cite{pmlr-v108-goldfeld20a} have been described in reports of the relevant literature.

Most of the previously described works have attempted to add regularizers onto the objective function. Some works address the fact that the tight mass-conservation constraint in the OT problem does not work well in some applications where weights and mass need not be preserved. For this particular problem, a {\it constraint-relaxed} approach has been proposed recently by loosening such strict constraints. This approach has gained great success for applications such as color transfer \cite{Rabin2014AdaptiveCT} and multi-label learning \cite{frogner2015learning}. However, it still exhibits a slow convergence property.

Envisioning the development of a faster solver producing sparser solutions in the OT problem, and particularly addressing its convex {\it semi-relaxed} formulation, this paper is the first to present a block-coordinate Frank-Wolfe (BCFW) algorithm with theoretical analysis. The FW algorithm (a.k.a. the conditional gradient method) is a class of linear convex programming methods calling a linear optimization oracle \cite{Frank_1956}. The key advantage of this algorithm is that its {\it projection-free} property is generally more efficient than projection operations when the dimension of the data is large. Thus, the FW algorithm is one of the most popular approaches in the OT problem \cite{DiscreteSira2013,Rakotomamonjy_arXiv_2015,Courty_PAMI_2017,Paty_ICML_2019}. In addition, the output solutions of the FW algorithm can be sparse, which are beneficial in many applications. However, because this algorithm must call linear oracle for all columns of the transport matrix at every iteration, its computational burden is problematic when the matrix size is extremely large. Hence, we further combine a coordinate descent approach, which selects one column randomly every iteration, resulting in much smaller computation cost, and also in achieving faster convergence \cite{Wright_2015_MP}. The block coordinate approach is also discussed in the literature of the OT problem \cite{Perrot_NIPS_2016,Redko_NeurIPS_2020}. Although this approach has already been discussed in the literature for various problems \cite{lacostejulien}, its concrete convergence for the relaxed OT problem remains unclear. Therefore, this paper can offer several important theoretical contributions. 
\begin{itemize}
\item Our convergence analysis yields an upper-bound of the curvature constant {\it without} relying on an oracle, as described in an earlier paper by \cite{lacostejulien}. Then, we directly exploit a variable block on the semi-relaxed domain and give iteration complexities for $\epsilon$-optimality with FW and BCFW algorithms for the semi-relaxed OT problem. 

\item Our analysis of the duality gap reveals that the {\it linearization duality gap}, a special case of the Fenchel duality gap, is equivalent to the Lagrangian duality gap. We derive the Lagrangian dual for the semi-relaxed OT problem. We prove this equivalence. This linearization duality gap certifies the quantity of the current approximation for monitoring the convergence. This point can be exploited for the stopping criterion in our proposed algorithms.

\item This paper proposes three fast variants of the proposed BCFW, i.e., the BCFW algorithms with pairwise-steps and away-step, and gap-adaptive sampling. For the latter, a convergence analysis is also provided. 

\item Numerical evaluations on the color transfer problem gives detailed analysis of the proposed BCFW, and show the effectiveness of the proposed BCFW in the semi-relaxed OT problem. 
\end{itemize}

The paper is organized as explained hereinafter. Section 2 presents preliminary descriptions of optimal transport, (semi-)relaxed optimal transport, and the block-coordinate Frank-Wolfe (BCFW) algorithm. Section 3 presents details of our proposed BCFW algorithm for the semi-relaxed optimal transport problem. The theoretical analysis for the convergence, duality gap, and computational complexity are also provided. In Section 5. we discuss three fast variants of the proposed BCFW algorithm with away-steps, pairwise-steps, and gap-adaptive sampling. Finally, in Section 6, numerical comparisons with existing methods are provided with results suggesting superior performance of the proposed BCFW algorithms. The proposed BCFW codes are implemented in MATLAB. Concrete proofs of theorems, additional numerical results and source codes are provided as supplementary materials.

\section{Preliminary and related work}

Herein, $\mathbb{R}^n$ denotes $n$-dimensional Euclidean space. Also, $\mathbb{R}^n_+$ denotes the set of vectors in which all elements are non-negative. $\mathbb{R}^{m \times n}$ denotes the set of $m \times n$ matrices and $\mathbb{R}^{m \times n}_+$ denotes the set of $m \times n$ matrices in which all elements are non-negative. We present vectors as bold lower-case letters $\vec{a},\vec{b},\vec{c},\dots$ and matrices as bold-face upper-case letters $\mat{A},\mat{B},\mat{C},\dots$. The $i$-th element of $\vec{a}$ and the element at the $(i,j)$ position of $\mat{A}$ are represented respectively as $a_i$ and ${A}_{i,j}$. When a matrix $\mat{A}$ is denoted as $(\vec{a}_1,\dots,\vec{a}_n)$, $\vec{a}_i$ represents the $i$-th column vector of $\mat{A}$. $\vec{e}_i$ is the canonical standard unit vector, of which the $i$-th element is 1. Others are zero. $\vec{1}_n \in \mathbb{R}^n$ is the $n$-dimensional vector in which all the elements are one.
The probability simplex is denoted as $\Delta_m = \lbrace \vec{a} \in \mathbb{R}^m : \sum_{i}a_i = 1\rbrace$. $\vec{\delta}_{\vec{a}}$ is the delta function at the vector $\vec{a}$. 
$\langle \cdot,\cdot\rangle$ is the Euclidean dot-product between vectors. For two matrices of the same size \mat{A} and \mat{B}, $\langle \mat{A},\mat{B}\rangle={\rm tr}(\mat{A}^T\mat{B})$ is the Frobenius dot-product. We denote the set $\{1, \ldots, n\}$ by $[n]$.

\subsection{Optimal transport problem}
The OT problem derives from the Monge problem, which seeks an optimal mapping between two probability distributions as
 $\vec{\nu}=\!\sum_{i=1}^m a_i\vec{\delta}_{{x}_i}$, $\vec{\mu}=\!\sum_{i=1}^n b_i\vec{\delta}_{{y}_i}$ given as
\begin{eqnarray*}
\defmin_{T} && \sum_{i=1}^m\ d(\vec{x}_i,T(\vec{x}_i)) \\
 \mathrm{subject\ to}&&  b_j =\ \sum_{\scriptsize i:T(\vec{x}_i)=\vec{y}_j} a_i,\quad \forall j \in [m],\vec{}
\end{eqnarray*}
where $d(\cdot, \cdot)$ is the cost function between two points. Both mapping and constraints are discrete, thus, the Monge problem is difficult to solve directly. To this difficulty, Kantorovich proposed a formulation by which the constraints are continuous \cite{Kantorovich_1942}. Concretely, given a cost matrix $\mat{C}$, the problem is defined as
\begin{equation}
	\label{eq:FormulationOptimalTransport}
	 \defmin_{\scriptsize {\mat{T} \in \mathcal{U}(\vec{a},\vec{b})}}\ \langle \mat{T},\mat{C} \rangle,
\end{equation}
where the domain $\mathcal{U}(\vec{a},\vec{b})$ is defined as 
\begin{equation}
	\label{eq:TransportPolytope}
	\mathcal{U}(\vec{a},\vec{b}) = \lbrace \mat{T} \in \mathbb{R}^{m \times n}_{+}: \mat{T}\vec{1}_n = \vec{a},\mat{T}^{T}\vec{1}_m = \vec{b} \rbrace.
\end{equation}
This domain $\mathcal{U}(\vec{a},\vec{b})$ requires the {\it mass-conservation 
 constraints} or the {\it marginal constraints} between two probabilities $\vec{a}$ and $\vec{b}$. The obtained optimal transport matrix $\mat{T}^*$ brings powerful distances between distributions defined as
\begin{equation*}
	\mathcal{W}_p(\vec{\nu},\vec{\mu}) = \langle \mat{T}^*,\mat{C} \rangle^{\frac{1}{p}},
\end{equation*}
which is called the $p$-th order {\it Wasserstein distance} \cite{Villani_2008_OTBook}. Especially, when $p=1$, the distance is equivalent to the Earth Mover Distance (EMD) \cite{Levina_ICCV_2001}. Many problems appearing in machine learning and statistical learning are definable in the OT problem. Interested readers are referred to \cite{Peyre_2019_OTBook} for a more comprehensive survey.

\subsection{Relaxed optimal transport}
\label{sec:RelaxedProblem}
As discussed in Section \ref{Seq:Intro}, solving large-scale linear programming problems is challenging in terms of the computational costs of obtaining solutions \cite{DisplacementOT}. Furthermore, the strict mass-conservation constraints might cause dreadful degradation of performance in some application. For example, Ferradans et al. reported that tight mass conservation does not reflect the color difference between images in a color transfer problem \cite{DiscreteSira2013}. This subsection introduces two categories of relaxed formulations of the OT problems.

\vspace*{0.2cm}
\noindent
{\bf Domain constraint relaxation.} One approach is to relax the domain constraint \cite{DiscreteSira2013}. Ferradans et al. propose allowing each point of \mat{X} to be transported to multiple points of \mat{Y} and vice versa.
This is defined as
\begin{equation*}
\label{eq:RelaxedOptimalTransport}
\defmin_{\scriptsize {\mat{T} \in \mathcal{S}_{\kappa}}}\ \langle \mat{T},\mat{C} \rangle,
\end{equation*}
where a relaxed domain $\mathcal{S}_{\kappa}$ is defined as 
\begin{eqnarray*}
	\label{eq:RelaxedTransportPolytope}
	\nonumber \mathcal{S}_{\kappa} = \lbrace \mat{T} \in \mathbb{R}^{n \times n}_+ : k_X\vec{1}_n \leq \mat{T}\vec{1}_n\leq K_X\vec{1}_n,
k_y\vec{1}_n \leq \mat{T}^T\vec{1}_n\leq K_Y\vec{1}_n, \vec{1}_n^T\mat{T}\vec{1}_n = M\rbrace,
\end{eqnarray*}
and constants $(k_X,K_X,k_Y,K_Y,M)$ are hyper-parameters. This method enables the transport matrix to increase or decrease the mass between two points. A noteworthy point is that the relaxed domain retains the linear constraints as the original. For that reason, existing solvers of linear programming are applicable. Rabin et al. extend it to propose the relaxed weighted OT, which loosens the column constraints \cite{Rabin2014AdaptiveCT}. There also exist other relaxed formulations considering only $\mat{T}\vec{1}_n = \vec{a}$ or $\mat{T}^T\vec{1}_m = \vec{b}$ as
\begin{equation*}
\defmin_{\displaystyle \scriptsize {\mat{T}\vec{1}_n = \vec{a}}}\ \langle \mat{T}, \mat{C} \rangle \quad {\rm or}\quad
\defmin_{\scriptsize \mat{T}^T\vec{1}_m = \vec{b}}\ \langle \mat{T}, \mat{C} \rangle.
\vec{}
\end{equation*}
These optimal solutions are summation of minimum costs of each row or column vector. Therefore, they are solvable faster than linear programming. In practice, this method is useful for document classification \cite{pmlr-v37-kusnerb15}. Its extended formulation has recently been developed in the context of style transfer \cite{kolkin2019style,FaststyleTransfer}. They attempt to define the relaxed earth mover distance (REMD) as the maximum of above formulations, and combine it with neural networks.

\vspace*{0.2cm}
\noindent
{\bf Regularized constraint relaxation.} In another line of attempts, the penalty of the domains defined in (\ref{eq:TransportPolytope}) is added to the objective function \cite{blondel2018smooth}. The relaxation of the marginal constraints is effective when only partial transport is allowed. Relaxing both marginal constraints in (\ref{eq:TransportPolytope}) yields the following relaxed formulation as
\begin{equation*}
	\label{eq:SmoothRelaxedOptimalTransport}
	\defmin_{\scriptsize {\mat{T}\geq \vec{0}}}\ \langle \mat{T},\mat{C} \rangle + \frac{1}{2}\Phi(\mat{T}\vec{1}_n,\vec{a})+\frac{1}{2}\Phi(\mat{T}^T\vec{1}_m,\vec{b}),
\end{equation*}
where $\Phi(\vec{x}, \vec{y})$ is a smooth divergence measure function. 

We also have an alternative formulation, which relaxes one of the two constraints in (\ref{eq:TransportPolytope}). This is a {\it semi-relaxed} problem, defined as the following.
\begin{equation}
	\label{eq:SmoothSemiRelaxedOptimalTransport}
	\defmin_{\scriptsize {\mat{T}\geq \vec{0},\mat{T}^T\vec{1}_m = \vec{b}}}\ \langle \mat{T},\mat{C} \rangle + \Phi(\mat{T}\vec{1}_n,\vec{a}).
\end{equation}
This setting is useful in color transfer. Rabin et al. also propose the weighted regularization term $\|\kappa - \vec{1}_n\|_1$ and the relaxed weighted OT so that the ratio of the source image becomes close to that of the reference image \cite{Rabin2014AdaptiveCT}. Benamou proposes a similar formulation, which is solvable using the augmented Lagrangian \cite{refId0}. Ferradans et al. propose a regularized and relaxed problem particularly addressing both color transfer and barycenter \cite{DiscreteSira2013}. They use the proximal splitting method and the coordinate descent method. 
Moreover, using the Kullback--Leibler (KL) divergence as $\Phi(\vec{x},\vec{y})$, a multi-label prediction problem is proposed, which is solved by using a Sinkhorn-like algorithm because of the entropy-regularized term \cite{frogner2015learning}. However, the KL divergence is not unstable because of divergence at zero \cite{chizat2018scaling}. 
Furthermore, some relaxed methods address cardinality-penalized problems. Instead of cardinality of solutions, Carli et al. approximate them by exploiting the rank regularization, sum-of-norm relaxation, and maximum norm relaxation for effective clustering \cite{carli2013convex}.

\subsection{Block-coordinate Frank-Wolfe algorithm}
\label{sec:FrankWolfeAlgorithm}
The Frank-Wolfe (FW) algorithm is a constraint convex optimization method. It is known to be a linear approximation algorithm that uses conditional gradient \cite{Frank_1956}. Although the FW algorithm is known to converge to optimal solutions at a {\it sublinear} rate, its {\it projection-free} property is preferred in the case where the convex constraint is simple and the feasible point can be found easily. More specifically, at every iteration, the feasible point $\vec{s}$ is found first by minimizing the {\it linearization} of $f$ over the convex feasible set $\mathcal{M}$. To find the feasible point $\vec{s}$, we solve the following subproblem.
\begin{equation}
\label{eq:GeneralFrankWolfeSubProblem}
\vec{s} =  \defargmin_{\scriptsize \vec{s}' \in \mathcal{M}}\ \langle \vec{s}',\nabla f(\vec{x}^{(k)}) \rangle.
\end{equation}
In that equation, $\vec{x}^{(k)}$ represents the $k$-th current point. Because the domain $\mathcal{M}$ is the convex set and the objective is linear for $\vec{s}$, it is possible to solve (\ref{eq:GeneralFrankWolfeSubProblem}) using linear programming. Finally, the next iterate $\vec{x}^{(k+1)}$ is obtainable by a convex combination as $\vec{x}^{(k+1)}=(1-\gamma) \vec{x}^{(k)}+\gamma \vec{s}$, where $\gamma$ is the stepsize. Consequently, the generated iterates can be maintained inside the feasible set $\mathcal{M}$ if the initial point $\vec{x}^{(0)}$ is in $\mathcal{M}$.

One shortcoming of the FW algorithm is that solving the minimization problem must be performed at each iteration. To address this issue, if domain $\mathcal{M}$ can be {\it block-separable} as the Cartesian product $\mathcal{M}=\mathcal{M}^{(1)}\times \mathcal{M}^{(2)}\times\dots \times \mathcal{M}^{(n)} \subset \mathbb{R}^{m}$ over $n \geq 1$, then we can perform a {\it single cheaper} update of only $\mathcal{M}^{(i)}$ instead of on an entire of $\mathcal{M}$. In this line of algorithms, the block-coordinate Frank-Wolfe (BCFW) algorithm has been proposed, for example, in the structural SVM problem \cite{lacostejulien} and in the MAP inference \cite{Swoboda_CVPR_2019}. This algorithm is applicable to the constrained convex problem of the form 
\begin{equation*}
\label{eq:GeneralCartesianformulation}
\defmin_{\scriptsize \vec{x} \in \mathcal{M}^{(1)}\times \mathcal{M}^{(2)}\times\dots \times \mathcal{M}^{(n)}} f(\vec{x}).
\end{equation*}
We assume that each factor $\mathcal{M}^{(i)}$ is convex, with $m = \sum_{i=1}^nm_i$. We solve the subproblem on the factor which is selected randomly. As a result, the BCFW algorithm can be implemented in cheaper iteration. When $n=1$, this algorithm is reduced to the FW algorithm.

\section{Block-coordinate Frank-Wolfe algorithm for semi-relaxed optimal transport problem}
\label{Sec:BaselineBCFW}

The present paper particularly addresses the semi-relaxed problem of (\ref{eq:SmoothSemiRelaxedOptimalTransport}) with $\Phi(\vec{x},\vec{y})=\frac{1}{2\lambda}\|\vec{x}-\vec{y}\|_2^2$ because it is not only smooth but also convex. The problem of interest is formally defined as
\begin{equation}
\label{eq:EuculideanSquareRelaxedProblem}
\defmin_{\scriptsize \substack{\displaystyle{\mat{T}\geq \vec{0}},\\ \mat{T}^T\vec{1}_m = \vec{b}}}
\left\{f(\mat{T}):=\langle \mat{T},\mat{C} \rangle + \frac{1}{2\lambda}\|\mat{T}\vec{1}_n-\vec{a}\|_2^2\right\},
\end{equation}
where $\lambda$ is a {\it relaxation} parameter. The domain is transformed into 
\begin{equation}
	\label{eq:EuclideanRelaxedCartesian}
	\mathcal{M} = {b}_1\Delta_m \times {b}_2\Delta_m \times \cdots \times {b}_n \Delta_m,
\end{equation}
where ${b}_i\Delta_m$ represents the simplex of the summation ${b}_i$.

\subsection{Algorithm description}

After describing Frank-Wolfe (FW) algorithm, we elaborate on a block-coordinate Frank-Wolfe (BCFW) algorithm for the semi-relaxed optimal transport problem.

\vspace*{0.2cm}
\noindent
{\bf Frank-Wolfe (FW) algorithm.} We first consider the FW algorithm for this problem, and then propose a faster block-coordinate Frank-Wolfe algorithm. 
The gradient $\nabla f(\mat{T}) \in \mathbb{R}^{mn}$ is given as
\begin{equation*}
	\label{eq:RelaxedGradient}
	\nabla f(\mat{T}) = 
		\left(
			\begin{array}{c}
				\vec{c}_1 \\
				\vdots \\
				\vec{c}_i \\
				\vdots \\
				\vec{c}_n \\
			\end{array}
		\right)
		+\frac{1}{\lambda}
		\left(
				\begin{array}{c}
					\mat{T}\vec{1}_n -\vec{a}\\
					\vdots \\
					\mat{T}\vec{1}_n -\vec{a}\\
					\vdots \\
					\mat{T}\vec{1}_n -\vec{a}\\
				\end{array}
		\right),
\end{equation*}
where $\displaystyle{\nabla f_i(\mat{T}):=\vec{c}_i+1/\lambda \cdot (\mat{T}\vec{1}_n -\vec{a})} \in \mathbb{R}^m$ represents the gradient on the $i$-th variable block ${b}_i\Delta_m$. The subproblem (\ref{eq:GeneralFrankWolfeSubProblem}) is equivalent to
\begin{equation}
	\label{eq:SemiRelaxedSubproblemImplement}
	\vec{s}_i = {b}_i\vec{e}_j = {b}_i\defargmin_{\scriptsize {\vec{e}_k \in \Delta_m ,k \in [m]}} \langle \vec{e}_k,\nabla_i f(\mat{T}^{(k)}) \rangle,
\end{equation}
where $j \in [m]$ and $\vec{e}_j$ is the extreme point on probability simplex \cite{AwayStepClarkson}. In other words, we just find the index of the minimal elements of the gradient of the variable blocks. The computational cost of the subproblem (\ref{eq:SemiRelaxedSubproblemImplement}) is greatly improved. The detailed computational cost analysis is described in {Section \ref{Sec:ComputationalComplexityAnalysis}}.

After finding the points $\mat{S}=(\vec{s}_1, \vec{s}_2, \ldots, \vec{s}_n) \in \mathbb{R}^{m \times n}$, we search an optimal stepsize $\gamma$. One classical way in the FW algorithm is a {\it decay} stepsize (DEC), where $\gamma=2/(k+2)$ with the iteration number $k$. A line-search algorithm can be also applicable. Concretely, we solve ${\min_{\gamma \in [0,1]}f((1-\gamma)\vec{x}+\gamma \vec{s})}$, and calculate $\gamma$ directly since the objective of the semi-relaxed problem is quadratic. As for the stopping criterion, we monitor the duality gap $g(\mat{T})$ that will be defined in {\bf Theorem \ref{thm:RelaxedFenchelDualityGap}} in Section \ref{sec:DualityGapAnalysis}, and stop the algorithm when $g(\mat{T}) < \epsilon$, where $\epsilon$ is an approximation precision parameter.

\begin{algorithm}[t]
\caption{Block-coordinate Frank-Wolfe (BCFW) for semi-relaxed OT}      
\label{alg:BCFW-SROT}
\begin{algorithmic}[1]       
\Require{$\mat{T}^{(0)} = (\vec{t}^{(0)}_1,\dots,\vec{t}^{(0)}_n) \in {b}_1\Delta_m \times \cdots \times {b}_n\Delta_m$}
\For {$k=0 \dots K$}
\State Select index $i \in [n]$ randomly
\State Compute $\vec{s}_i = b_i \defargmin_{\scriptsize {\vec{e}_k \in \Delta_m}, k \in [m]}\ \langle \vec{e}_k,\nabla_if(\mat{T}^{(k)}) \rangle$

\State{Compute stepsize $\gamma$ as
\[
	\hspace*{-0.2cm}\gamma  = 
	\begin{cases}
	\gamma_{\rm LS},\hfill {\rm (line-search\ in\ (\ref{eq:SemiRelaxedBCFWStep}))}\\
	\displaystyle{\frac{2n}{k+2n}}, \quad\quad\quad\hfill {\rm (decay\ rule)}
	\end{cases}	
\]
}
\State {Update $\vec{t}^{(k+1)}_j\ \forall j \in [n]$ as 
\[
	\vec{t}^{(k+1)}_j  = 
	\begin{cases}
	\vec{t}^{(k)}_j, & ({\rm for\ } j \neq i)\\
	(1-\gamma)\vec{t}^{k}_i + \gamma \vec{s}_i, & ({\rm otherwise})
	\end{cases}	
\]
}
\EndFor
\end{algorithmic}
\end{algorithm}

\vspace*{0.2cm}
\noindent
{\bf Block-coordinate Frank-Wolfe (BCFW) algorithm.} We now propose an application of the block-coordinate Frank-Wolfe algorithm to the semi-relaxed problem considering that the feasible set $\mathcal{M}$ can be separable as the cartesian product. The procedure of {Algorithm \ref{alg:BCFW-SROT}} most closely resembles that of the FW algorithm, but they are slightly different. It is necessary to solve the subproblem on the variable block selected randomly at every iteration. More concretely, the subproblem is identical to (\ref{eq:SemiRelaxedSubproblemImplement}), but we solve the subproblem only for the $i$-th column, which is selected randomly. Then, all the other columns of $\mat{T}$ remain the same. Regarding the stepsize calculation, we use the formula $\gamma=2n/(k+2n)$, which is necessary for the convergence guarantee, as shown in {\bf Theorem \ref{thm:relaxedBCFWconvergence}}. Similarly to the FW algorithm, an exact line-search (ELS) algorithm can be used. Nevertheless, the optimal stepsize in the BCFW algorithm differs from that of the FW algorithm (\ref{AppenEq:SemiRelaxedFWStep}) because the BCFW algorithm only requires the updated column vector on the variable block. Consequently, the optimal stepsize $\gamma_{\rm LS}$ is calculated as 

\begin{equation}
\label{eq:SemiRelaxedBCFWStep}
\gamma_{\rm LS} = \frac{\displaystyle{\lambda\langle \vec{t}^{(k)}_i-\vec{s}_i,\vec{c}_i\rangle+\langle \vec{t}^{(k)}_i - \vec{s}_i,\mat{T}^{(k)}\vec{1}_n - \vec{a} \rangle}}{\displaystyle{ \|\vec{t}^{(k)}_i - \vec{s}_i\|^2}},
\end{equation}
where $\vec{t}_i$ is the $i-$th column of \mat{T}, and $\vec{s}_i$ is the solution of the $i$-th subproblem in (\ref{eq:SemiRelaxedSubproblemImplement}). As we will discuss in {\bf Theorem \ref{thm:RelaxedFenchelDualityGap}}, the duality gap can be used for the stopping criterion, and in a practical implementation, we monitor the value of the duality gap because the subproblem is solved at every iteration. It is noteworthy that, in the BCFW algorithm, calculating the value of the duality gap from the solution of the value of (\ref{eq:SemiRelaxedSubproblemImplement}) is impossible because the solutions of the subproblems on all the variable blocks are needed. Therefore, calculating the duality gap, if attempted every iteration, engenders huge increases of runtime, consequently resulting in loss of the benefit of the cheaper iteration complexity in BCFW. Consequently, in our practical implementation, we monitor the duality gap every $n$ iterations, of which period is equal to that of the FW algorithm.

Lastly, we consider two rules for choosing (sampling) a column at each iteration: the uniform random order and the random permutation. The former randomly selects $i \in [n]$, of which convergence analysis is given in Section \ref{Sec:ConvergenceAnalysis}. The latter runs a cyclic order on a permuted index, for example $(1\rightarrow 2\rightarrow 3)\rightarrow (3\rightarrow2\rightarrow1) \rightarrow (2\rightarrow1\rightarrow3)\rightarrow\cdots$ when $n=3$. Those algorithms are, hereinafter,  denoted, respectively, as BCFW-U and BCFW-P. Another sampling strategy using the duality gap is discussed in Section \ref{Sec:BCFW-GA}, which is called BCFW-GA.

\subsection{Theoretical results}
This section explains convergence analysis of the FW and BCFW algorithms proposed in the preceding subsection. We then discuss the relation between the linearization duality gap as a special case of the Fenchel duality gap and Lagrange duality gap. This discussion provides equivalence between them in this semi-relaxed OT problem. Finally, we also summarize their computational complexity. 

\subsubsection{Convergence analysis}
\label{Sec:ConvergenceAnalysis}

We analyze theoretically the worst convergence iteration of the FW and BCFW algorithms. The result for the FW algorithm is provided in the supplementary material. The result for the proposed BCFW algorithm is given below. We first define the curvature constant $C_f^\otimes$ as follows:
\begin{Def}[Curvature constant for cartesian product \cite{lacostejulien}]
\label{def:CurvatureBCFW}
When a domain $\mathcal{M}$ has a structure of the cartesian product $\mathcal{M}^{(1)}\times \mathcal{M}^{(2)} \times \cdots \times \mathcal{M}^{(n)}$, its curvature constant is defined as $C_f^\otimes := \sum_{i=1}^nC^{(i)}_f$, where $C^{(i)}_f$ is 
\begin{equation*}
\label{eq:BlockCurvature}
C^{(i)}_f := \!\!\!\!\!\!\!\!\sup_{\scriptsize
		\substack{
		\displaystyle{\mat{T} \in \mathcal{M},\vec{s}_i} \in \mathcal{M}^{(i)},\\
		\displaystyle{\gamma \in [0,1]},\\
		\displaystyle{\mat{Y} = \mat{T} + \gamma(\vec{s}_{[i]} - \vec{t}_{[i]})}
}}\!\!\!\! \frac{2}{\gamma^2}(f(\mat{Y})-f(\mat{T})-\langle \vec{y}_i -\vec{t}_i,\nabla_i f(\mat{T})\rangle),
\end{equation*}
where $\vec{x}_{[i]} $ refers to the zero-padding of $\vec{x}_i$. 
\end{Def}
Then, we have the convergence of BCFW:
\begin{Thm}
\label{thm:relaxedBCFWconvergence}
Let $\mat{T}^*$ is the optimal solution of the semi-relaxed OT problem in (\ref{eq:EuculideanSquareRelaxedProblem}). Consider Algorithm \ref{alg:BCFW-SROT} under the initial point of \mat{T} as $\mat{T}^{(0)}=(b_1\vec{e}_1,\dots, b_i\vec{e}_1,\dots, b_n\vec{e}_1)$ with a decay stepsize rule $k=\frac{2n}{k+2n}$. Then, we have 
$\mathbb{E}[f(\mat{T}^{(j)})] - f(\mat{T}^{*}) \leq \frac{2n}{k+2n}(C_f^\otimes+h_0)$, 
where $h_0=f(\mat{T}^{(0)})-f(\mat{T}^*)$, and where $C_f^\otimes$ is the curvature constant with $\leq \frac{4}{\lambda}$. 
Additionally, given an approximation precision constant $\epsilon$, if $\|\mat{C} \|_{\infty} \leq \frac{2}{\lambda}$, Algorithm \ref{alg:BCFW-SROT} requires at most the number of $\mathcal{O}(\frac{n}{\lambda \epsilon})$ for its convergence. Otherwise, it requires the additional number of $\frac{2nh_0}{\epsilon} \leq \frac{2n({\small \|\mat{C}\|_{\infty}})}{\epsilon}$.
\end{Thm}

For its proof, we first bound the curvature constant $C_f^\otimes$ by taking into account the twice differentiability of $f(\mat{T})$ and the simplex structure. $g(\mat{T}^{(0)})$ is also upper-bounded from $\| \mat{C}\|_{\infty}$ and the assumption of $\mat{T}^{(0)}$. Finally, we derive the upper-bound of the complexity. It should be noted that some additional iterations are needed when $\|\mat{C} \|_{\infty} > \frac{2}{\lambda}$.
The full proof is given in the supplementary materiel.

\subsubsection{Linearization duality gap and stopping criterion}
\label{sec:DualityGapAnalysis}

The {\it linearization duality} is a special case of the {\it Fenchel duality} in the FW algorithm \cite{lacostejulien,JaggiMartin2013}, and its duality gap at the points $\vec{x}$ is given as
\begin{equation*}
	\label{eq:FenchelDualityGap}
	g(\vec{x})=\max_{\scriptsize \vec{s}' \in \mathcal{M}}\ \langle \vec{x} - \vec{s}', \nabla f(\vec{x}) \rangle = \langle \vec{x} - \vec{s}, \nabla f(\vec{x}) \rangle,
\end{equation*} 
where $\mathcal{M}$ is convex. Note that adding $f(\vec{x})$ onto the linearization duality is equivalent to the {\it Wolf duality} \cite{AwayStepClarkson}. For the semi-relaxed OT problem, we specifically give the equivalence between the linearization gap, denoted as $g(\mat{T})$, and the Lagrangian duality gap as shown below. 
\begin{Thm}
\label{thm:RelaxedFenchelDualityGap}
Consider the semi-relaxed problem in (\ref{eq:EuculideanSquareRelaxedProblem}). The linearization duality gap is provided as
\begin{equation*}
	\label{eq:RelaxedFenchelDualityGap}
	g(\mat{T})= \langle \mat{T}-\mat{S},\mat{C}\rangle + \frac{1}{\lambda} \langle \mat{T}\vec{1}_n - \mat{S}\vec{1}_n,\mat{T}\vec{1}_n - \vec{a} \rangle,
\end{equation*} 
where \mat{S} is the solution of the subproblem (\ref{eq:SemiRelaxedSubproblemImplement}). Then, the linearization duality gap $g(\mat{T})$ is equivalent to the Lagrangian duality gap of the semi-relaxed problem.
\end{Thm}
The full proof is given in the supplementary materiel, but its proof sketch is the following:
The dual problem is first derived as
\begin{eqnarray*}
\label{eq:WolfeDualSemiRelaxed}
\defmax_{\scriptsize\mat{T}}\ 
f(\mat{T}) - \sum_{i=1}^n\langle \vec{t}_i , \nabla_if(\mat{T}) \rangle + \sum_{i=1}^{n}{b}_i\max_{j \in [m]} (\nabla_i f(\mat{T}))_j.
\end{eqnarray*}
Then, we consider the Lagrangian duality gap $g_L(\mat{T})$ as the difference between the objective and dual objective of the semi-relaxed problem. Finally, we show that $g_L(\mat{T})$ is equal to $g(\mat{T})$ defined in this theorem. From this theorem, we can use the function $g(\mat{T})$ as both the linearization duality gap and the Lagrangian duality gap. Therefore, $g(\mat{T})$ is suitable for the stopping criterion of the algorithms.

\subsubsection{Computational complexity analysis of baseline algorithm of BCFW}
\label{Sec:ComputationalComplexityAnalysis}
Next we analyze the subproblem of the semi-relaxed OT problem. The subproblem is solvable using linear programming. However, the computational cost is $\mathcal{O}((mn)^3\log (mn))$ because the transport matrix $\mat{T}$ is vectorized as the $mn$-dimension for linear programming. The column vector of the transport matrix $\mat{T}$ is independent of other column vectors of the semi-relaxed OT problem. 
Therefore, because it is possible to solve the subproblem on $n$ variable blocks, the computational complexity of the subproblem~(\ref{eq:GeneralFrankWolfeSubProblem}) can be reduced to $\mathcal{O}(nm^3 \log m)$. The subproblem (\ref{eq:SemiRelaxedSubproblemImplement}) in the proposed FW algorithm is defined on the Cartesian product of the probability simplex. Therefore, it is equivalent to the problem (\ref{eq:SemiRelaxedSubproblemImplement}) \cite{JaggiMartin2013}. As a result, the computational complexities of linear programming $\mathcal{O}(m^3\log m)$ are reduced to $\mathcal{O}(m)$, which speeds up the time.
The BCFW algorithm requires only one variable block selected randomly at every iteration, whereas the FW algorithm must solve the $n$ variable block. The BCFW algorithm has the same convergence as that of the FW algorithm. As a result, the computational complexities of the BCFW algorithm are more improved than those of the FW algorithm. 
We further analyze of the computational complexities of fast variants of the proposed BCFW algorithm in Section \ref{Sec:ComputationalComplexityAnalysisFastVariants}.

\section{Fast variants of BCFW}
\label{Sec:FastVariantsBCFW}

\subsection{Variants of pairwise-steps  (BCAFW) and away-steps (BCPFW)}
\label{Sec:FWandAW}

As discussed in the previous section, the Frank Wolfe (FW) algorithm exhibits sublinear, thus several improvements have been investigated to accelerate this rate \cite{WolfeBook1970}. Among them, this subsection addresses and follows a strategy that replaces the FW direction with different directions, which are called the pairwise-steps \cite{WolfeBook1970} and the away-steps \cite{AwaystepMitchell}. These modifications achieve a linear rate without the strongly convexity of the objective \cite{lacostejulien2015global}. More specifically, we exploit the away-steps and the pairwise-steps in block-coordinate method \cite{Osokin_ICML_2016,PairwiseFranc}. We denote the BCFWs with the away-steps and with the pairwise-steps as BCAFW and BCPFW, respectively.


The BCFW algorithm generates convex combinational points from {\it atoms} in each variable block. Thus, there might exist select {\it non-desirable} atoms, and this leads to sublinear convergence rates of the FW and BCFW algorithms. To avoid this situation, the away-steps and pairwise-steps have been proposed. They remove unnecessary atoms from an {\it active set} on each variable block. By following the work \cite{Osokin_ICML_2016}, this paper combines the proposed BCFW algorithm with the away-steps and the pairwise-steps, and attempts to improve the rate of convergence.

Let $\mathcal{S}_{i}$ be the active set on the $i$-th ($i\in [n]$) variable block, which is defined as
\begin{equation}
\label{eq:ActiveSet}
	\mathcal{S}_{i} = \lbrace \vec{e}_j \in \Delta_m : \alpha_{\scriptsize{\vec{e}_j}} > 0, j \in [n] \rbrace,
\end{equation}
where $\alpha_{\vec{e}_j}$ is the coefficient of the $j$-th extreme point $\vec{e}_j$. This is because each variable block in the semi-relaxed problem is the probability simplex, and the extreme points is in $\lbrace \vec{e}_1,\vec{e}_2,\dots,\vec{e}_n \rbrace$. We then consider a new subproblem in order to remove the atoms, which is defined as
\begin{equation*}
\label{eq:AwaySubproblem}
	\vec{v}_{i} = \defargmax_{\vec{v}' \in \mathcal{S}_i}\ \langle  \vec{v}', \vec{c}_i + \frac{1}{\lambda}(\mat{T}\vec{1}_n - \vec{a} )\rangle.
\end{equation*} 
This problem can be solved in the same way as the subproblem (\ref{eq:SemiRelaxedSubproblemImplement}) because $\mathcal{S}_i \subset \lbrace \vec{e}_1,\vec{e}_2,\dots,\vec{e}_n \rbrace$. Defining two directions, i.e., the {\it FW} direction $\vec{d}_{\rm FW}=\vec{s}_i - \vec{t}^{(k)}_i$ and the {\it Away} direction $\vec{d}_{\rm Away}=\vec{t}^{(k)}_i - \vec{v}_i $, respectively, we select the one reducing the objective function value more. Then, we find the stepsize $\gamma$ satisfying ${\min_{\gamma \in [0,\gamma_{\rm max}]}f((1-\gamma)\vec{x}+\gamma \vec{s})}$. This stepsize $\gamma_{\rm LS}$ is calculated by replacing $ \vec{t}^{(k)}_i-\vec{s}_i$ in (\ref{eq:SemiRelaxedBCFWStep}) with $\vec{d}$, which is given by
\begin{equation}
\label{Eq:SemiRelaxedBCAFWStep}
\gamma_{\rm LS} = -\frac{\lambda \langle \vec{d},\vec{c}_i \rangle + \langle \vec{d}, \mat{T}^{(k)}\vec{1}_n - \vec{a} \rangle}{ \|\vec{d}\|^2},
\end{equation}
where $\vec{d}$ can be $\vec{d}_{\rm FW}$ or $\vec{d}_{\rm Away}$. Then we must update not only the selected column vector $\vec{t}^{(k)}_i$ but also the selected active set $\mathcal{S}^{(k)}_i$. 

We similarly consider the block-coordinate Pairwise Frank-Wolfe (BCPFW), of which procedure is similar to that of BCAFW. The main difference is the updating direction. While the BCAFW algorithm combines the current point $\vec{t}^{(k)}_i$ with the direction, the BCPFW only uses two atoms $\vec{s}_i$ and $\vec{v}_i$. We set the direction $\vec{d}_{\rm Pair} = \vec{s}_i - \vec{v}_i$ and $\gamma_{\rm max} = \alpha_{\vec{v}_i}$. This operation develops the movement between only two atoms and improves the convergence rate.

The overall algorithms of BCAFW and BCFPFW are summarized in Algorithms \ref{AppenAlg:BCAFW-SROT} and \ref{AppenAlg:BCPFW-SROT}, respectively.

\subsection{A variant of adaptive sampling (BCFW-GA)}
\label{Sec:BCFW-GA}

This subsection, furthermore, focuses on another approach to fasten the convergence speed of the proposed algorithm, which is an {\it adaptive sampling} scheme that is popular approach in block-coordinate methods \cite{Nesterov2012stochastic,perekrestenko2017faster,needell2014stochastic_s,zhao2015stochastic}. This paper particularly addresses the approach considering the duality gap, and denotes the method as the BCFW algorithm with {\it gap-adaptive} sampling (BCFW-GA).

\subsubsection{Algorithm description}

The BCFW algorithm operates on the block-separable domain as in (\ref{eq:EuclideanRelaxedCartesian}), and updates one single column that is randomly selected. Therefore, the convergence rate of such coordinate-descent-based algorithms heavily depends on its sampling method, i.e., sampling probability distribution over the coordinates (columns). While the BCFW-U and BCFW-P algorithms proposed earlier use a uniform sampling, a new variant of the BCFW algorithm in this section attempt to faster the convergence by exploiting the weighted distributions that are generated by the duality gap in each column, i.e., {\it column-wise} duality gap. 
In the literature, many efforts have been done in this particular direction for this decade. This paper specifically follows the same line of the researches of \cite{Osokin_ICML_2016, perekrestenko2017faster} because they address the duality gap whereas others mainly focus on the Lipschitz constants of the gradients \cite{Nesterov2012stochastic,perekrestenko2017faster,needell2014stochastic_s,zhao2015stochastic}. 

The main idea behind our proposed approach is as follows: The columns with larger duality gaps admit higher improvement to the objective function value, thus, such columns should be sampled more often. In this way, we try to make more significant progress than the uniform-sampling method. For this purpose, after update of $\vec{t}_i$, the proposed BCFW-GA updates the duality gap for each column. Here, note that $g(\mat{T})$ is given as 
\begin{eqnarray*}
	g(\mat{T}) &=& \langle \mat{T}- \mat{S}, \mat{C} \rangle + \frac{1}{\lambda}\langle \mat{T}\vec{1}_n - \mat{S}\vec{1}_n, \mat{T}\vec{1}_n -\vec{a} \rangle\\
	 &=&  \sum_{i=1}^n (\vec{t}_i - \vec{s}_i)^T \vec{c}_i
	+  \frac{1}{\lambda}\langle \sum_{i=1}^n(\vec{t}_i-\vec{s}_i), \mat{T}\vec{1}_n \!- \!\vec{a} \rangle 
	 =   \sum_{i=1}^n g_i(\mat{T}),
\end{eqnarray*}
where $g_i(\mat{T})$ is given by
\begin{eqnarray}
\label{AppenEq:Gi}
	g_i(\mat{T}) &=& \langle \vec{t}_i - \vec{s}_i, \vec{c}_i \rangle + \frac{1}{\lambda} \langle \vec{t}_i - \vec{s}_i, \mat{T}\vec{1}_n - \vec{a} \rangle. \quad \quad \forall i \in [n].
\end{eqnarray}
Therefore, updating the column-wise duality gap $g_i(\mat{T})$ every iteration, we select an index $i$ at random in proportion to the probability generated from $(g_1(\mat{T}), g_2(\mat{T}), \ldots, g_n(\mat{T}))$.

In the meantime, the update of $g_i(\mat{T})$ apparently depends on \mat{T}. Hence, every time one single $\vec{t}_i$ is updated, we need to re-calculate $g_i(\mat{T})$ of all other $(n-1)$ columns to obtain its correct probability. Nevertheless, this is intractable, and wastes the benefit of the block coordinate approach. Therefore, in practice, at every $M \times n$ iterations, we periodically update $g_i(\mat{T})$ of all the columns to obtain their exact values. This update is specifically called the {\it global update} in this paper, and the loop of this global update is called an {\it outer iteration}. In contract to the outer iteration, the update of single $g_i(\mat{T})$ within the cycle of the global update is called an {\it inner iteration}. Within the global update period, i.e., the inner iteration, we store the calculated $g_i(\mat{T})$ for each $i$-th column, and do not perform the global update for the other columns. For the update of $g_j(\mat{T})$ of the $j$-th column $(j\neq i)$, we utilize the stored latest (but outdated) $g_j(\mat{T})$. Hence, we expect that, when $M$ is reasonably small, the convergence can be achieved, otherwise not. 

The overall algorithm of BCFW-GA is summarized in Algorithm \ref{AppenAlg:BCFW-GA-SROT}.

\subsubsection{Convergence analysis of BCFW-GA}
We give a convergence analysis of BCFW-GA, which is a straightforward extension to the semi-relaxed OT problem from that of the structured SVM problem in \cite{Osokin_ICML_2016}. 
\begin{Thm}(Total complexity analysis of BCFW-GA (Algorithm \ref{AppenAlg:BCFW-GA-SROT}))
\label{Thm:ComplexityBCFW-GA-SROT}
Let $\mat{T}^*$ is the optimal solution of the semi-relaxed OT problem. Consider Algorithm \ref{AppenAlg:BCFW-GA-SROT} under the initial point of $\mat{T}$ as $\mat{T}^{(0)} = (b_1\vec{e}_1,\dots,b_n\vec{e}_n)$ with a decay stepsize rule $\gamma = \frac{2n}{k+2n}$. Then, if $\|\mat{C}\|_{\infty} \leq \frac{2}{\lambda}$, Algotihm \ref{AppenAlg:BCFW-GA-SROT} requires $\mathcal{O}(\frac{8n}{\epsilon \lambda})$ at best and $\mathcal{O}(\frac{8n\sqrt{n}}{\epsilon \lambda})$ at worst. On the other hand, if $\|\mat{C}\|_{\infty} > \frac{2}{\lambda}$, it requires $\mathcal{O}(\frac{2n\|\mat{C}\|_\infty}{\epsilon \lambda}+\frac{8}{\epsilon \lambda \sqrt{n}})$ at best and $\mathcal{O}(\frac{2n\|\mat{C}\|_\infty}{\epsilon}+\frac{8n\sqrt{n}}{\epsilon \lambda})$ at worst.
\end{Thm}

The followings are remarks.

\begin{Remark}(Comparison to {\bf Theorem \ref{thm:relaxedBCFWconvergence}} in BCFW)

BCFW has
\[
\epsilon \geq \frac{2n}{k+2n} \cdot \left(\frac{4}{\lambda} + \frac{4}{\lambda} \right)
			 \Longleftrightarrow   k \geq \frac{8n}{\lambda \epsilon} +  \frac{8n}{\lambda \epsilon} - 2n
			 \overset{\rm approx.}{\Longleftrightarrow}   k \geq \frac{8n}{\lambda \epsilon} +  \underbrace{\frac{8n}{\lambda \epsilon}}_{(**)} = \frac{16n}{\lambda \epsilon}.
\]
Comparing this to the result of BCFW-GA, we confirm that the term (**) of BCFW is replaced with $\frac{8}{\epsilon \lambda \sqrt{n}}$ in (\ref{Eq:TotalCompBCFG-GA-best}) of BCFW-GA. Therefore, we find that BCFW-GA can reduce the complexity of BCFW roughly by half in the best case. It is, however, that, in the worst case, it increases by $\mathcal{O}(\frac{8n\sqrt{n}}{\epsilon \lambda })$. 
\end{Remark}

As for the the worst case, we have a similar discussion as \cite{Osokin_ICML_2016} below.
\begin{Remark}
Recall that the BCFW algorithm has $\mathbb{E}[f(\mat{T}^{(k)})] - f(\mat{T}^{*}) \leq \frac{2n}{k+2n}(C_f^\otimes+h_0)$ in 
{\bf Theorem \ref{thm:relaxedBCFWconvergence}}, 
and BCFW-GA has 
$\mathbb{E}[f(\mat{T}^{(k)})] - f(\mat{T}^*) \leq \frac{2n}{k+2n}(C^\otimes_f\chi^\otimes + h_0)$ as in (\ref{eq:GapSamplingInequality}) in {\bf Theorem \ref{Thm:ConvAnaBCFW-GA-SROT}}. Comparing two inequalities, BCFW-GA has the additional coefficient $\chi^\otimes $. 
Therefore, we have
\begin{eqnarray*}
	\frac{\sqrt{n}}{\chi(g_:(\vec{x}^{(k)}))^3} \leq 1 \Longleftrightarrow \chi(g_:(\vec{x}^{(k)})) \geq n^{\frac{1}{6}}.
\end{eqnarray*}
\end{Remark}

\subsection{Computational complexity analysis of fast variants of BCFW}
\label{Sec:ComputationalComplexityAnalysisFastVariants}

The algorithmic difference of BCAFW and BCPFW against the baseline BCFW is the additional procedures of the away-steps and the pairwise-steps. These steps have little effect on the complexity. Defining the cardinality of $\mathcal{S}_i$ as $|\mathcal{S}_i|$, the complexity of the away-steps and that of pairwise-steps is $\mathcal{O}(|\mathcal{S}_i|)$ because their procedure is equivalent to the subproblem (\ref{eq:SemiRelaxedSubproblemImplement}). Therefore, the total complexity of those algorithms is $\mathcal{O}(m+|\mathcal{S}_i|)$ at every iteration.  Moreover its cardinality satisfies $1 \leq |\mathcal{S}_i| \leq m$, and thus, the computational costs of BCAFW and BCPFW are approximately $\mathcal{O}(n)$.

The complexity of BCFW-GA is worse than that of BCFW because it needs to adaptively re-compute the probability based on the column-wise duality gap $g_i(\mat{T})$ at every inner or outer iteration. We first consider only the update of the total duality gap at every outer iteration, i.e., epoch. Sampling needs to compute the cumulative sum of its column-wise duality gap, and its computational cost is $\mathcal{O}(n)$. Therefore, the total complexity of gap sampling is $\mathcal{O}(n(m+n))$ at every epoch, including the calculation of the column-wise duality gap. When the total duality gap is updated at every iteration, its computational cost is actually equal to those of updating the column-wise duality gap. Therefore the computation at each outer iteration requires the total complexity $\mathcal{O}(n(m+n))$. In \cite{NesterovSubgradient2014,shalevshwartz2016minimizing}, the computation of cumulative sums is improved by use of tree structure which is built in $\mathcal{O}(n\log n)$.  It computes the cumulative sums and the update of the $i$-th column-wise duality gap in $\mathcal{O}(\log n)$. Therefore, it allows us to compute gap sampling in $\mathcal{O}(n\log n + nm)$. But, we do not explore this structure in the numerical evaluations.

\section{Numerical evaluations}
\label{Sec:NumericalEvalations}

This section evaluates the performances of the proposed BCFW algorithm and the fast variants of BCFW. We first evaluate convergence behaviors of the proposed baseline algorithm of BCFW discussed in Section \ref{Sec:BaselineBCFW}. Then, the comparison evaluations among the fast variants of BCFW proposed in Section \ref{Sec:FastVariantsBCFW} are performed. Finally, we  evaluate the color-transferred images visually using the baseline algorithm of BCFW. It should be noted that, hereinafter, this section uses BCFW-U for the baseline algorithm of BCFW unless otherwise stated. We denote the optimal transport matrix of the {\it unregularized} and {\it unrelaxed} linear programming in (\ref{eq:FormulationOptimalTransport}) and the obtained matrix of (\ref{eq:EuculideanSquareRelaxedProblem}) as $\mat{T}_{\rm LP}^*$ and $\mat{T}$, respectively.
The evaluation metrics are defined as explained below. 
(i) objective function value: $f(\mat{T})$,
(ii) duality gap value: $g(\mat{T})$,
(iii) marginal constraint error: $e_{c} = \| \mat{T}\vec{1}_n - \vec{a} \| + \| \mat{T}^T\vec{1}_m - \vec{b} \|$,
(iv) sparsity: the ratio of zero elements in \mat{T},
(v) transport matrix error: $e_{m} = \| \mat{T} - \mat{T}_{\rm LP}^* \| / \| \mat{T}_{\rm LP}^* \|$, and 
(vi) value error: $e_{v} = | \langle \mat{T},\mat{C} \rangle -\langle \mat{T}_{\rm LP}^*,\mat{C} \rangle |/|\langle \mat{T}_{\rm LP}^*,\mat{C} \rangle|$.
The algorithms are initialized from the same initialization point $\mat{T}^{(0)}$, of which first row is set $\vec{b}$. The algorithms are stopped when the iteration count reaches $1000$ epochs unless otherwise stated. We selected the relaxation parameters $\lambda= \{10^{-7},10^{-8},10^{-9},10^{-10},10^{-11}\}$ from our preliminary evaluations. All the experiments are executed on a 3.7 GHz Intel Core i5 PC with 64 GB RAM. Finally, this experiment uses two public domain images, which are source image ``Gangshan District" by Boris Smokrovic, and reference image, and ``Minesota landscape arboretum" by Shannon Kunkle. 

\subsection{Configurations for color transfer problem}
This experiment addresses the OT-based color transfer problem \cite{ColortransferOT}, which is an effective application of the semi-relaxed formulation. We mainly follow the configurations introduced in \cite{blondel2018smooth}. Given two images that have three dimensions of RGB, we first extract image features of the images. The algorithm of the feature extraction uses the $k$-means algorithm for image quantization. We used the {\it litekmeans} package. After executing $k$-means with a predefined number of classes, all the pixels in the image are assigned into each class. Averaging all the pixel values assigned in each class yields weight vectors, i.e., centroids. By following the procedure, we obtain $m$ color centroids $\vec{x}_1, \vec{x}_2, \ldots, \vec{x}_m \in \mathbb{R}^3$. Additionally, we obtain a color histogram $\vec{a} \in \Delta_m$ by counting of the assigned pixels for $m$ classes. Similarly, $n$ color centroids $\vec{y}_1, \vec{y}_2, \ldots, \vec{y}_n \in \mathbb{R}^3$ and $\vec{b} \in \Delta_n$ are obtained. Finally, the empirical distributions are obtained as 
	$\alpha = \sum_{i=1}^m{a}_i\delta_{\vec{x}_i},\beta = \sum_{i=1}^n{b}_i\delta_{\vec{y}_i}$.
The cost matrix \mat{C} in (\ref{eq:FormulationOptimalTransport}) is calculated as ${C}_{i,j} = \|\vec{x}_i-\vec{y}_j\|_2$. After obtaining a transport matrix $\mat{T} \in \mathbb{R}_+^{m \times n}$ by solving the optimization problem using these values, the new $i$-th color centroid $ \hat{\vec{x}}_i$ is calculated using the following projection operator: 
\begin{equation}
\label{eq:barycentric_projection}
 \hat{\vec{x}}_i = \defargmin_{\scriptsize \vec{x} \in \mathbb{R}^3}\ \  \sum_{j=1}^n {T}_{i,j}\|  \vec{x}-\vec{y}_j\|_2 =\frac{\sum_{j=1}^n {T}_{i,j}{y}_j}{\sum_{j=1}^n{T}_{i,j}}.
\end{equation}
Finally, we recover a new {\it color-transformed} image by substituting $\hat{\vec{x}}_i$ into $\vec{x}_i$.

\subsection{Evaluations of baseline BCFW}
\label{Sec:EvalBaselineBCFW}

\subsubsection{Approximation error and convergence behavior}

This subsection evaluates the empirical approximation errors, and the convergence behaviors. The comparison algorithms are the projected gradient descent (PGD) method and the fast iterative shrinkage-thresholding algorithm (FISTA) \cite{Beck_2009_SIAMIS} for the semi-relaxed OT problem defined in (\ref{eq:EuculideanSquareRelaxedProblem}). As for our proposed algorithms, we used the decay stepsize rule (DEC), i.e., $\gamma=2/(k+2)$ and $\gamma=2n/(k+2n)$ in FW and BCFW, respectively. The exact line-search stepsize rule (ELS) $\gamma_{\rm LS}$ is also used. The corresponding FW and BCFW algorithms are denoted as FW-DEC and FW-ELS, and BCFW-U-DEC and BCFW-U-ELS, respectively.

\begin{figure}[t]
\begin{center}
	\begin{minipage}[t]{0.32\textwidth}
	\begin{center}
		\includegraphics[width=\textwidth]{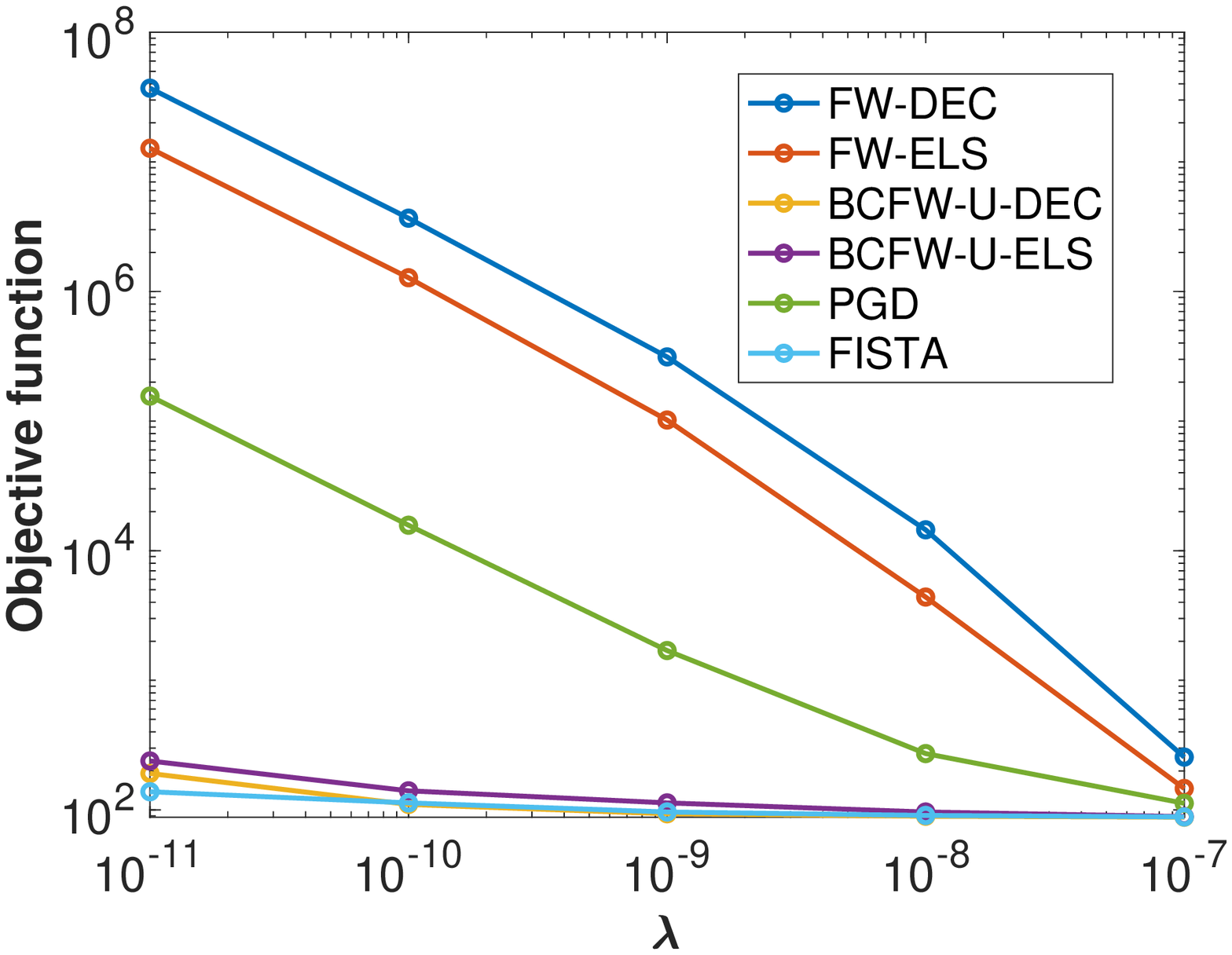}\\
		
		{\footnotesize (a) objective value : $f(\mat{T})$}
		
	\end{center} 
	\end{minipage}
	\hspace*{-0.2cm}		
	\begin{minipage}[t]{0.32\textwidth}
	\begin{center}
		\includegraphics[width=\textwidth]{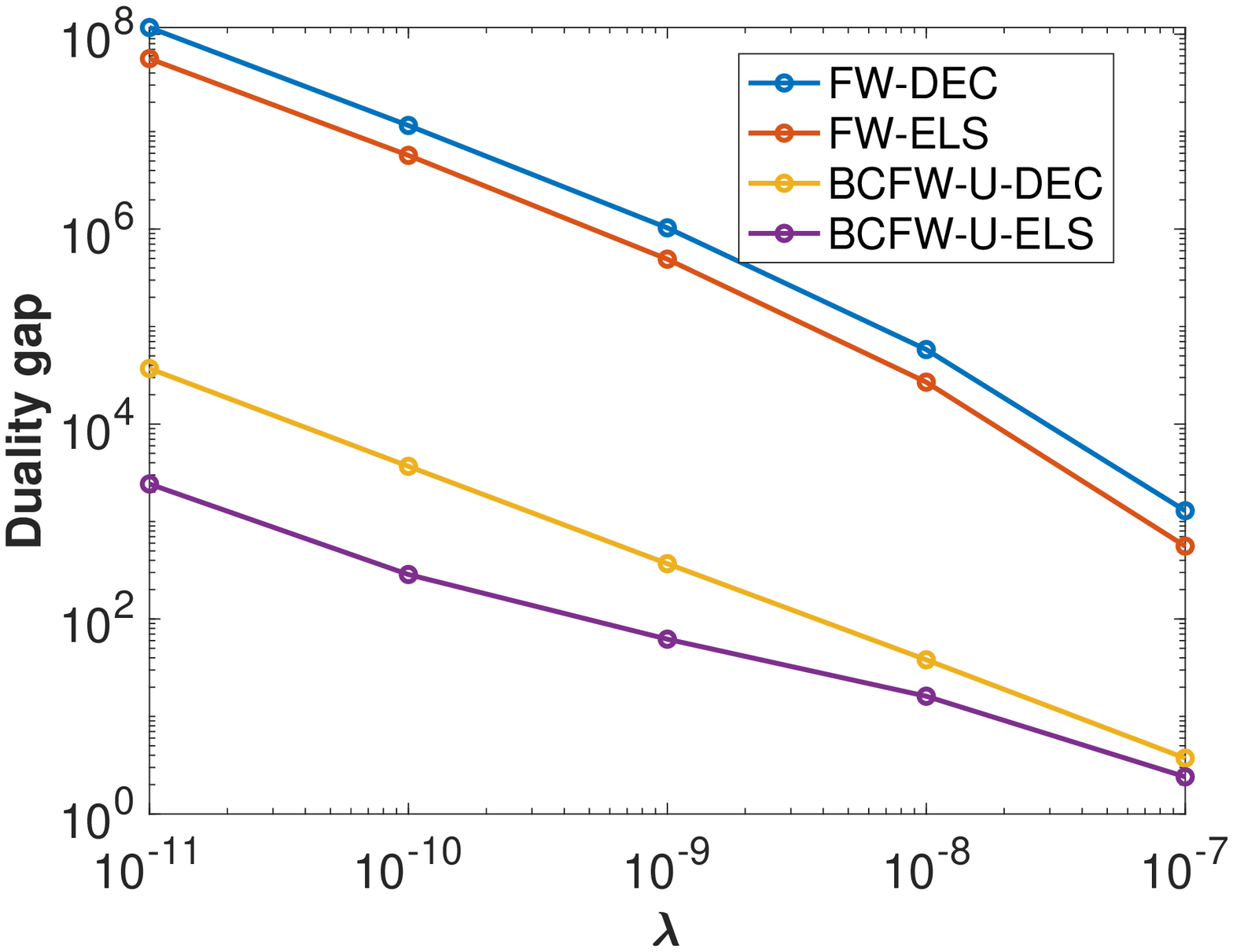}\\
		
		{\footnotesize (b) duality gap : $g(\mat{T})$}
		
	\end{center} 
	\end{minipage}tt
	\hspace*{-0.2cm}
	\begin{minipage}[t]{0.32\textwidth}
	\begin{center}
		\includegraphics[width=\textwidth]{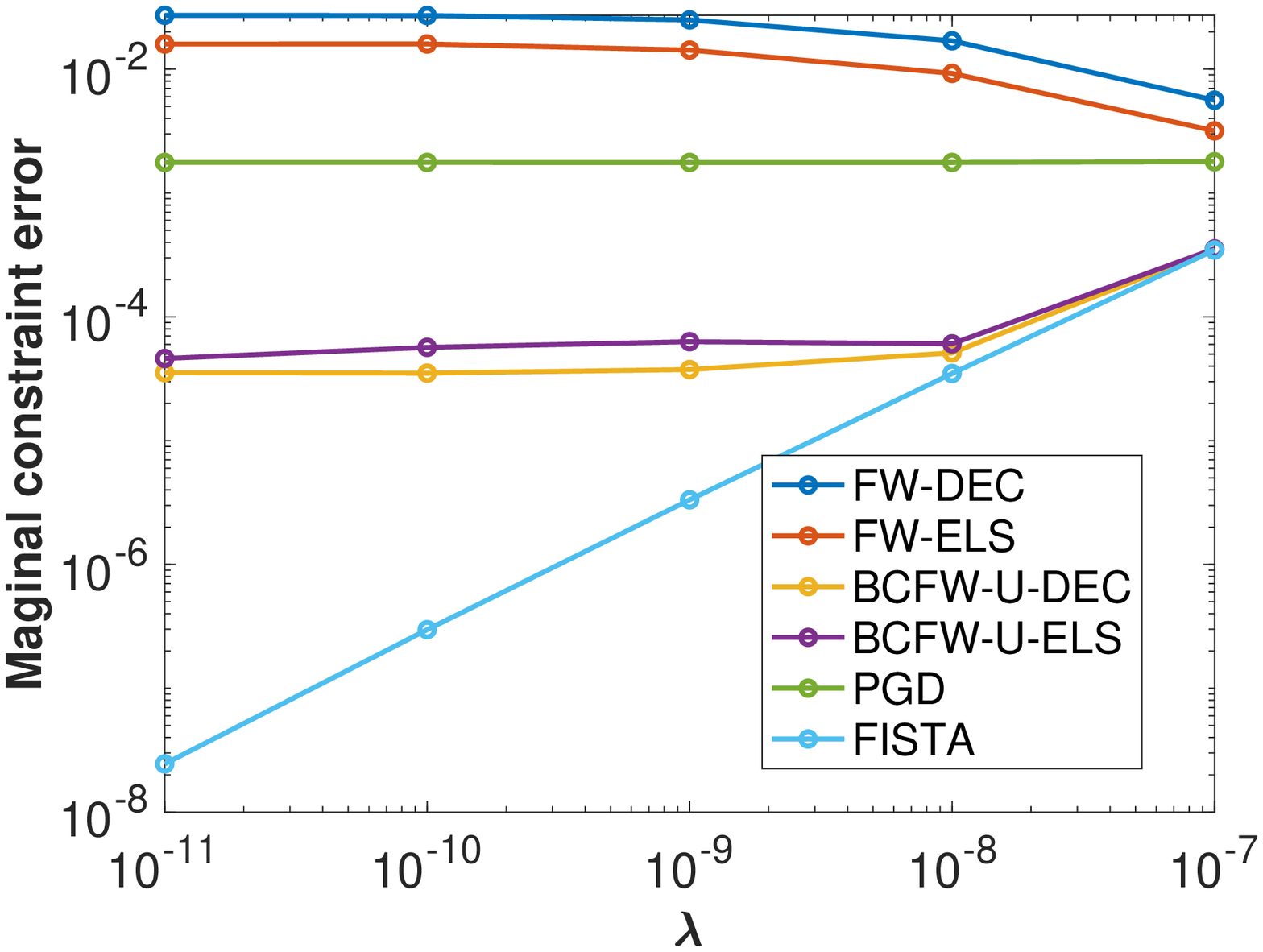}\\
		
		{\footnotesize (c) marginal constraint error : $e_c$}
		
	\end{center} 
	\end{minipage}
	\vspace*{0.4cm}
	
	\begin{minipage}[t]{0.32\textwidth}
	\begin{center}
		\includegraphics[width=\textwidth]{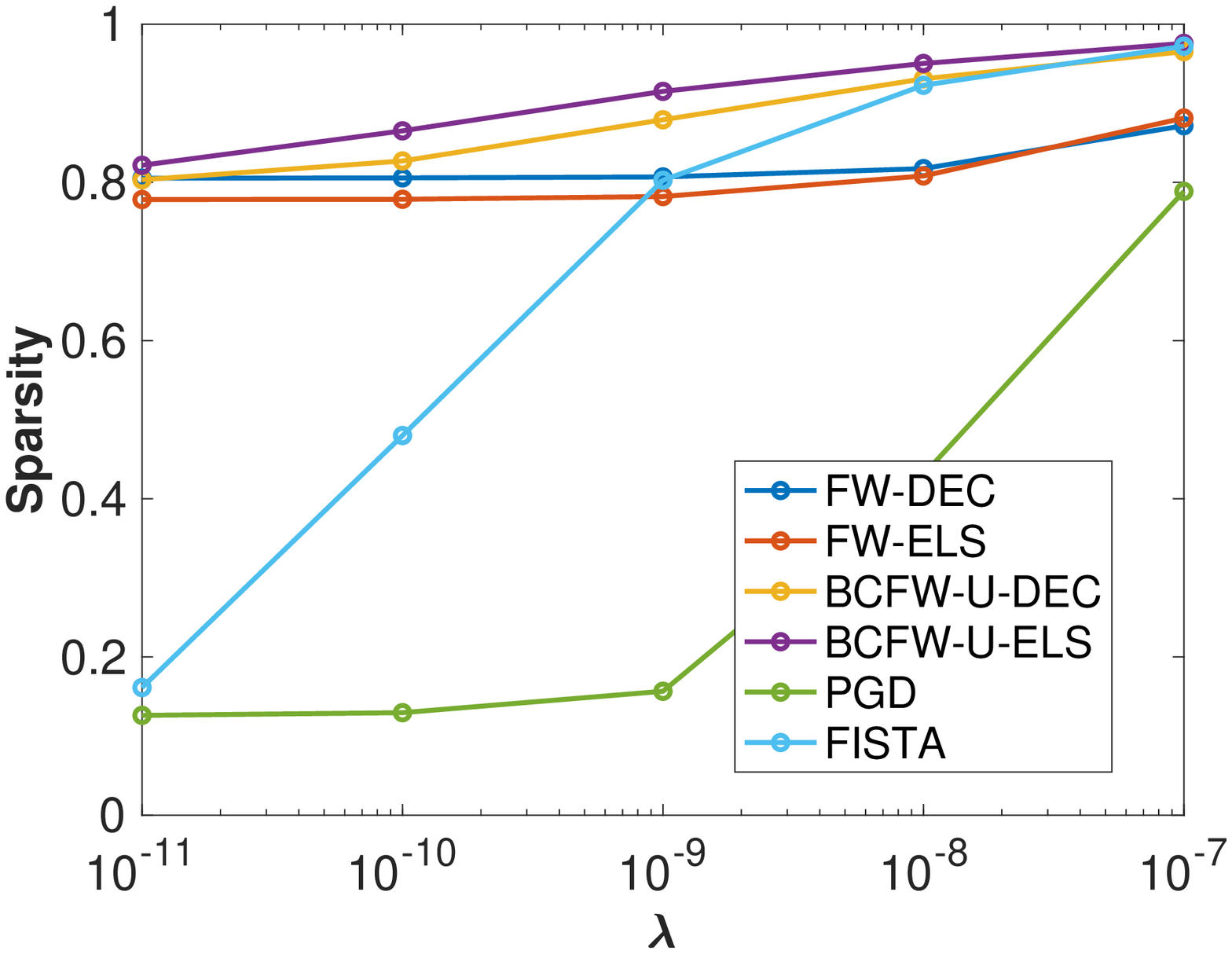}\\
		
		{\footnotesize (d) sparsity}
		
	\end{center} 
	\end{minipage}
	\hspace*{-0.2cm}		
	\begin{minipage}[t]{0.32\textwidth}
	\begin{center}
		\includegraphics[width=\textwidth]{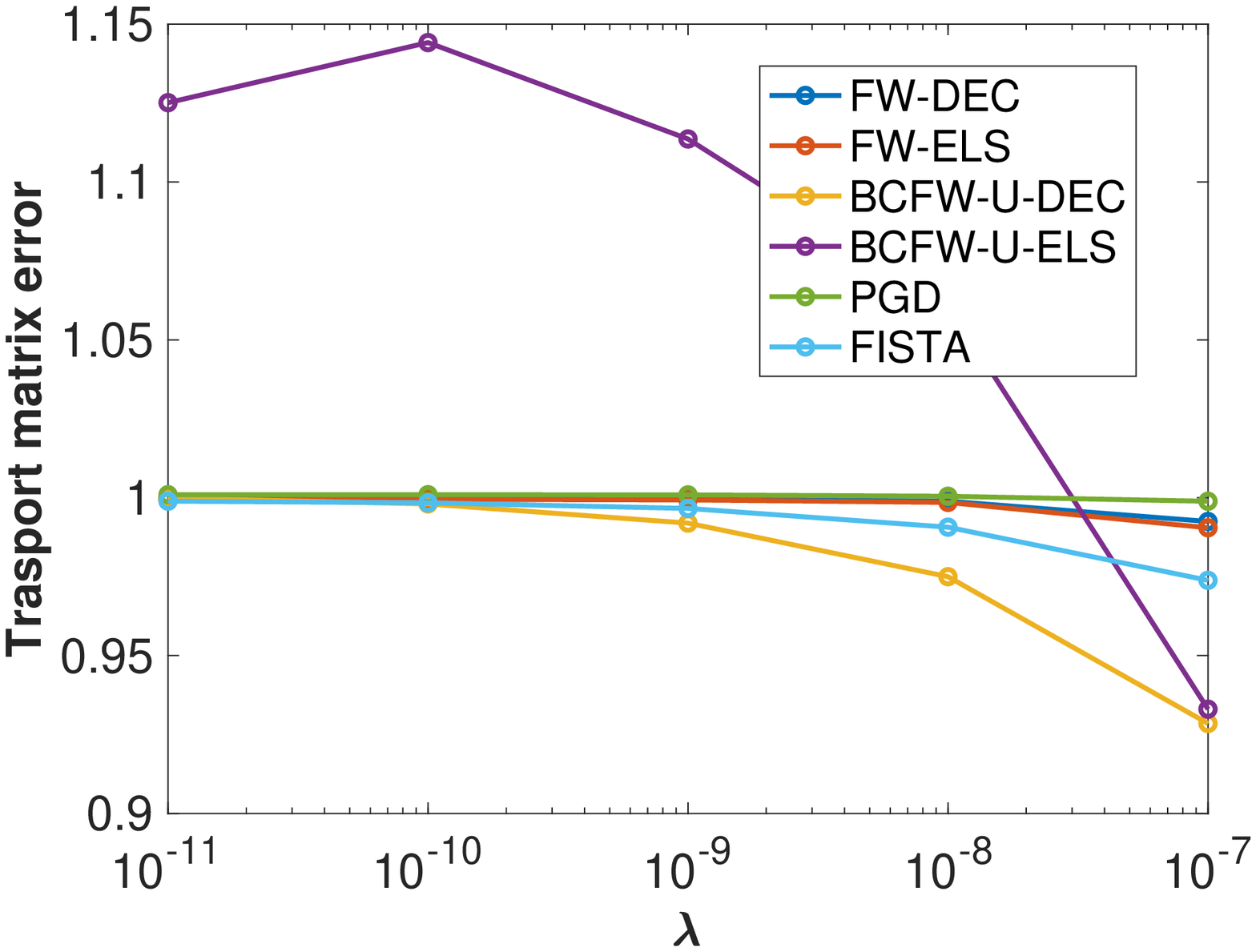}\\
		
		{\footnotesize (e) matrix error : $e_M$}
		
	\end{center} 
	\end{minipage}	
	\hspace*{-0.2cm}			
	\begin{minipage}[t]{0.32\textwidth}
	\begin{center}
		\includegraphics[width=\textwidth]{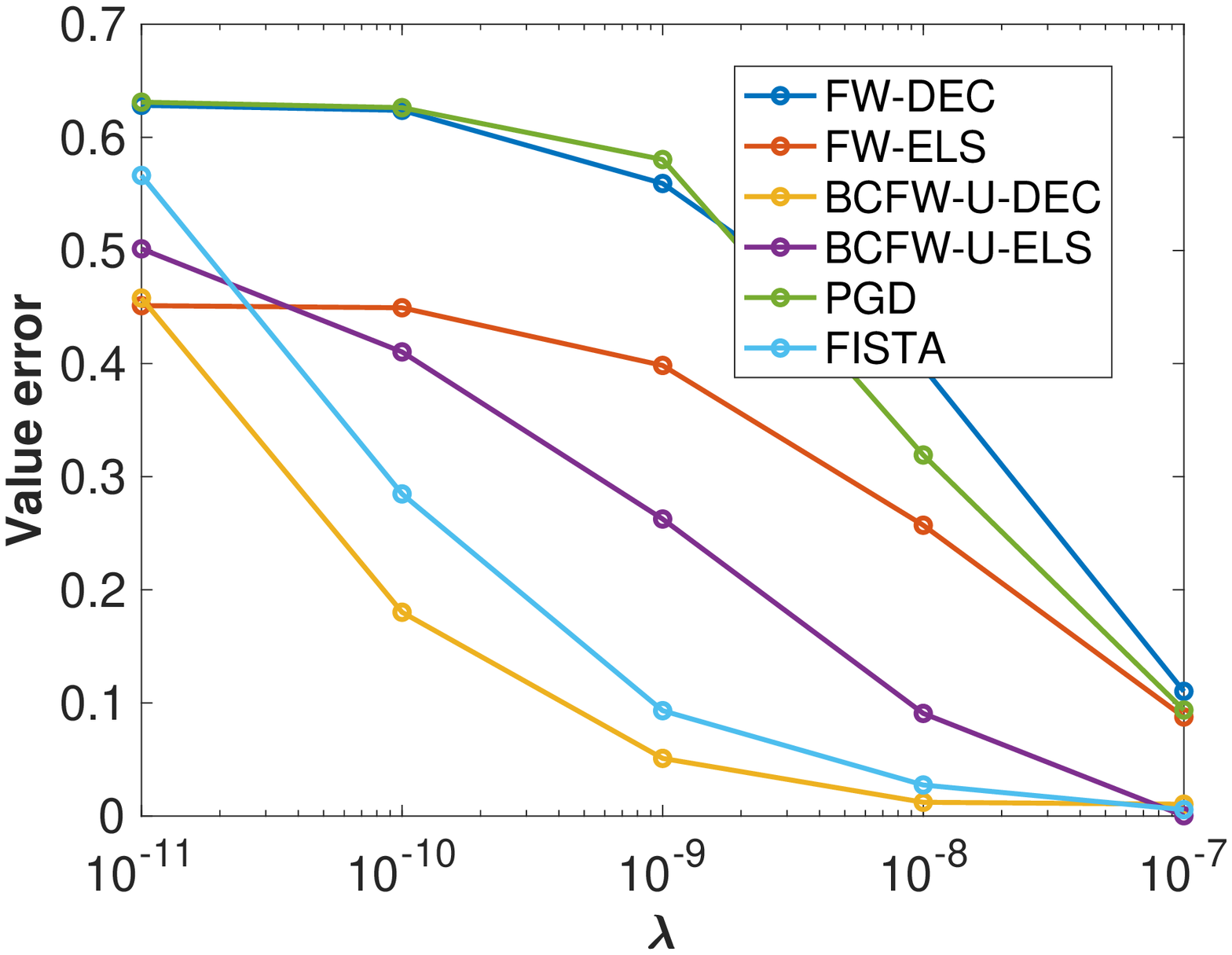}\\
		
		{\footnotesize (f) value error : $e_v$}
		
	\end{center} 
	\end{minipage}
	\vspace*{0.4cm}

	\begin{minipage}[t]{0.32\textwidth}
	\begin{center}
		\includegraphics[width=\textwidth]{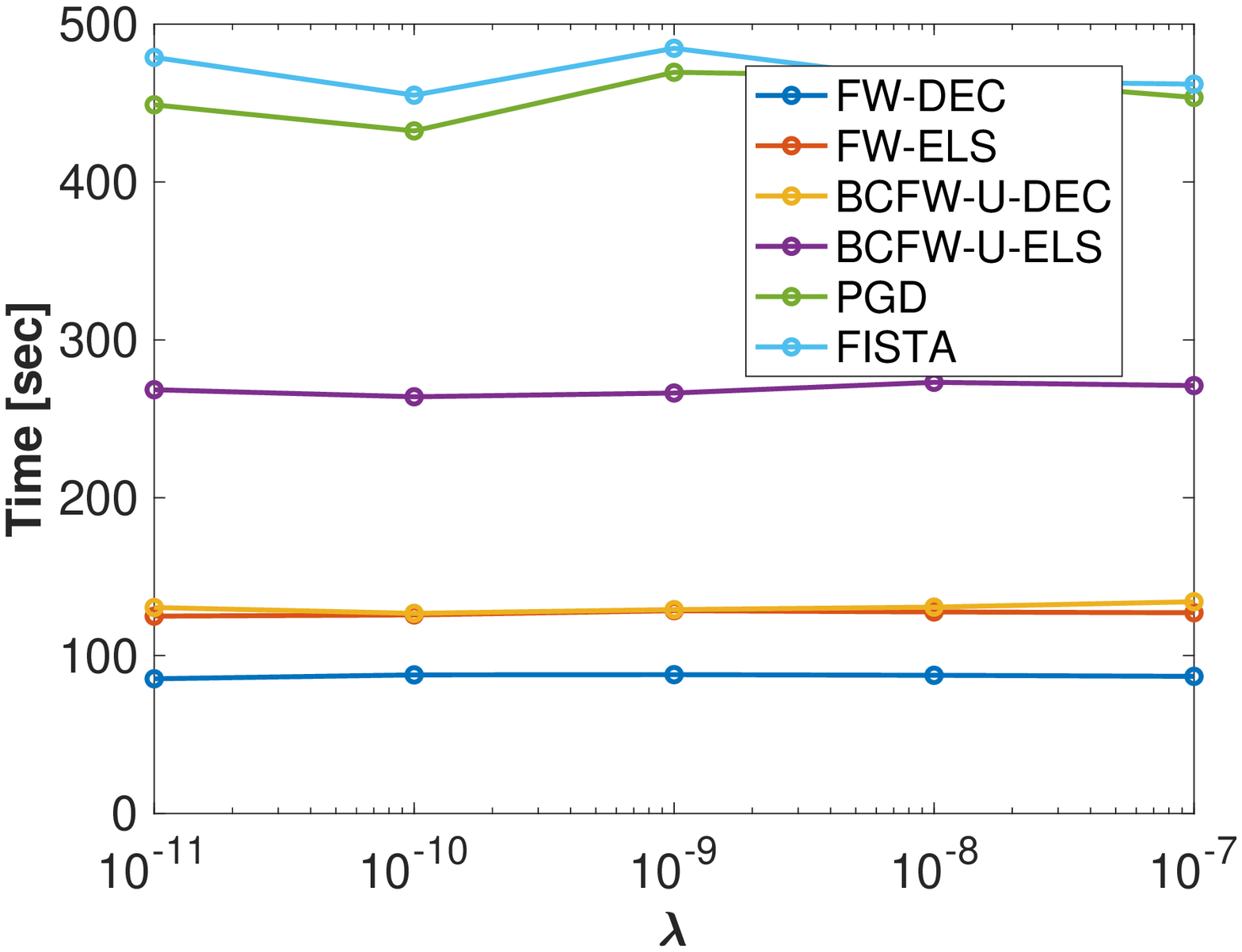}\\
		
		{\footnotesize (g) computational time}
		
	\end{center} 
	\end{minipage}
	\hspace*{-0.2cm}	
	\begin{minipage}[t]{0.32\textwidth}
	\begin{center}
	\end{center} 
	\end{minipage}
	\hspace*{-0.2cm}	
	\begin{minipage}[t]{0.32\textwidth}
	\begin{center}
	\end{center} 
	\end{minipage}	
\caption{Evaluations on different relaxation parameters $\lambda=\{10^{-7},10^{-8},10^{-9},10^{-10},10^{-11}\}$. }
\label{fig:PerformanceLambda}
\end{center}		
\end{figure}

\vspace*{0.2cm}
\noindent
{\bf Comparison across different $\lambda$.} Figure \ref{fig:PerformanceLambda} shows  comparison results of approximation errors across different $\lambda$s. From Figure \ref{fig:PerformanceLambda}(a) and (b), the objective value  and the duality gap indicate the smallest values when $\lambda=10^{-7}$ because the larger $\lambda$s cause smaller relaxation term, thus both values become smaller. Among the algorithm, BCFWs and FISTA give the smallest objective values as shown in (a), and BCFW-U-ELS yields the best duality gap as in (b). 
Regarding the marginal constraint error $e_{c}$ in (c), both the FW algorithms give the worst performances across all $\lambda$s. Both the FW and BCFW algorithms yield stably sparser solutions across different $\lambda$s, and both the BCFW algorithms give the sparsest solutions when $\lambda=10^{-7}$ as seen in (d). Similarly, (e) demonstrates that the cases with $\lambda=10^{-7}$ give the best transport matrix error $e_{m}$ across all the algorithms. From (f), the value errors $e_v$ in all the algorithms indicate the smallest values when $\lambda=10^{-7}$ because the smaller $\lambda$s cause bigger relaxation term, i.e., the second term in (\ref{eq:EuculideanSquareRelaxedProblem}). Specifically, both the BCFW-U-ELS algorithms give the best results. Finally, we can see in (g) that our proposed FW-DEC and BCFW-U-DEC are roughly $4$--$5$ times faster than PGD and FISTA. Also, both the FW-ELS and BCFW-U-ELS with the exact line-search rule are slower than those with the decay stepsize rule. 
Overall, the proposed BCFW-U-DEC/ELS algorithms stably outperform all the other algorithms across all $\lambda$s, and we find that they particularly give the best performances when $\lambda=10^{-7}$.

\begin{figure}[htbp]
\begin{center}
	\begin{minipage}[t]{0.32\textwidth}
	\begin{center}
		\includegraphics[width=\textwidth]{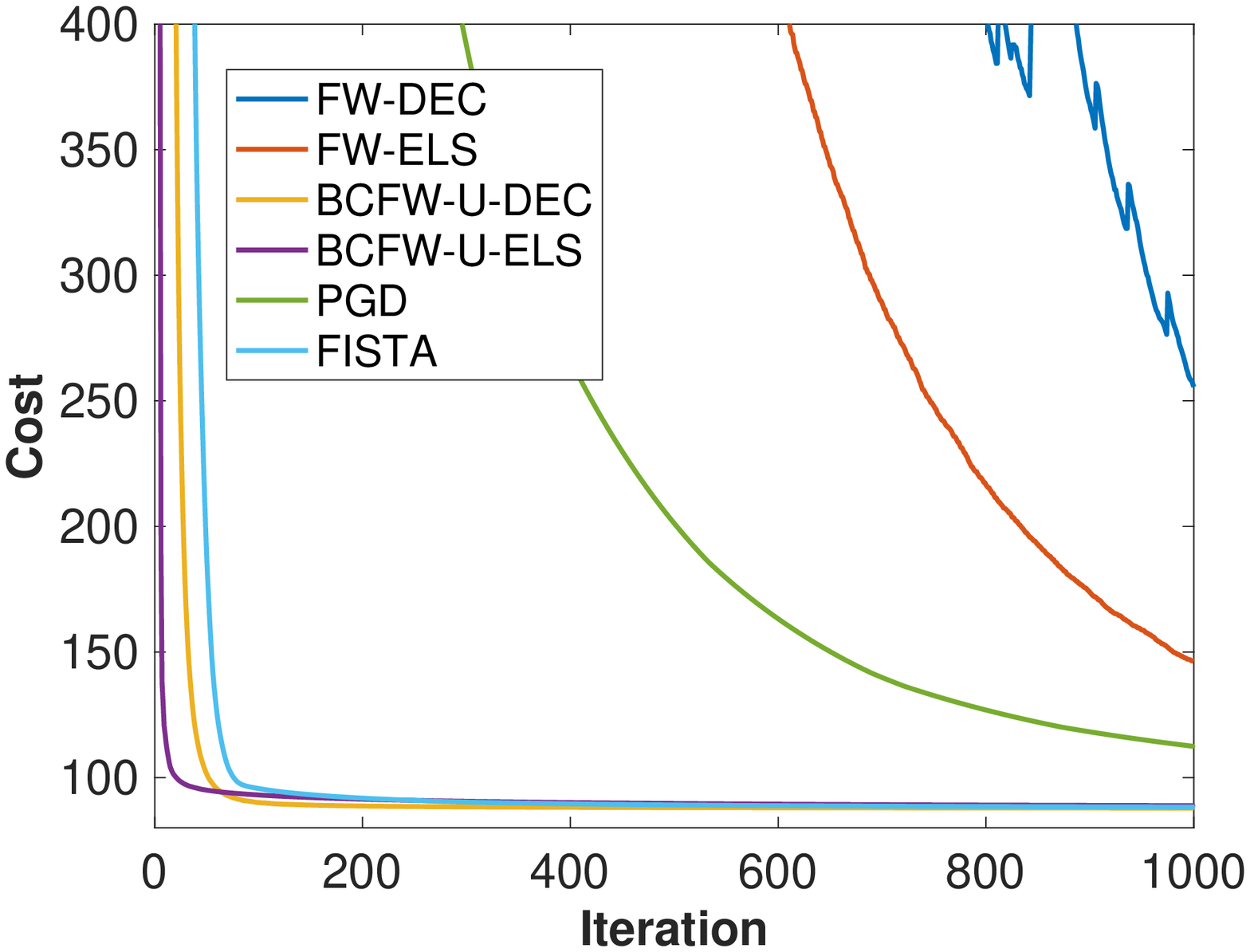}\\
		
		{\footnotesize (a) objective value : $f(\mat{T})$ }
		
	\end{center} 
	\end{minipage}
	\hspace*{-0.2cm}	
	\begin{minipage}[t]{0.32\textwidth}
	\begin{center}
		\includegraphics[width=\textwidth]{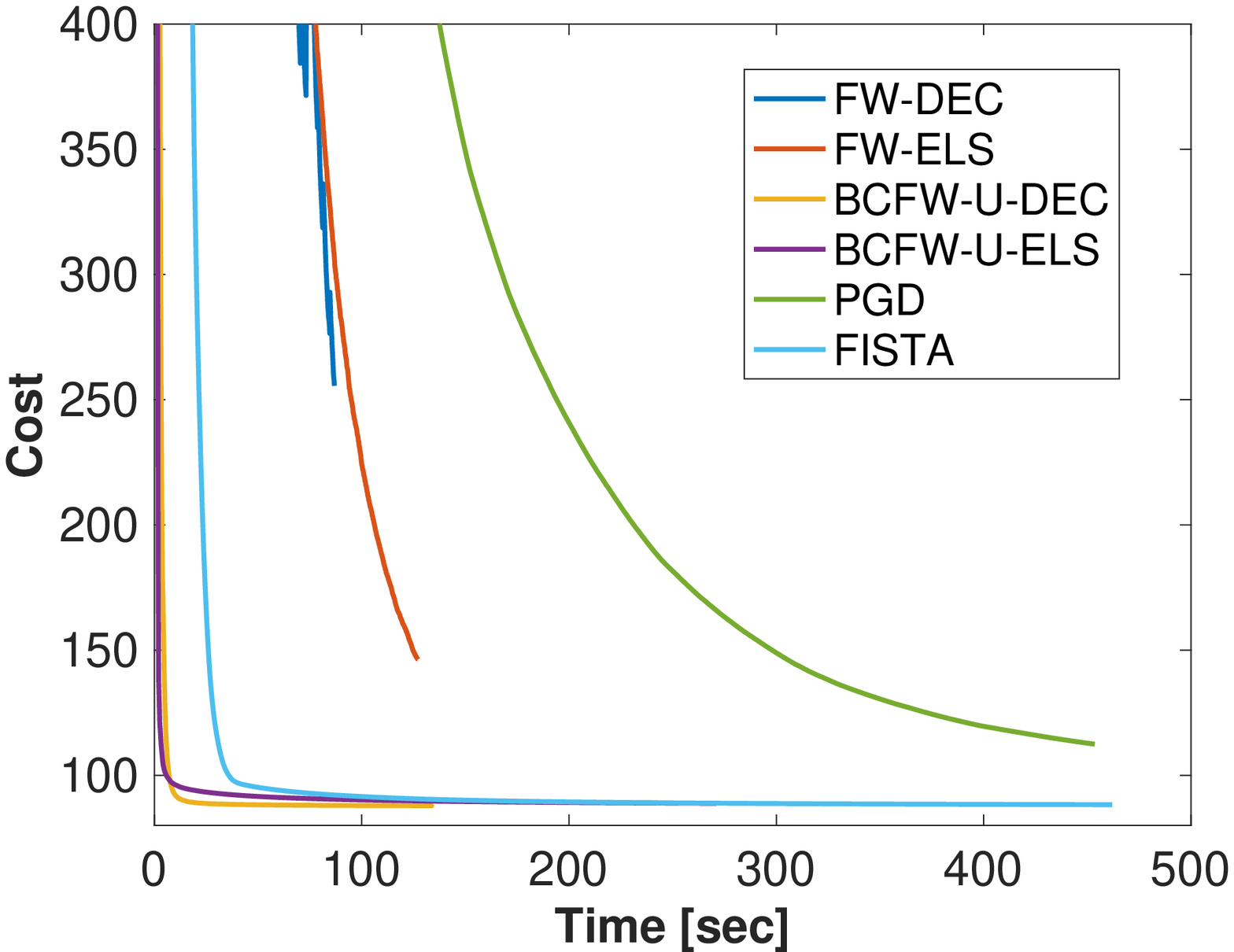}\\
		
		{\footnotesize (b) objective value (time): $f(\mat{T})$ }
		
	\end{center} 
	\end{minipage}
	\hspace*{-0.2cm}	
	\begin{minipage}[t]{0.32\textwidth}
	\begin{center}
		\includegraphics[width=\textwidth]{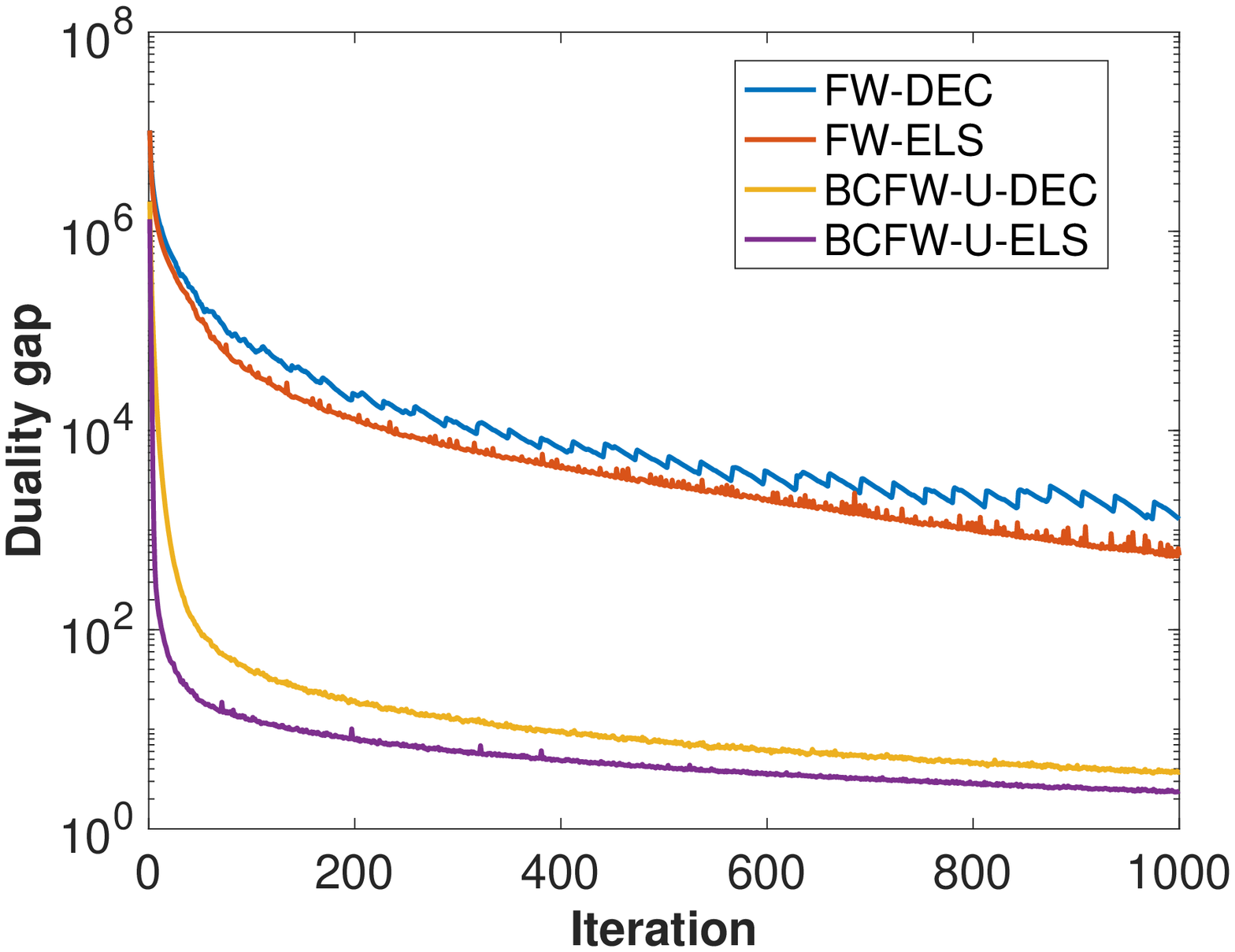}\\
		
		{\footnotesize (c) duality gap : $g(\mat{T})$}
		
	\end{center} 
	\end{minipage}
	\vspace*{0.4cm}

	\begin{minipage}[t]{0.32\textwidth}
	\begin{center}
		\includegraphics[width=\textwidth]{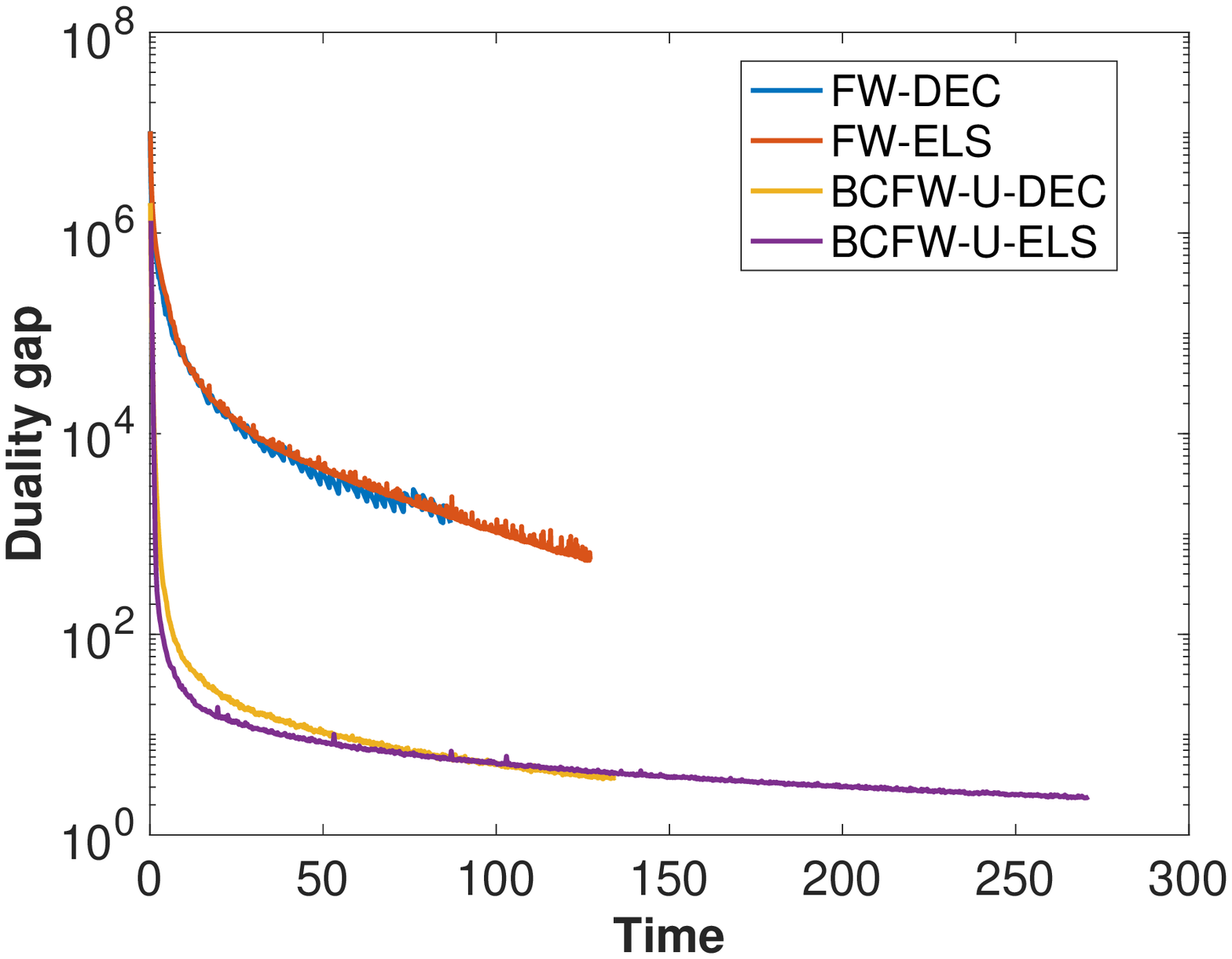}\\
		
		{\footnotesize (d) duality gap (time) : $g(\mat{T})$}
		
	\end{center} 
	\end{minipage}
	\hspace*{-0.2cm}	
	\begin{minipage}[t]{0.32\textwidth}
	\begin{center}
		\includegraphics[width=\textwidth]{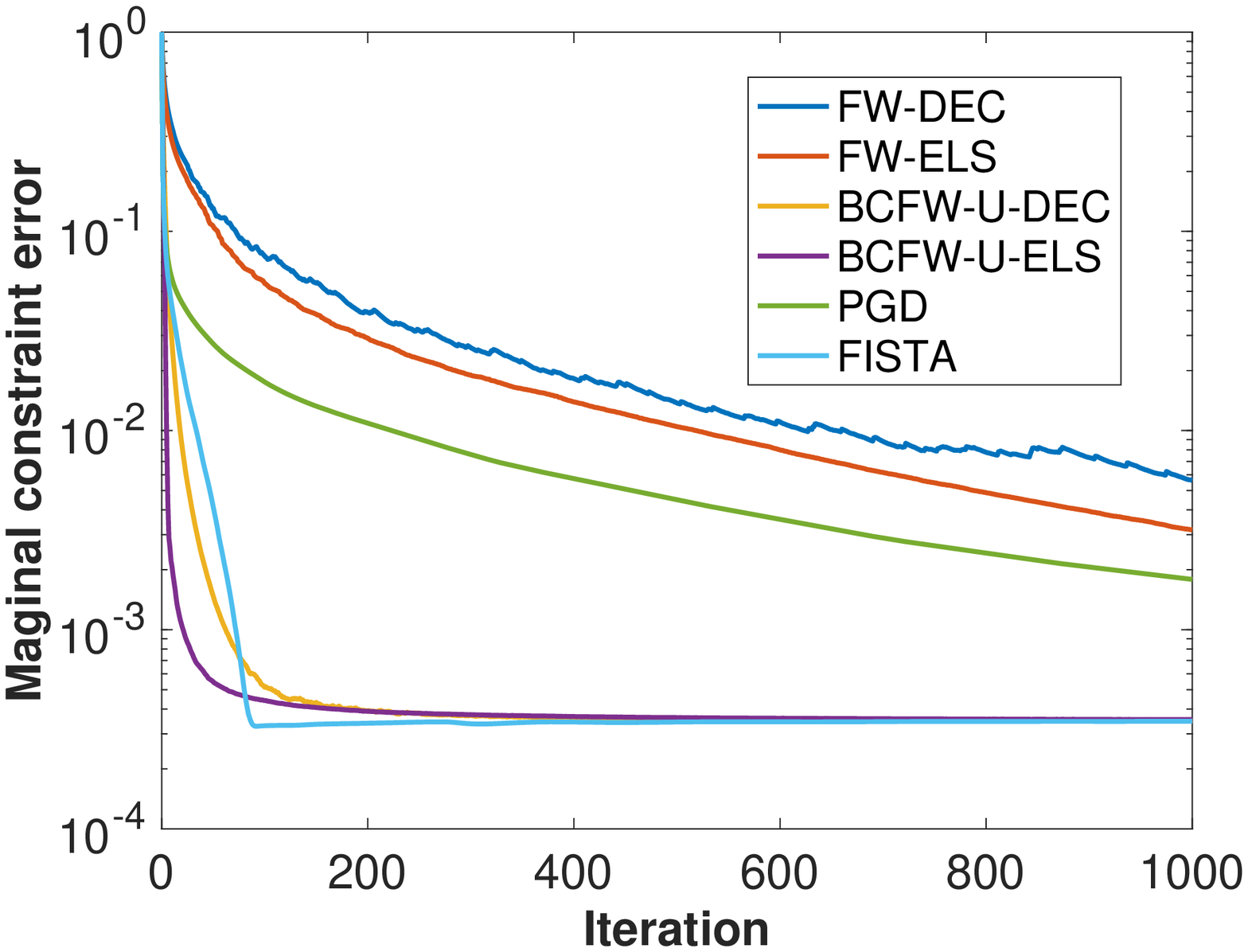}\\
		
		{\footnotesize (e) marginal constraint error : $e_{c}$}
		
	\end{center} 
	\end{minipage}
	\hspace*{-0.2cm}
	\begin{minipage}[t]{0.32\textwidth}
	\begin{center}
		\includegraphics[width=\textwidth]{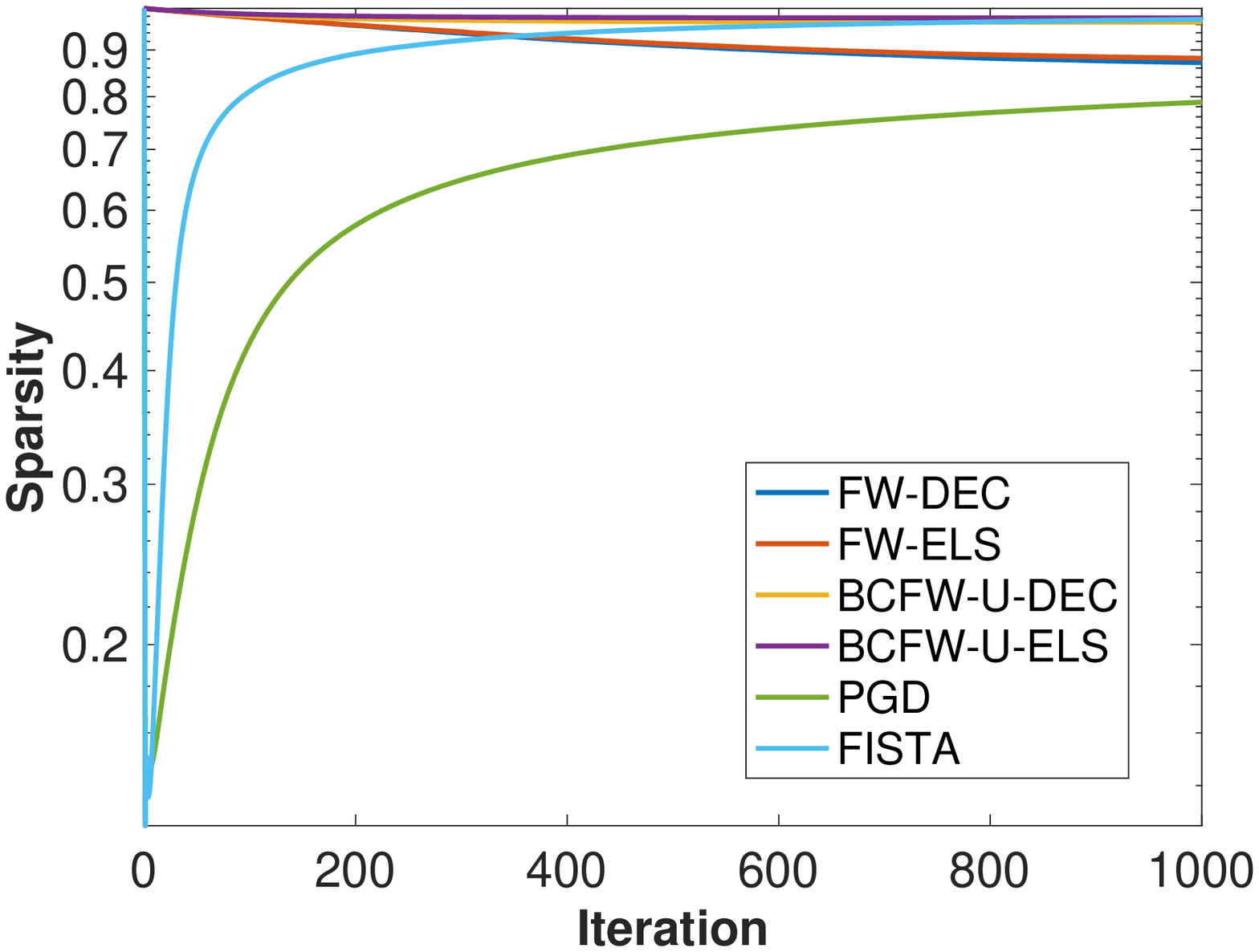}\\
		
		{\footnotesize (f) sparsity}
		
	\end{center} 
	\end{minipage}
	\vspace*{0.4cm}
		
	\begin{minipage}[t]{0.32\textwidth}
	\begin{center}
		\includegraphics[width=\textwidth]{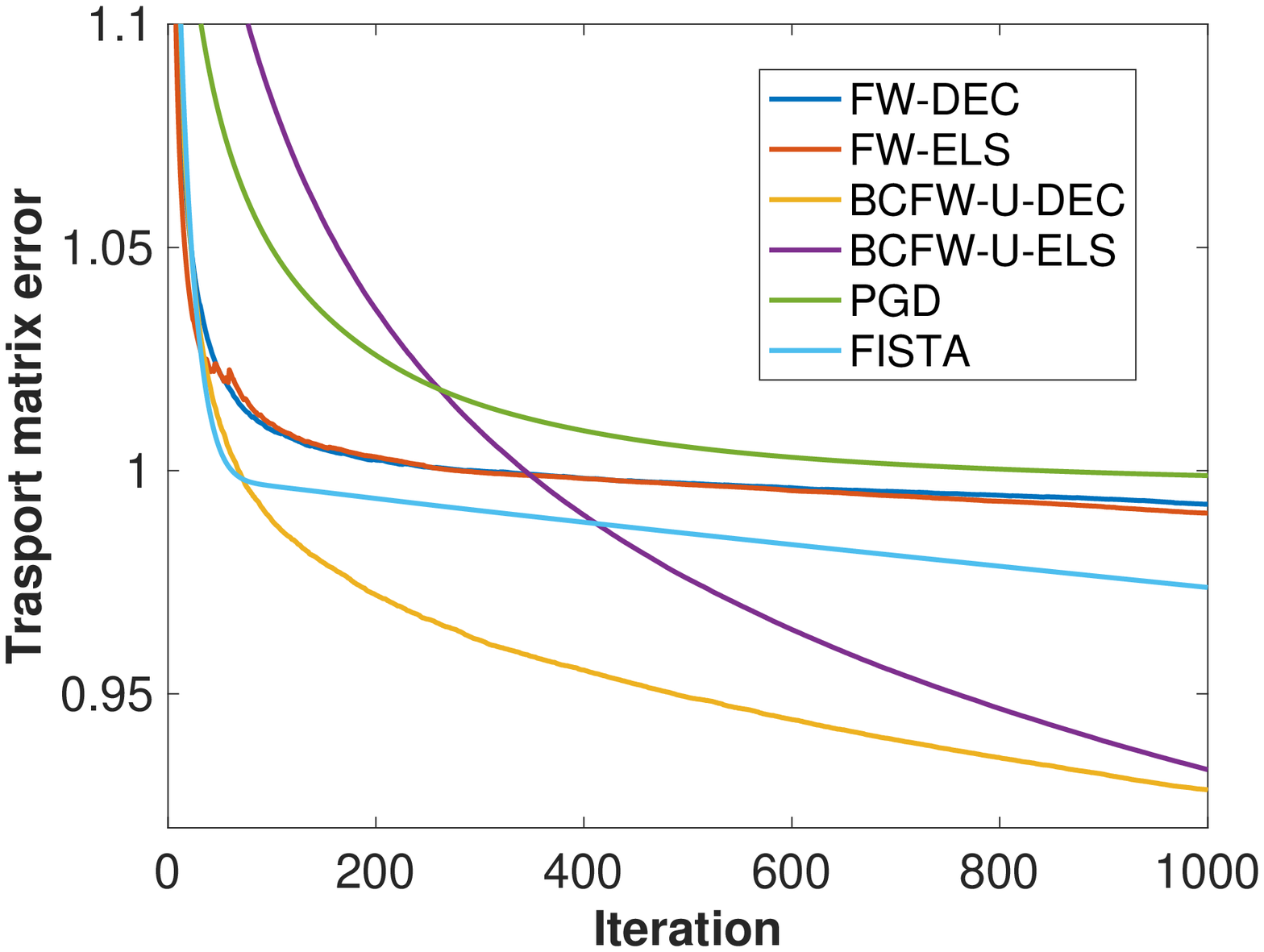}\\
		
		{\footnotesize (g) matrix error : $e_{m}$ }
		
	\end{center} 
	\end{minipage}
	\hspace*{-0.2cm}
	\begin{minipage}[t]{0.32\textwidth}
	\begin{center}
		\includegraphics[width=\textwidth]{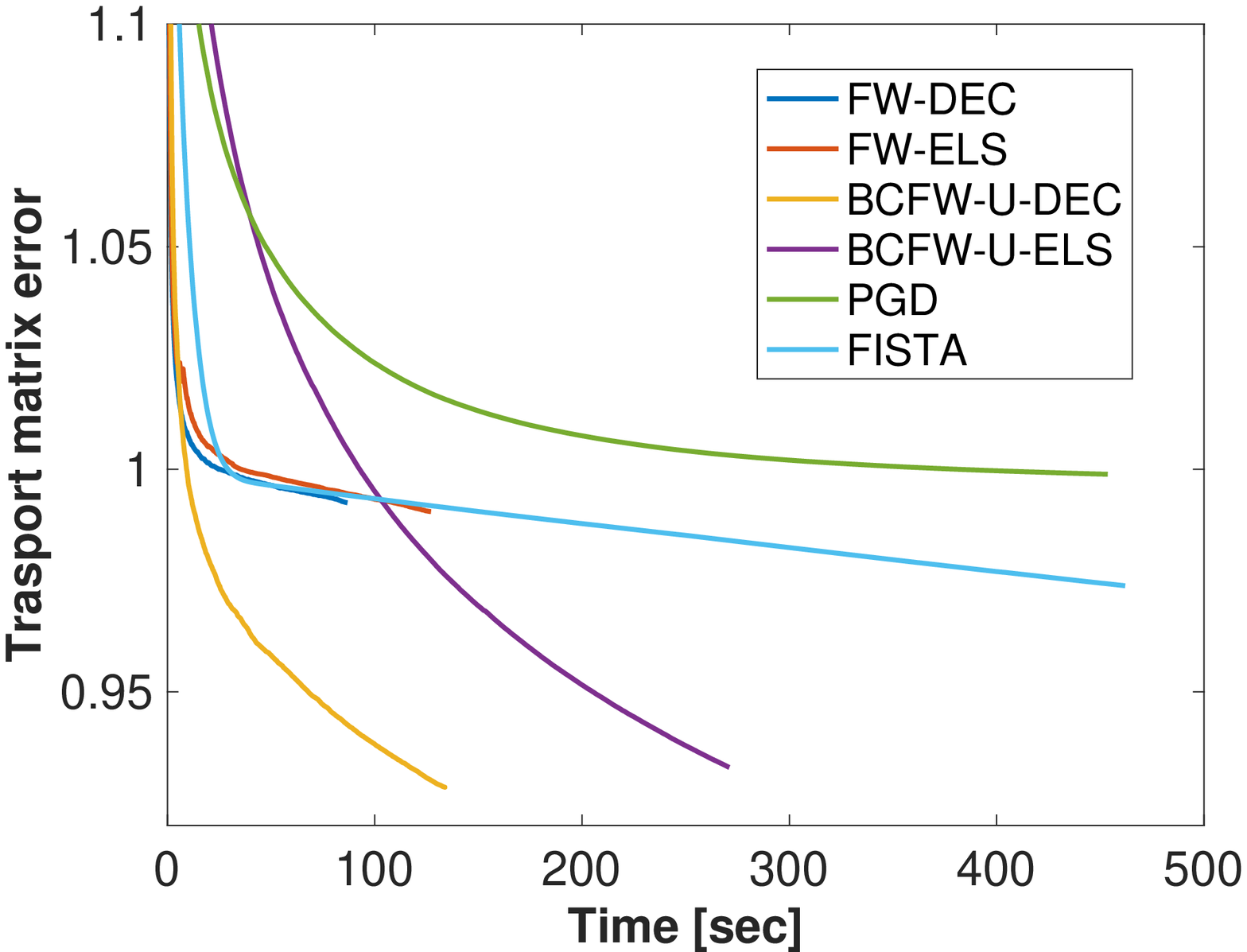}\\
		
		{\footnotesize (h) matrix error (time) : $e_{m}$ }
		
	\end{center} 
	\end{minipage}
	\hspace*{-0.2cm}	
	\begin{minipage}[t]{0.32\textwidth}
	\begin{center}
		\includegraphics[width=\textwidth]{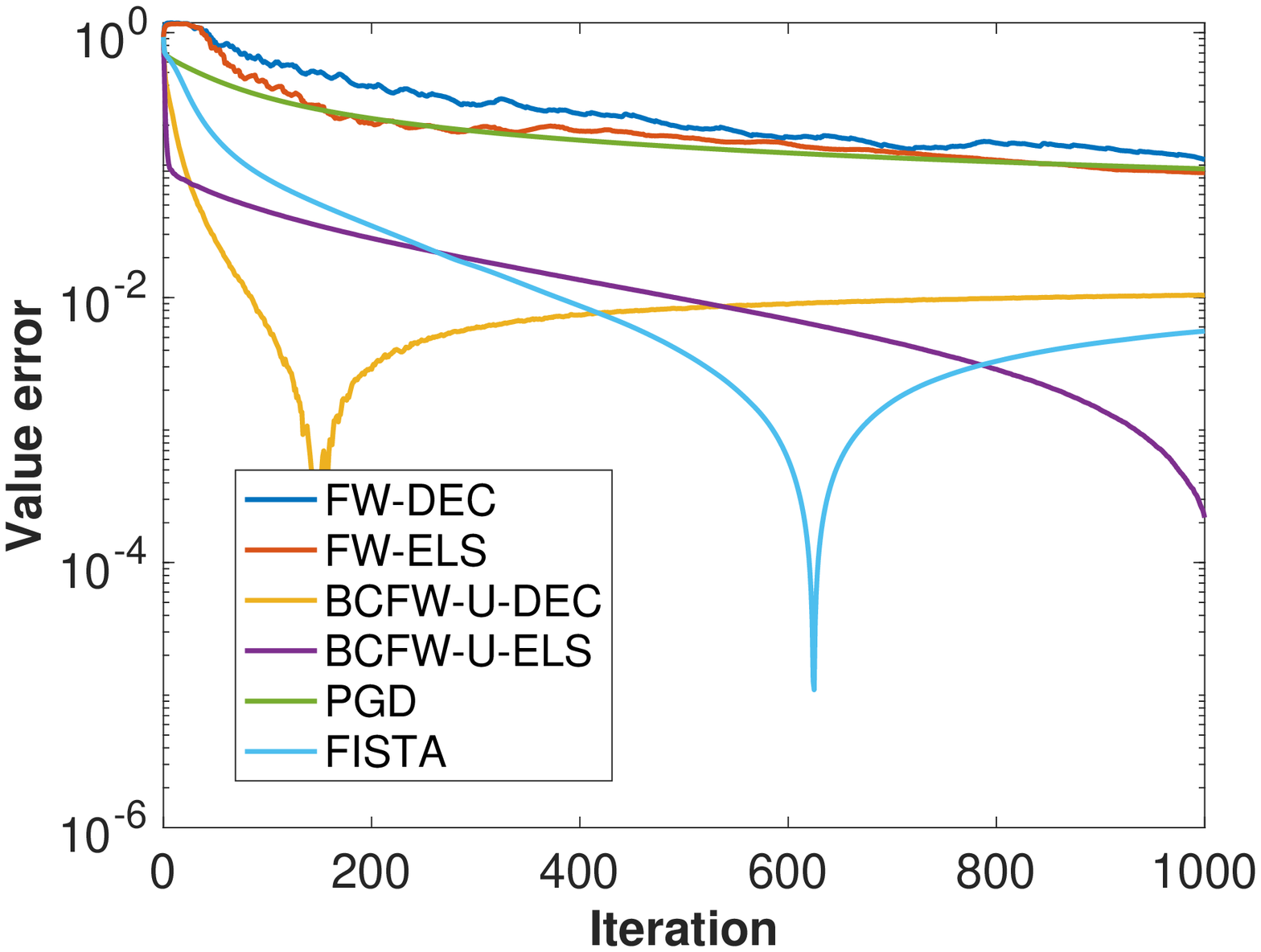}\\
		
		{\footnotesize (i) value error : $e_v$}
		
	\end{center} 
	\end{minipage}		
\caption{Evaluations on convergence ($\lambda = 10^{-7}$).}
\label{fig:ConvergencePerformance}
\end{center}
\end{figure}

\vspace*{0.2cm}
\noindent
{\bf Convergence behavior.} Addressing one specific case with $\lambda=10^{-7}$, we discuss  convergence performances of the approximation errors. Figure \ref{fig:ConvergencePerformance} shows the results. The objective value and duality gap in Figures \ref{fig:ConvergencePerformance}(a)-(d) reveal that the BCFW-U algorithms are faster than the FW algorithm. In terms of iteration, the exact line-search stepsize rules are superior to the decay stepsize rules in both the FW and BCFW-U algorithms, but the superiorities of the exact line-search become diminished in terms of computational time. Regarding the marginal constraint error $e_c$ in (e), FISTA gives the best performance thanks to the expensive orthogonal projection on the simplex, but both the BCFW algorithms give similar performances. As for the transport matrix error $e_{m}$ in (g) and (h), the BCFW-U algorithm with the decay stepsize rule gives the best performance. 
Lastly, it is understandable that the proposed algorithms stably generate sparser solutions due to the algorithm architecture. 

\subsubsection{Comparison with smoothed dual algorithms}
We further compared with the smoothed dual algorithm (Smoothed Dual) and its semi-dual variant (Smoothed SemiDual) \cite{blondel2018smooth}. Their cost functions are actually different from (\ref{eq:EuculideanSquareRelaxedProblem}), thus, we cannot directly compare them with ours. Nevertheless, the transport matrix error $e_{m}$ and the marginal constraint error $e_{c}$ would yield valuable insights. Additionally, we measured computational times for rough comparisons although the author's code\footnote{\url{https://github.com/mblondel/smooth-ot/}} is written in Python whereas ours are MATLAB. The squared 2-norm is used for the smoothed algorithms, and their regularization parameter $\gamma$ are set as $\gamma=\{10^{-2},10^0,10^{2}\}$, which is inside the the range of the original paper. We used the relaxation parameter $\lambda=\{10^{-11},10^{-9},10^{-7}\}$ for BCFW-U, which has the same range of orders as $\gamma$.  
Table \ref{tbl:comp_dual} summarizes $e_{m}$, $e_{c}$ and the computational time. From Table \ref{tbl:comp_dual}, the lowest $e_{m}$ is obtained by Smoothed SemiDual. This is because the smoothed methods regularize the original LP problem, and tend to output a similar $\mat{T}$ as $\mat{T}_{\rm LP}$. However, our proposed algorithms can give stabler $e_{m}$, and smaller marginal constraint errors $e_{c}$. More importantly, it should be emphasized that our proposed algorithms are extremely faster than the smoothed algorithms although the Smooth Dual and SemiDual algorithms call internally the L-BFGS-B solver of {\it scipy} library, which is a widely-used, reliable and fast solver. 

\begin{table}[t]
\begin{center}
\caption{Comparison to smoothed dual and smoothed semi-dual algorithms}
\label{tbl:comp_dual}
\renewcommand{\arraystretch}{1.2}
\begin{tabular}{l|l|l|l|l}
\hline
algorithms  &   \multicolumn{1}{|c|}{$\lambda,\gamma$} & \multicolumn{1}{|c|}{$e_{m}$} & \multicolumn{1}{|c|}{$e_{c}$}  & \multicolumn{1}{|c}{time [s]} \\
\hline
\hline 
BCFW-U-DEC & $10^{-11}$ & 1.00e+00 &3.54e--05, & 4.58e--01 \\
\cline{2-5}
 & $10^{-9}$ & 9.92e0-01& 3.75e--05,& 1.29e+02\\
\cline{2-5}
& $10^{-7}$ & 9.28e--01,& 3.50e-04 & 1.34e+02 \\
\hline 
BCFW-U-ELS & $10^{-11}$ & 1.13e+00 &4.61e--05 & 2.68e+02 \\
\cline{2-5}
 & $10^{-9}$ & 1.11e+00& 6.30e--05& 2.66e+02\\
\cline{2-5}
& $10^{-7}$ & 9.33e--01,& 3.54e-04&2.71e+02 \\
\hline
\hline
Smoothed Dual & $10^{-2}$ &4.65e+02 &1.25e+00&4.75e+02 \\
\cline{2-5}
 & $10^{+0}$ &4.67e--01 &1.25e+00&3.76e+02 \\
\cline{2-5}
 & $10^{+2}$ &1.24e+00 & 3.04e--02& 3.83e+02 \\
\hline
Smoothed SemiDual & $10^{-2}$ & 1.07e+00&1.24e--02&1.08e+03 \\
\cline{2-5}
 & $10^{+0}$ & 1.00e+00&1.23e+00&1.07e+03 \\
\cline{2-5}
 & $10^{+2}$ &5.82e--01& 2.60e--03& 1.09e+03 \\
\hline
\end{tabular}
\end{center}
\end{table}

\subsection{Evaluations of fast variants of BCFW}

\subsubsection{Evaluations on BCAFW and BCPFW}

\begin{figure}[htbp]
\begin{center}
	\begin{minipage}[t]{0.32\textwidth}
	\begin{center}
		\includegraphics[width=\textwidth]{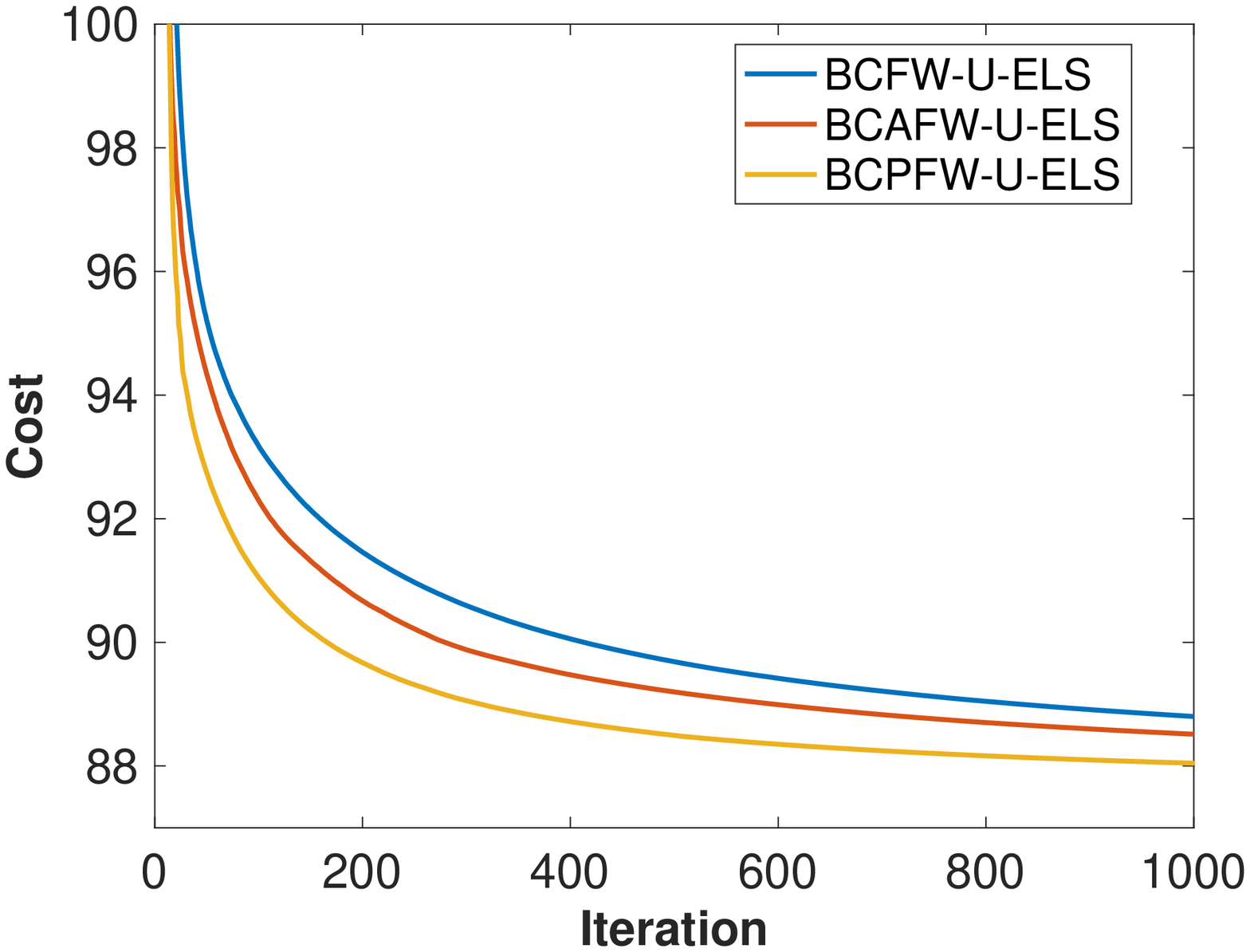}\\
		
		{\footnotesize (a) objective value : $f(\mat{T})$ }
		
	\end{center} 
	\end{minipage}
	\hspace*{-0.2cm}	
	\begin{minipage}[t]{0.32\textwidth}
	\begin{center}
		\includegraphics[width=\textwidth]{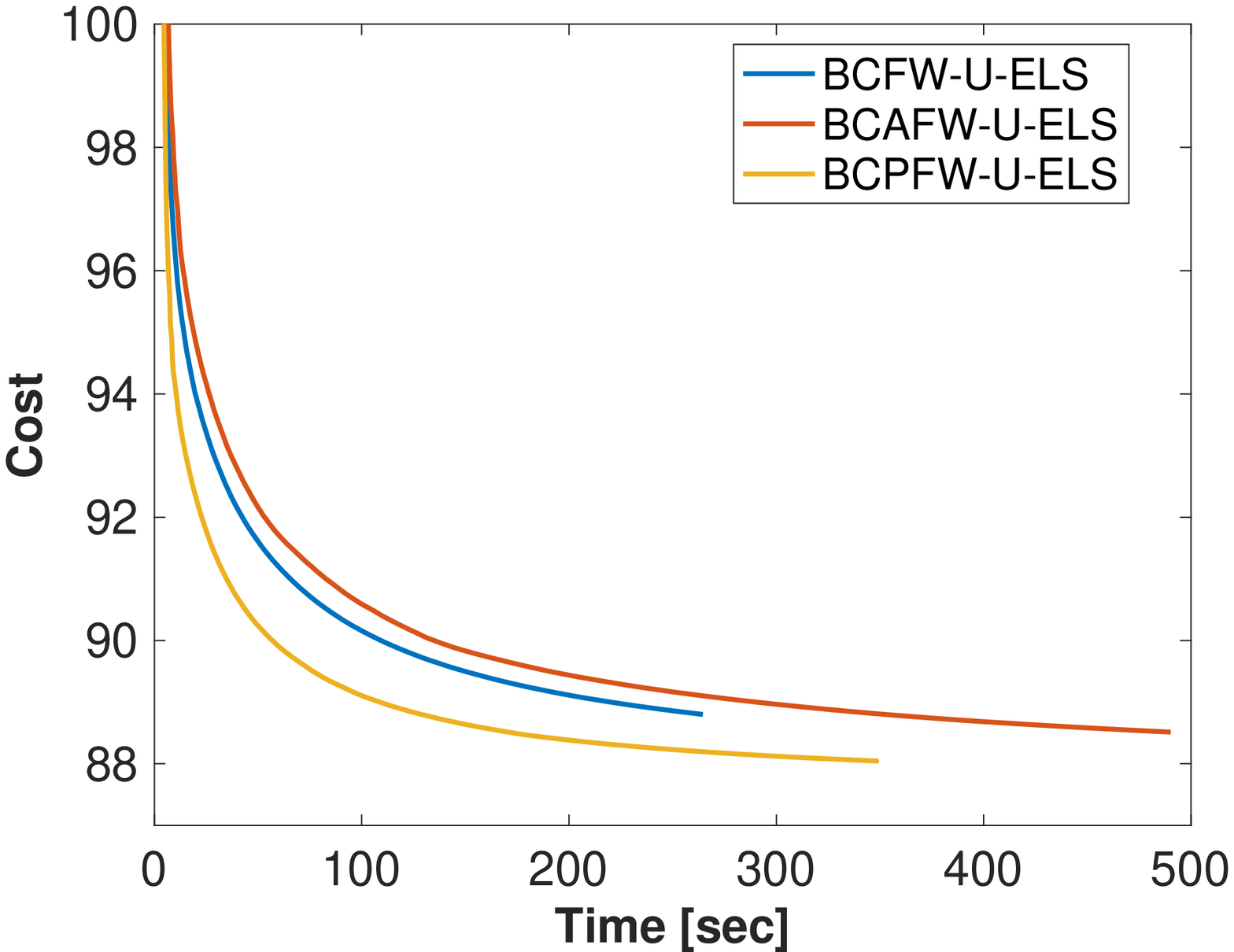}\\
		
		{\footnotesize (b) objective value (time): $f(\mat{T})$ }
		
	\end{center} 
	\end{minipage}
	\hspace*{-0.2cm}	
	\begin{minipage}[t]{0.32\textwidth}
	\begin{center}
		\includegraphics[width=\textwidth]{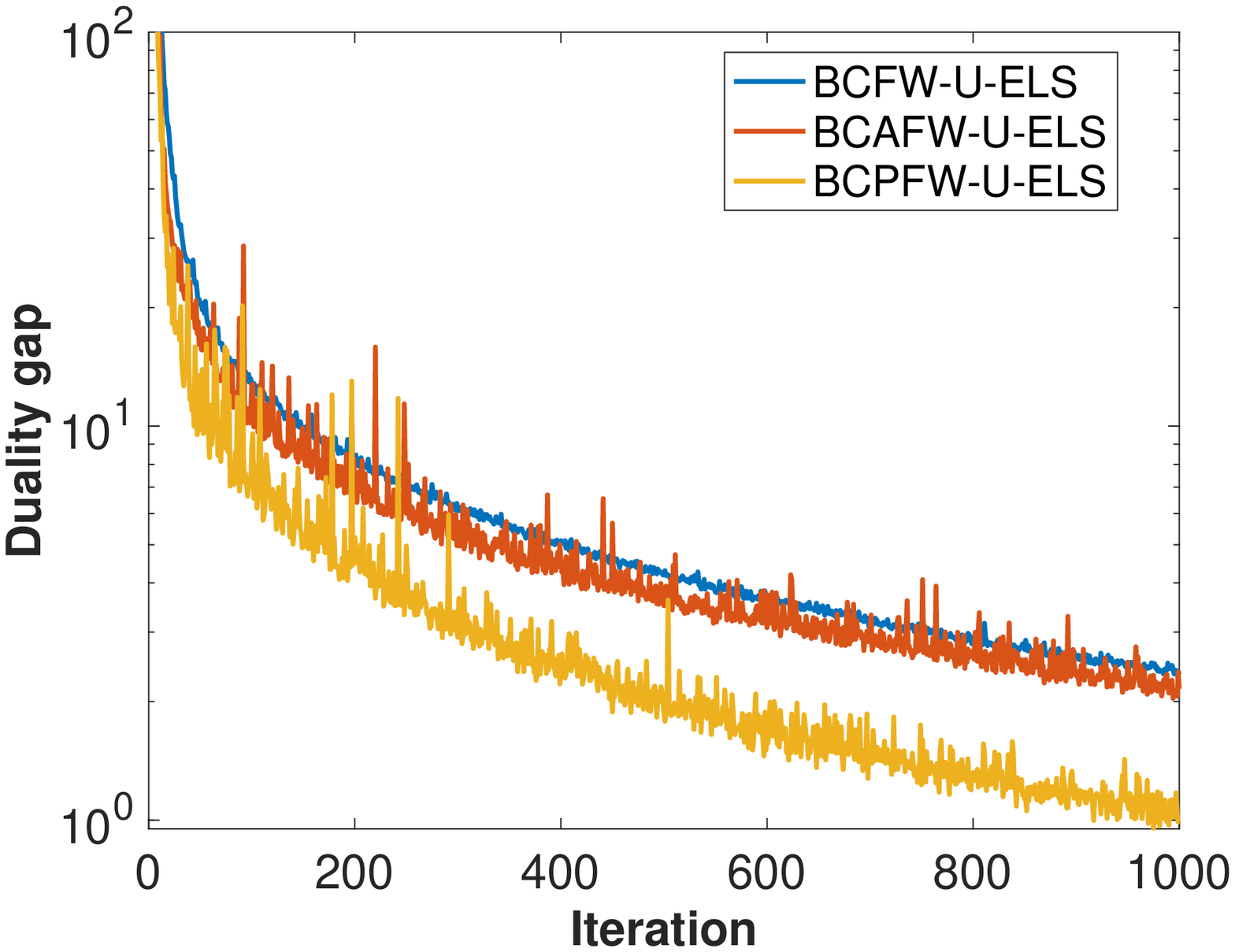}\\
		
		{\footnotesize (c) duality gap : $g(\mat{T})$}
		
	\end{center} 
	\end{minipage}
	\vspace*{0.2cm}

	\begin{minipage}[t]{0.32\textwidth}
	\begin{center}
		\includegraphics[width=\textwidth]{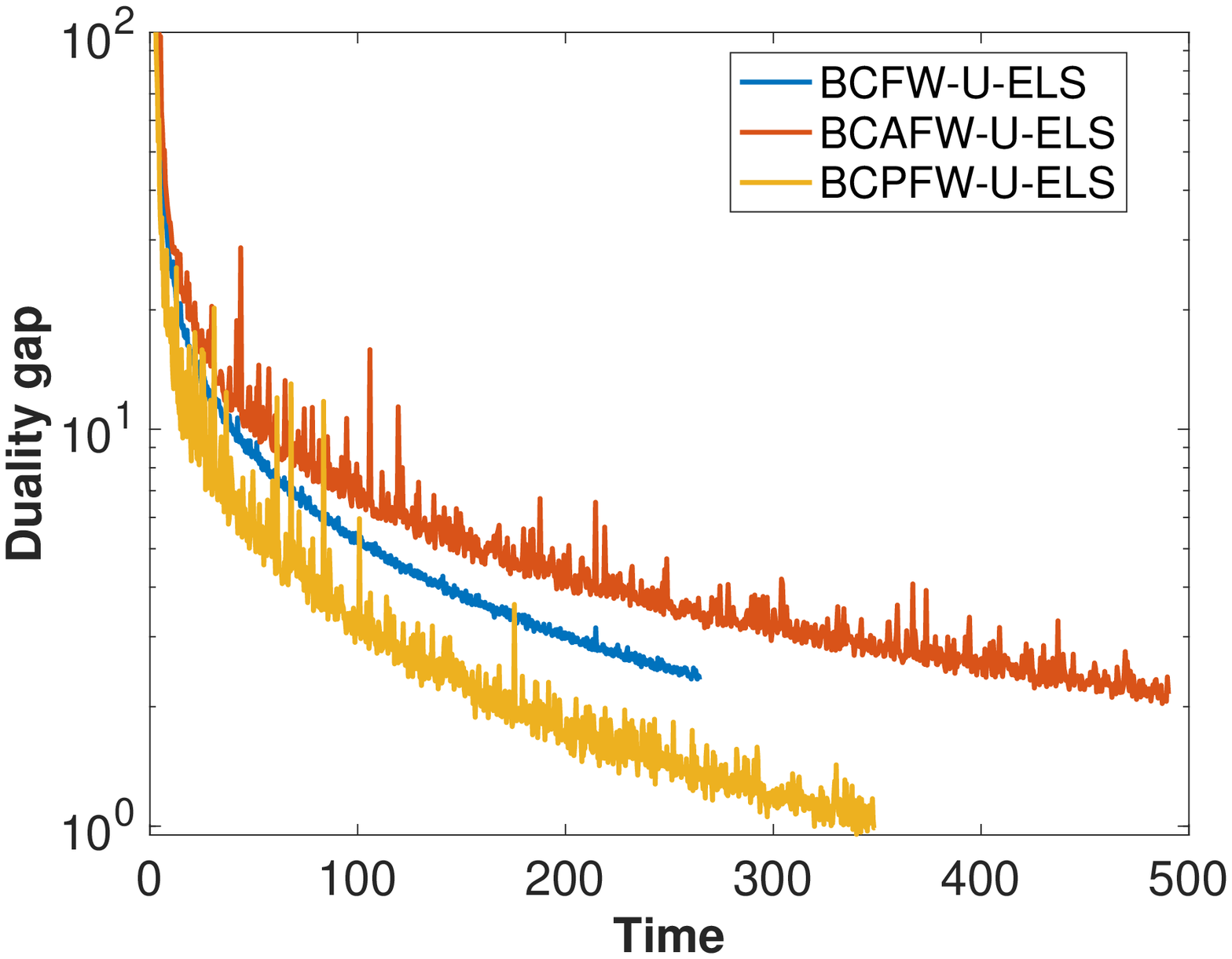}\\
		
		{\footnotesize (d) duality gap (time) : $g(\mat{T})$}
		
	\end{center} 
	\end{minipage}
	\hspace*{-0.2cm}	
	\begin{minipage}[t]{0.32\textwidth}
	\begin{center}
		\includegraphics[width=\textwidth]{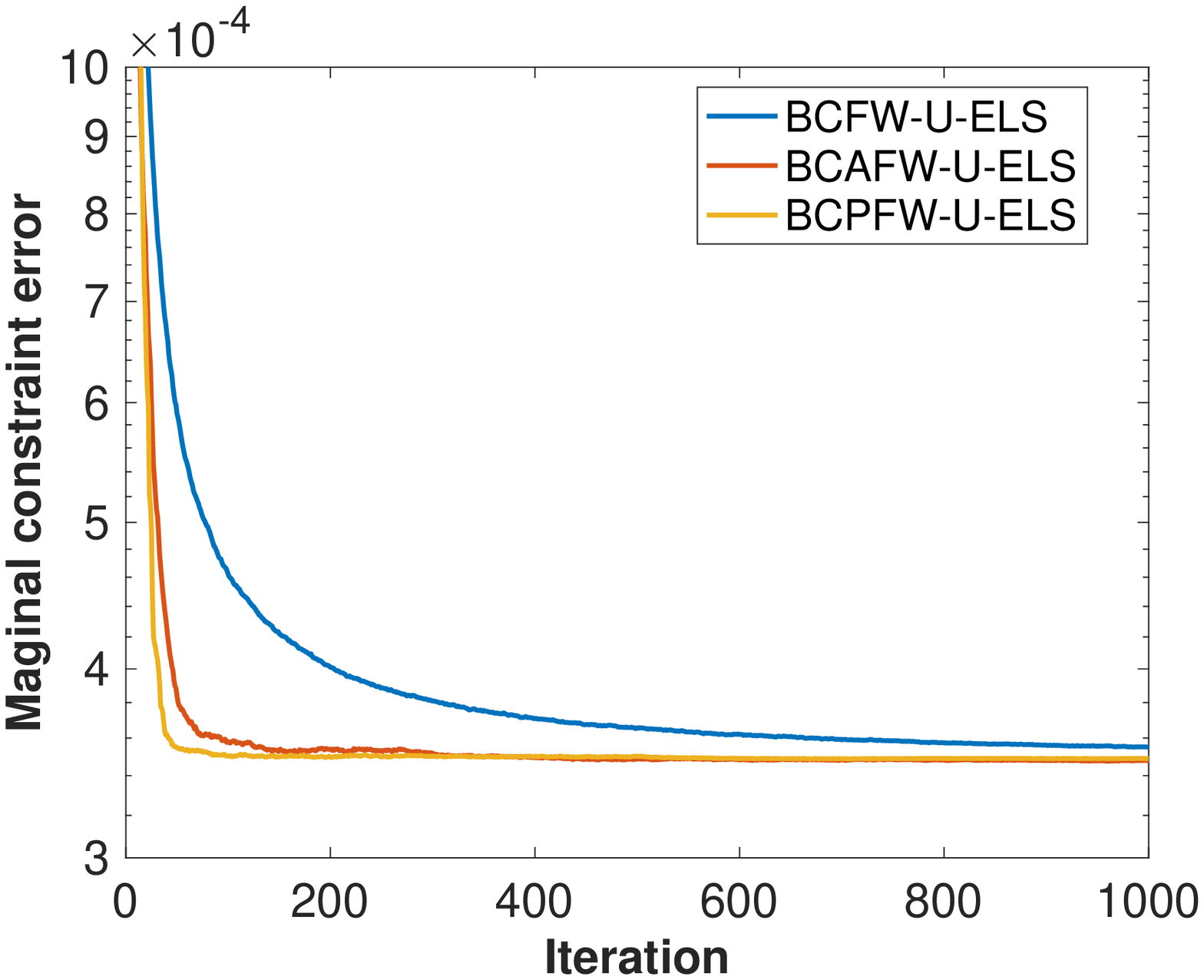}\\
		
		{\footnotesize (e) marginal constraint error : $e_{c}$}
		
	\end{center} 
	\end{minipage}
	\hspace*{-0.2cm}
	\begin{minipage}[t]{0.32\textwidth}
	\begin{center}
		\includegraphics[width=\textwidth]{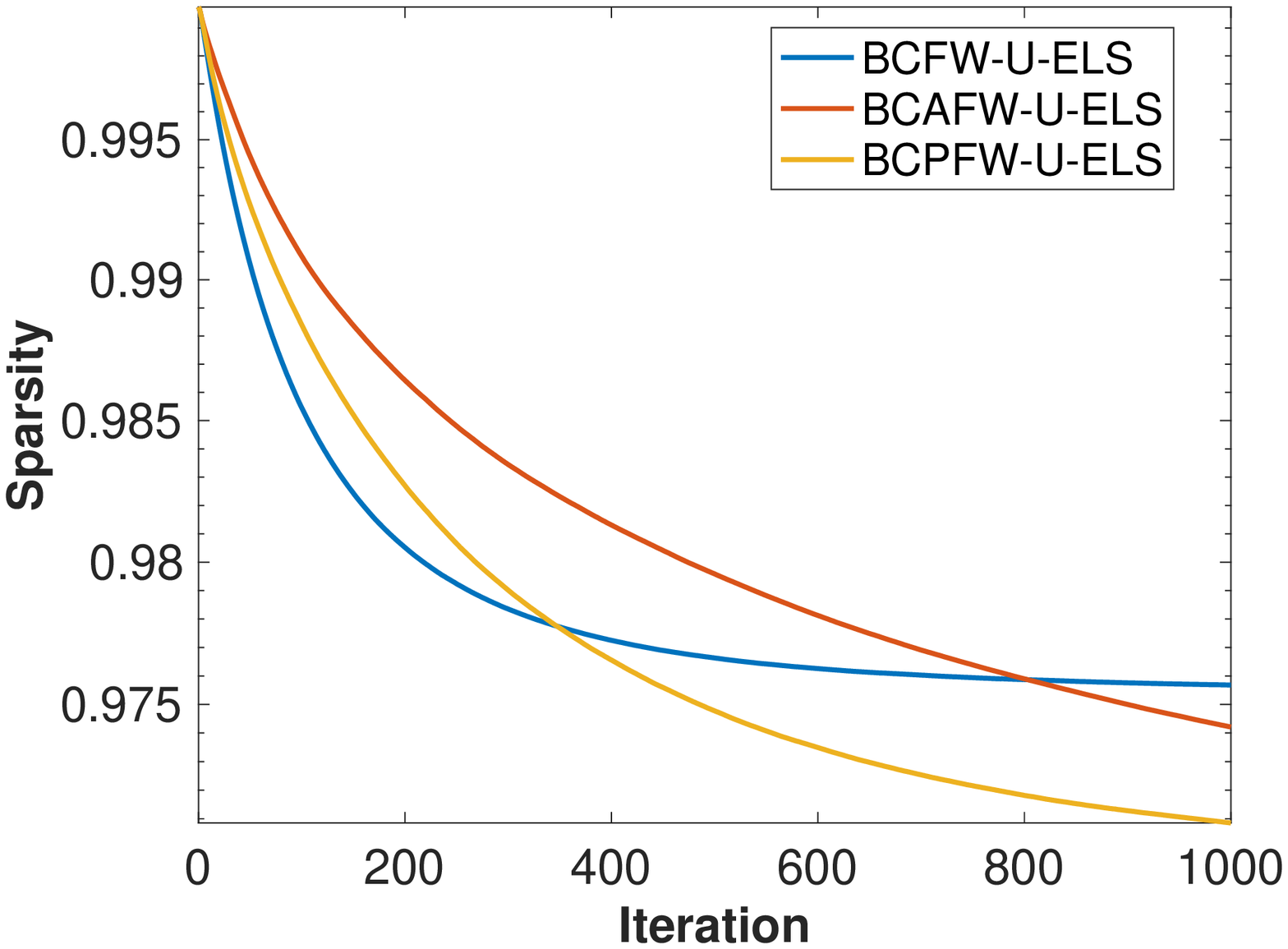}\\
		
		{\footnotesize (f) sparsity}
		
	\end{center} 
	\end{minipage}	
	\vspace*{0.2cm}

	\begin{minipage}[t]{0.32\textwidth}
	\begin{center}
		\includegraphics[width=\textwidth]{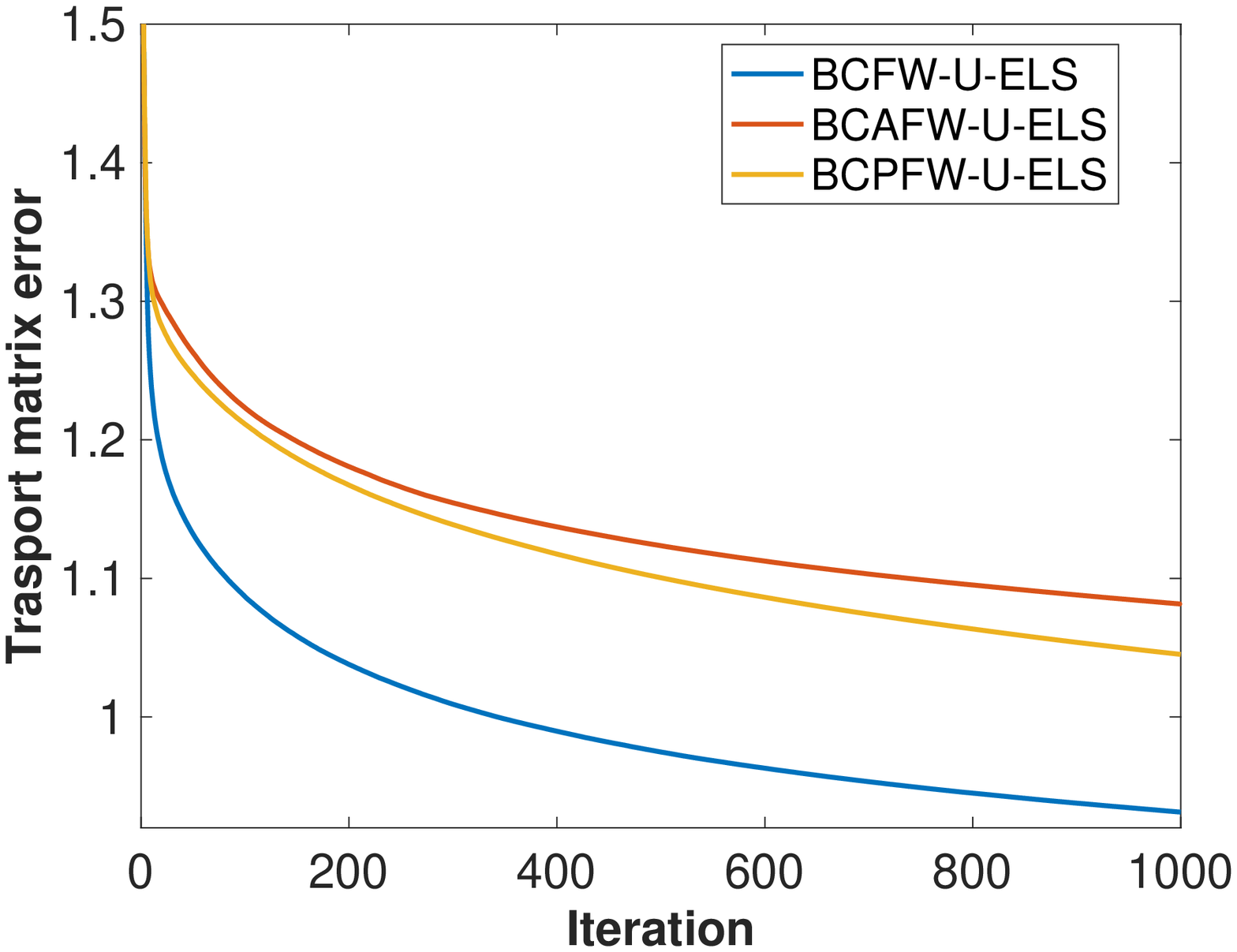}\\
		
		{\footnotesize (g) matrix error : $e_{m}$ }
		
	\end{center} 
	\end{minipage}
	\hspace*{-0.2cm}
	\begin{minipage}[t]{0.32\textwidth}
	\begin{center}
		\includegraphics[width=\textwidth]{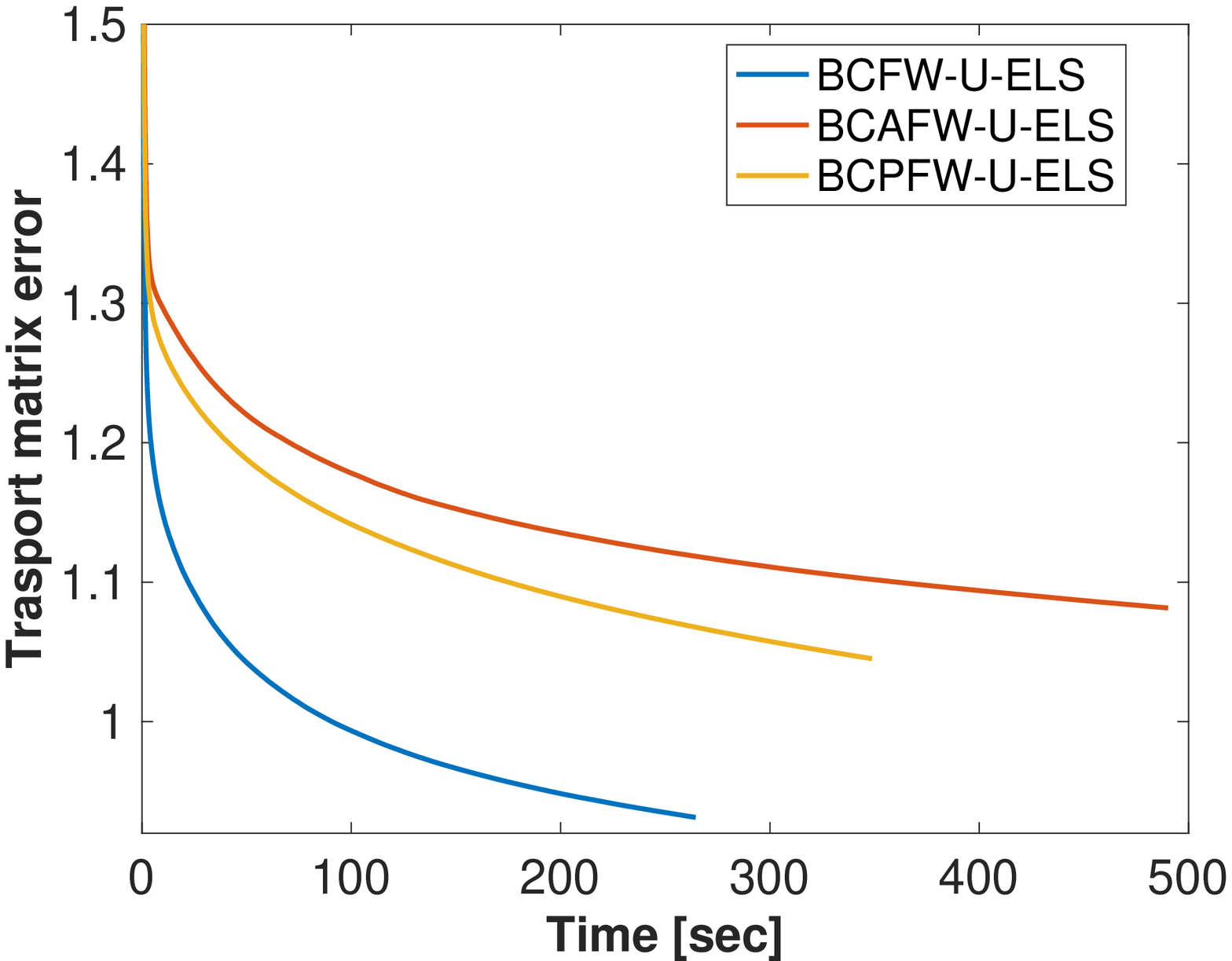}\\
		
		{\footnotesize (h) matrix error (time) : $e_{m}$ }
		
	\end{center} 
	\end{minipage}		
	\hspace*{-0.2cm}			
	\begin{minipage}[t]{0.32\textwidth}
	\begin{center}
		\includegraphics[width=\textwidth]{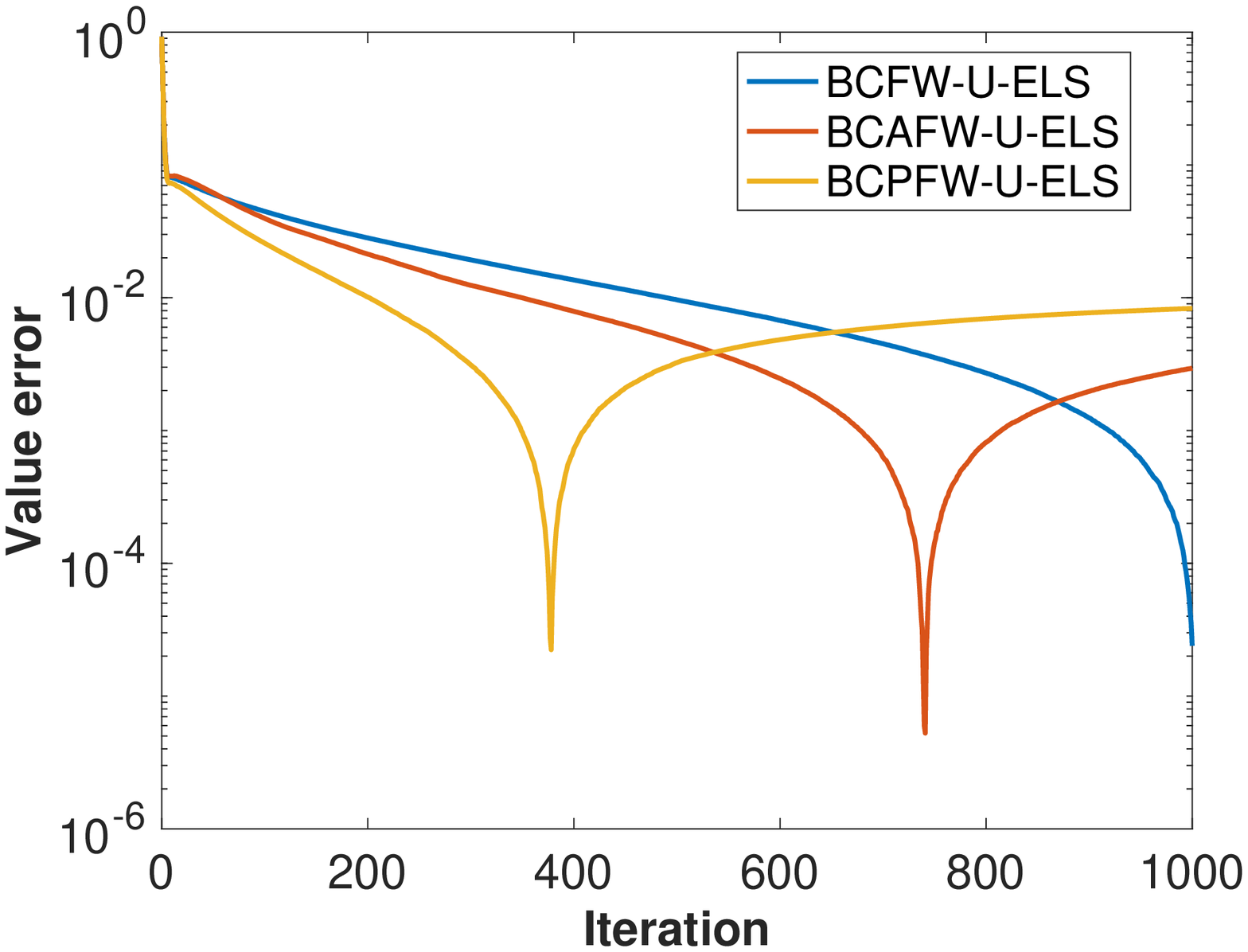}\\
		
		{\footnotesize (i) value error : $e_v$}
		
	\end{center} 
	\end{minipage}

\caption{Evaluations on convergence of fast variants of BCFW (BCAFW and BCPFW).}
\label{fig:FastVariantConvergencePerformance}
\end{center}
\end{figure}

This subsection evaluates the performance improvements by two fast variants discussed in Section \ref{Sec:FWandAW}, i.e., the BCFW with away-steps (BCAFW) and the BCFW with pairwise-steps (BCPFW).  
The experimental configurations are the same as in Section \ref{Sec:EvalBaselineBCFW}. Addressing the exact line-search stepsize rule  (ELS) and $\lambda=10^{-7}$, we discuss the convergence performances of the approximation errors. Figure \ref{fig:FastVariantConvergencePerformance} shows the results. The objective value and duality gap in Figures \ref{fig:FastVariantConvergencePerformance}(a)-(d) reveal that both the two fast variants BCAFW and BCPFW yield faster convergence in terms of iteration. BCPFW gives much better performances in terms of computational time, as well. As for the matrix error $e_m$ and the value error $e_v$, both the two faster variants are worse than the baseline BCFW. However, these two metrics are evaluated in terms of the solution $\mat{T}_{\rm LP}$, i.e., the solution of the non-relaxed linear programming problem. Thus, this degradation is caused by the difference of the two objective functions. 

\subsubsection{Evaluations on BCFW-GA}

\begin{figure}[htbp]
\begin{center}
	\begin{minipage}[t]{0.24\textwidth}
	\begin{center}
		
		\includegraphics[width=\textwidth]{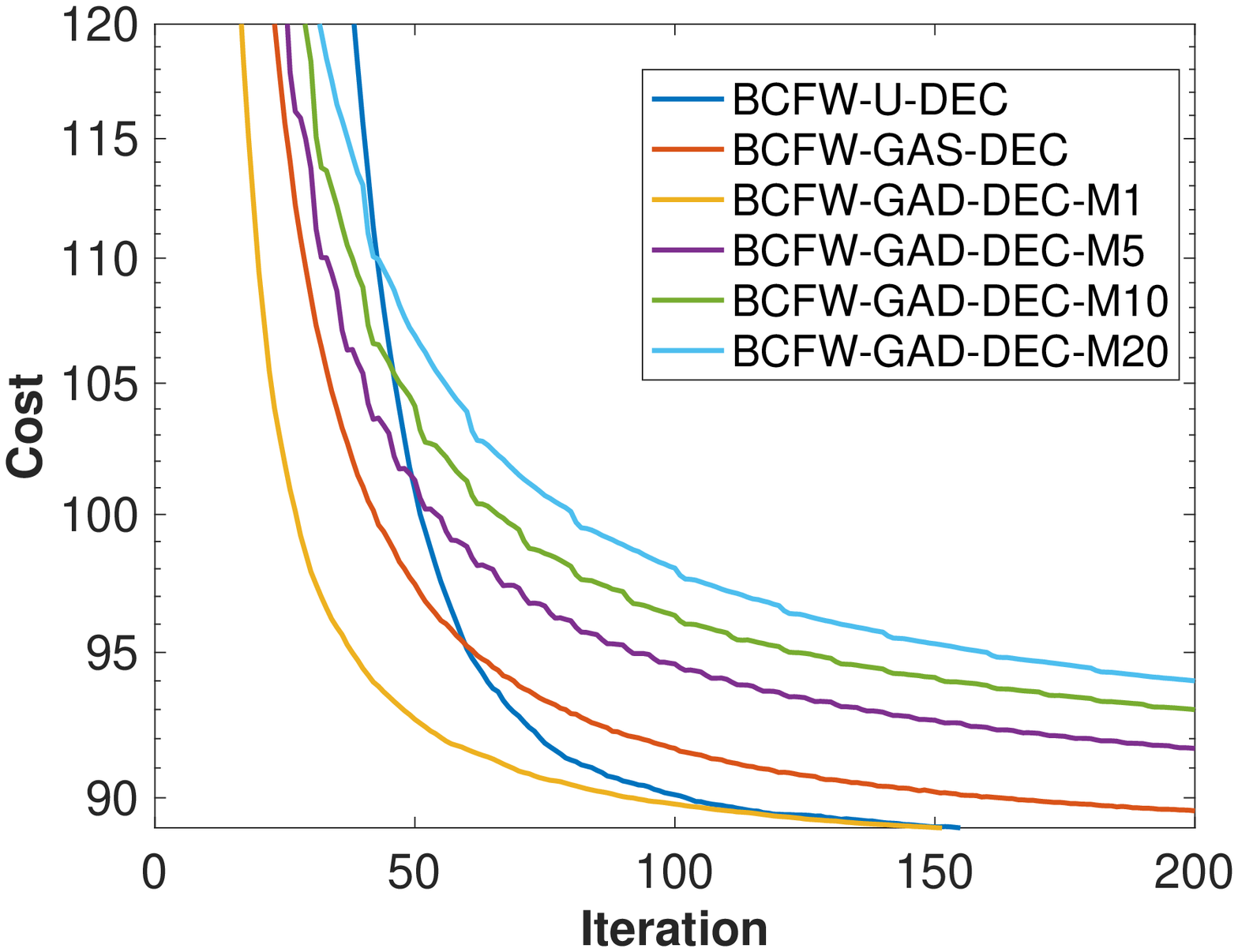}\\
		
		{\footnotesize (i) objective value : $f(\mat{T})$ }
		
	\end{center} 
	\end{minipage}
	\begin{minipage}[t]{0.236\textwidth}
	\begin{center}
		
		\includegraphics[width=\textwidth]{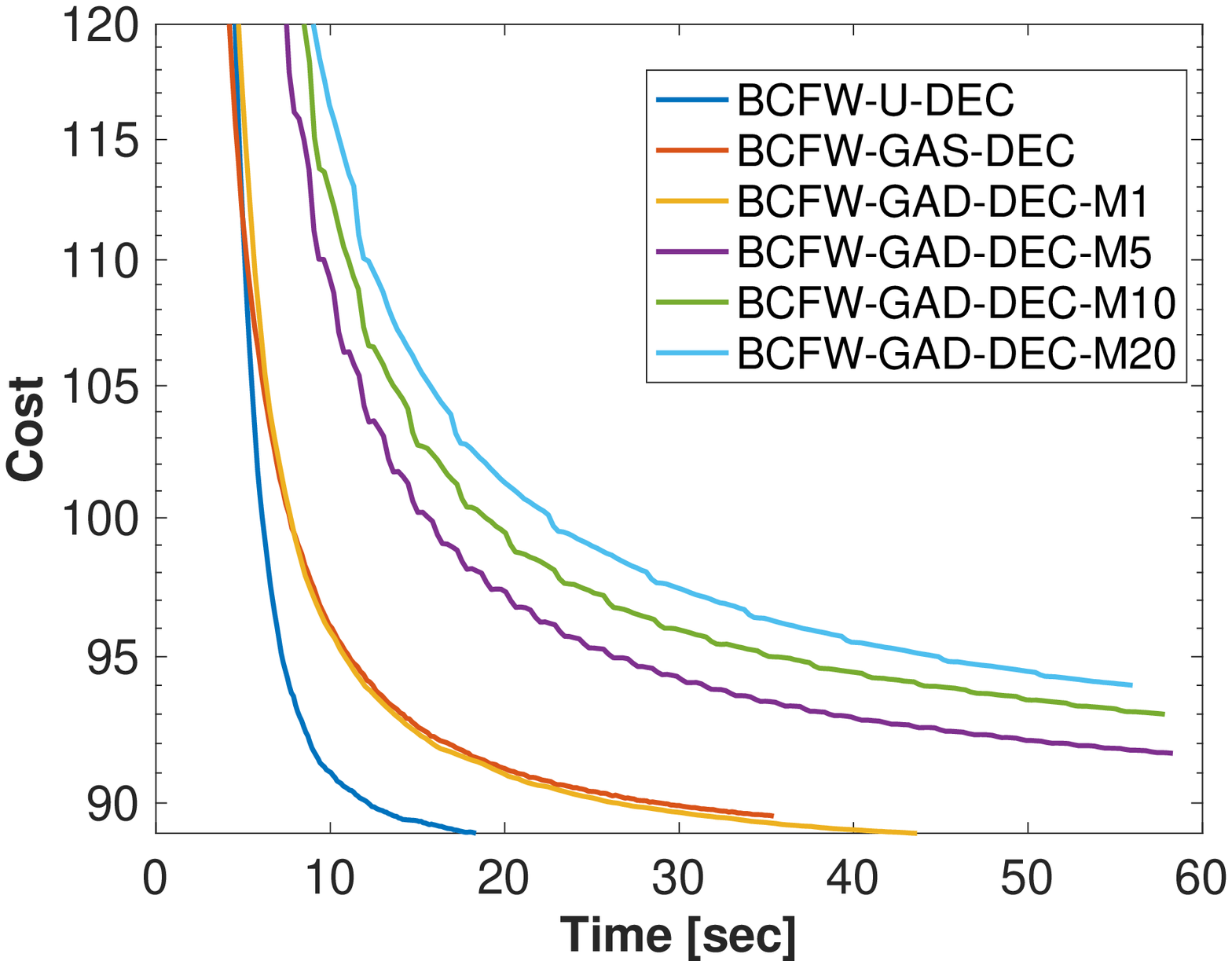}\\
		
		{\footnotesize  (ii) objective value (time)}
		
	\end{center} 
	\end{minipage}	
	\begin{minipage}[t]{0.24\textwidth}
	\begin{center}
		\includegraphics[width=\textwidth]{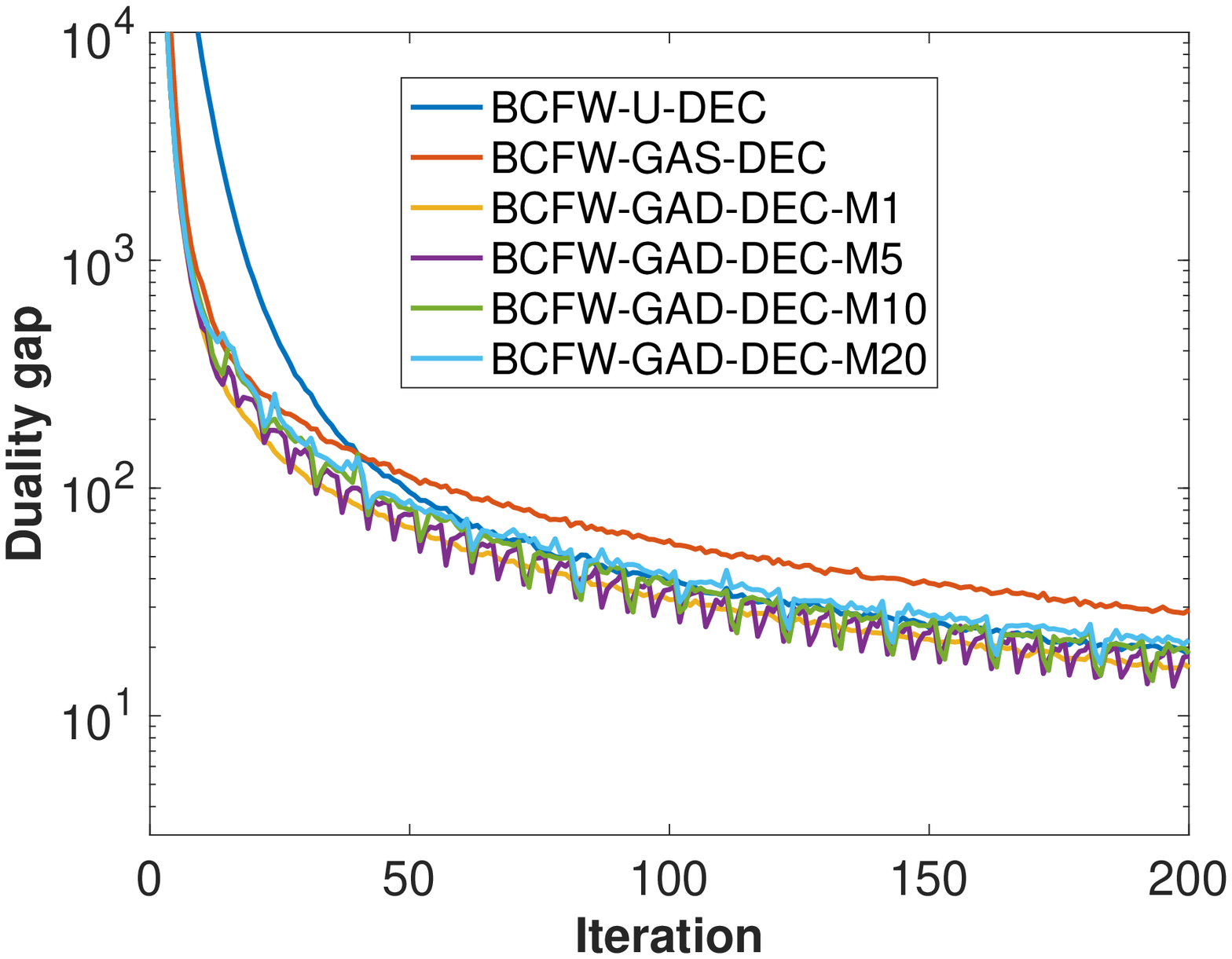}\\
		
		{\footnotesize  (iii) duality gap: $g(\mat{T})$}
		
	\end{center} 
	\end{minipage}
	\hspace*{0.1cm}
	\begin{minipage}[t]{0.23\textwidth}
	\begin{center}
		\includegraphics[width=\textwidth]{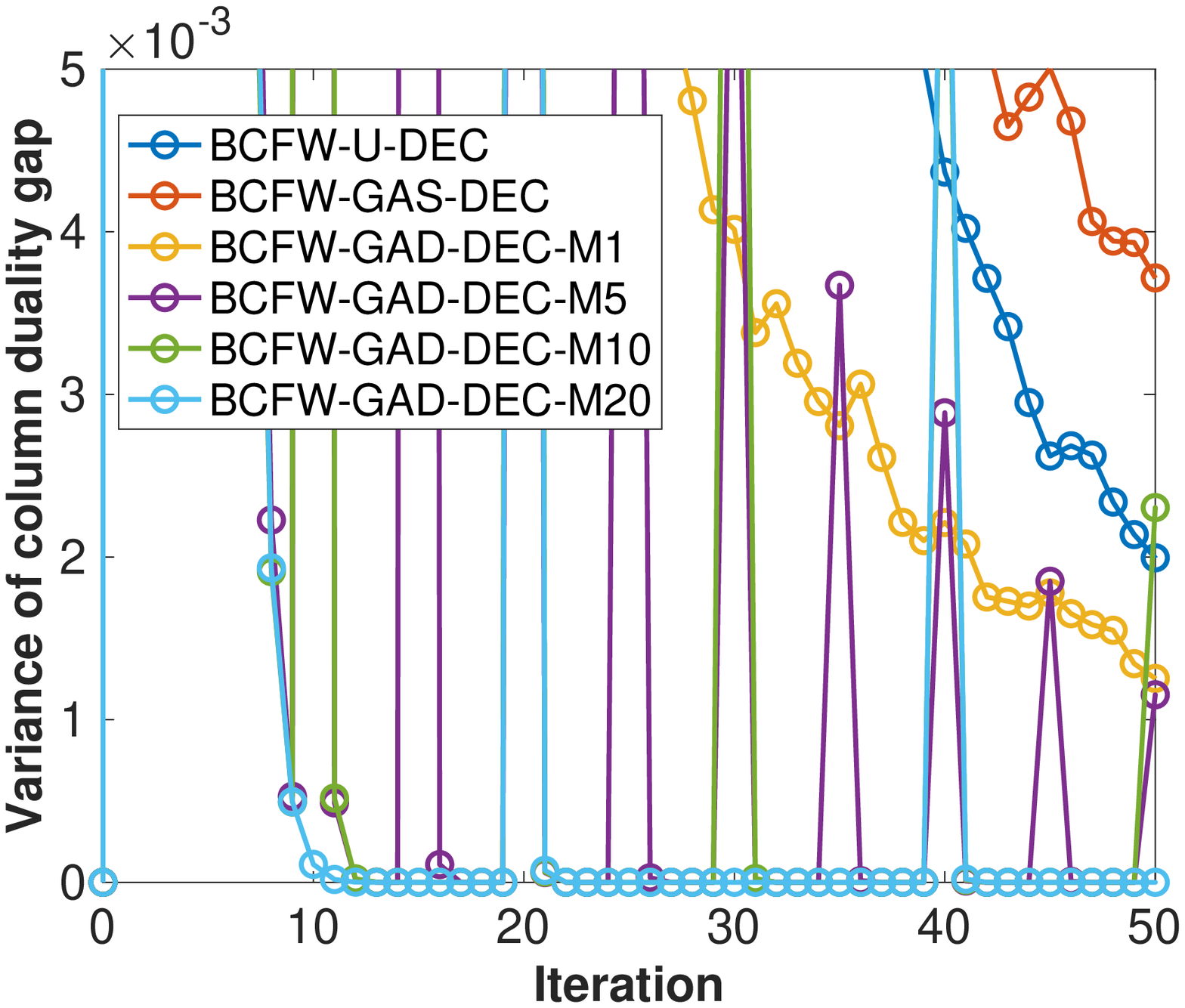}\\
		
		{\footnotesize  (iv) variance of $g_i(\mat{T})$}
		
	\end{center} 
	\end{minipage}
	\vspace*{0.2cm}	
	
	{\small (a) BCFW-U-DEC and BCFW-GA-DEC}
	\vspace*{0.6cm}
	
	\begin{minipage}[t]{0.24\textwidth}
	\begin{center}
		
		\includegraphics[width=\textwidth]{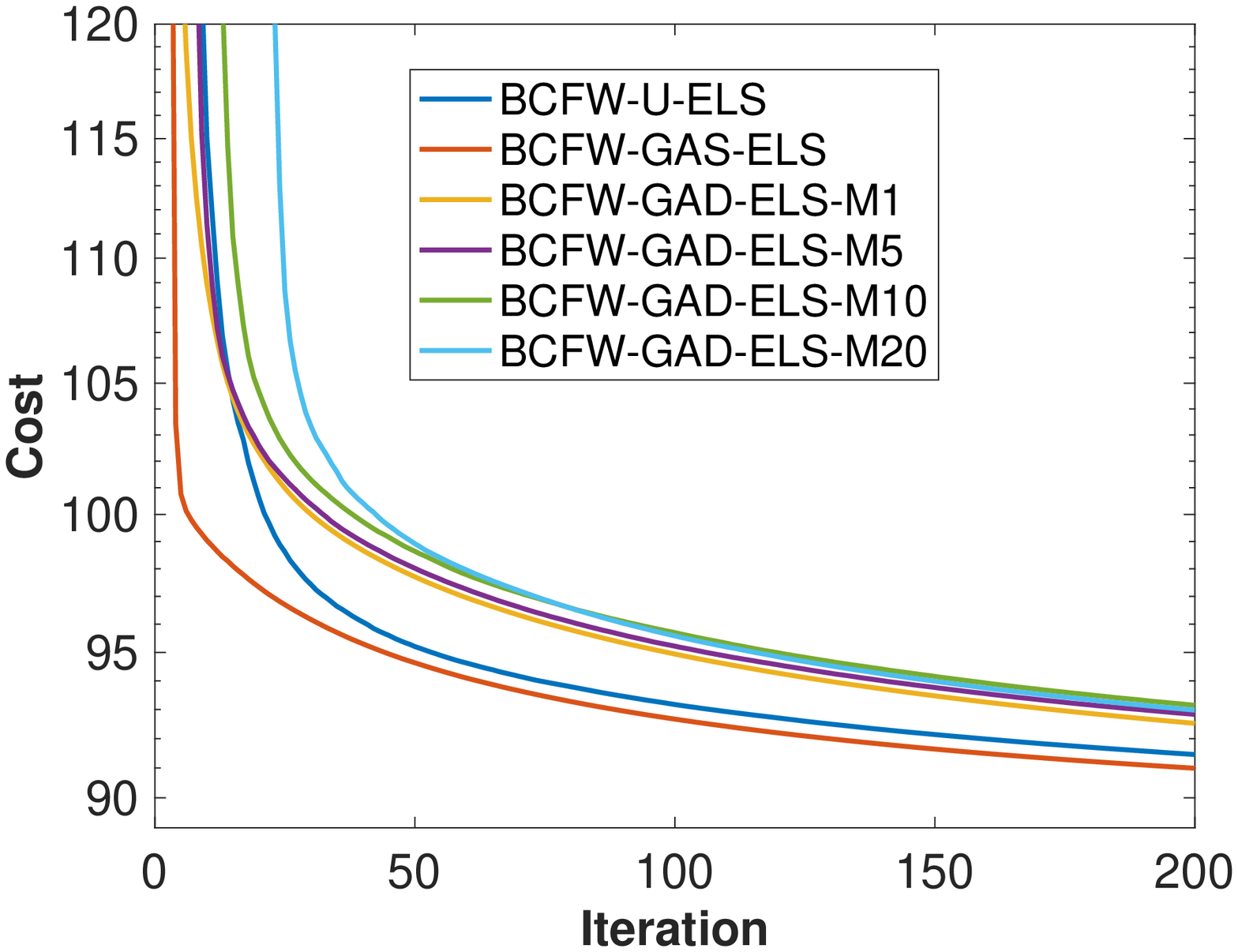}\\
		
		{\footnotesize  (i) objective value : $f(\mat{T})$ }
		
	\end{center} 
	\end{minipage}
	\begin{minipage}[t]{0.24\textwidth}
	\begin{center}
		
		\includegraphics[width=\textwidth]{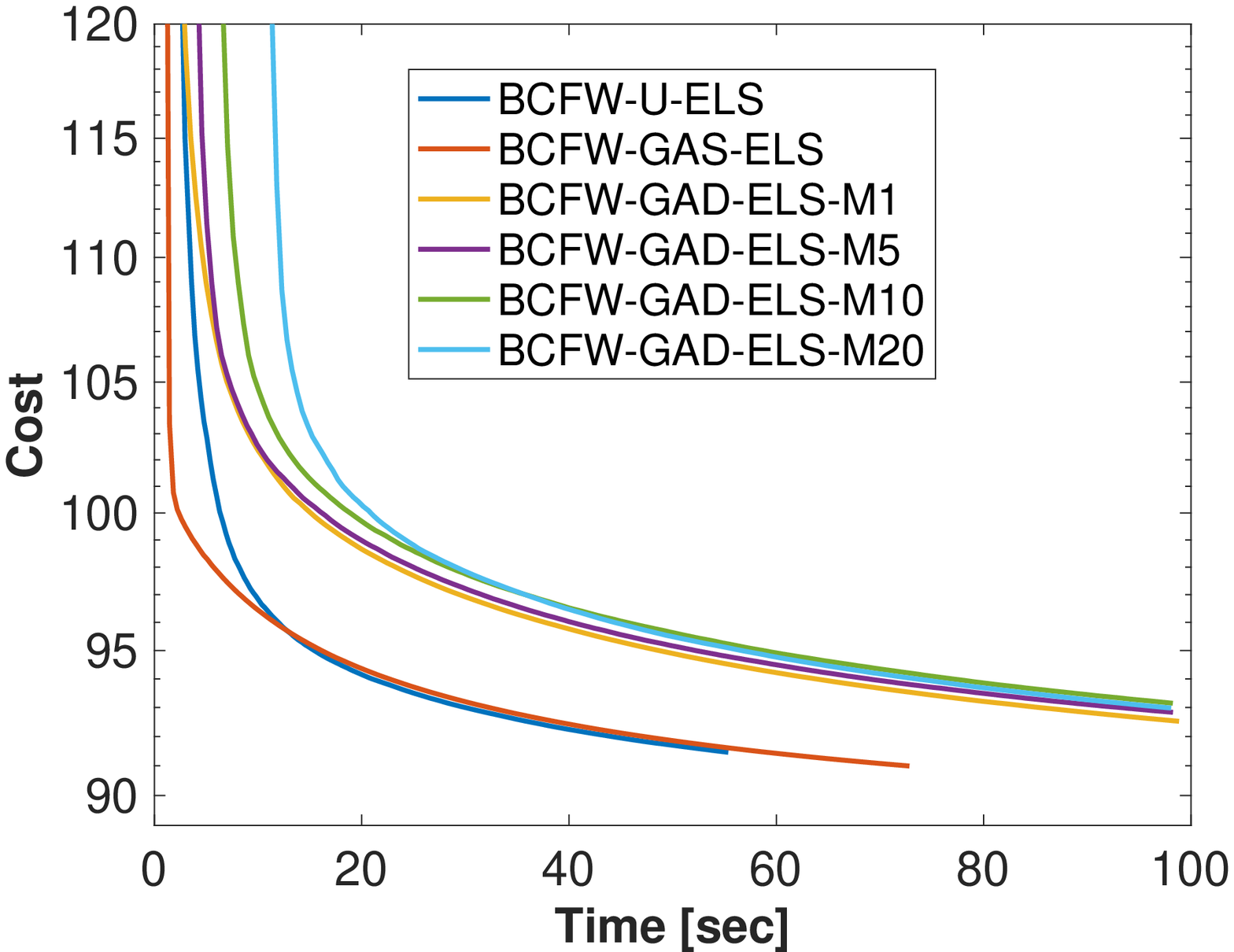}\\
		
		{\footnotesize  (ii) objective value (time)}
		
	\end{center} 
	\end{minipage}	
	\begin{minipage}[t]{0.24\textwidth}
	\begin{center}
		\includegraphics[width=\textwidth]{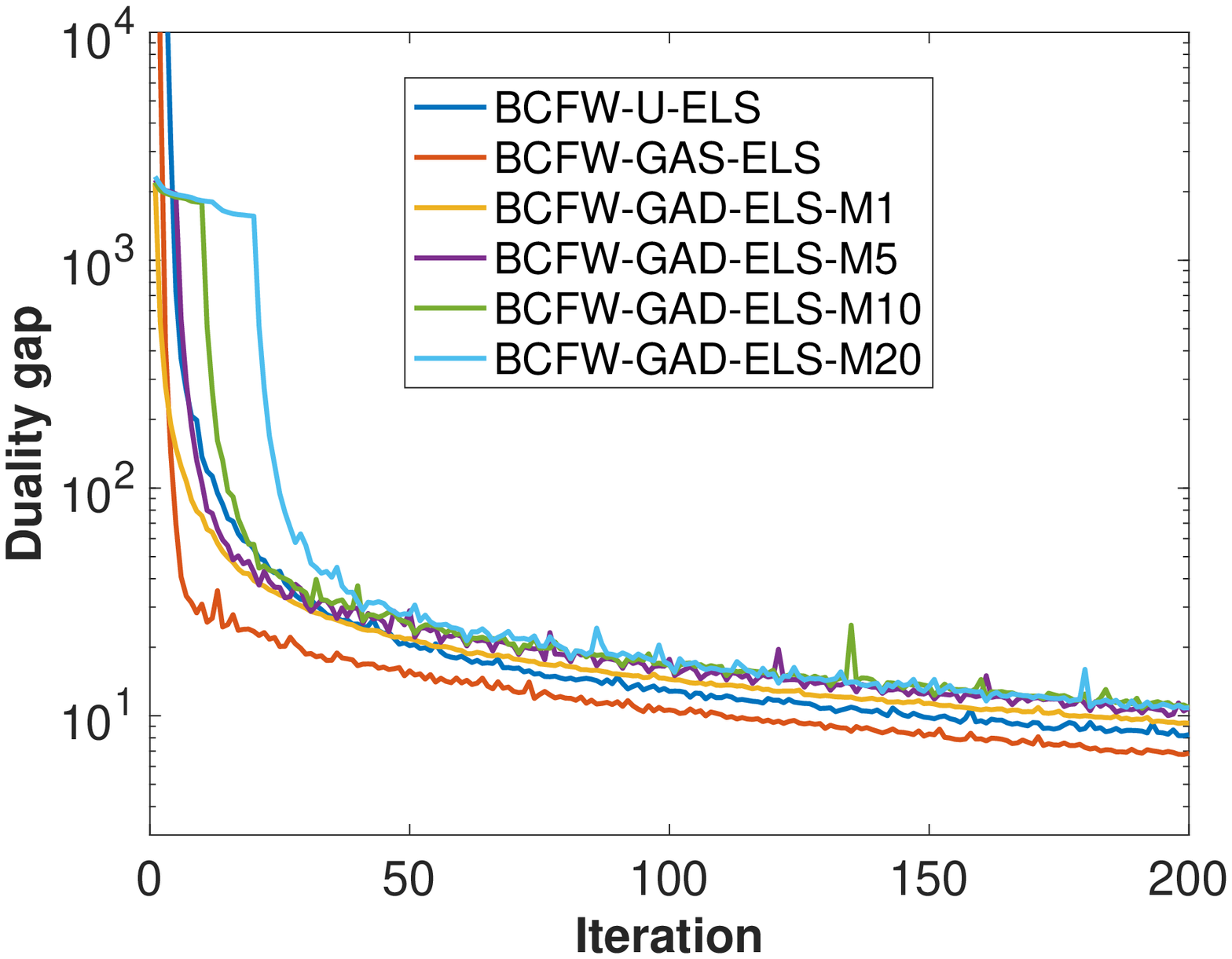}\\
		
		{\footnotesize  (iii) duality gap: $g(\mat{T})$}
		
	\end{center} 
	\end{minipage}
	\hspace*{0.1cm}
	\begin{minipage}[t]{0.23\textwidth}
	\begin{center}
		\includegraphics[width=\textwidth]{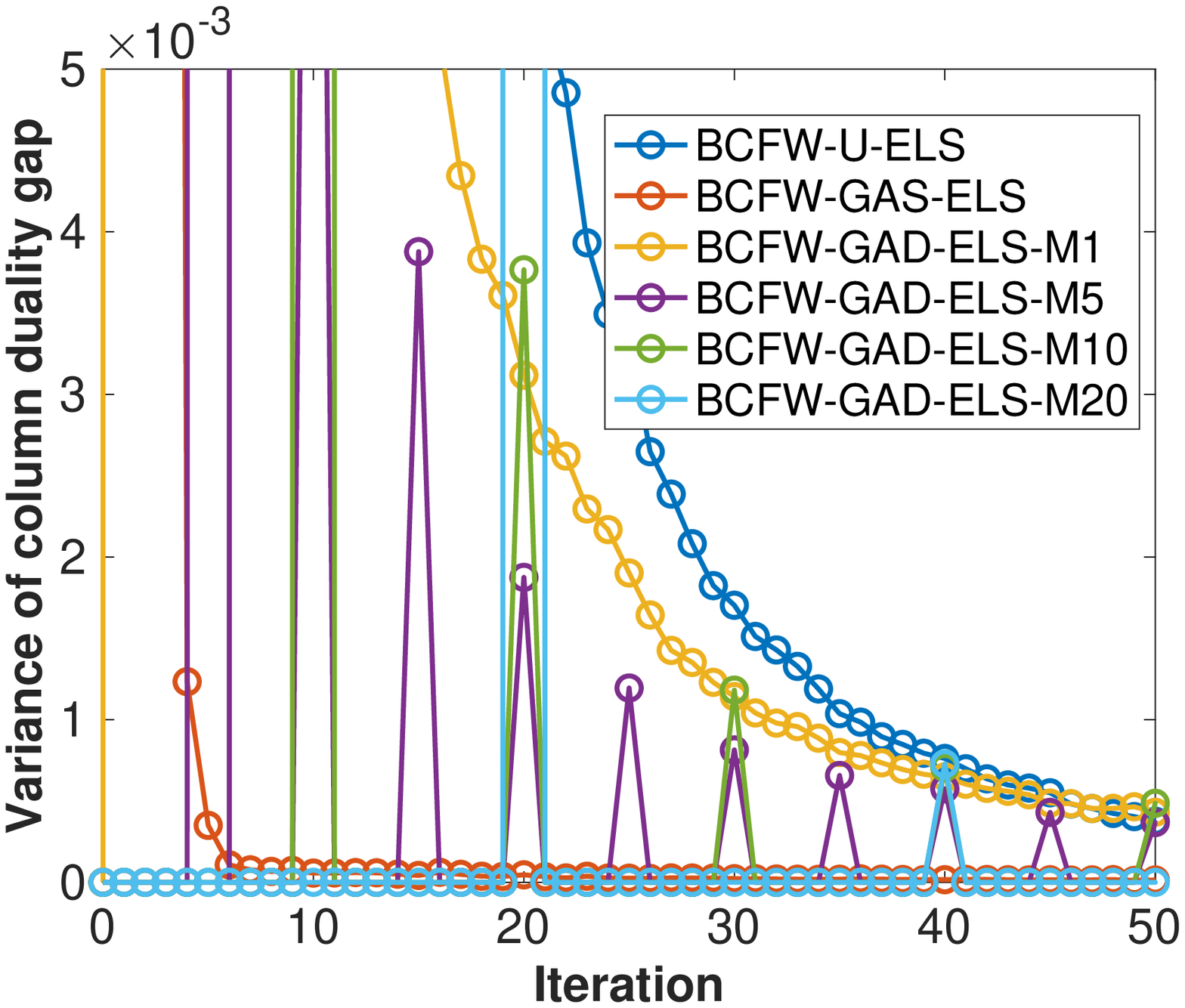}\\
		
		{\footnotesize  (iv) variances of $g_i(\mat{T})$}
		
	\end{center} 
	\end{minipage}	
	\vspace*{0.2cm}	
	
	{\small (b) BCFW-U-ELS and BCFW-GA-ELS}
	\vspace*{0.6cm}
	
	\begin{minipage}[t]{0.24\textwidth}
	\begin{center}
		
		\includegraphics[width=\textwidth]{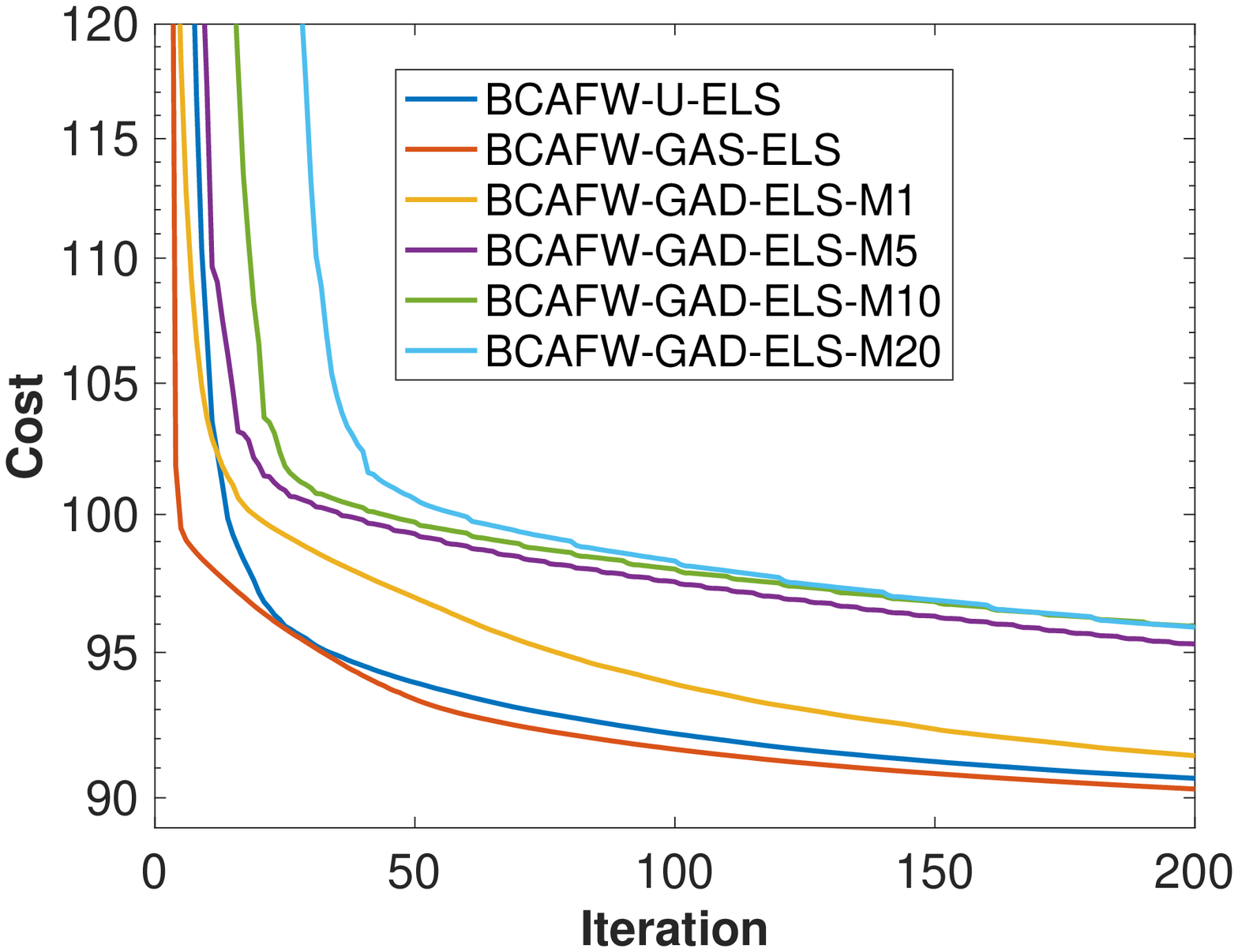}\\
		
		{\footnotesize  (i) objective value : $f(\mat{T})$ }
		
	\end{center} 
	\end{minipage}
	\begin{minipage}[t]{0.24\textwidth}
	\begin{center}
		
		\includegraphics[width=\textwidth]{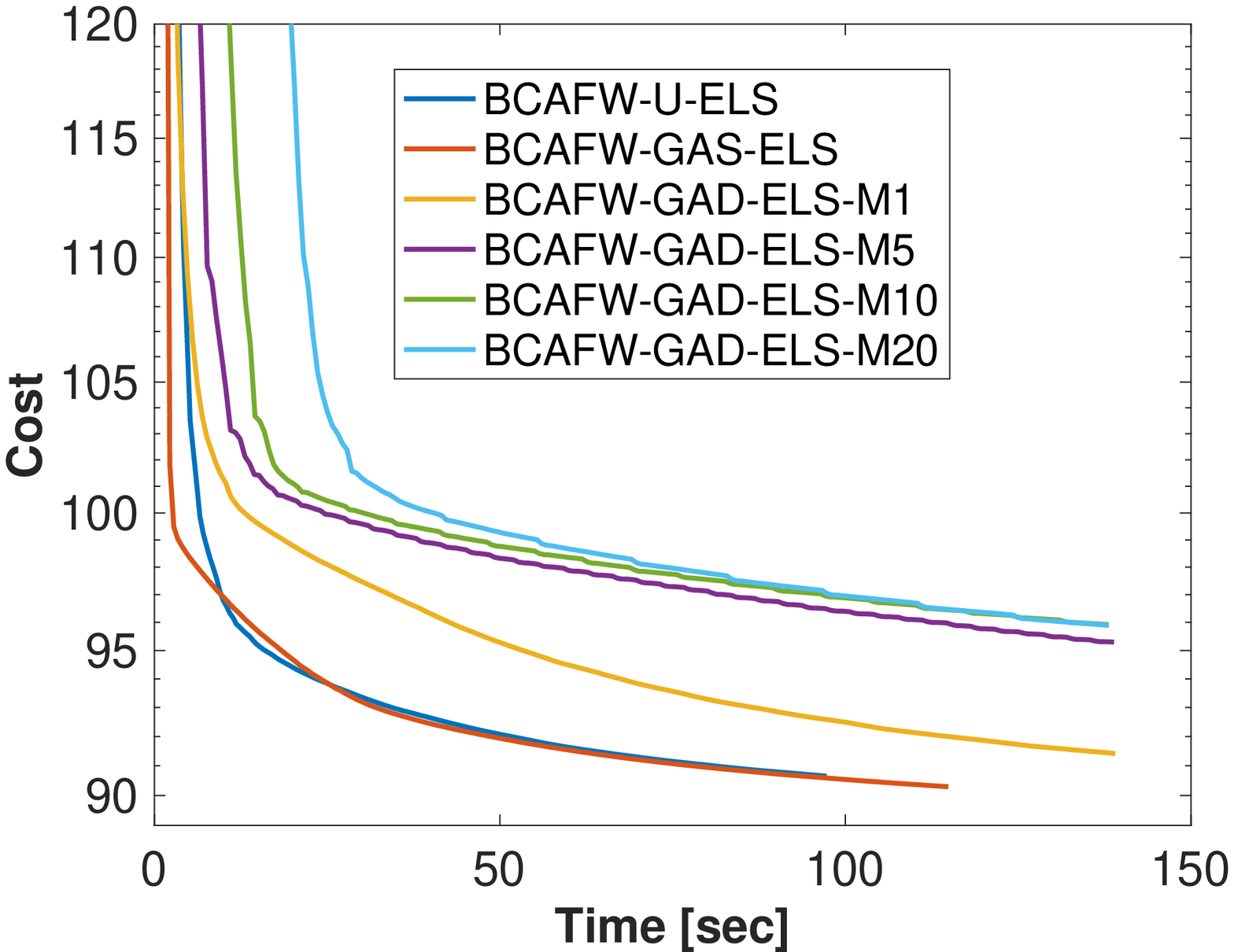}\\
		
		{\footnotesize  (ii) objective value (time)}
		
	\end{center} 
	\end{minipage}	
	\begin{minipage}[t]{0.24\textwidth}
	\begin{center}
		\includegraphics[width=\textwidth]{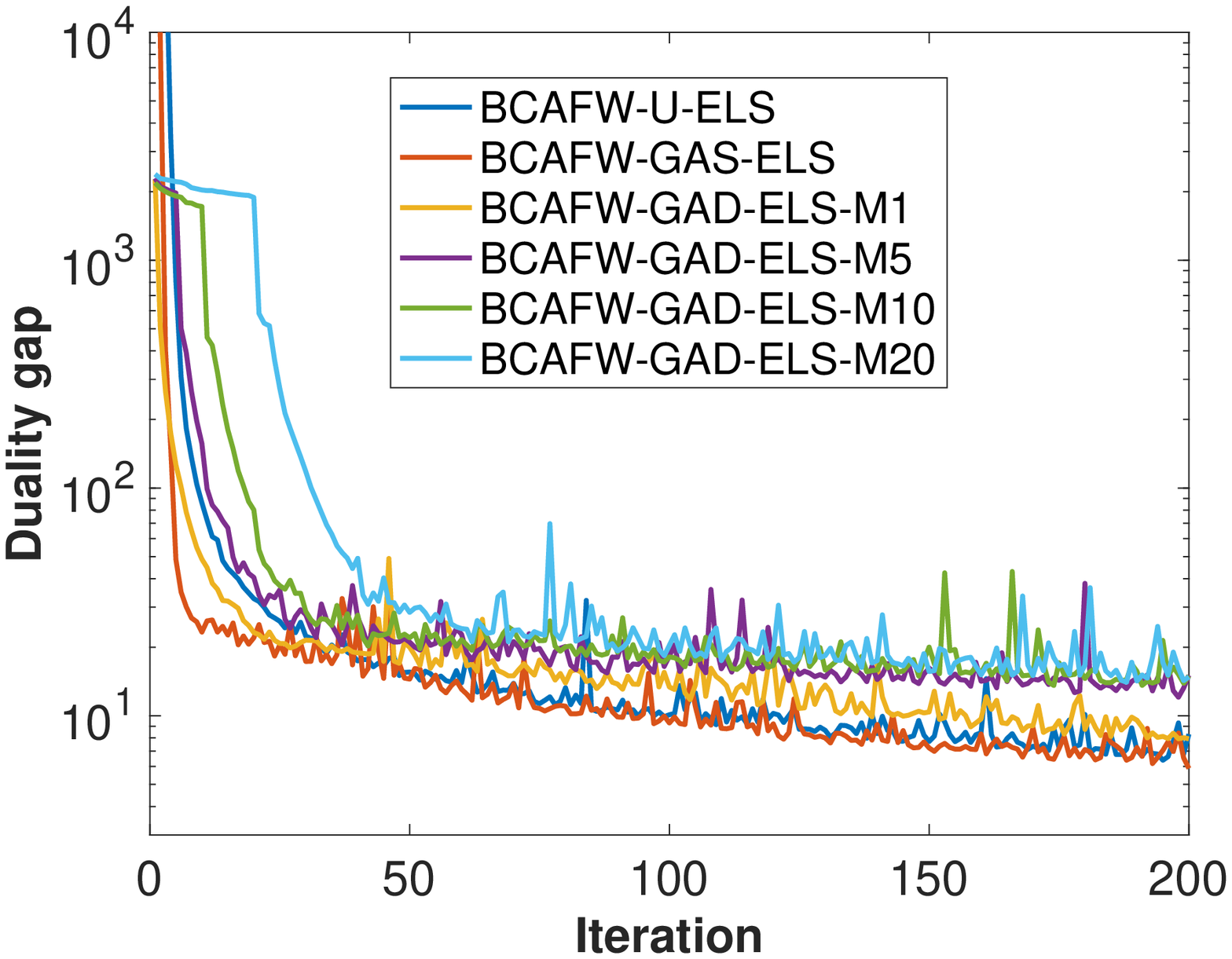}\\
		
		{\footnotesize  (iii) duality gap: $g(\mat{T})$}
		
	\end{center} 
	\end{minipage}
	\hspace*{0.1cm}
	\begin{minipage}[t]{0.23\textwidth}
	\begin{center}
		\includegraphics[width=\textwidth]{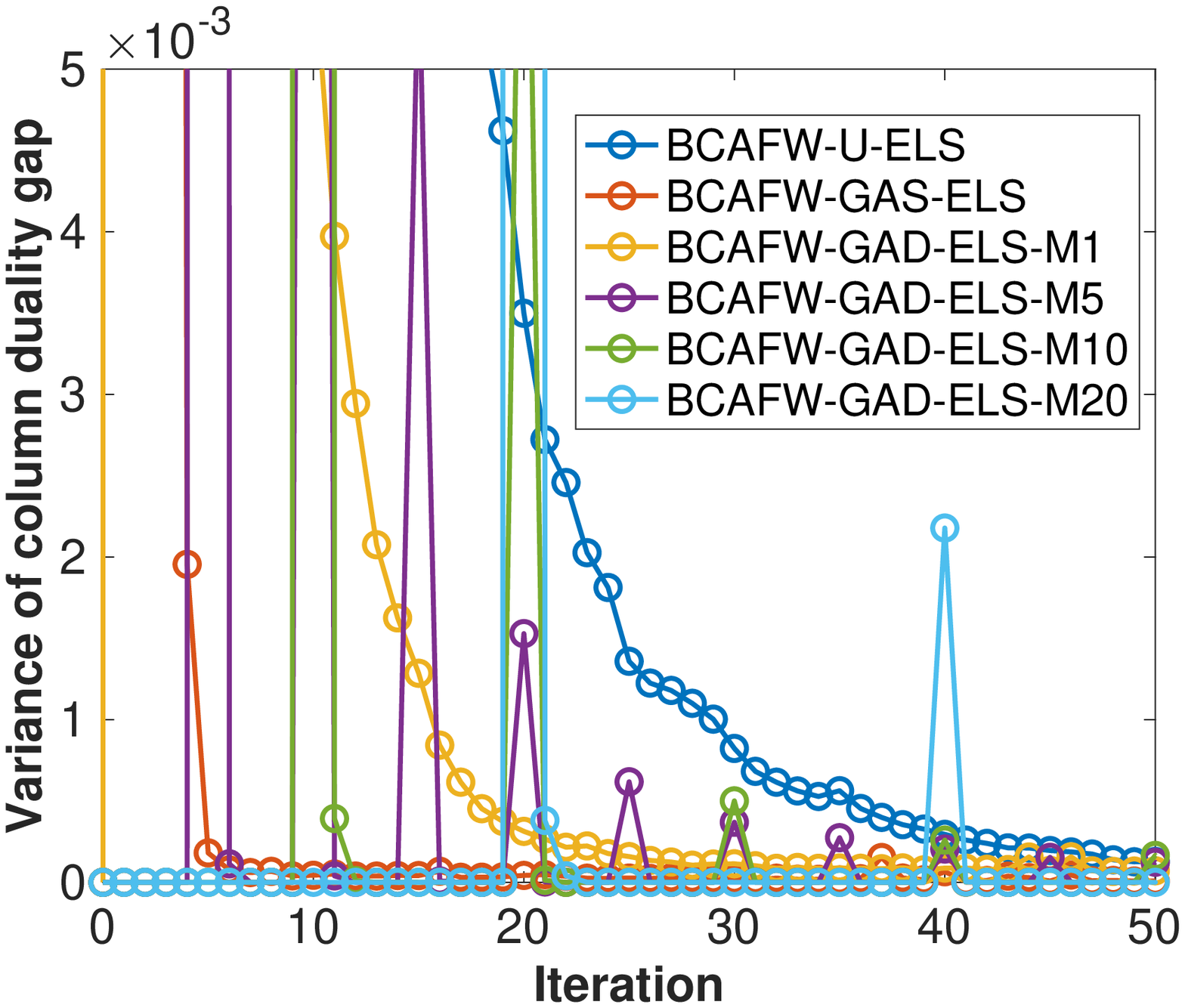}\\
		
		{\footnotesize  (iv) variances of $g_i(\mat{T})$}
		
	\end{center} 
	\end{minipage}
	\vspace*{0.2cm}		
	
	{\small (c) BCAFW-U-ELS and BCAFW-GA-ELS}
	\vspace*{0.6cm}		
	
	\begin{minipage}[t]{0.24\textwidth}
	\begin{center}
		
		\includegraphics[width=\textwidth]{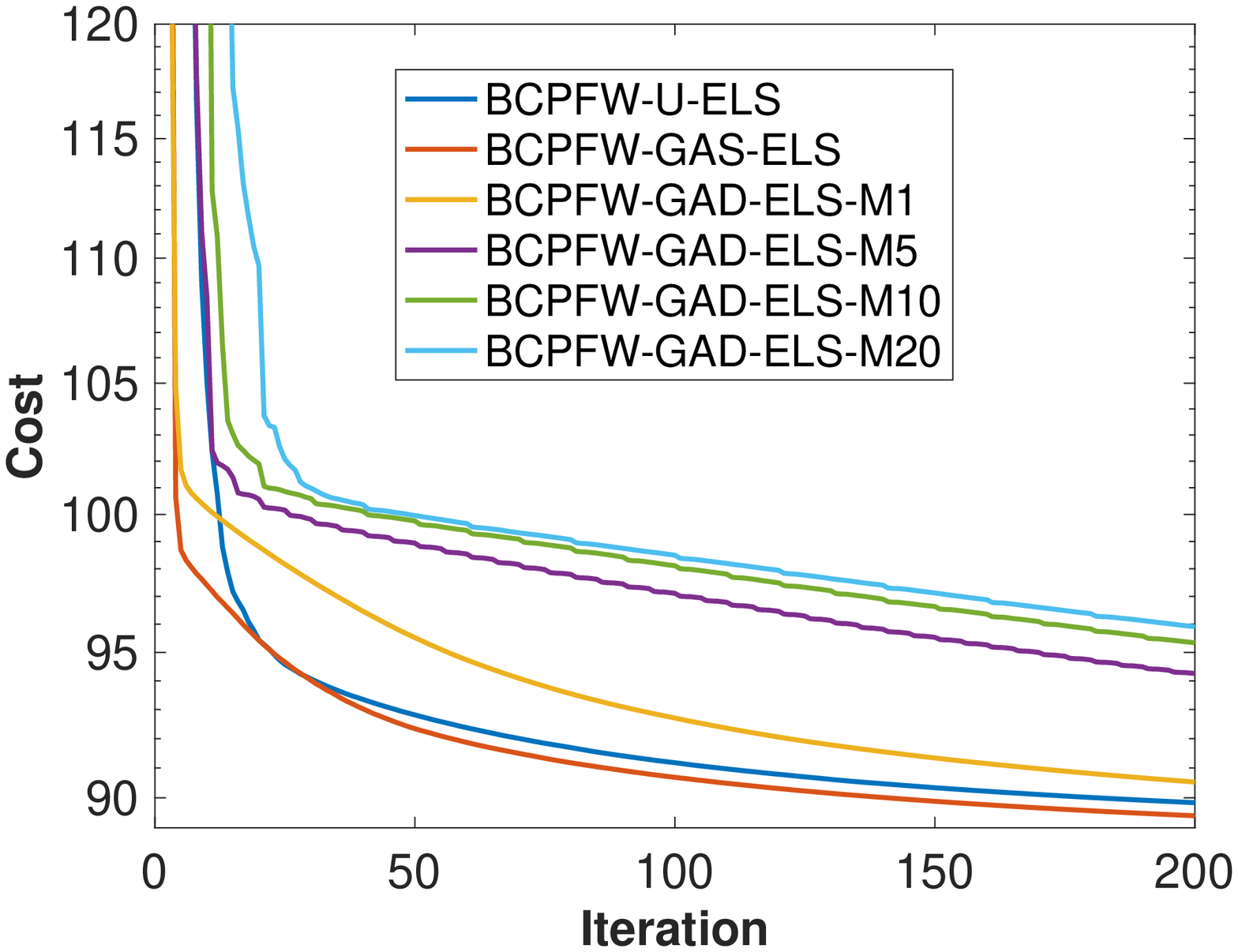}\\
		
		{\footnotesize  (i) objective value : $f(\mat{T})$ }
		
	\end{center} 
	\end{minipage}
	\begin{minipage}[t]{0.24\textwidth}
	\begin{center}
		
		\includegraphics[width=\textwidth]{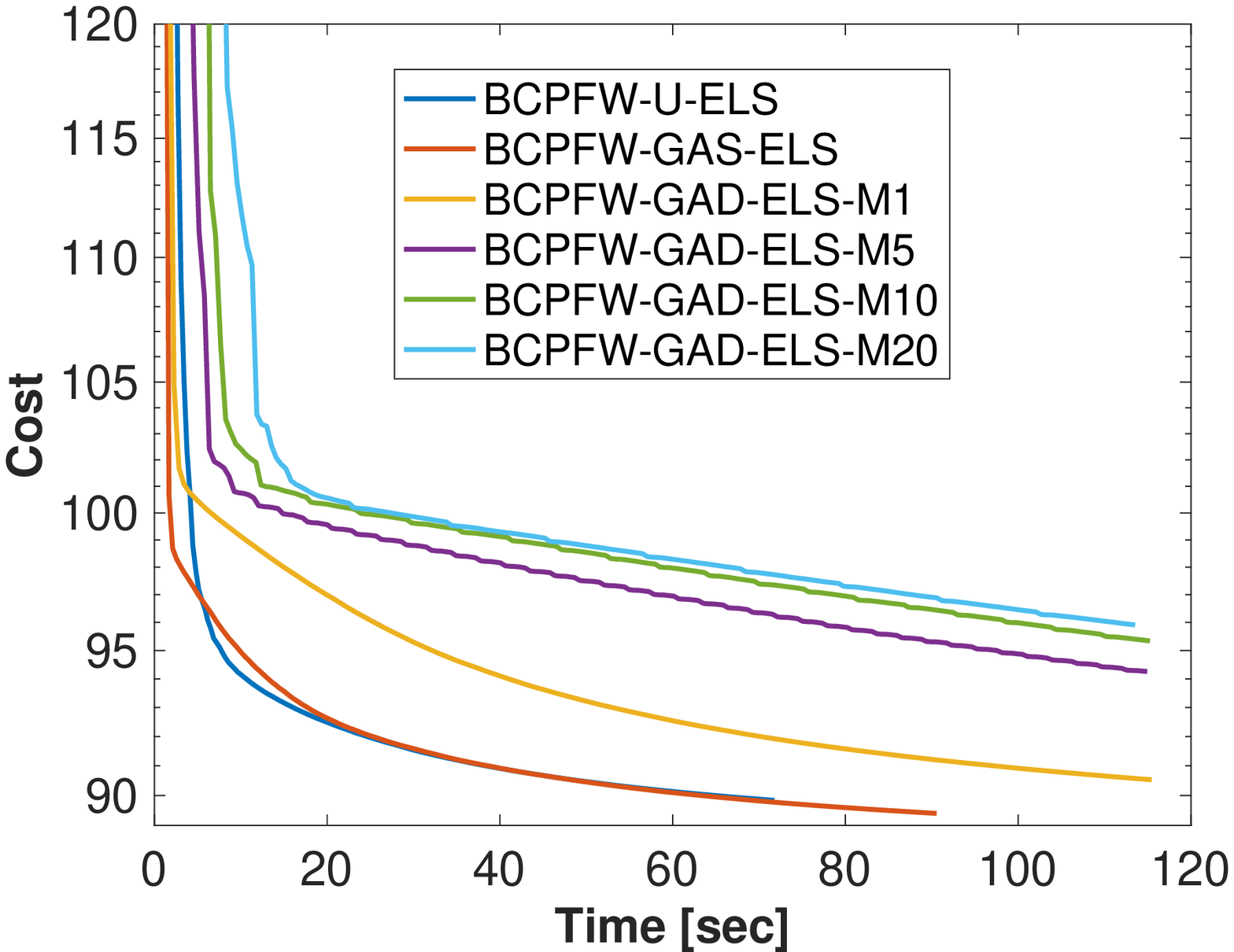}\\
		
		{\footnotesize  (ii) objective value (time)}
		
	\end{center} 
	\end{minipage}	
	\begin{minipage}[t]{0.24\textwidth}
	\begin{center}
		\includegraphics[width=\textwidth]{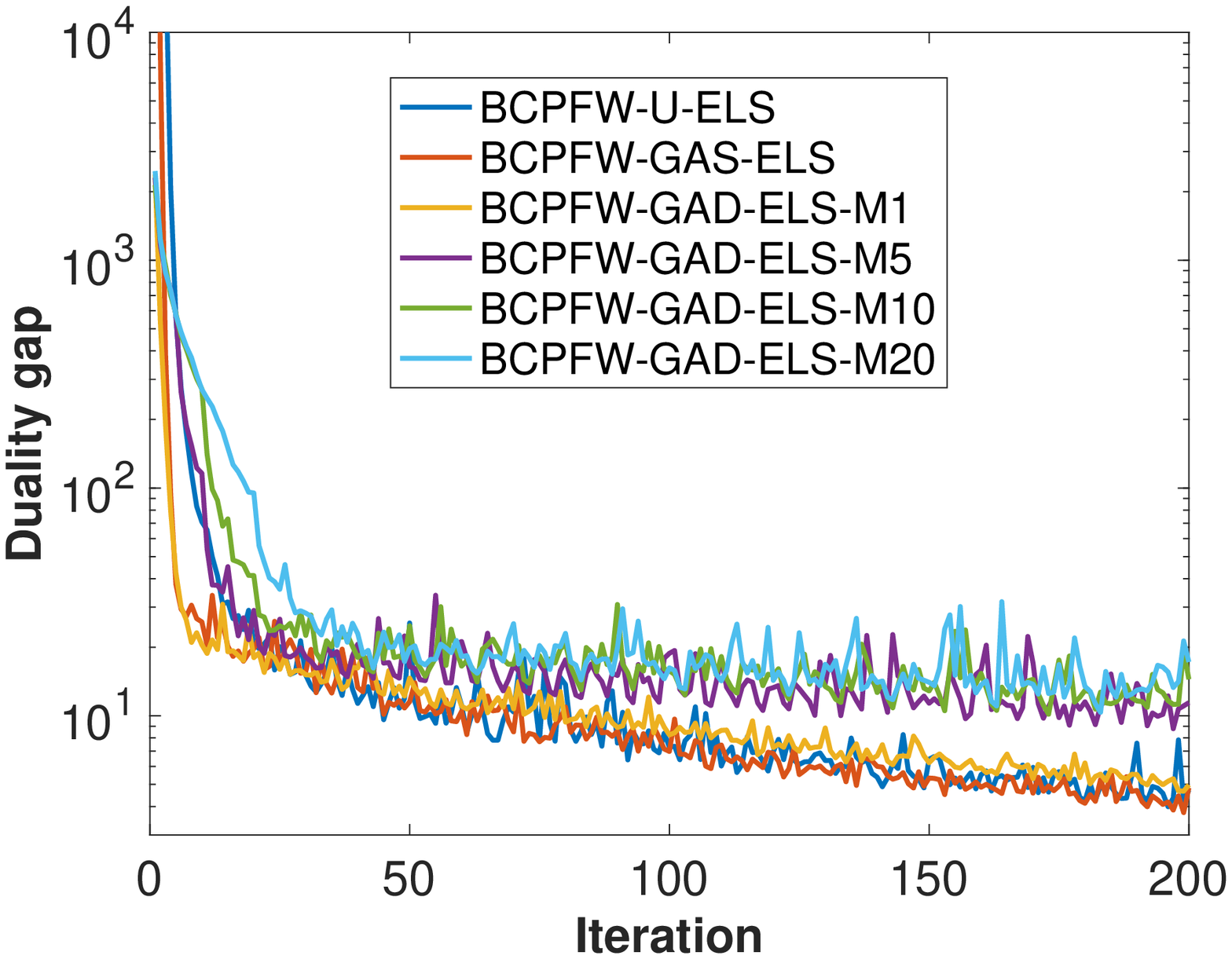}\\
		
		{\footnotesize  (iii) duality gap: $g(\mat{T})$}
		
	\end{center} 
	\end{minipage}
	\hspace*{0.1cm}
	\begin{minipage}[t]{0.23\textwidth}
	\begin{center}
		\includegraphics[width=\textwidth]{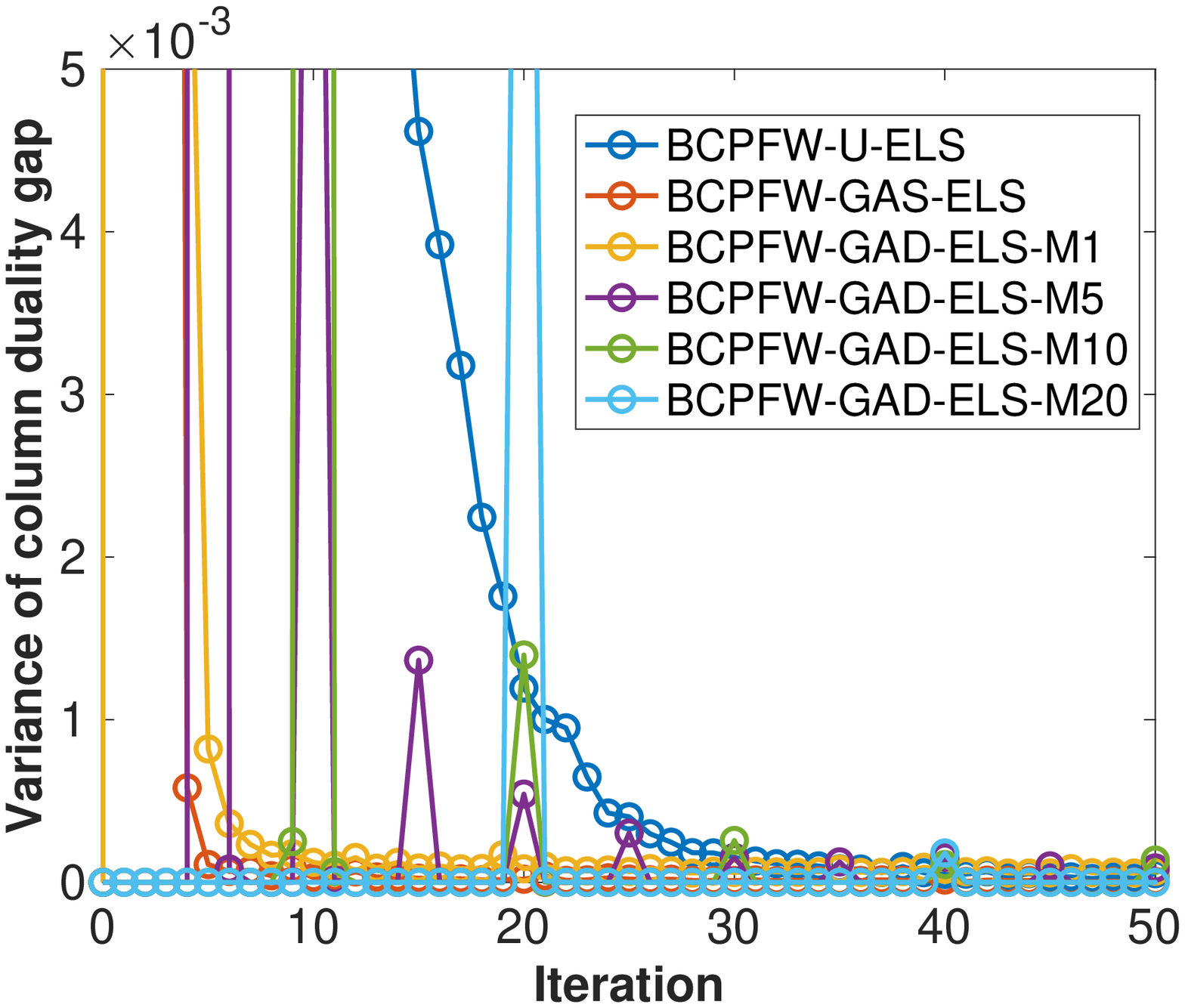}\\
		
		{\footnotesize  (iv) variances of $g_i(\mat{T})$}
		
	\end{center} 
	\end{minipage}
	\vspace*{0.2cm}		
	
	{\small (d) BCPFW-U-ELS and BCPFW-GA-ELS}
	
\caption{Evaluations on duality-adaptive sampling.}
\label{fig:CompAdaptiveSampling}
\end{center}
\end{figure}

This subsections evaluates the effectiveness of the proposed gap-adaptive sampling (BCFW-GA).  
We denote the BCFW-GA with different $M$ as ``BCFW-GAD-\{DEC,ELS\}-Mx''. We also consider the BCFW-GA that does not execute the internal update of the column-wise duality gap $g_i(\mat{T})$, but updates them globally at every outer iteration, i.e., $n$ internal iterations. This is denoted as ``BCFW-GAS-\{DEC,ELS\}''. The baseline method is BCFW-U. Figure \ref{fig:CompAdaptiveSampling} shows the results, of which $x$-axis is from $0$ to $200$ for detailed considerations by addressing the beginning of the (outer) iteration.  We set $\lambda = 10^{-7}$.

As for the case with the decay stepsize rule in (a), BCFW-GAD-DEC-M1 give the fastest convergence in the objective value, and gives more stable dual gap than other  BCFW-GAD-DEC-Mx. Also, BCFW-GAS-DEC yields the wort result. However, with respect to computational time, this superiority vanishes, and the uniform uniform-sampling BCFW-U-DEC yields the smallest objective function value. 

The gap-adaptive sampling strategy relies on the column-wise duality gap $g_i(\mat{T})$ for $i \in [n]$. Therefore, when these variances are large with truly correct estimations, we expect that we can make more significant progress by sampling more often the columns with larger sub-optimality gap. On the other hand, if the variances are small, the performance of the gap-adaptive sampling method is similar to that of the uniform-sampling method. From this consideration, we present the variances of the column-wise duality gap $g_i(\mat{T})$ in each outer iteration in (iv). In order to investigate the behavior of the variances in more detail, the range of $x$-axis is set the first $50$ iterations. From (iv), the variances of BCFW-GAD-ELS-M1 becomes quickly much smaller than that of BCFW-U. This reveals that the adaptive-sampling method can reduce drastically the sub-optimality gap at the beginning of iterations. As for the BCFW-GAD-ELS-M\{5,10,20\} methods, their variances drop to zero values very quickly during $M$ period within a few (1 or 2) iterations, and then go up larger values at the next $M$ cycle. From these results, after a few iterations, these methods start to sample nearly uniformly, where any decrease of the sub-optimality gap cannot be gained. Here, it should be emphasized that these variances are not true ones because the column-wise duality gaps are outdated and contain accumulated errors until the next global update is performed. Consequently, the BCFW with the gap-adaptive sampling with the global update is effective for the reduction of the duality gap with respect to iteration number. However, as pointed out above, it pays a high price for expensive computation for updating the column-wise duality gap and the weighted sampling. 

When the case with the exact line-search stepsize rule in (b), the observations are slightly different from those above. From (i) and (ii) in (b), the BCFW-GAD-ELS-M1 does not improve any performances of the uniform-sampling BCFW-U-ELS, rather it is inferior to the uniform sampling algorithm. Furthermore,  BCFW-GAS-ELS, which does not update the column-wise duality gap in the inner iteration, gives the best performances, but those gains are fairly small. From (iv), its variance of $g_i(\mat{T})$ is much smaller than others. In fact, it is not zero, but, is monotonically decreasing. Therefore, it is understandable that BCFW-GAS-ELS adaptively selects the columns with larger column-wise duality gap $g_i(\mat{T})$, and effectively reduces the total duality gap $g(\mat{T})$. We also see similar observations in the BCAFW and BCPFW algorithms in (c) and (d), respectively. It is, however, that the improvements cannot be seen with respect to computational time. 
Consequently, the exact line-search stepsize rule is effective, and the effectiveness of the gap-adaptive sampling is fairly limited.

\subsection{Evaluations on color-transformed images}

This section specifically evaluates the impact of the relaxation parameter $\lambda$ in the color transfer problem. Then, we discuss a suitable setting to obtain the images that are visually natural.

\subsubsection{Synthetic image data}

\begin{figure}[htbp]
\begin{center}
	\begin{minipage}[t]{0.32\textwidth}
	\begin{center}
		\includegraphics[width=\textwidth]{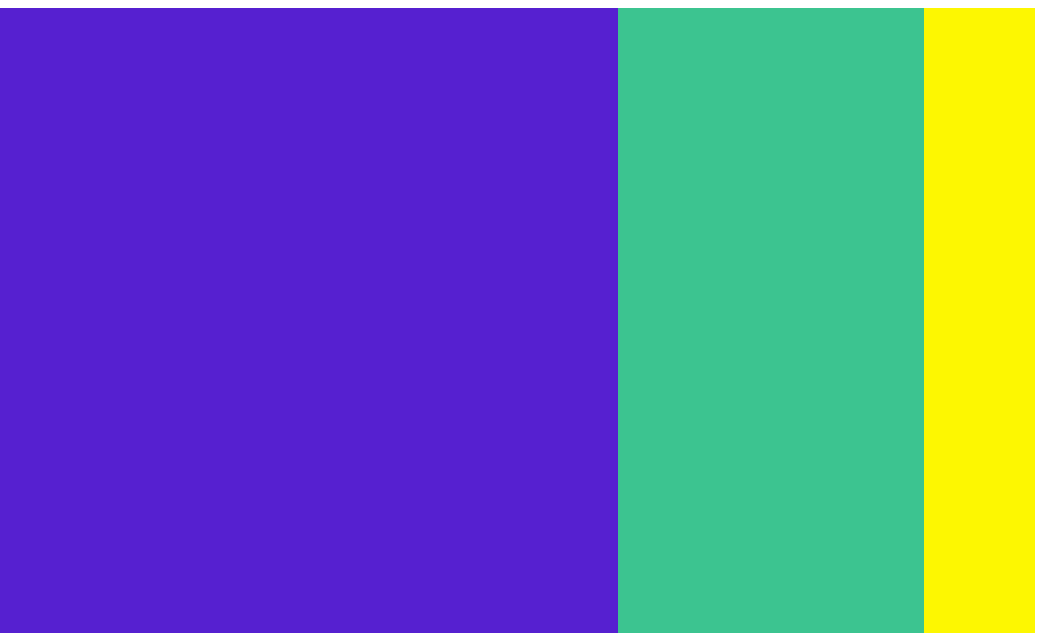}\\
		\vspace*{-0.1cm}
		
		{\footnotesize  (a) source}
		
	\end{center} 
	\end{minipage}
	\begin{minipage}[t]{0.32\textwidth}
	\begin{center}
		\includegraphics[width=\textwidth]{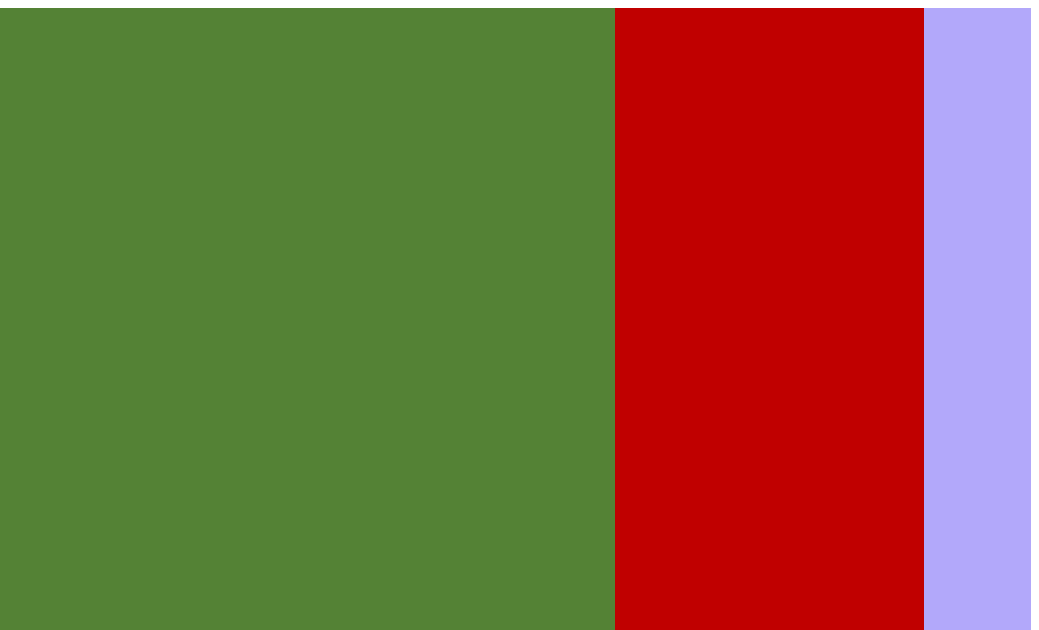}\\
		\vspace*{-0.1cm}
				
		{\footnotesize  (b) reference}
		
	\end{center} 
	\end{minipage}
	\begin{minipage}[t]{0.32\textwidth}
	\begin{center}
		\includegraphics[width=\textwidth]{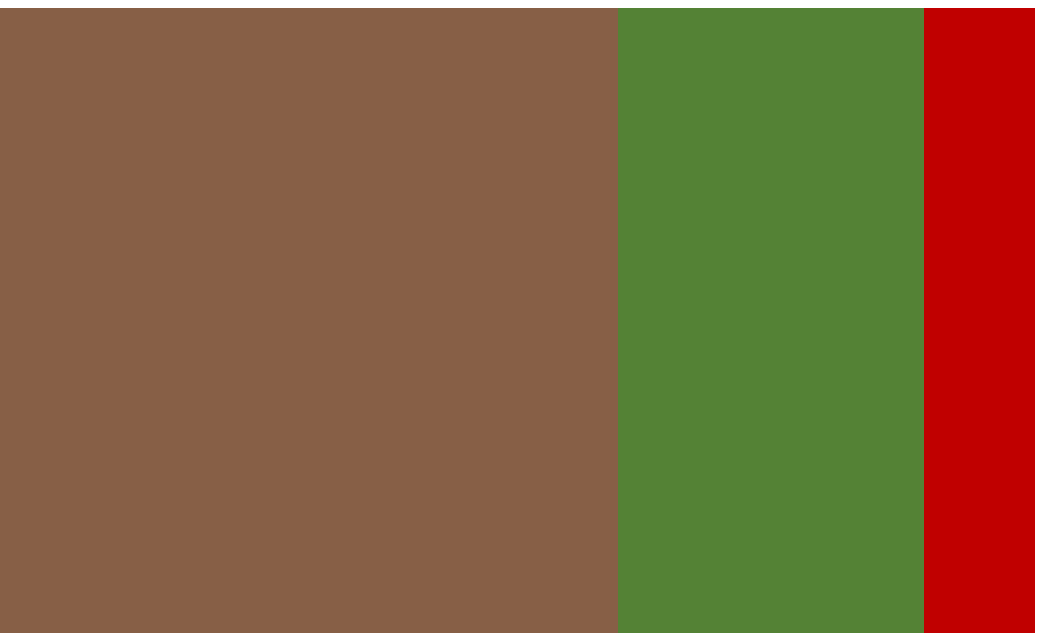}\\
		\vspace*{-0.1cm}
				
		{\footnotesize  (c) color-transferred image by LP}
		
	\end{center} 
	\end{minipage}
\caption{Source and reference synthetic images, and color-transferred image by LP  ($n=m=3$).}
\label{fig:CTOriSyntheImages}
\end{center} 
\end{figure}

In this section, two synthetic images are created, which are the source image (a) and the reference image (b) as shown in Figure \ref{fig:CTOriSyntheImages}. These two images contain three colors, where $n=m=3$, and $\vec{a}\approx(0.1,0.3,0.6)^T$ and $\vec{b}\approx(0.6,0.3,0.1)^T$. In this setting, BCFW-U does not necessarily sample all the columns at every outer iteration due to the very extremely small $m$ and $n$. Thus, this poses difficulties to analyze the transitions of the color-transferred images and the transport matrices. Therefore, we used BCFW-P-DEC insted, which runs a cyclic order on a permuted index at every outer iteration, i.e., epoch. We compare two cases $\lambda = \{10^{-6},10^{-3}\}$.

\begin{figure}[htbp]
\begin{center} 
	\begin{minipage}[t]{0.24\textwidth}
	\begin{center}
		\includegraphics[width=\textwidth]{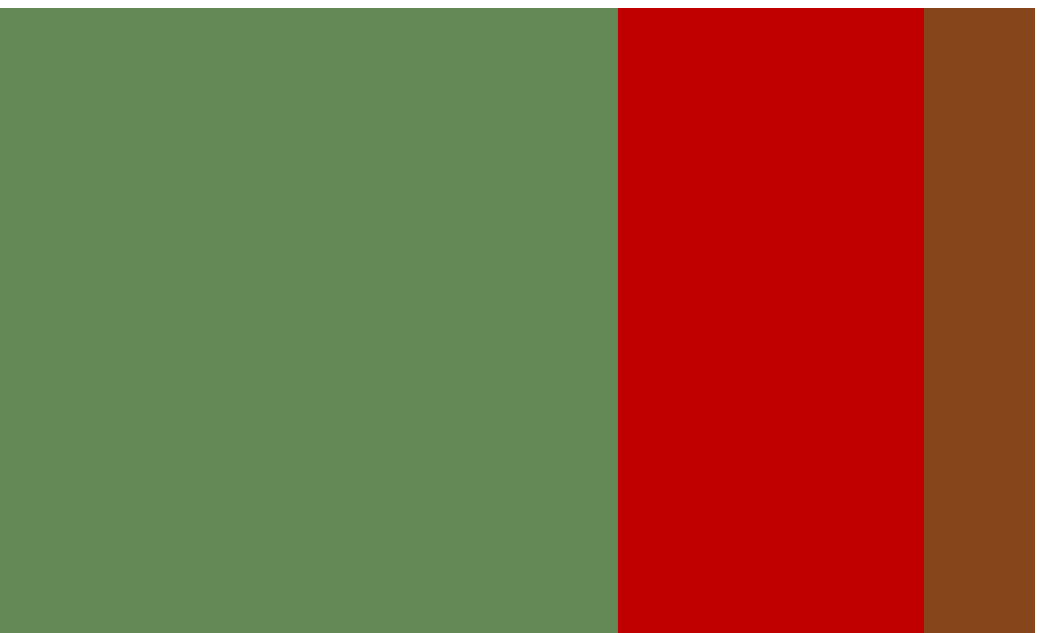}\\
		
		{\footnotesize  $k = 1$}		
	\end{center} 
	\end{minipage}
	\begin{minipage}[t]{0.24\textwidth}
	\begin{center}
		\includegraphics[width=\textwidth]{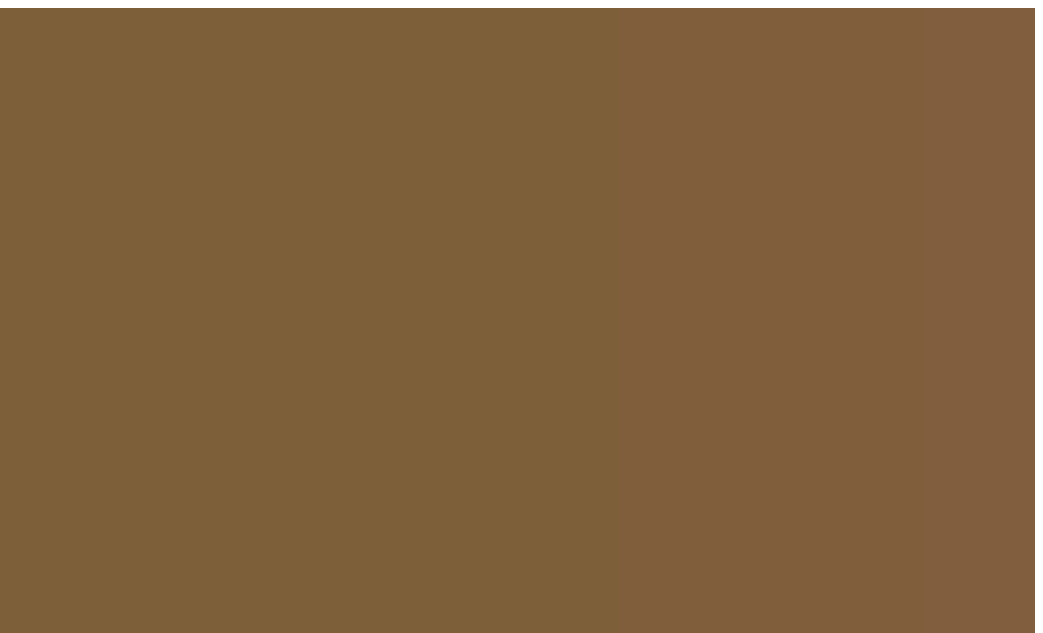}\\
		
		{\footnotesize  $k = 140$}			
	\end{center} 
	\end{minipage}	
	\begin{minipage}[t]{0.24\textwidth}
	\begin{center}
		\includegraphics[width=\textwidth]{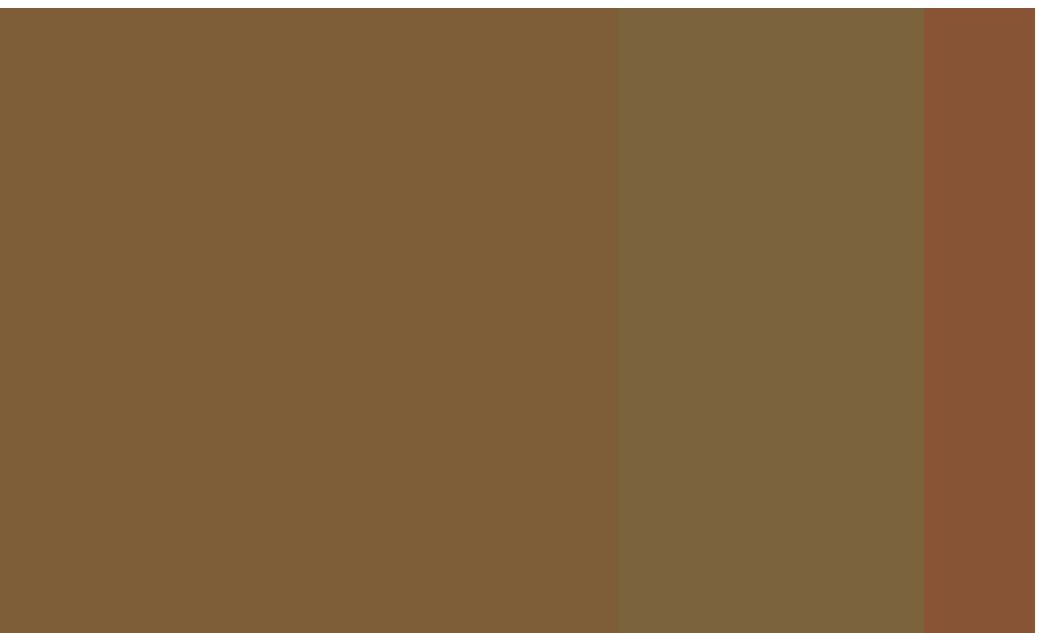}\\
		
		{\footnotesize  $k = 10^3$}		
	\end{center} 
	\end{minipage}	
	\begin{minipage}[t]{0.24\textwidth}
	\begin{center}
		\includegraphics[width=\textwidth]{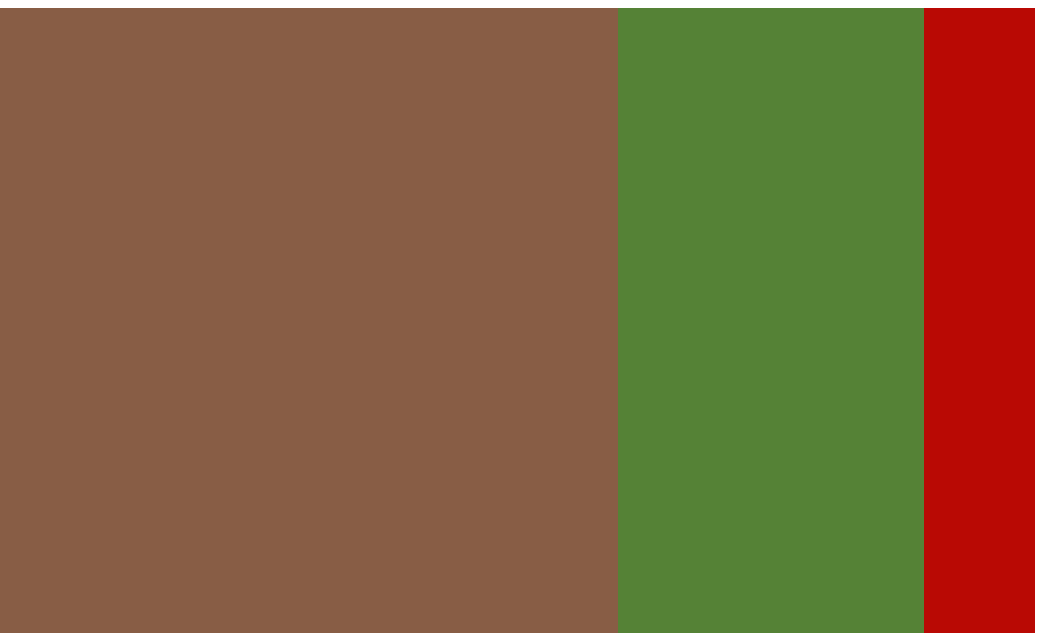}\\
			
		{\footnotesize  $k = 5\times 10^4$}		
	\end{center} 
	\end{minipage}	
	\vspace*{0.2cm}
	
	{\small  (a) transition of color-transferred images}			
	\vspace*{0.2cm}
	
	\begin{minipage}[t]{0.24\textwidth}
	\begin{center}
	{\scriptsize
	\[
		\mat{T}^{(1)}\!\!=\!\!  \left(\!\!
		\begin{array}{ccc}
		0.085 \!\!&\!\! 0.075 \!\!&\!\! 0.000 \\
		0.000 \!\!&\!\! 0.220 \!\!&\!\! 0.000 \\
		0.510 \!\!&\!\! 0.000 \!\!&\!\! 0.100 \\
		\end{array}
	\!\!\right)
	\]
	}
	\end{center} 
	\end{minipage}
	\begin{minipage}[t]{0.24\textwidth}
	\begin{center}
	{\scriptsize
	\[
		\mat{T}^{(140)}\!\!=\!\!  \left(\!\!
		\begin{array}{ccc}
		0.060 \!\!&\!\! 0.034 \!\!&\!\! 0.014 \\
		0.170 \!\!&\!\! 0.093 \!\!&\!\! 0.034 \\
		0.360 \!\!&\!\! 0.170 \!\!&\!\! 0.057 \\
		\end{array}
	\!\!\right)
	\]
	}
	\end{center} 
	\end{minipage}	
	\begin{minipage}[t]{0.24\textwidth}
	\begin{center}
	{\scriptsize
	\[
		\mat{T}^{(10^3)}\!\!=\!\!  \left(\!\!
		\begin{array}{ccc}
		0.055 \!\!&\!\! 0.041 \!\!&\!\! 0.011 \\
		0.190 \!\!&\!\! 0.078 \!\!&\!\! 0.033 \\
		0.360 \!\!&\!\! 0.180 \!\!&\!\! 0.060 \\
		\end{array}
	\!\!\right)
	\]
	}
	\end{center} 
	\end{minipage}	
	\begin{minipage}[t]{0.24\textwidth}
	\begin{center}
	{\scriptsize
	\[
		\ \ \mat{T}^{(5\cdot10^{\tiny 5})}\!\!=\!\!  \left(\!\!
		\begin{array}{ccc}
		0.007 \!\!&\!\! 0.100 \!\!&\!\! 0.000 \\
		0.290 \!\!&\!\! 0.000 \!\!&\!\! 0.001 \\
		0.300 \!\!&\!\! 0.300 \!\!&\!\! 0.100 \\
		\end{array}
	\!\!\right)
	\]
	}
	\end{center} 
	\end{minipage}
	\vspace*{0.2cm}
	
	{\small  (b) transition of transport matrices $\mat{T}^{(k)}$}

	\vspace*{0.7cm}	
	
	\begin{minipage}[t]{0.24\textwidth}
	\begin{center}
	{\scriptsize heat-map of \vec{b}}
	
		\hspace*{-0.42cm}\includegraphics[width=0.87\textwidth]{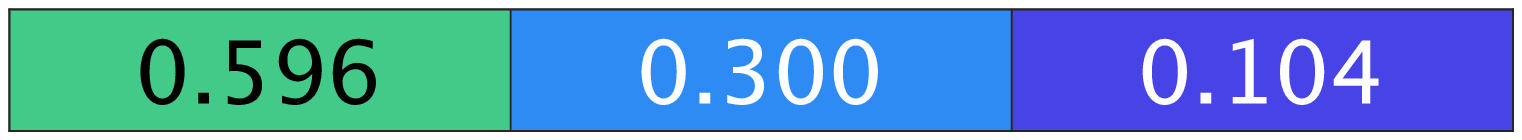}
			
		\hspace*{0.0cm}\includegraphics[width=\textwidth]{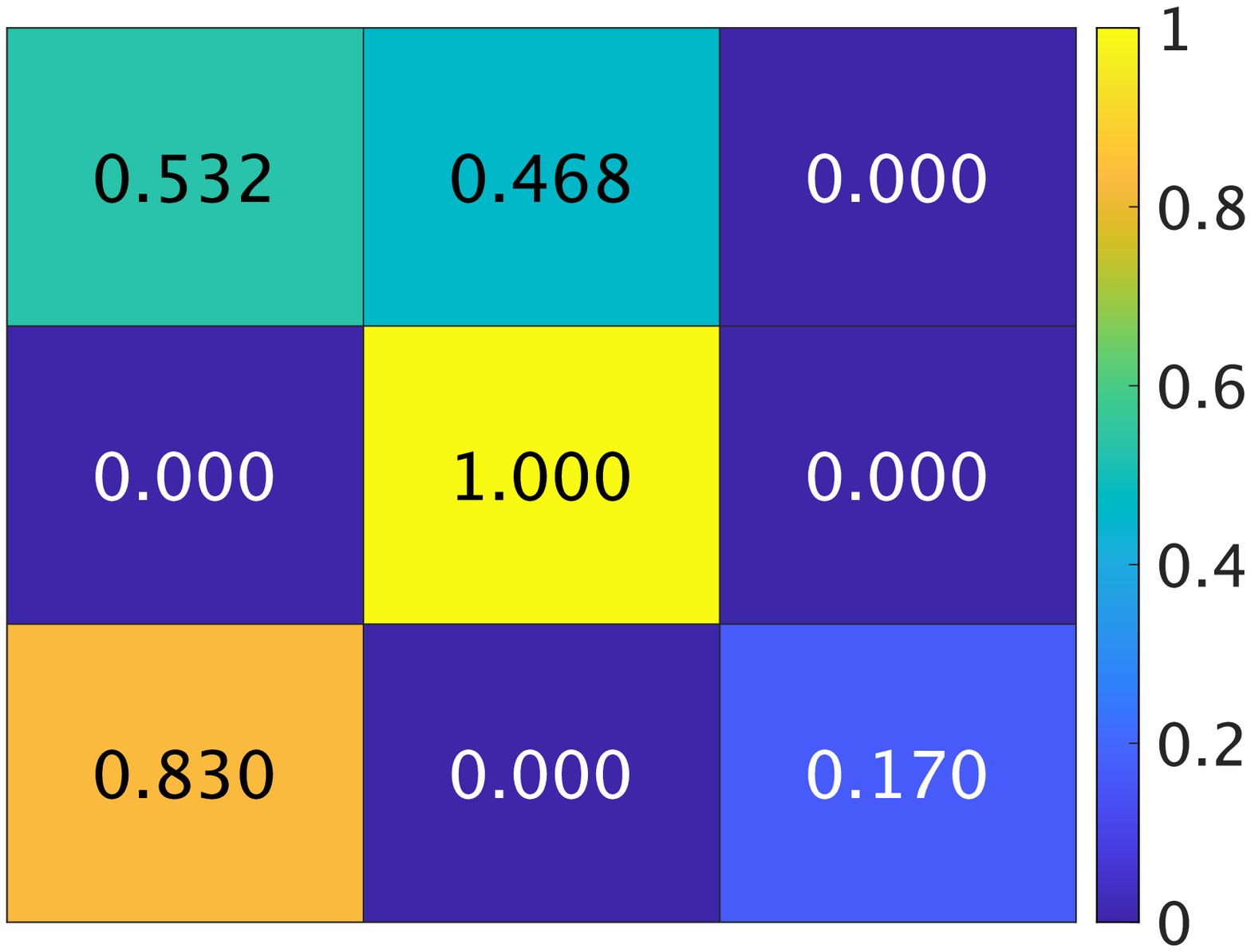}\\
			
	{\footnotesize  $k = 1$}		
	\end{center} 
	\end{minipage}
	\begin{minipage}[t]{0.24\textwidth}
	\begin{center}
	{\scriptsize heat-map of \vec{b}}
		
		\hspace*{-0.42cm}\includegraphics[width=0.87\textwidth]{results/images/ThreeColor/Reference_Distribution.eps}
			
		\hspace*{0.0cm}\includegraphics[width=\textwidth]{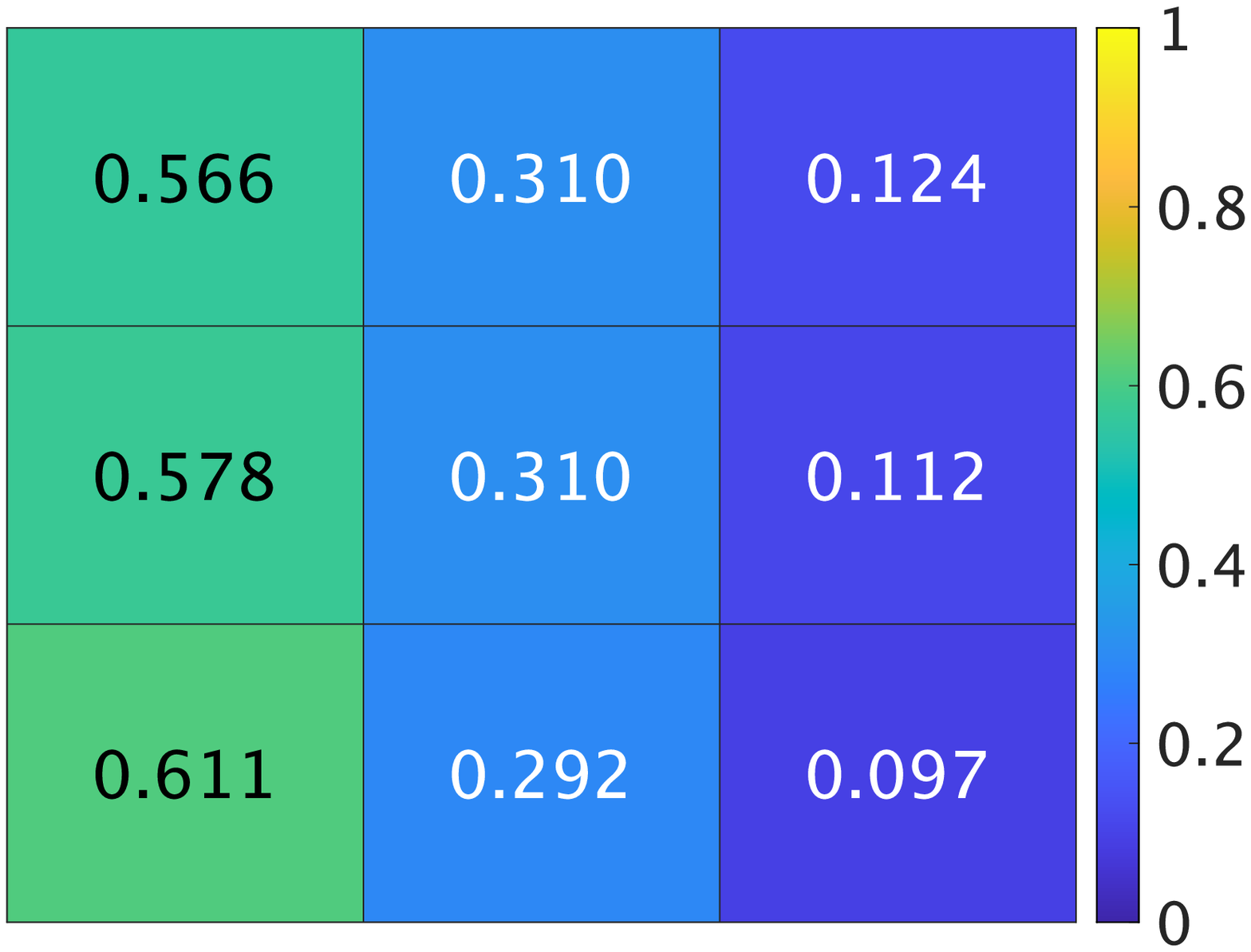}\\
			
	{\footnotesize  $k = 140$}	
	\end{center} 
	\end{minipage}	
	\begin{minipage}[t]{0.24\textwidth}
	\begin{center}
	{\scriptsize heat-map of \vec{b}}
		
		\hspace*{-0.42cm}\includegraphics[width=0.87\textwidth]{results/images/ThreeColor/Reference_Distribution.eps}	
			
		\hspace*{0.0cm}\includegraphics[width=\textwidth]{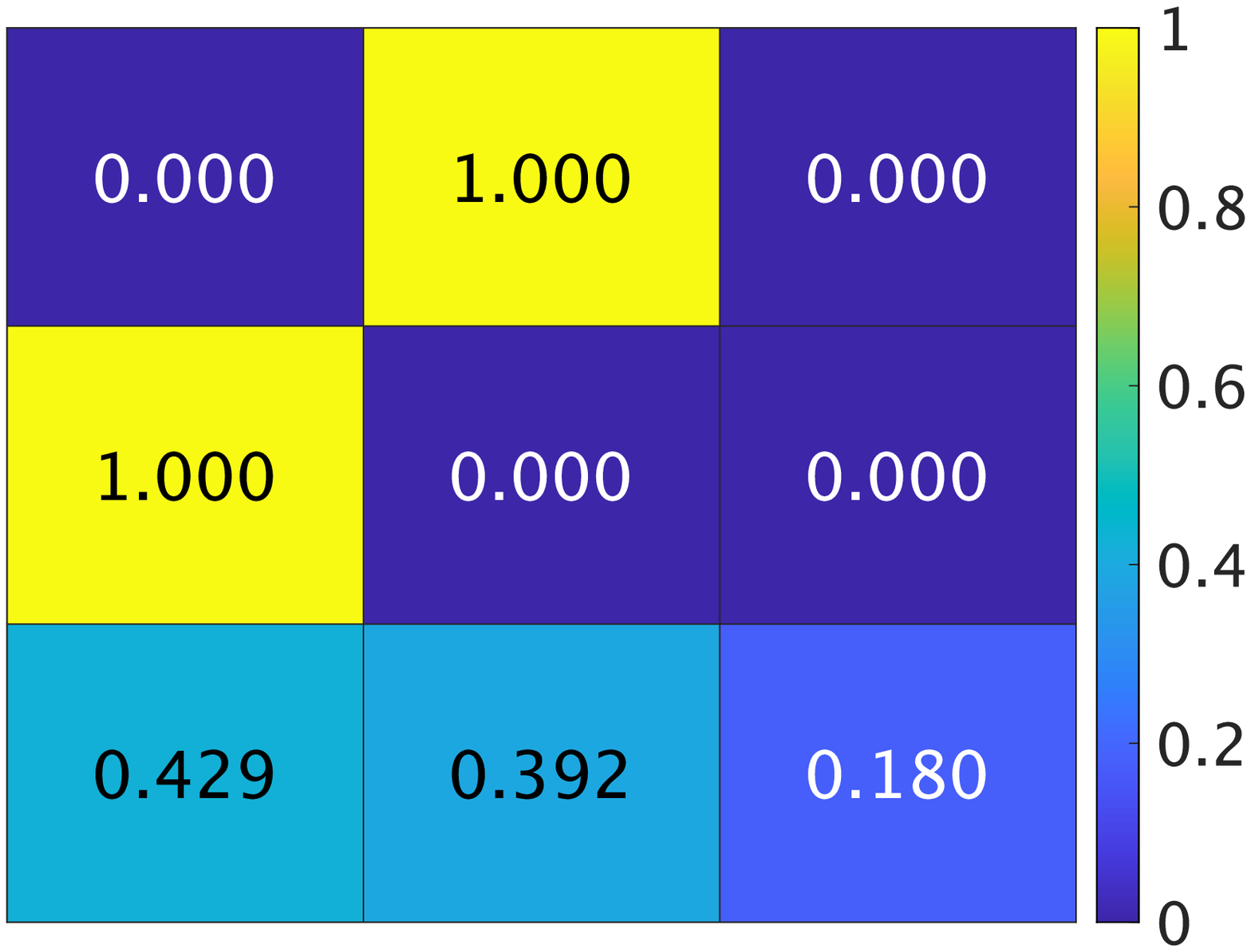}\\
			
	{\footnotesize  $k = 10^3$}		
	\end{center} 
	\end{minipage}	
	\begin{minipage}[t]{0.24\textwidth}
	\begin{center}
	{\scriptsize heat-map of \vec{b}}
		
		\hspace*{-0.42cm}\includegraphics[width=0.87\textwidth]{results/images/ThreeColor/Reference_Distribution.eps}	
			
		\hspace*{0.0cm}\includegraphics[width=\textwidth]{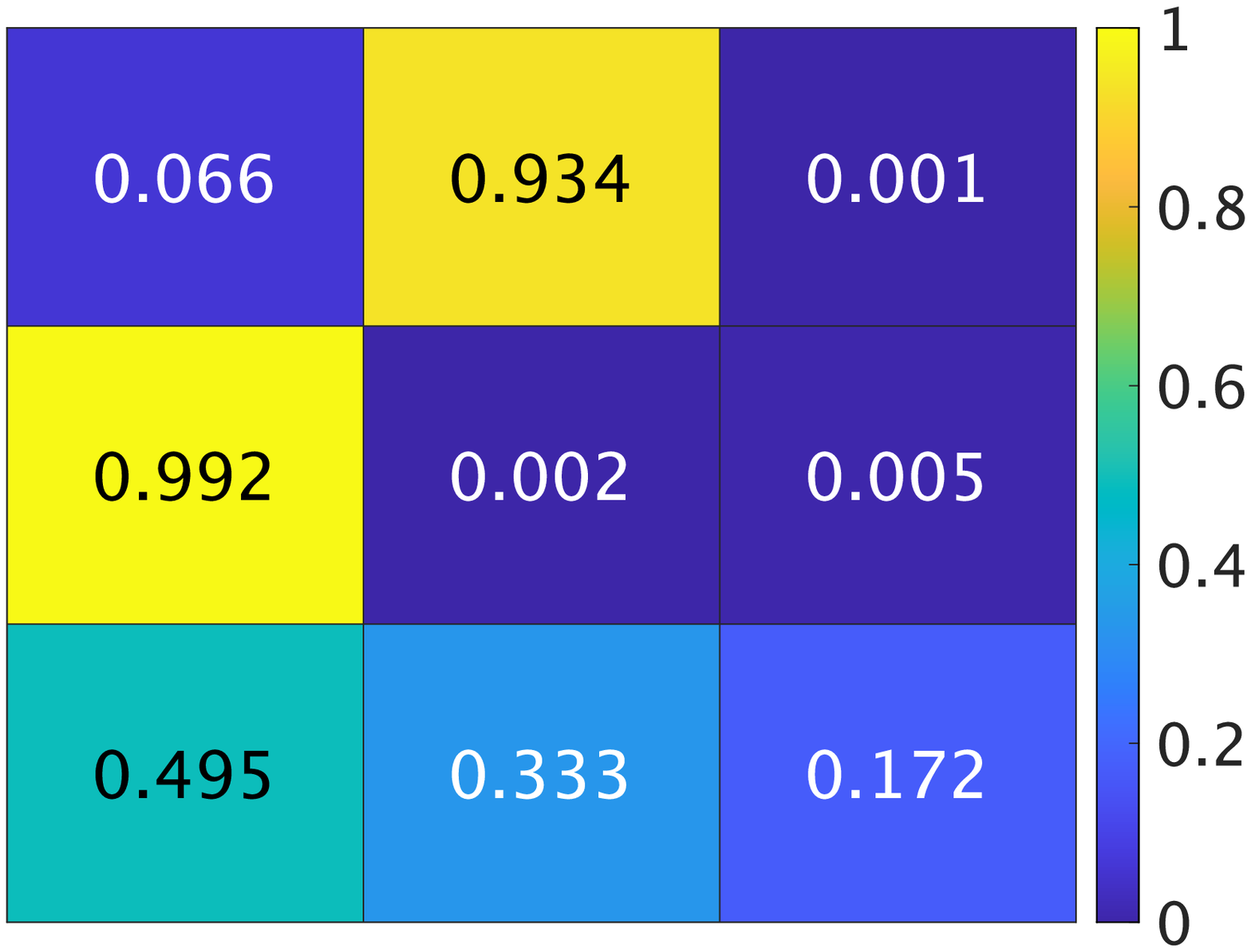}\\
			
	{\footnotesize  $k = 5\times 10^4$}	
	\end{center} 
	\end{minipage}
	\vspace*{0.2cm}
	
	{\small  (c) heat-map of  row-wise normalized transport matrices of $\mat{T}^{(k)}$ with \vec{b}}
	\vspace*{0.6cm}
	
	\begin{minipage}[t]{0.33\textwidth}
	\begin{center}
		\includegraphics[width=\textwidth]{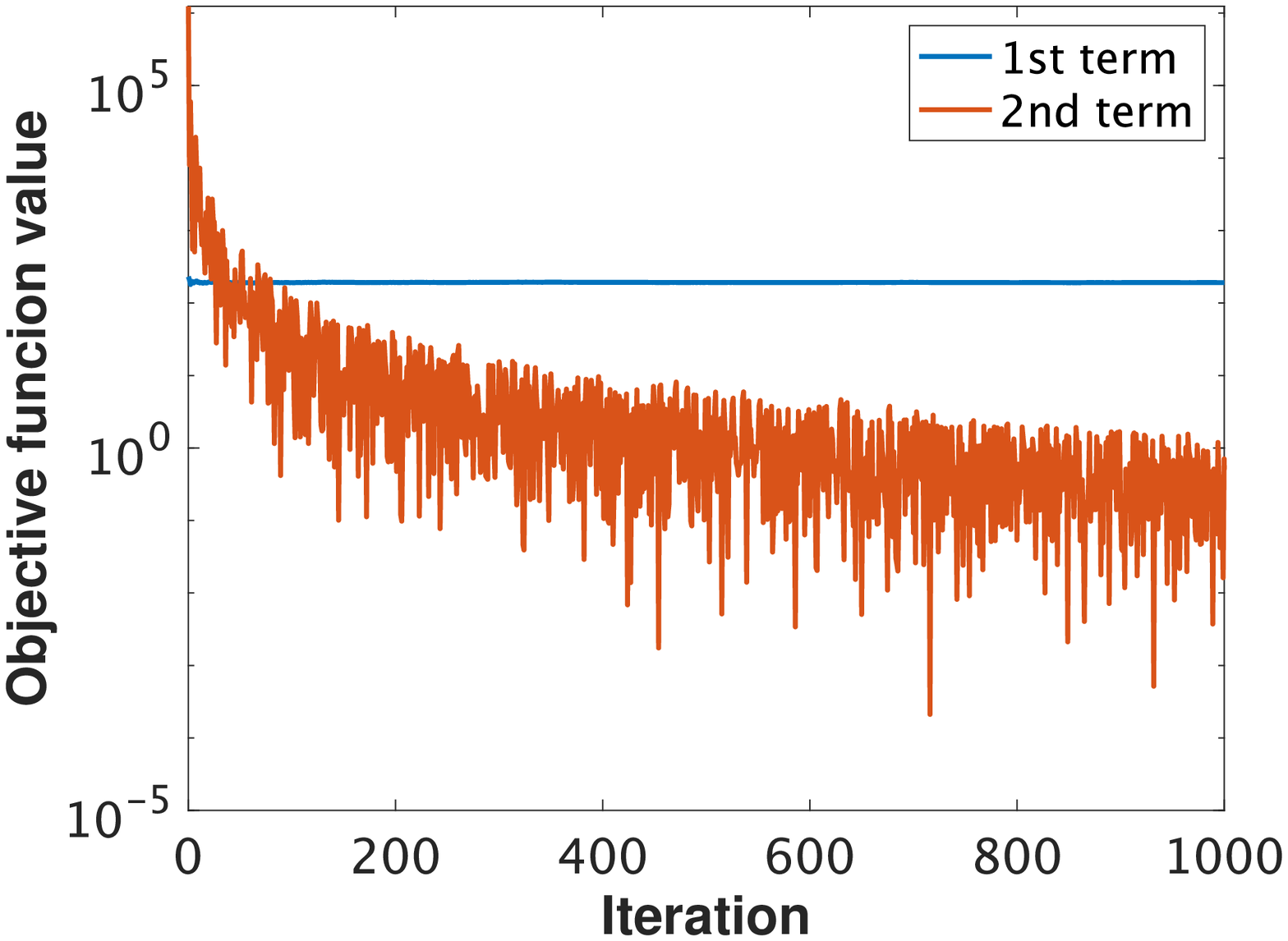}\\
		
		{\small  (d) objective value}	
	\end{center} 
	\end{minipage}
	\begin{minipage}[t]{0.33\textwidth}
	\begin{center}
		\includegraphics[width=\textwidth]{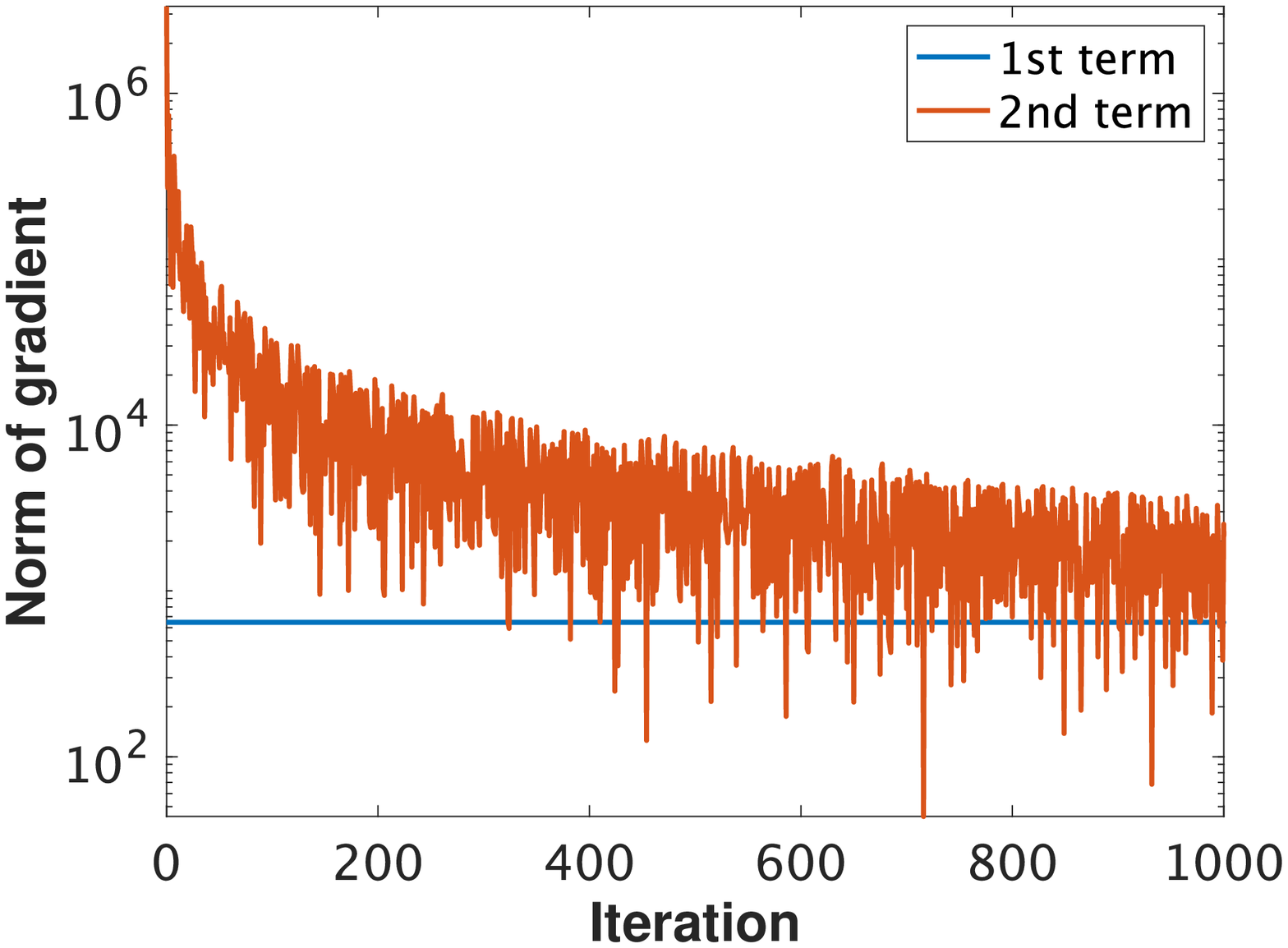}\\
		
		{\small  (e) norm of gradient}	
	\end{center} 
	\end{minipage}

\caption{Transition of color-transferred images, transport matrices, normalized transport matrices, objective value, and norm of gradient ($\lambda=10^{-6}$).}
\label{fig:CTImages_lambda6}
\end{center}		
\end{figure}

\begin{figure}[htbp]
\begin{center} 
	\begin{minipage}[t]{0.24\textwidth}
	\begin{center}
		\includegraphics[width=\textwidth]{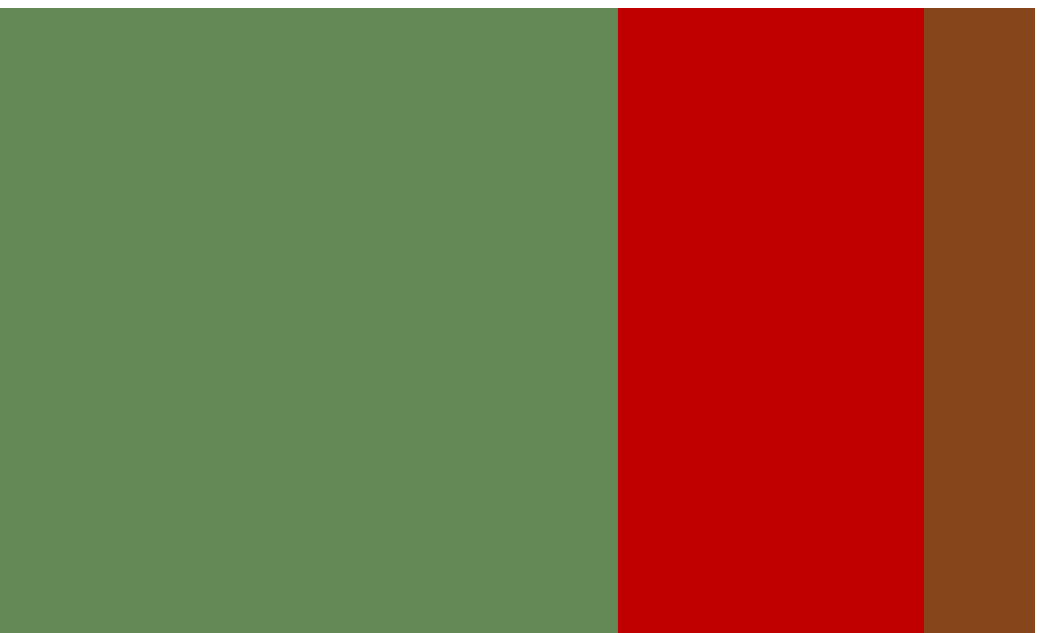}\\
		
		{\footnotesize  $k = 1$}	
	\end{center} 
	\end{minipage}
	\begin{minipage}[t]{0.24\textwidth}
	\begin{center}
		\includegraphics[width=\textwidth]{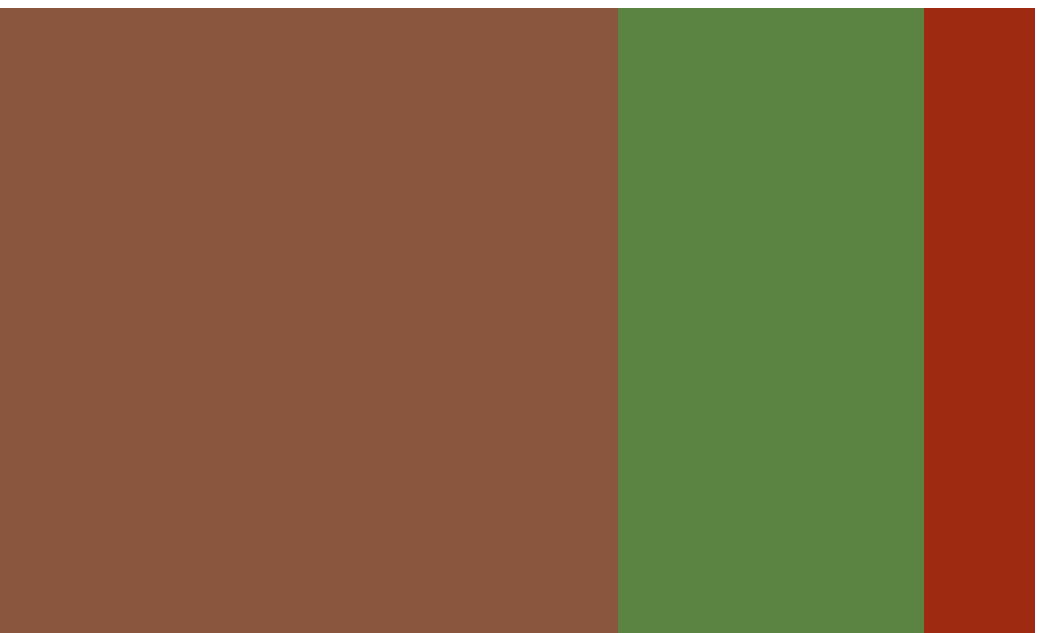}\\
		
		{\footnotesize  $k = 10$}		
	\end{center} 
	\end{minipage}	
	\begin{minipage}[t]{0.24\textwidth}
	\begin{center}
		\includegraphics[width=\textwidth]{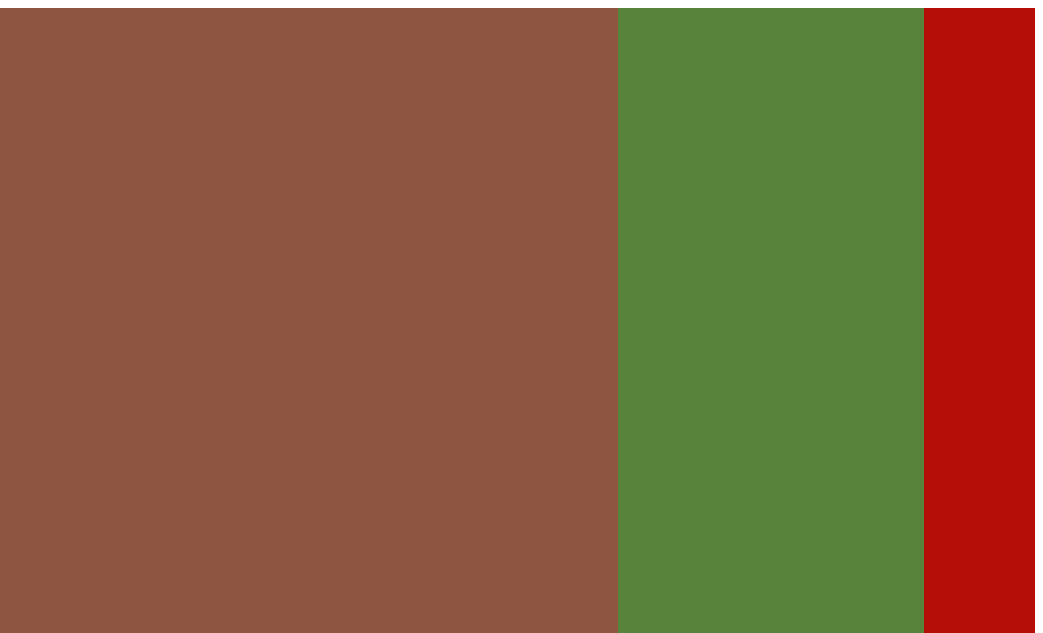}\\
		
		{\footnotesize  $k = 30$}		
	\end{center} 
	\end{minipage}	
	\begin{minipage}[t]{0.24\textwidth}
	\begin{center}
		\includegraphics[width=\textwidth]{results/images/ThreeColor/lambda_3/BCFW-P_iter30_lambda3.eps}\\
		
		{\footnotesize  $k = 10^3$}		
	\end{center} 
	\end{minipage}	
	\vspace*{0.2cm}
	
	{\small  (a) transition of color-transferred images}			
	\vspace*{0.2cm}
	
	\begin{minipage}[t]{0.24\textwidth}
	\begin{center}
	{\scriptsize
	\[
		\mat{T}^{(1)}\!\!=\!\!  \left(\!\!
		\begin{array}{ccc}
		0.085 \!\!&\!\! 0.075 \!\!&\!\! 0.000 \\
		0.000 \!\!&\!\! 0.220 \!\!&\!\! 0.000 \\
		0.510 \!\!&\!\! 0.000 \!\!&\!\! 0.100 \\
		\end{array}
	\!\!\right)
	\]
	}
	\end{center} 
	\end{minipage}
	\begin{minipage}[t]{0.24\textwidth}
	\begin{center}
	{\scriptsize
	\[
		\mat{T}^{(10)}\!\!=\!\!  \left(\!\!
		\begin{array}{ccc}
		0.076 \!\!&\!\! 0.018 \!\!&\!\! 0.000\\
		0.320 \!\!&\!\! 0.006 \!\!&\!\! 0.022 \\
		0.200 \!\!&\!\! 0.280 \!\!&\!\! 0.083 \\
		\end{array}
	\!\!\right)
	\]
	}
	\end{center} 
	\end{minipage}	
	\begin{minipage}[t]{0.24\textwidth}
	\begin{center}
	{\scriptsize
	\[
		\mat{T}^{(30)}\!\!=\!\!  \left(\!\!
		\begin{array}{ccc}
		0.001 \!\!&\!\! 0.080\!\!&\!\! 0.000 \\
		0.340 \!\!&\!\! 0.000 \!\!&\!\! 0.010 \\
		0.250 \!\!&\!\! 0220 \!\!&\!\! 0.095 \\	
		\end{array}
	\!\!\right)
	\]
	}
	\end{center} 
	\end{minipage}	
	\begin{minipage}[t]{0.24\textwidth}
	\begin{center}
	{\scriptsize
	\[
		\mat{T}^{(10^{\tiny 3})}\!\!=\!\!  \left(\!\!
		\begin{array}{ccc}
		0.000\!\!&\!\! 0.072 \!\!&\!\! 0.000 \\
		0.350 \!\!&\!\! 0.000 \!\!&\!\! 0.000 \\
		0.250 \!\!&\!\! 0.230 \!\!&\!\! 0.100 \\
		\end{array}
	\!\!\right)
	\]
	}
	\end{center} 
	\end{minipage}
	\vspace*{0.2cm}
	
	{\small  (b) transition of transport matrices $\mat{T}^{(k)}$}

	\vspace*{0.7cm}	
	
	\begin{minipage}[t]{0.24\textwidth}
	\begin{center}
	{\scriptsize heat-map of \vec{b}}
	
		\hspace*{-0.42cm}\includegraphics[width=0.87\textwidth]{results/images/ThreeColor/Reference_Distribution.eps}	
		
		\hspace*{0.0cm}\includegraphics[width=\textwidth]{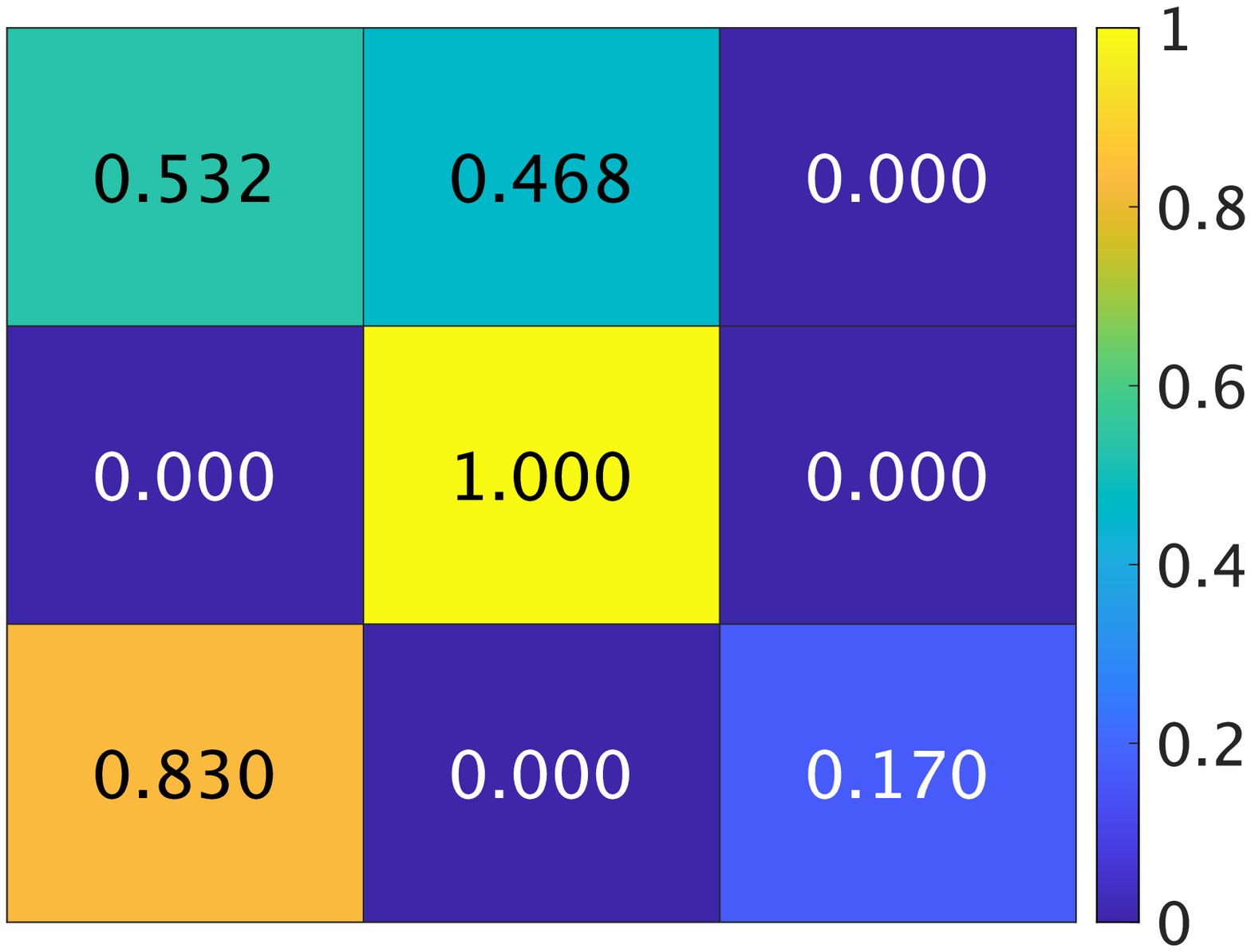}\\
		
	{\footnotesize  $k = 1$}		
	\end{center} 
	\end{minipage}
	\begin{minipage}[t]{0.24\textwidth}
	\begin{center}
	{\scriptsize heat-map of \vec{b}}
	
		\hspace*{-0.42cm}\includegraphics[width=0.87\textwidth]{results/images/ThreeColor/Reference_Distribution.eps}	
			
		\hspace*{0.0cm}\includegraphics[width=\textwidth]{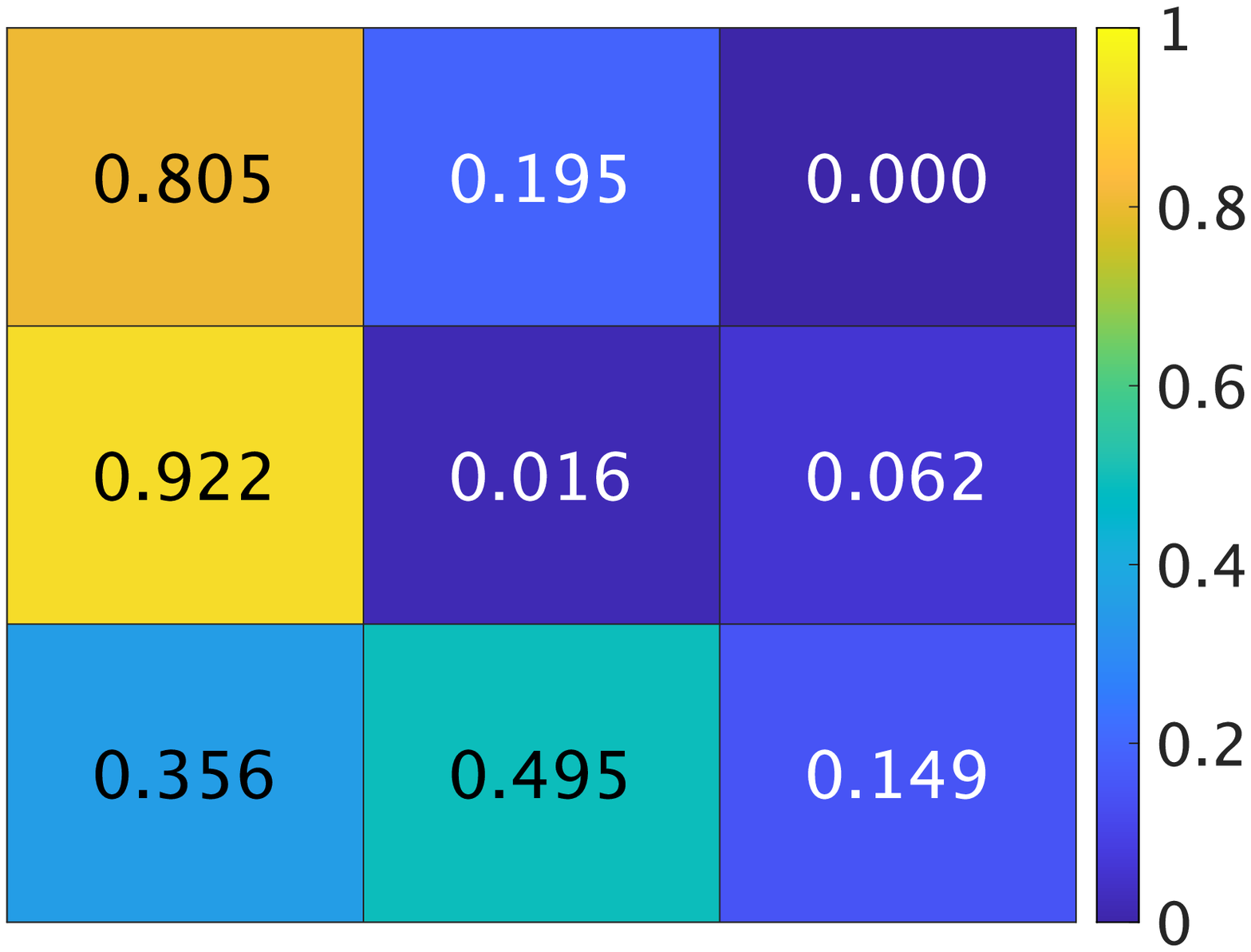}\\
	
	{\footnotesize  $k = 10$}		
	\end{center} 
	\end{minipage}	
	\begin{minipage}[t]{0.24\textwidth}
	\begin{center}
	{\scriptsize heat-map of \vec{b}}
	
		\hspace*{-0.42cm}\includegraphics[width=0.87\textwidth]{results/images/ThreeColor/Reference_Distribution.eps}	
		
		\hspace*{0.0cm}\includegraphics[width=\textwidth]{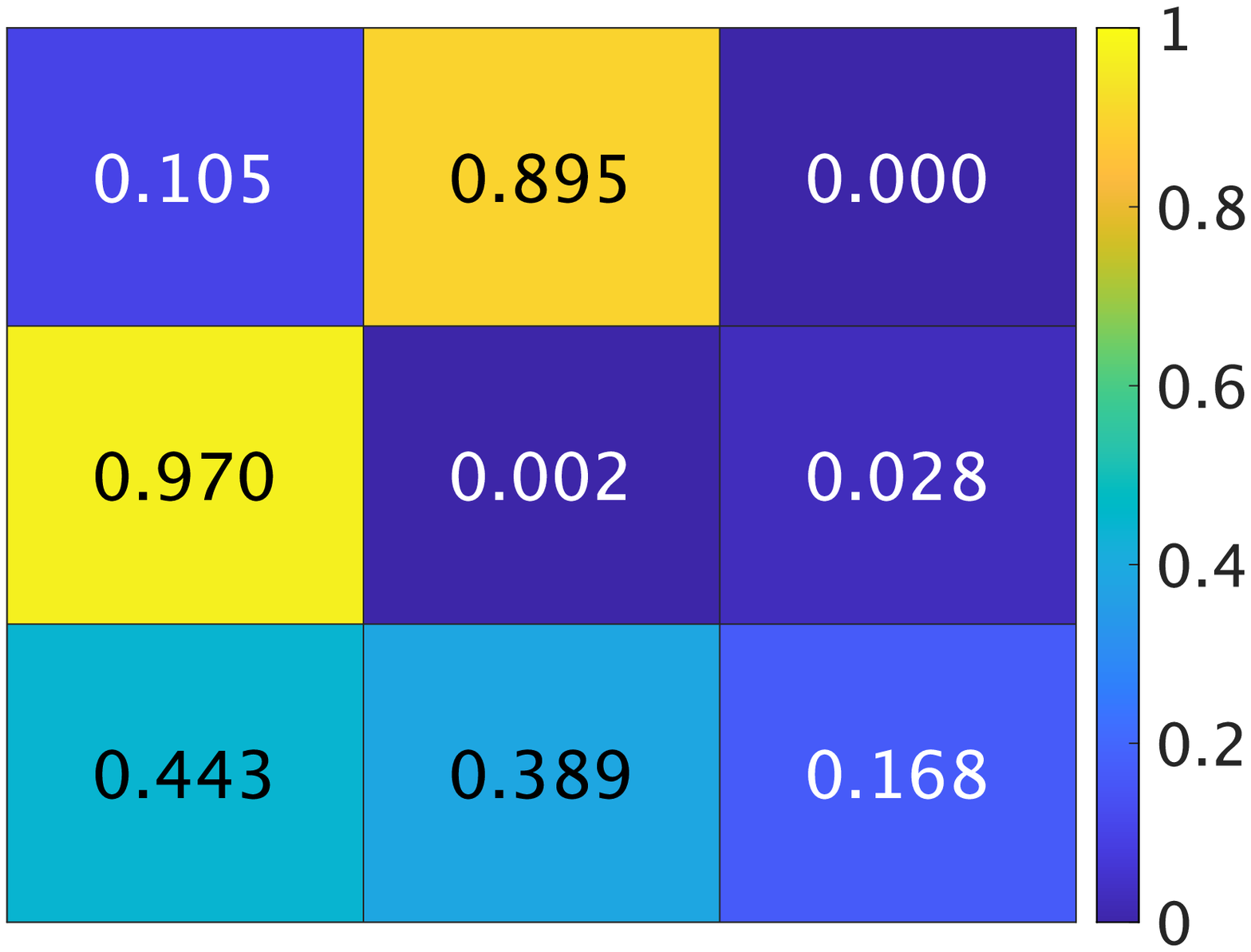}\\
	
	{\footnotesize  $k = 30$}	
	\end{center} 
	\end{minipage}	
	\begin{minipage}[t]{0.24\textwidth}
	\begin{center}
	{\scriptsize heat-map of \vec{b}}
	
		\hspace*{-0.42cm}\includegraphics[width=0.87\textwidth]{results/images/ThreeColor/Reference_Distribution.eps}	
		
		\hspace*{0.0cm}\includegraphics[width=\textwidth]{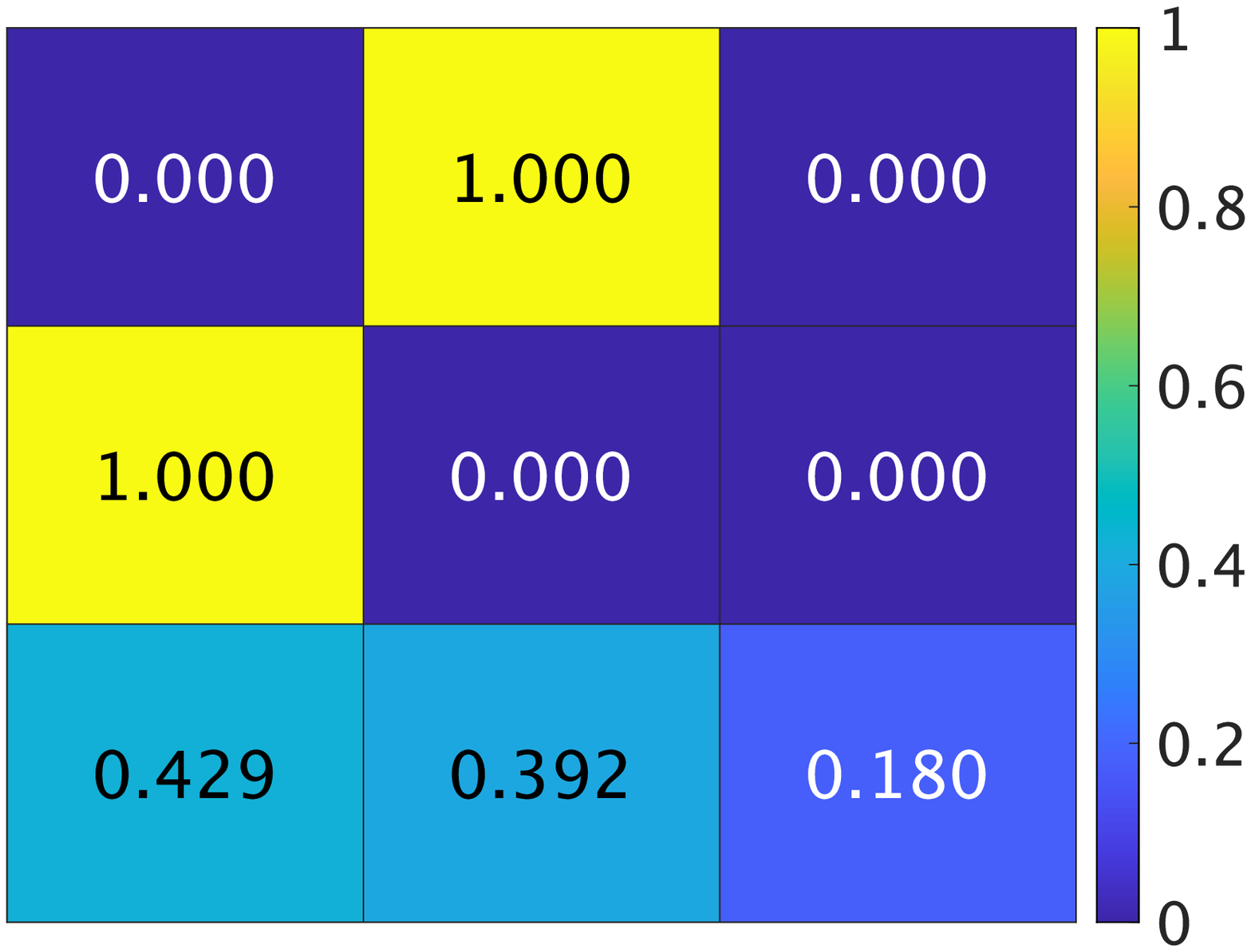}\\
	
	{\footnotesize  $k = 10^3$}	
	\end{center} 
	\end{minipage}
	\vspace*{0.2cm}
	
	{\small  (c) heat-map of row-wise normalized transport matrices of $\mat{T}^{(k)}$ and \vec{b}}
	\vspace*{0.6cm}
	
	\begin{minipage}[t]{0.33\textwidth}
	\begin{center}
		\includegraphics[width=\textwidth]{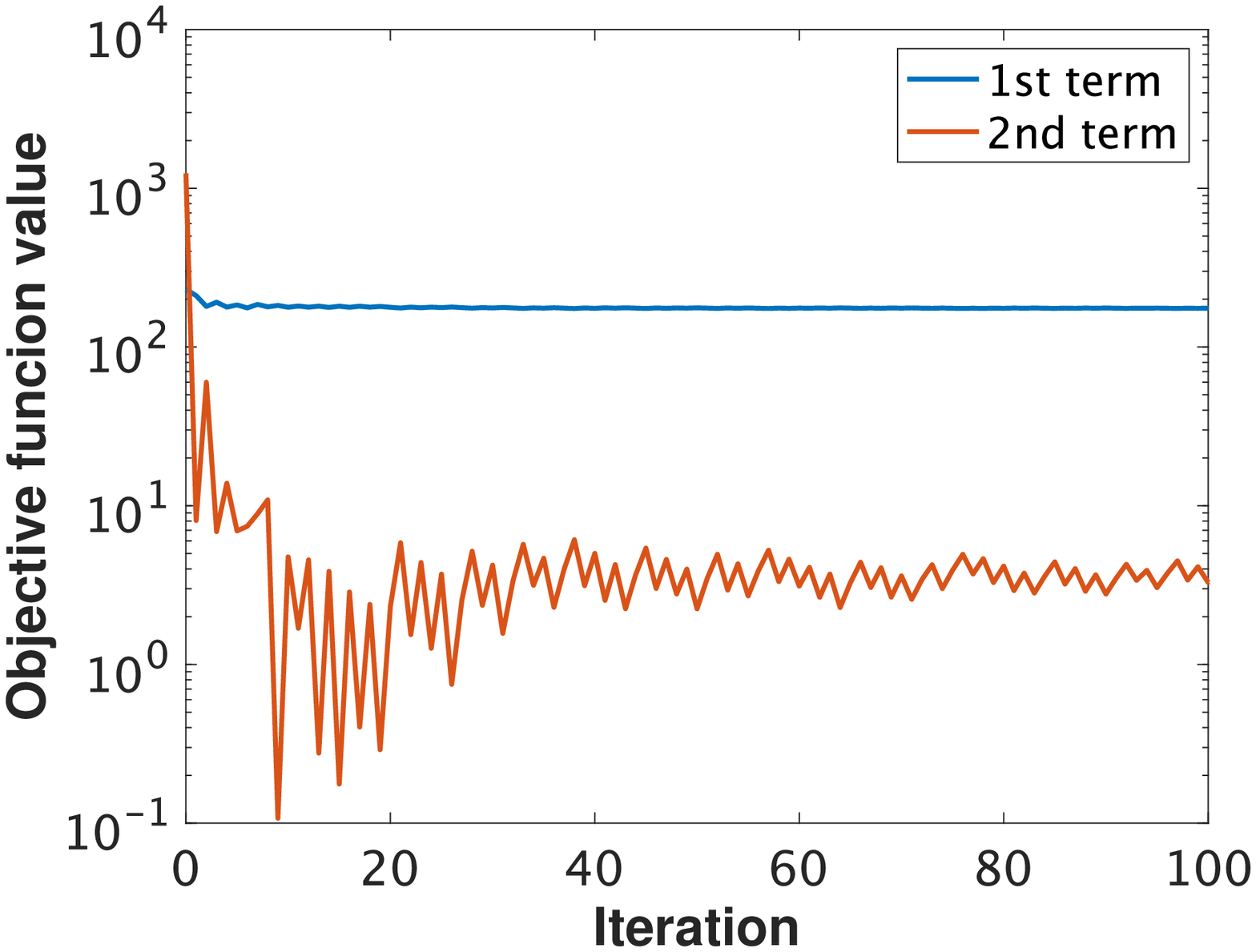}\\
		
		{\small  (d) objective value}	
				
	\end{center} 
	\end{minipage}	
	\begin{minipage}[t]{0.33\textwidth}
	\begin{center}
		\includegraphics[width=\textwidth]{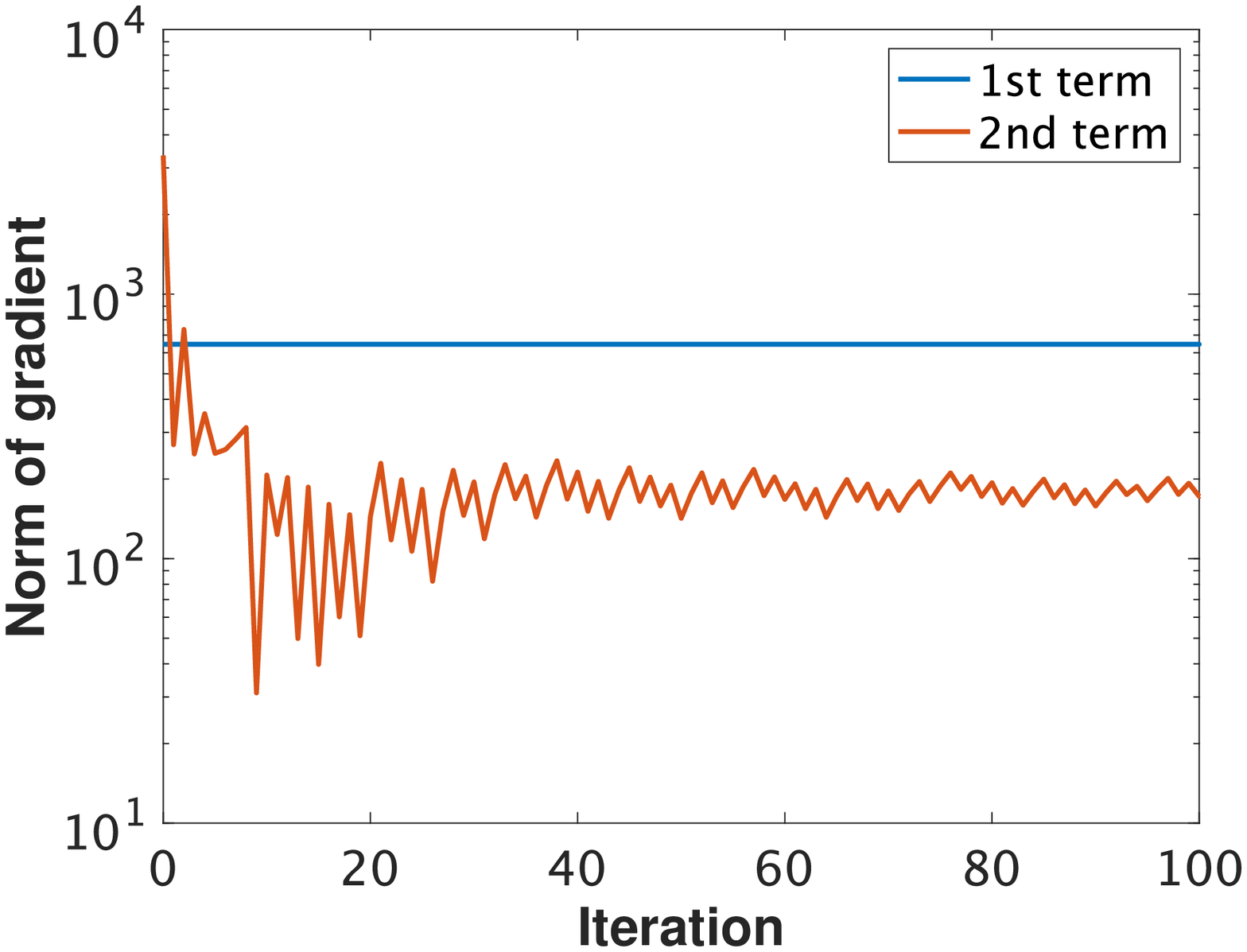}\\
		
		{\small  (e) norm of gradient}			
	\end{center} 
	\end{minipage}	
	
\caption{Transition of color-transferred images, transport matrices, normalized transport matrices, objective value, and norm of gradient ($\lambda=10^{-3}$).}
\label{fig:CTImages_lambda3}
\end{center}		
\end{figure}


\vspace*{0.2cm}
\noindent
{\bf Small relaxation parameter ($\lambda = 10^{-6}$) in Figure \ref{fig:CTImages_lambda6}.} At the beginning of the iterations, the first term $\langle \mat{T},\mat{C} \rangle$ of the objective function $f(\mat{T}):=\langle \mat{T},\mat{C} \rangle + \frac{1}{2\lambda}\|\mat{T}\vec{1}_n-\vec{a}\|_2^2$ is mostly ignored in terms of {\it optimization} as shown in (d), and has little impact on the solution because of small $\lambda$. During this phase, the cost matrix $\mat{C}$ does not have any impact, and the optimization process proceeds to reduce the second term $\frac{1}{2\lambda}\|\mat{T}\vec{1}_n - \vec{a}\|_2^2$. This can be verified in (d), which presents the objective function value and the norm of gradient of these two separated terms. As seen in (d), it takes a few hundred of iterations to decrease the function value of the second term. Therefore, if the initial value of $\mat{T}^{(0)}$ satisfies the constraint of $\vec{b}$, i.e., $\mat{T}^T\vec{1}_m=\vec{b}$, the optimization process tends to operate in the vertical direction of \mat{T}, i.e., the row direction, in order to satisfy the given distribution of $\vec{a}$. As a result, {\it the ratio of the elements in each row vector moves closer to that of $\vec{b}$}. More specifically, the $i$-th row vector of $\mat{T}$, i.e., $({T}_{i,1}, {T}_{i,2}, \ldots, {T}_{i,n})$ approaches $a_i \vec{b}^T$. In fact, the ratio of $({T}_{i,1}, {T}_{i,2}, {T}_{i,3})$ of the $i$-th ($i \in \{1,2,3\}$) row of \mat{T} is close to $6:3:1$ at $k=\{140,1000\}$ in (b). 

Here, recalling that the projection operator as in (\ref{eq:barycentric_projection}), we understand that the new centroids $\hat{\vec{x}}_i$ of the $i$-th image depends on only $T_{i,j} (\forall j \in [n])$, more precisely, the {\it ratio of $T_{i,j}$}. It should be also noted that the centroid of the source image does {\it not} have any impact on creating $\hat{\vec{x}}_i$. Therefore, because the ratios of all the row vectors get similar each other, the obtained new centroid $\hat{\vec{x}}_i (i \in [n])$ mostly resemble across all $i \in [n]$. This can be verified that the images at $k=140$ in figure (a) consist of a (nearly) single (brown) color. Indeed, this brown color is the $\vec{b}$-weighted mixed color of the three reference colors in Figure \ref{fig:CTOriSyntheImages}(b). Furthermore, figure (c) shows the heat-map of a {\it row-wise normalized} transport matrix, of which each row is normalized such that its sum is equal to 1. As seen in (c), all the rows at each column at $k=140$ have mostly the same colors with similar values, and these colors are nearly identical to those of \vec{b}. 

However, when the iteration proceeds and when $\frac{1}{2\lambda}\|\mat{T}\vec{1}_n - \vec{a}\|_2^2$ relatively decreases compared to the first term, the first term $\langle \mat{T},\mat{C} \rangle$ is gaining its influence on the optimization process. Then, the process proceeds towards the similar result of linear programming (LP). This can be verified the image at $k=5\times 10^4$ in (a). 

\vspace*{0.2cm}
\noindent
{\bf Large relaxation parameter ($\lambda = 10^{-3}$) in Figure \ref{fig:CTImages_lambda3}.} This case starts to reduce the first and second term simultaneously from the beginning as shown in (d) and (e), and proceeds directly towards a close solution of LP. From (d), the objective function value and the norm of gradient for the second term go down quickly, and the values of the first term get dominant after only a few iterations. We see that the image at $k=1000$ is quite similar to that of LP in Figure \ref{fig:CTOriSyntheImages}(c). Also, this result reveals that its convergence is much faster than the case with $\lambda = 10^{-6}$. This is due to that the BCFW algorithm requires a large number of iterations when small $\lambda$ because of the large penalty term. 

\subsubsection{Real-world image data}

\begin{figure}[htbp]
\begin{center}
	\begin{minipage}[t]{0.32\textwidth}
	\begin{center}
		\includegraphics[width=\textwidth]{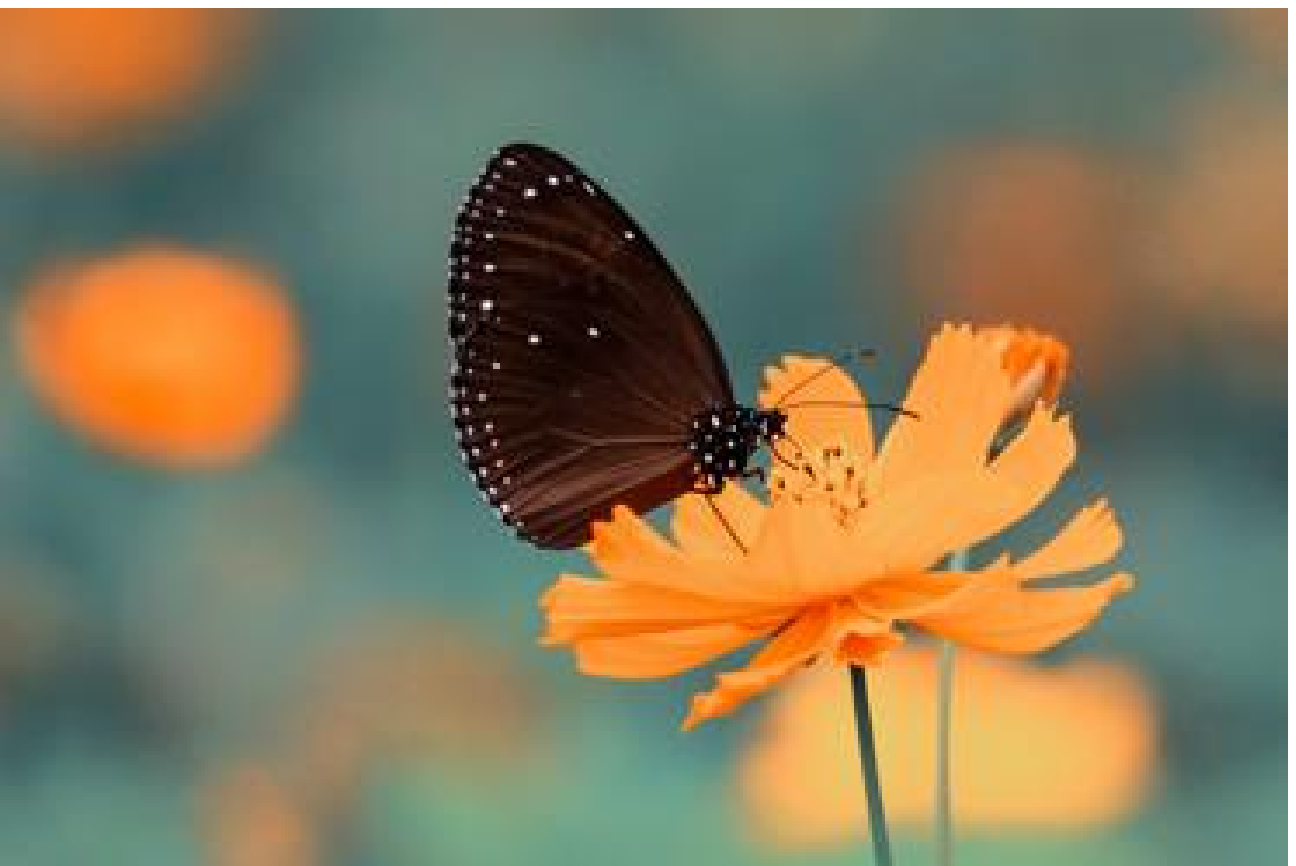}\\
		\vspace*{-0.1cm}
		
		{\footnotesize  (a) source}
		
	\end{center} 
	\end{minipage}
	\begin{minipage}[t]{0.32\textwidth}
	\begin{center}
		\includegraphics[width=\textwidth]{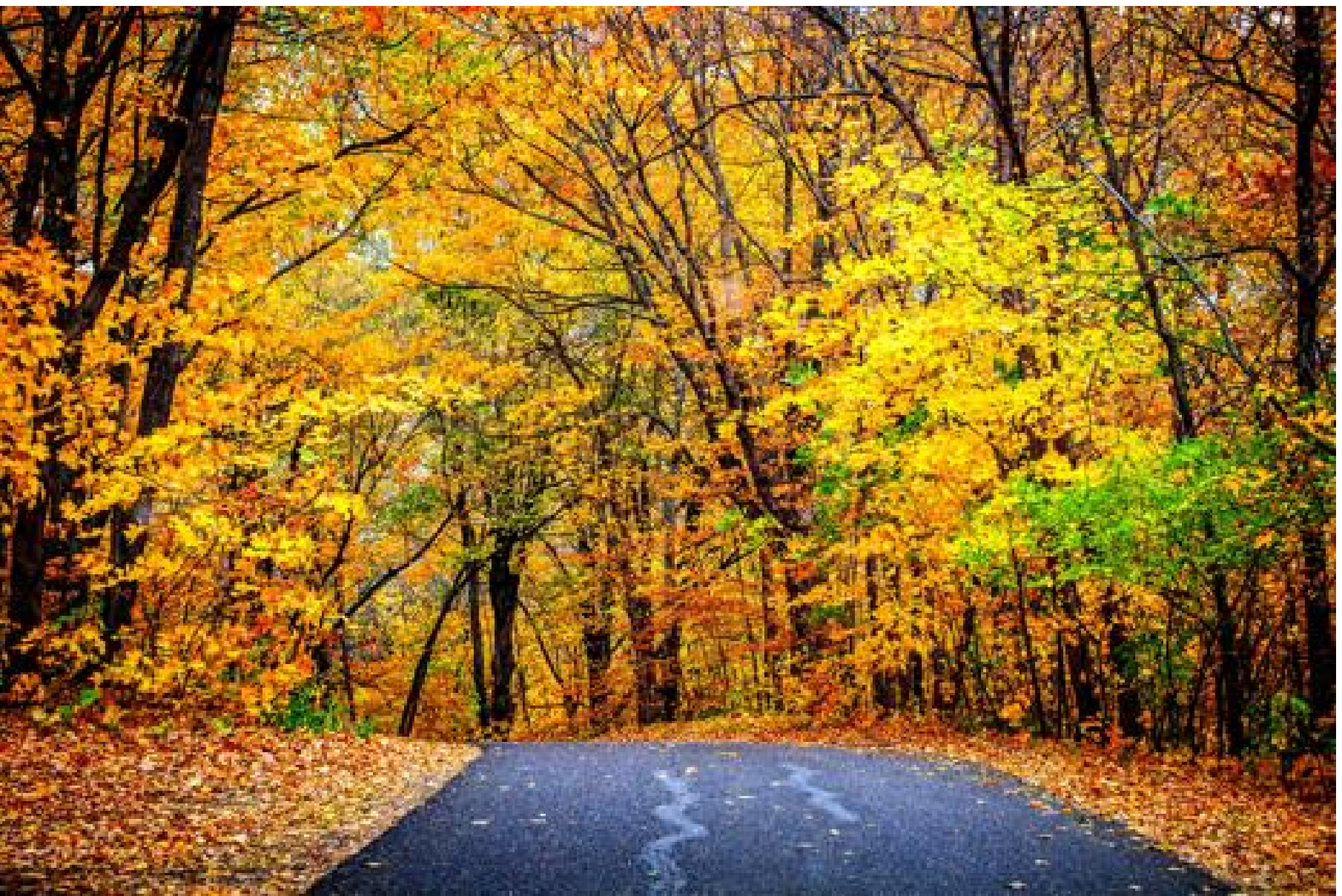}\\
		\vspace*{-0.1cm}
				
		{\footnotesize  (b) reference}
		
	\end{center} 
	\end{minipage}
	\begin{minipage}[t]{0.32\textwidth}
	\begin{center}
		\includegraphics[width=\textwidth]{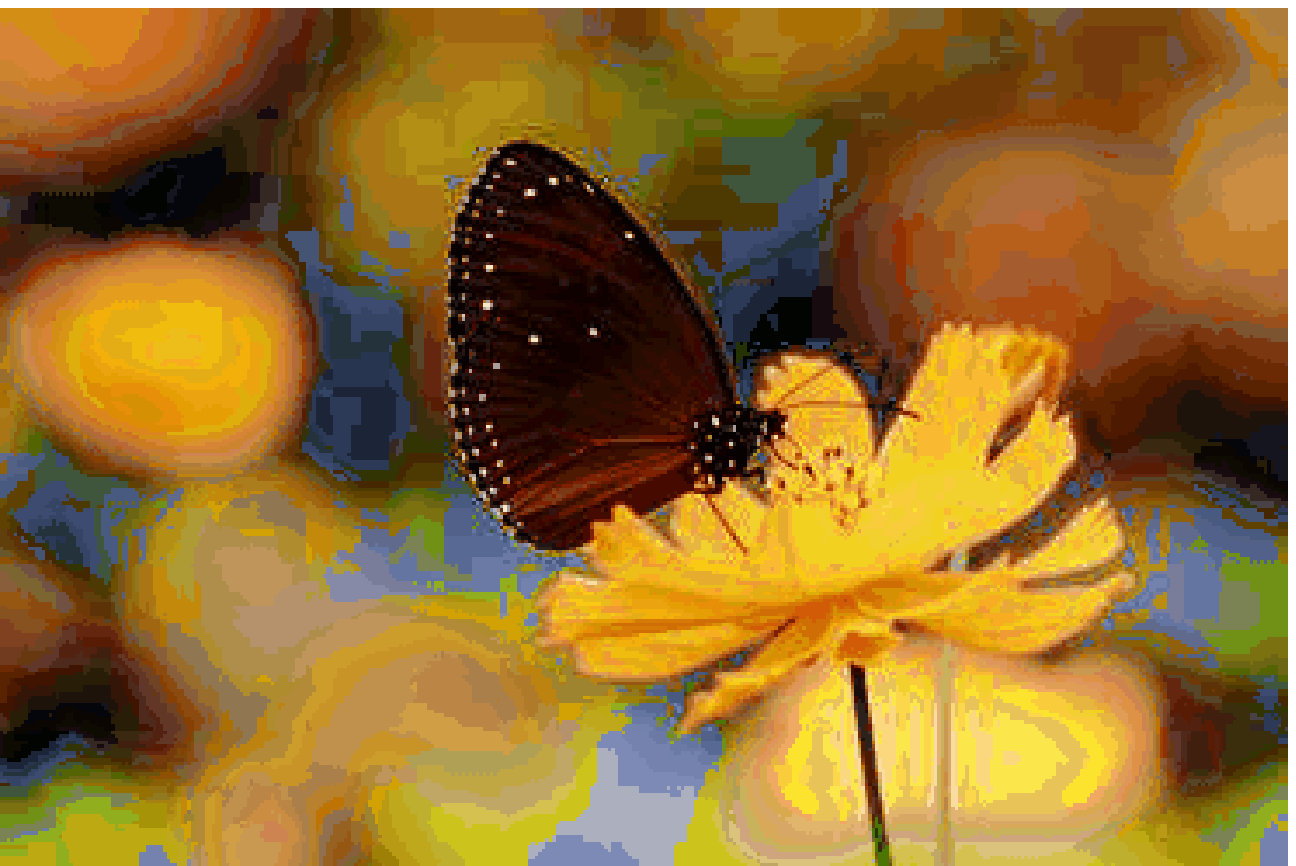}\\
		\vspace*{-0.1cm}
				
		{\footnotesize  (c) color-transferred image by LP}
		
	\end{center} 
	\end{minipage}
\caption{Source and reference real-world images, and color-transferred image by LP.}
\vspace*{0.4cm}

\label{fig:CTRealworldImages}
\end{center}
\end{figure}

We continue to discuss the impact of $\lambda$ on the color-transferred image using real-world images. The two images used here are shown in Figures \ref{fig:CTRealworldImages}(a) and (b), respectively. The color-transferred image generated by linear programming (LP) for the original OT problem is also shown in (c). 

\begin{figure}[htbp]
\begin{center}
\begin{minipage}[t]{0.24\textwidth}
	\begin{center}
		\includegraphics[width=\textwidth]{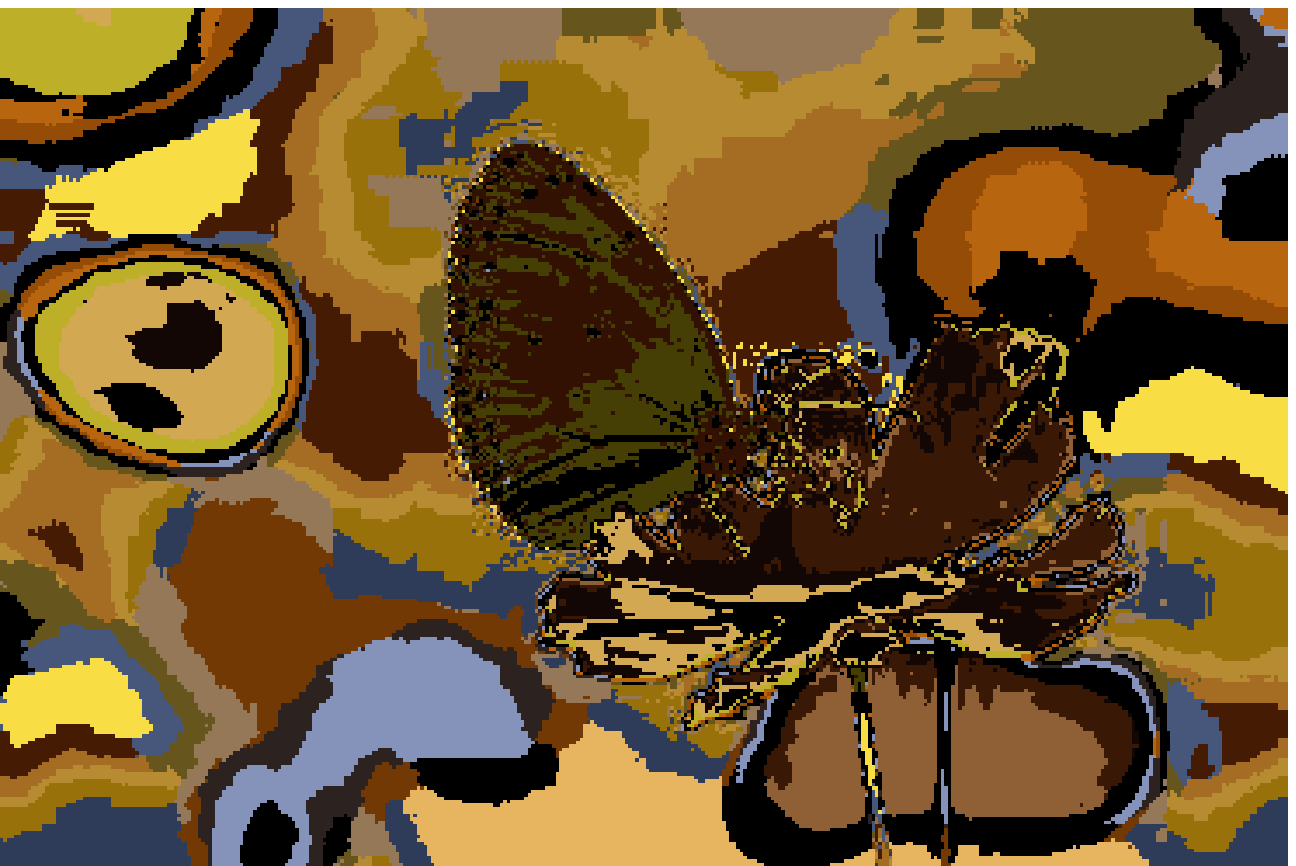}\\
				
		{\footnotesize  $k = 1$}			
	\end{center} 
	\end{minipage}
	\begin{minipage}[t]{0.24\textwidth}
	\begin{center}
		\includegraphics[width=\textwidth]{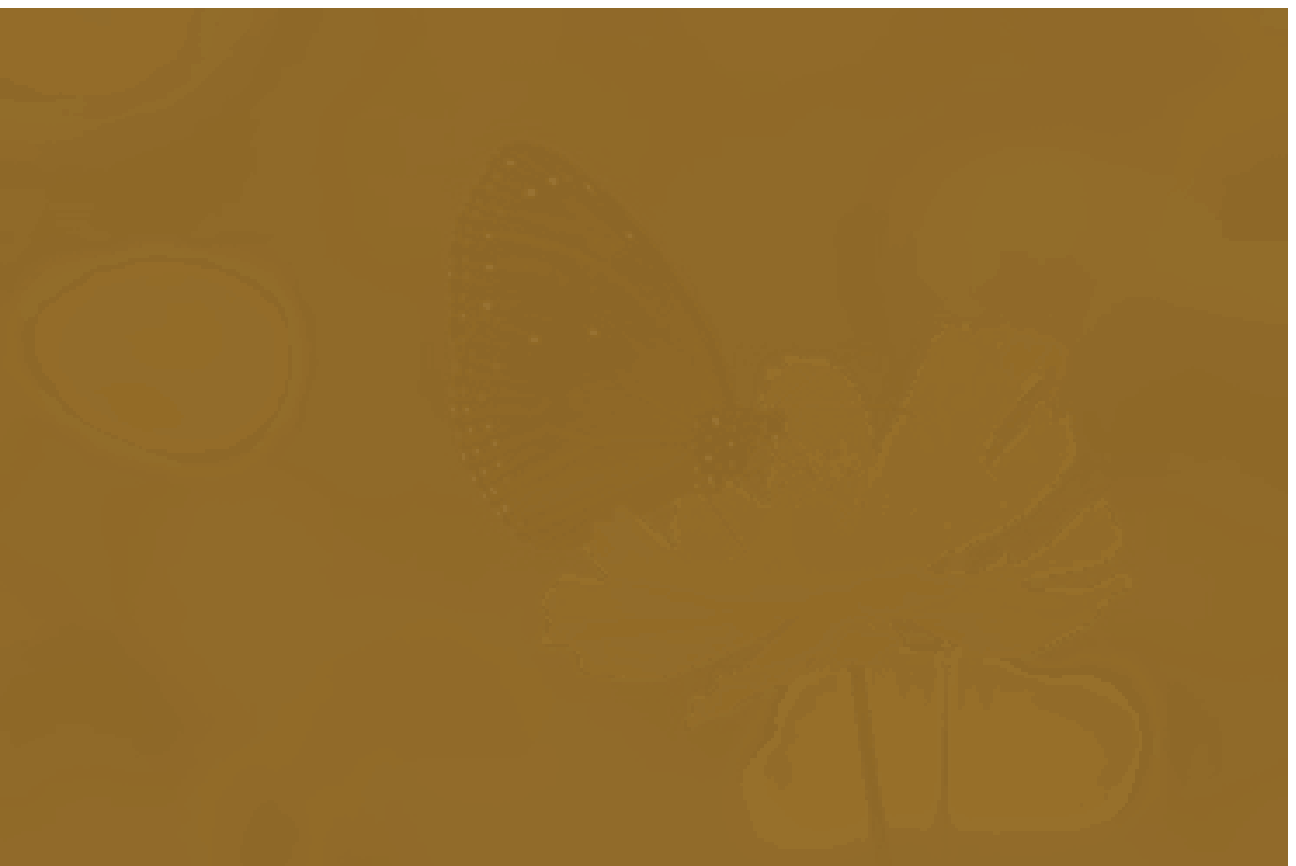}\\
				
		{\footnotesize  $k = 3300$}		
	\end{center} 
	\end{minipage}	
	\begin{minipage}[t]{0.24\textwidth}
	\begin{center}
		\includegraphics[width=\textwidth]{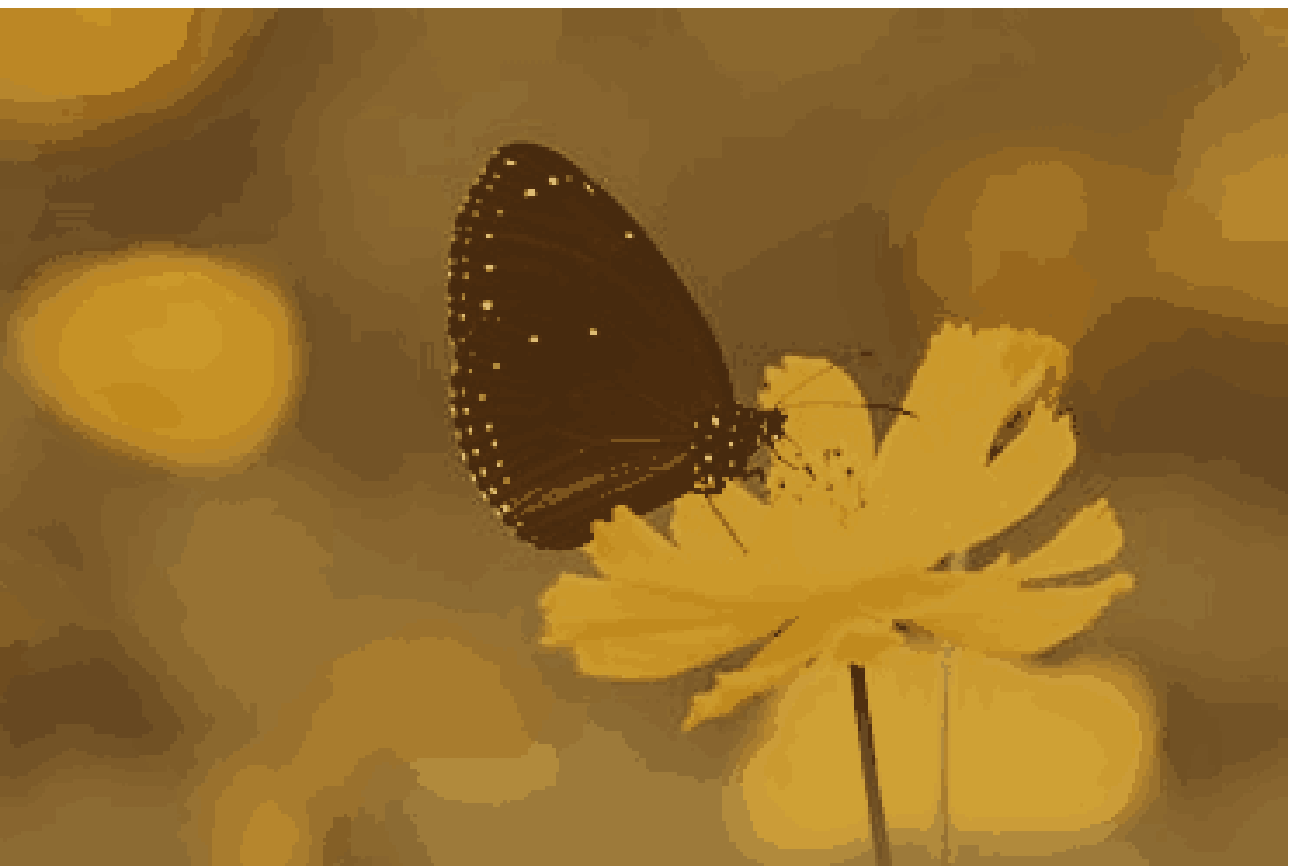}\\
				
		{\footnotesize  $k = 10^5$}				
	\end{center} 
	\end{minipage}	
	\begin{minipage}[t]{0.24\textwidth}
	\begin{center}
		\includegraphics[width=\textwidth]{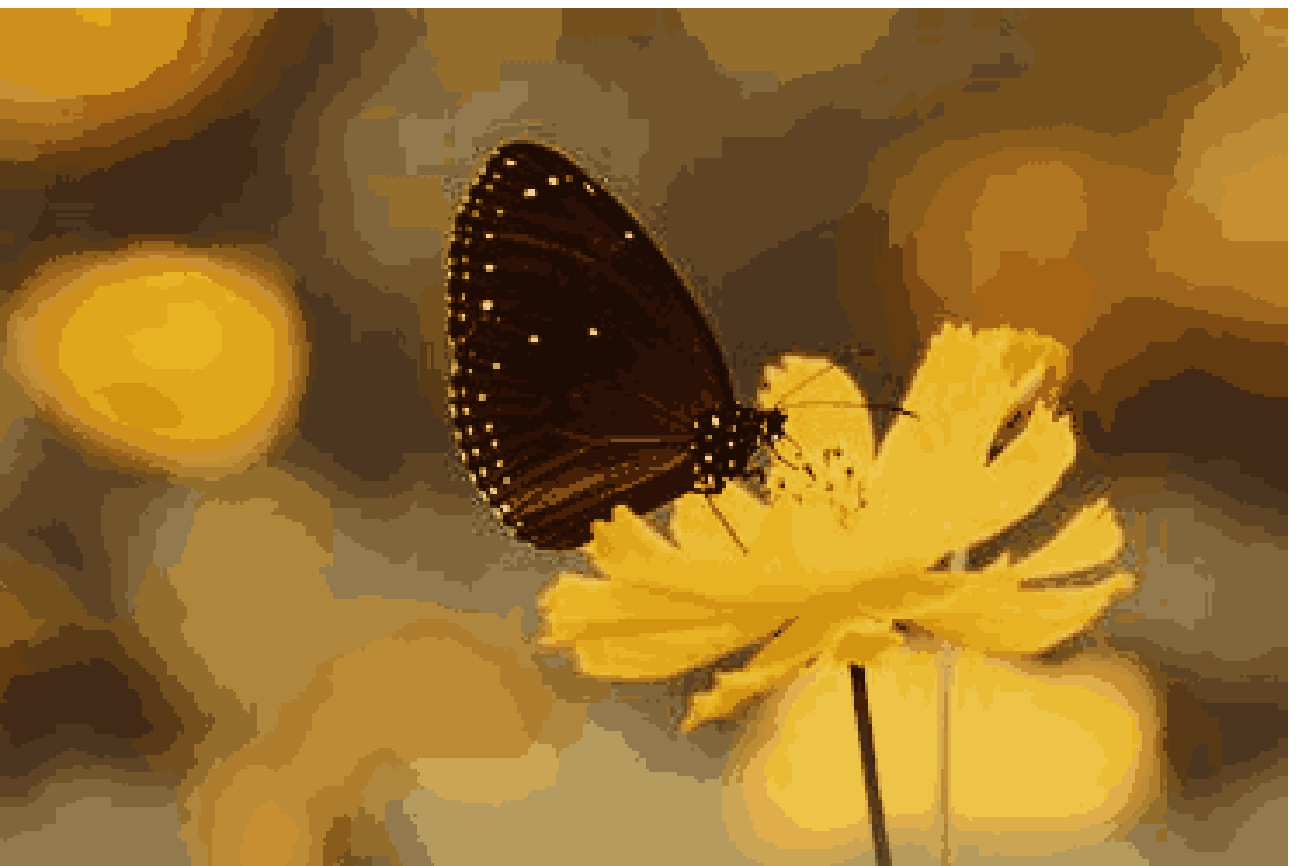}\\
					
		{\footnotesize  $k = 10^6$}		
				
	\end{center} 
	\end{minipage}	
	\vspace*{0.2cm}
	
	{\small  (a) transition of color-transferred images ($\lambda=10^{-9}$)}			
	\vspace*{-0.2cm}
	
	\begin{minipage}[t]{0.24\textwidth}
	\begin{center}
		\hspace*{-0.15cm}\includegraphics[width=\textwidth]{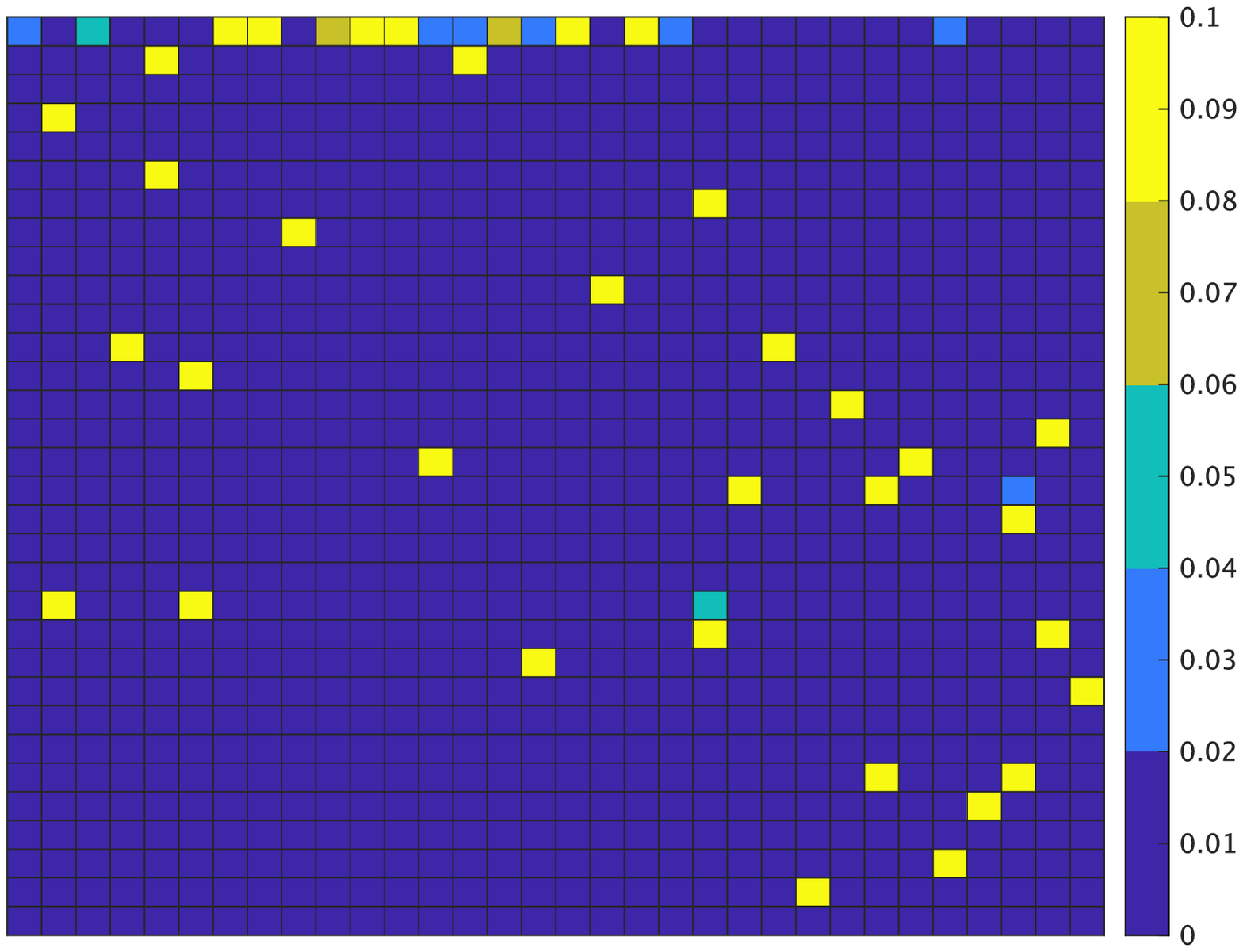}\\
		\vspace*{0.1cm}
		
		{\footnotesize  $k = 1$}
		
	\end{center} 
	\end{minipage}
	\begin{minipage}[t]{0.24\textwidth}
	\begin{center}
		\hspace*{-0.15cm}\includegraphics[width=\textwidth]{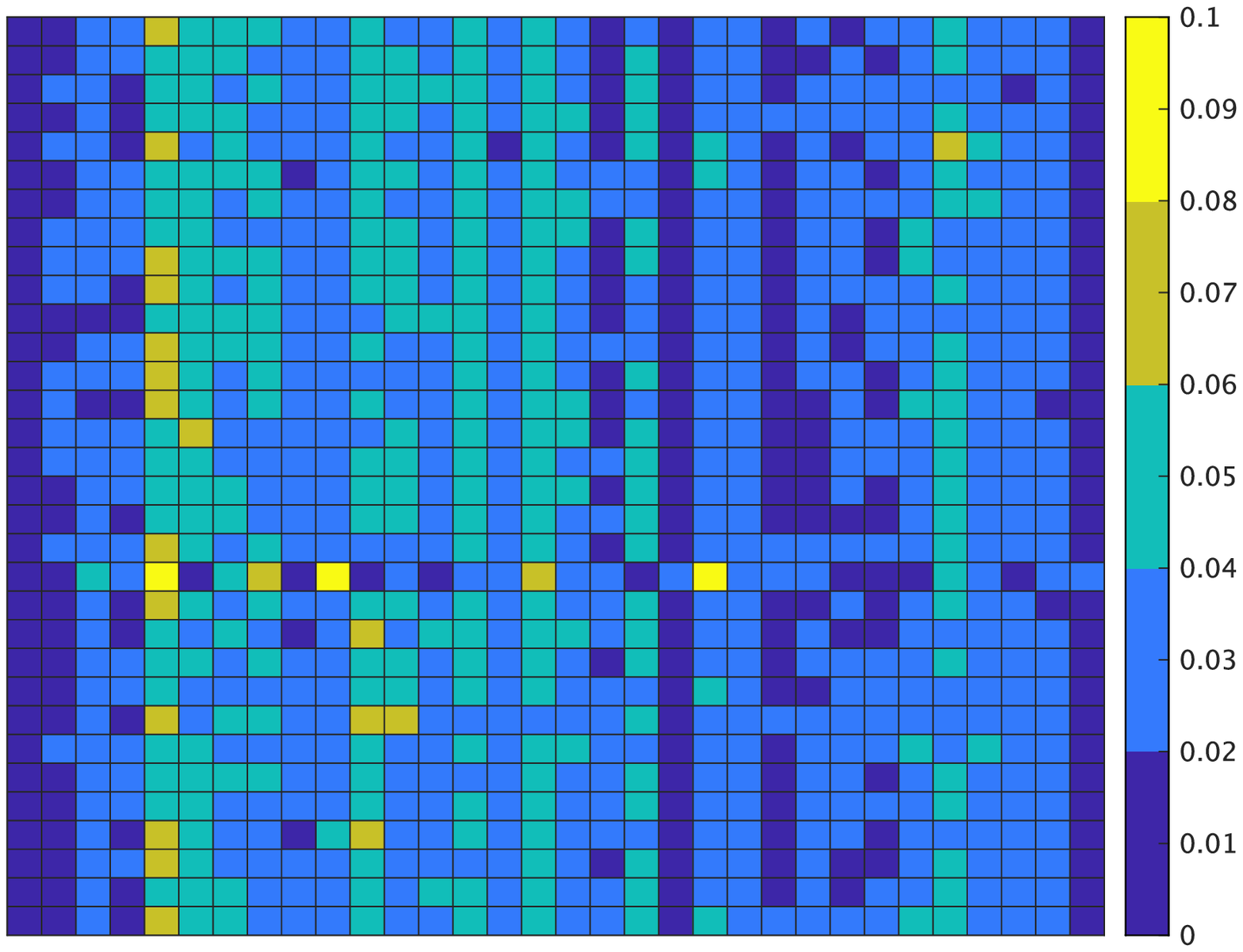}\\
		\vspace*{0.1cm}
				
		{\footnotesize  $k = 3000$}
		
	\end{center} 
	\end{minipage}
	\begin{minipage}[t]{0.24\textwidth}
	\begin{center}
		\hspace*{-0.15cm}\includegraphics[width=\textwidth]{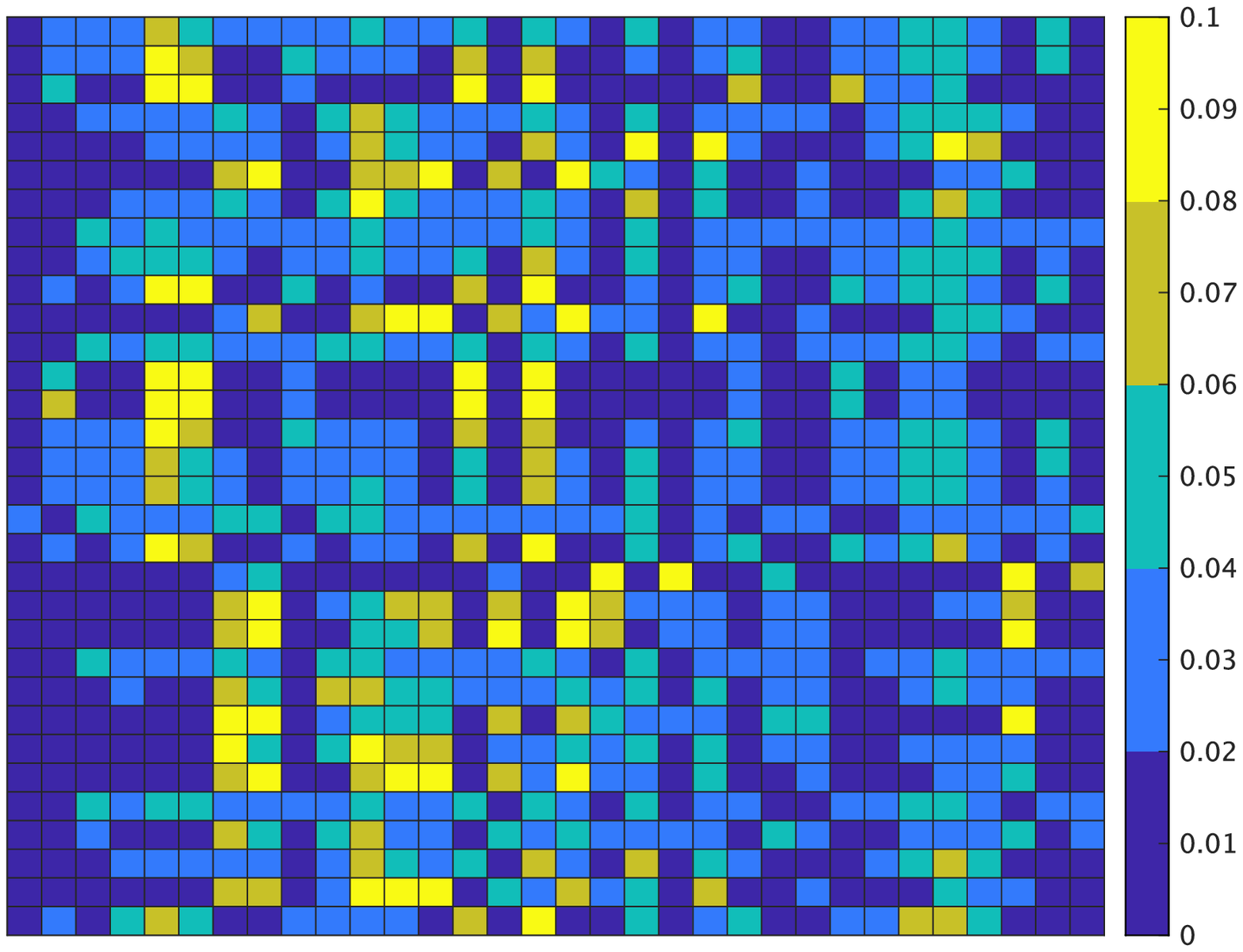}\\
		\vspace*{0.1cm}
				
		{\footnotesize  $k = 10^5$}
		
	\end{center} 
	\end{minipage}	
	\begin{minipage}[t]{0.24\textwidth}
	\begin{center}
		\hspace*{-0.15cm}\includegraphics[width=\textwidth]{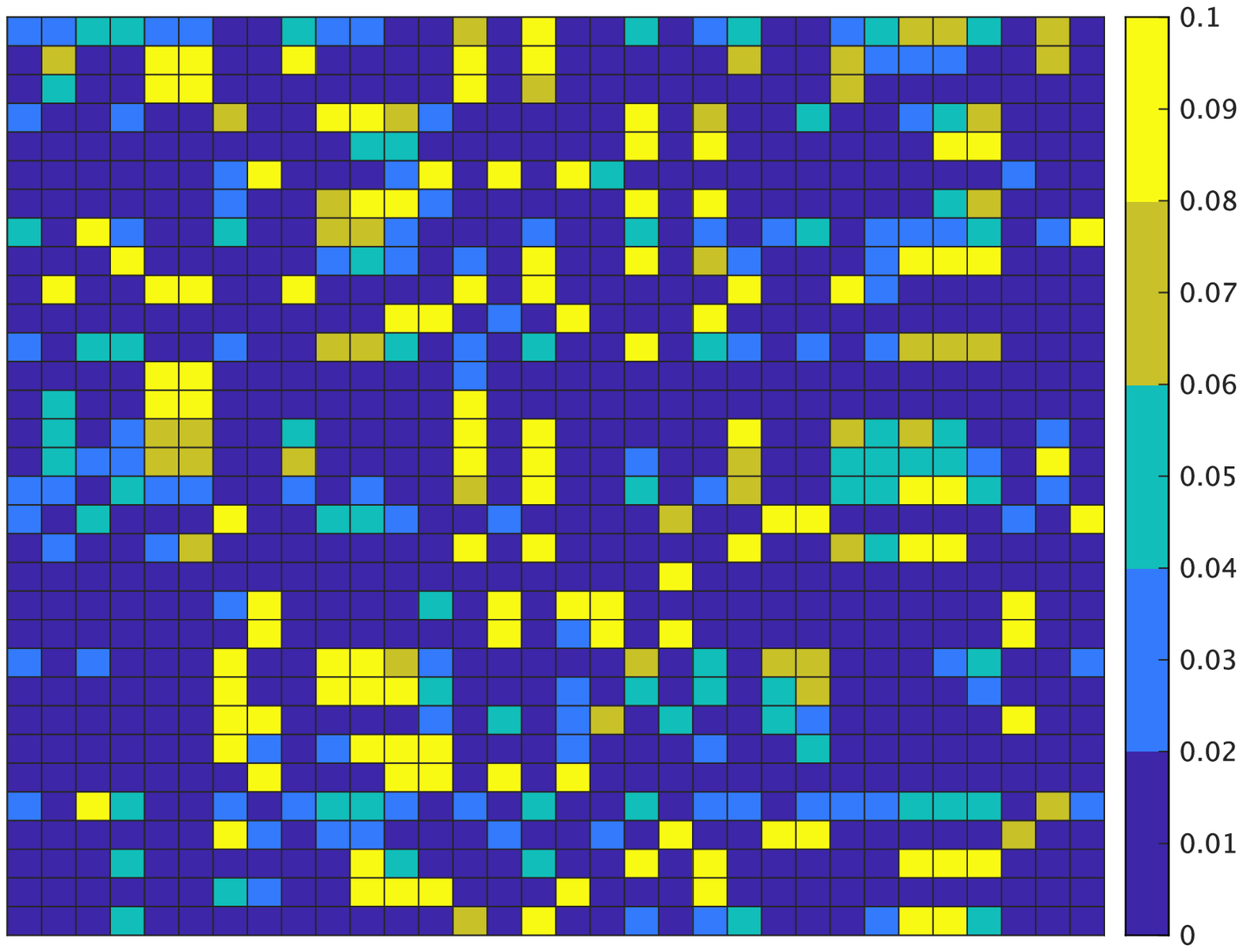}\\
		\vspace*{0.1cm}
				
		{\footnotesize  $k = 10^6$}		
	\end{center} 
	\end{minipage}
	\vspace*{0.2cm}
	
	{\small  (b) transition of heat-map of row-wise normalized transport matrices $\mat{T}^{(k)}$ ($\lambda=10^{-9}$)}
	
	\vspace*{1.5cm}
	
	\begin{minipage}[t]{0.24\textwidth}
	\begin{center}
		\includegraphics[width=\textwidth]{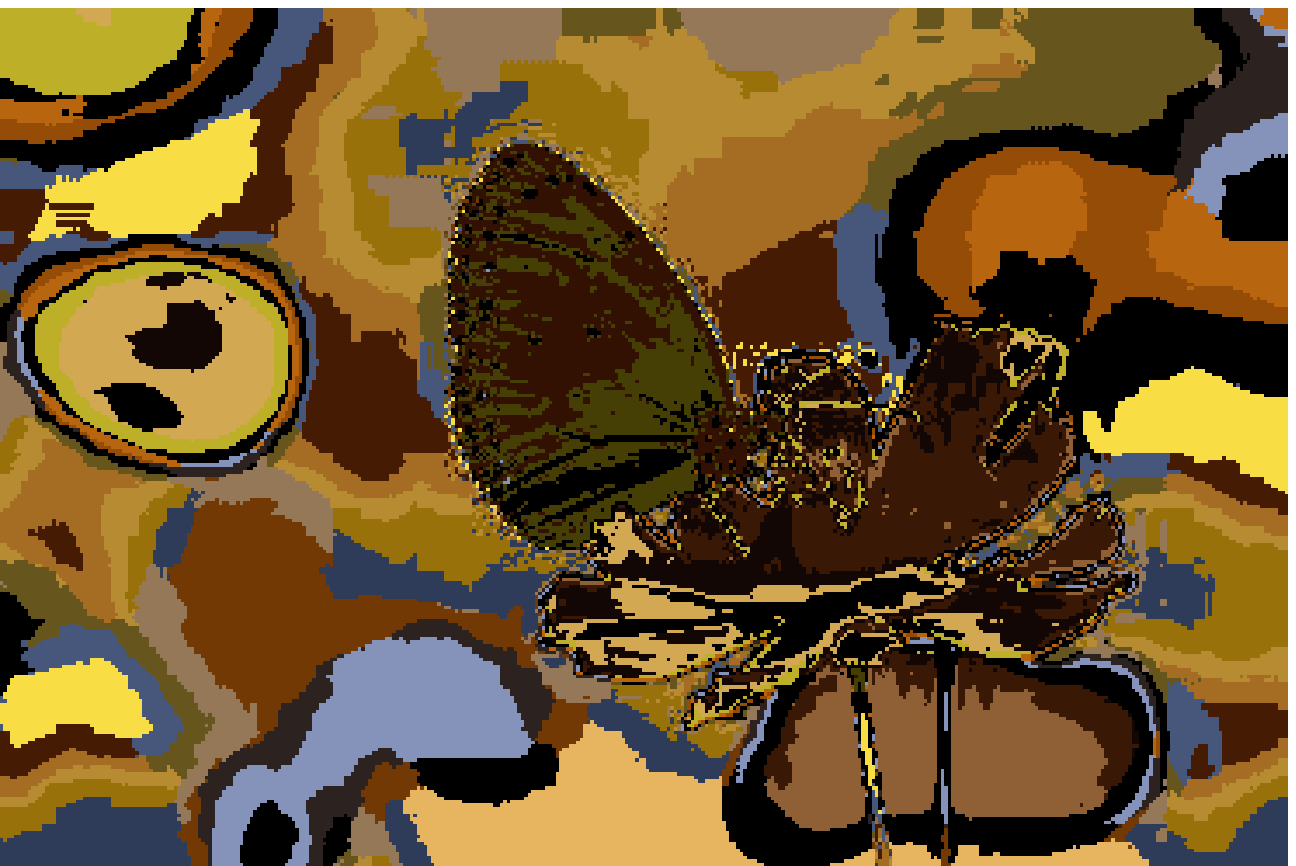}\\
				
		{\footnotesize  $k = 1$}		
	\end{center} 
	\end{minipage}
	\begin{minipage}[t]{0.24\textwidth}
	\begin{center}
		\includegraphics[width=\textwidth]{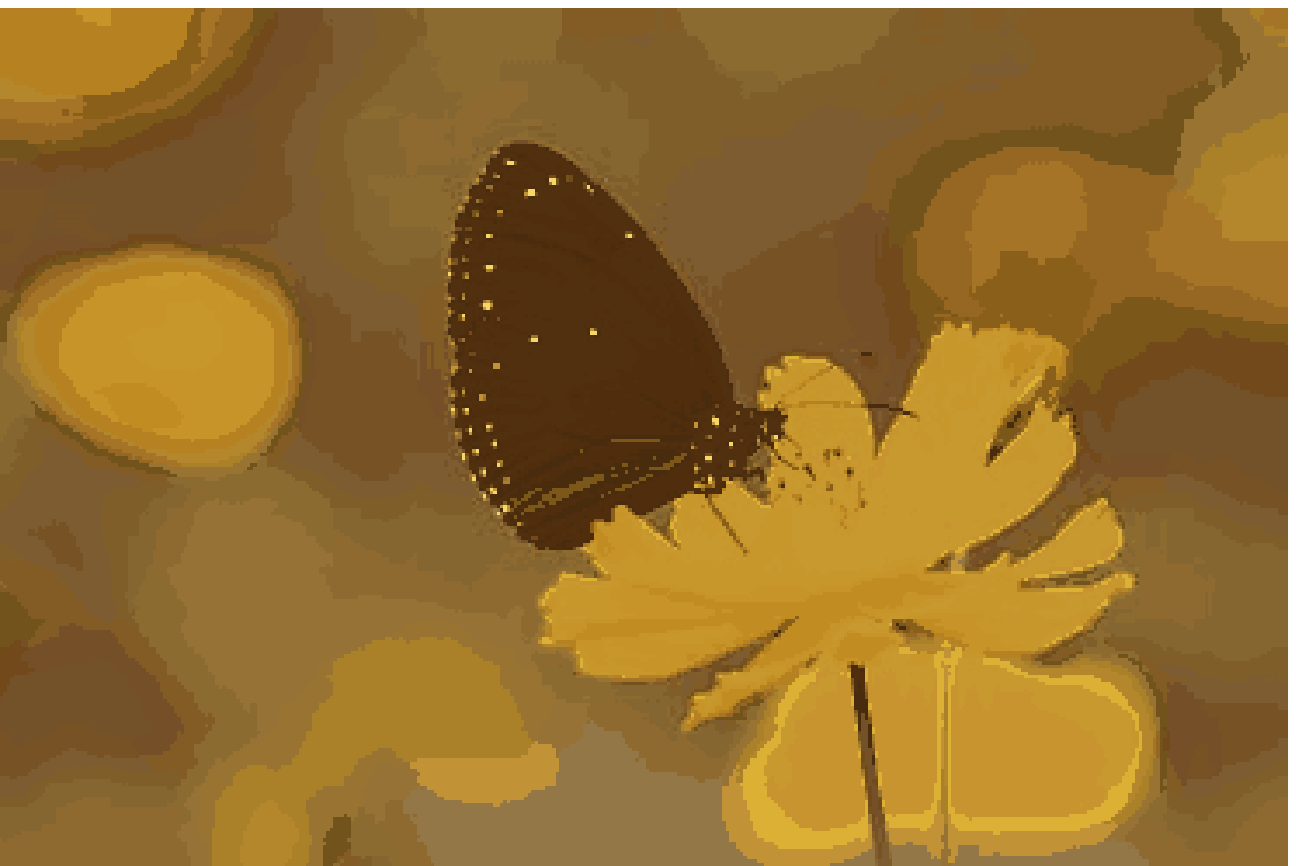}\\
				
		{\footnotesize  $k = 100$}		
	\end{center} 
	\end{minipage}	
	\begin{minipage}[t]{0.24\textwidth}
	\begin{center}
		\includegraphics[width=\textwidth]{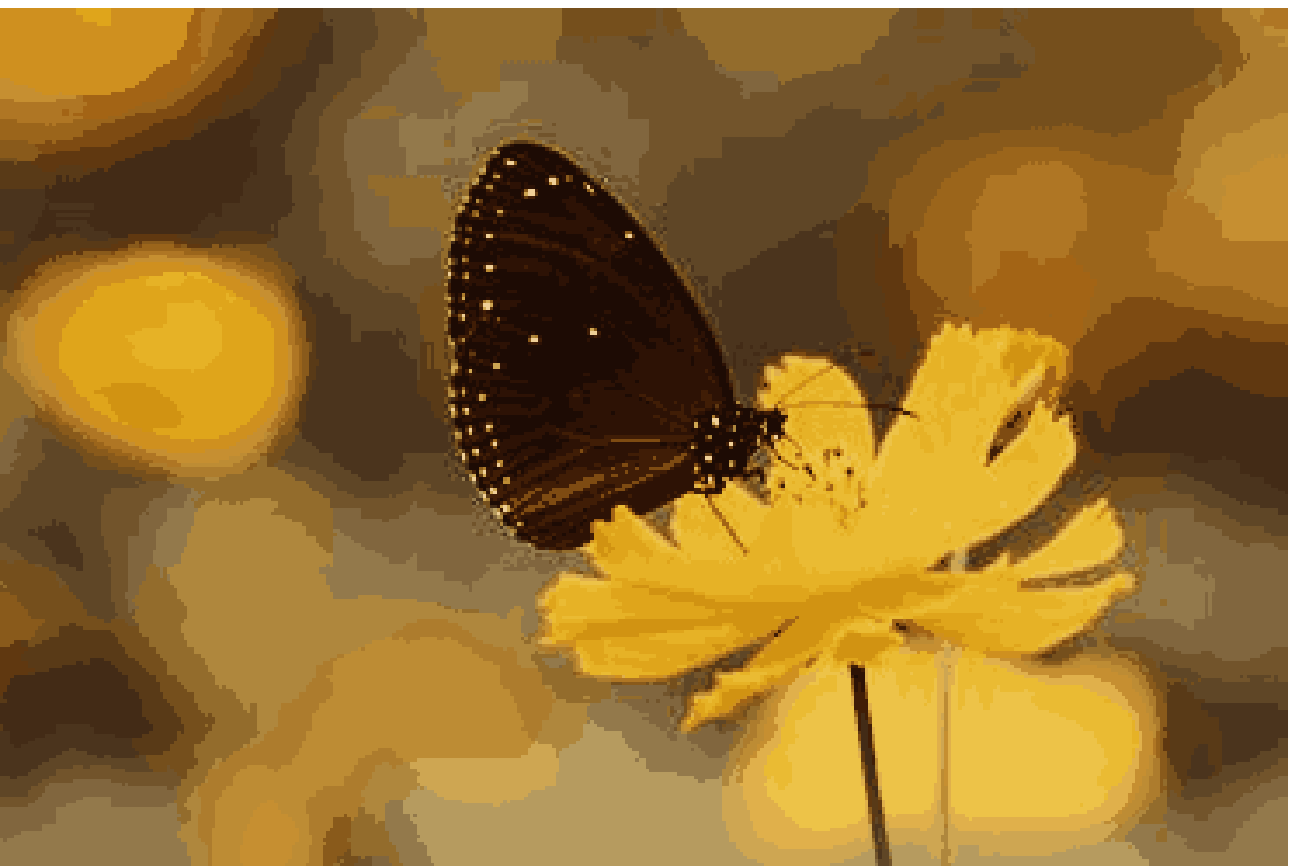}\\
				
		{\footnotesize  $k = 10^3$}			
	\end{center} 
	\end{minipage}	
	\begin{minipage}[t]{0.24\textwidth}
	\begin{center}
		\includegraphics[width=\textwidth]{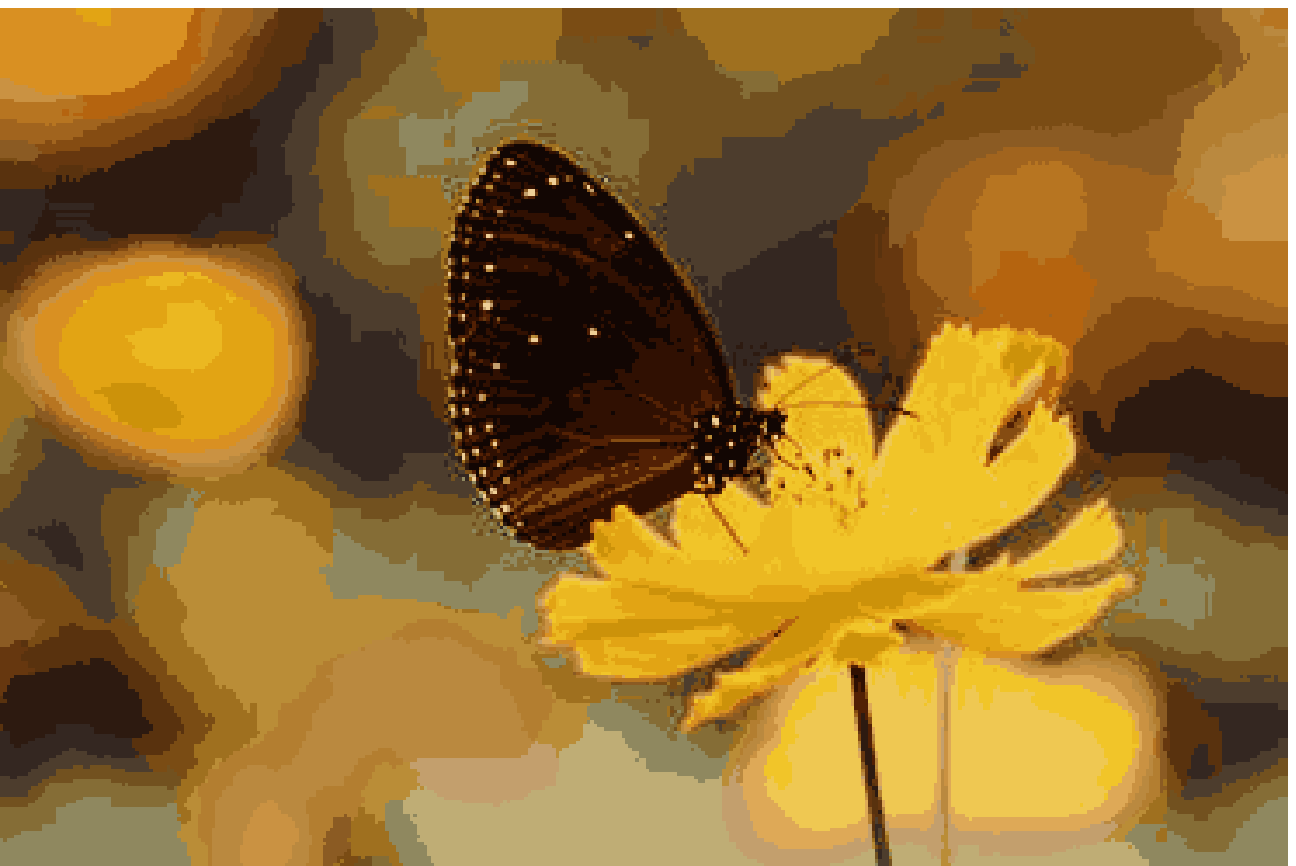}\\
				
		{\footnotesize  $k = 10^4$}			
	\end{center} 
	\end{minipage}	
	\vspace*{0.2cm}
	
	{\small  (c) transition of color-transferred images ($\lambda=10^{-6}$)}		
	\vspace*{-0.2cm}
	
	\begin{minipage}[t]{0.24\textwidth}
	\begin{center}
		\hspace*{-0.15cm}\includegraphics[width=\textwidth]{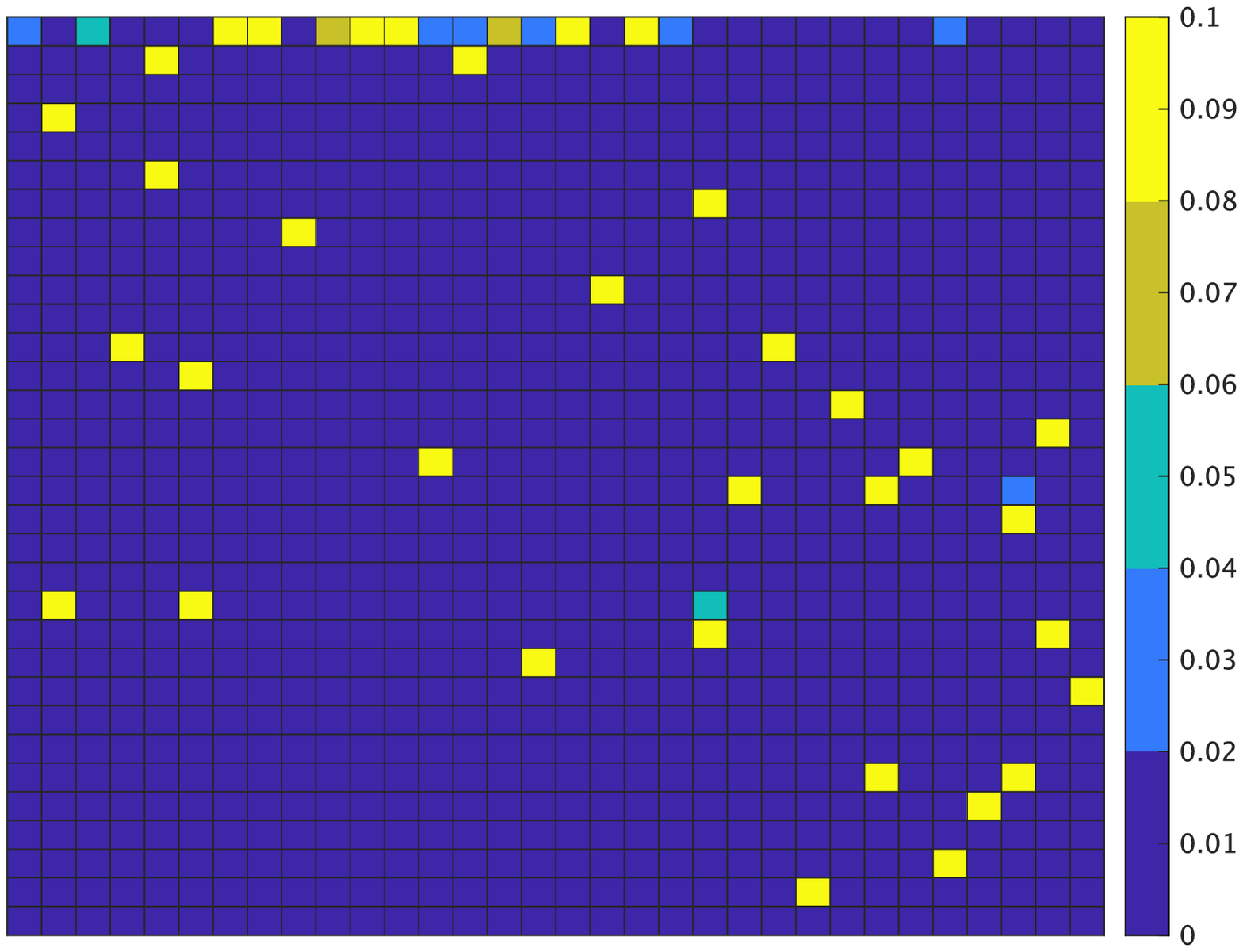}\\
		\vspace*{0.1cm}
				
		{\footnotesize  $k = 1$}
		
	\end{center} 
	\end{minipage}
	\begin{minipage}[t]{0.24\textwidth}
	\begin{center}
		\hspace*{-0.15cm}\includegraphics[width=\textwidth]{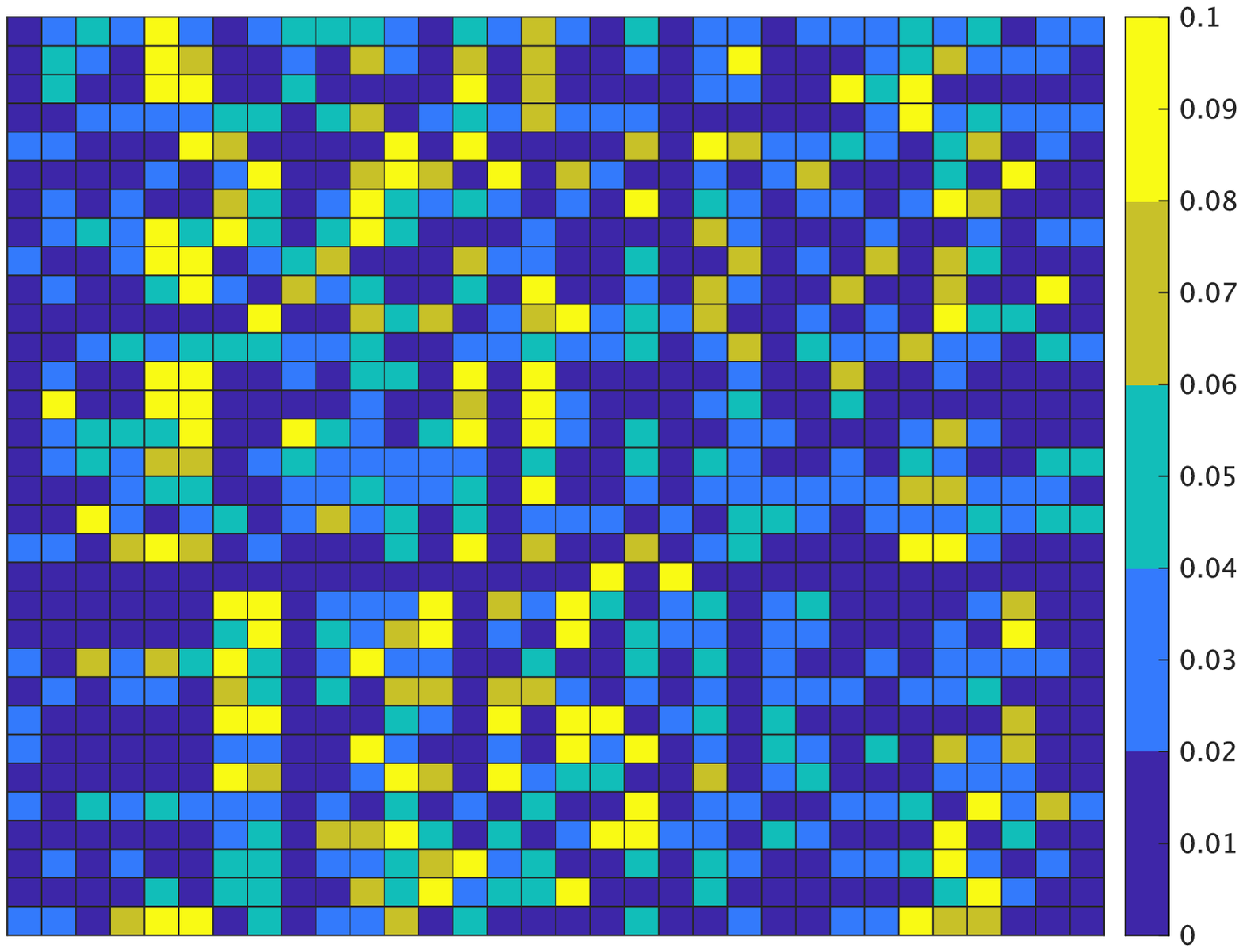}\\
		\vspace*{0.1cm}
				
		{\footnotesize  $k = 100$}
	\end{center} 
	\end{minipage}	
	\begin{minipage}[t]{0.24\textwidth}
	\begin{center}
		\hspace*{-0.15cm}\includegraphics[width=\textwidth]{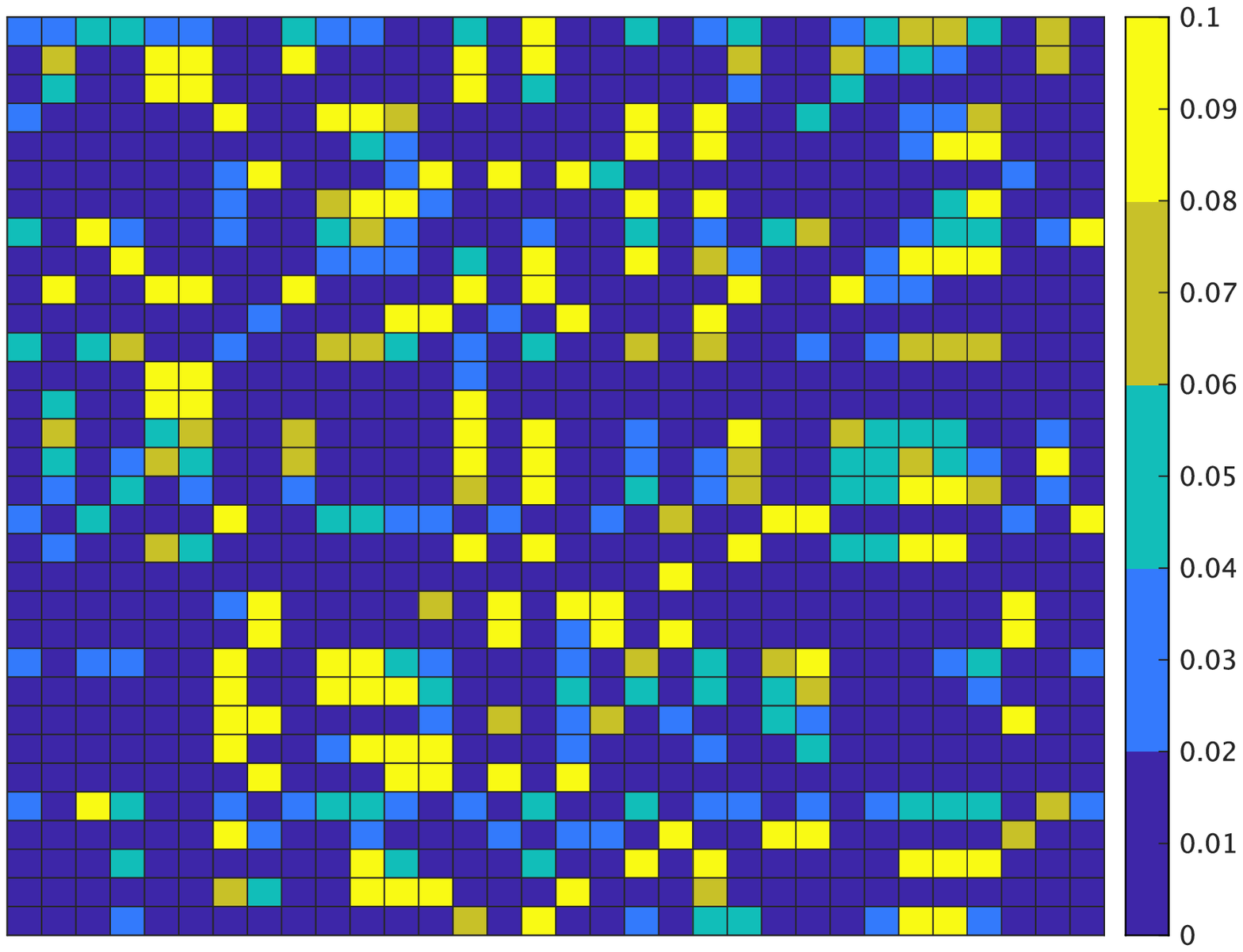}\\
		\vspace*{0.1cm}
				
		{\footnotesize  $k = 10^3$}
	\end{center} 
	\end{minipage}	
	\begin{minipage}[t]{0.24\textwidth}
	\begin{center}
		\hspace*{-0.15cm}\includegraphics[width=\textwidth]{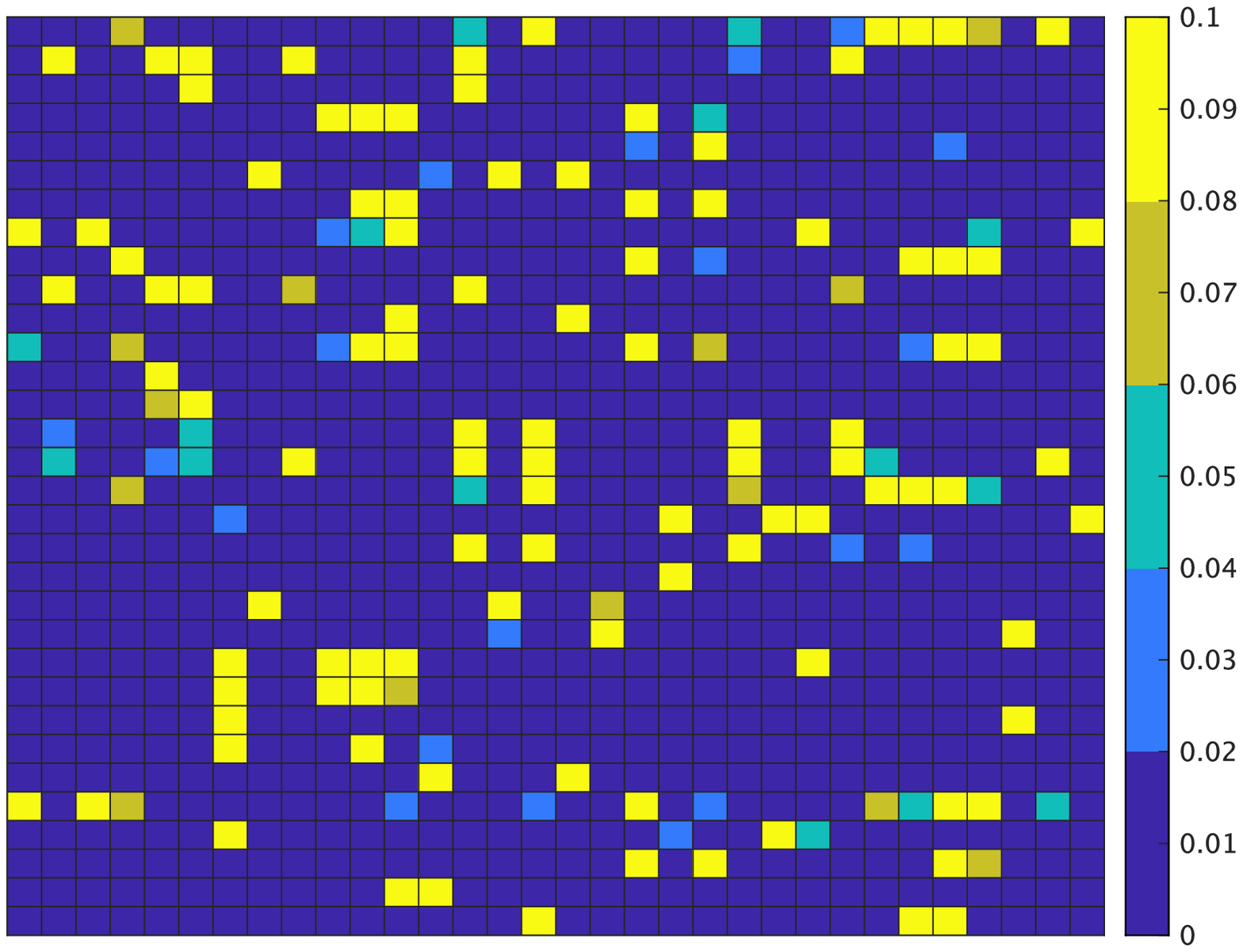}\\
		\vspace*{0.1cm}
				
		{\footnotesize  $k = 10^4$}		
	\end{center} 
	\end{minipage}
	\vspace*{0.2cm}
	
	{\small (d) transition of heat-map of row-wise normalized transport matrices  $\mat{T}^{(k)}$ ($\lambda=10^{-6}$)}
	
\centering
\caption{Transition of color-transferred images and heat-map of row-wise normalized transport matrices ($m=n=32$).}
\vspace*{0.2cm}
\label{fig:CTImages_N32}
\end{center}
\end{figure}

\begin{figure}[htbp]
\begin{center}	
	\begin{minipage}[t]{0.24\textwidth}
	\begin{center}
		\includegraphics[width=\textwidth]{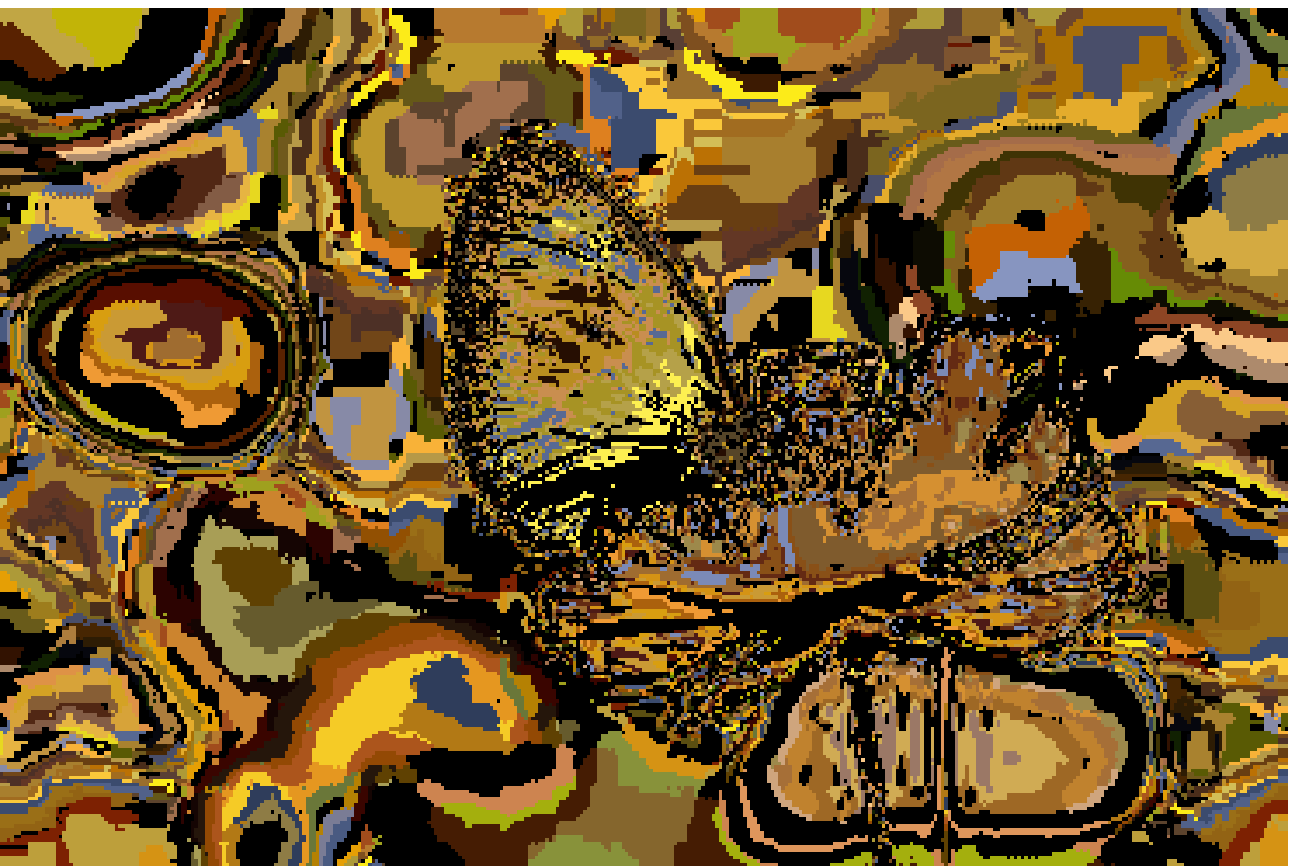}\\
		
		{\footnotesize  $k = 1$}
	\end{center} 
	\end{minipage}
	\begin{minipage}[t]{0.24\textwidth}
	\begin{center}
		\includegraphics[width=\textwidth]{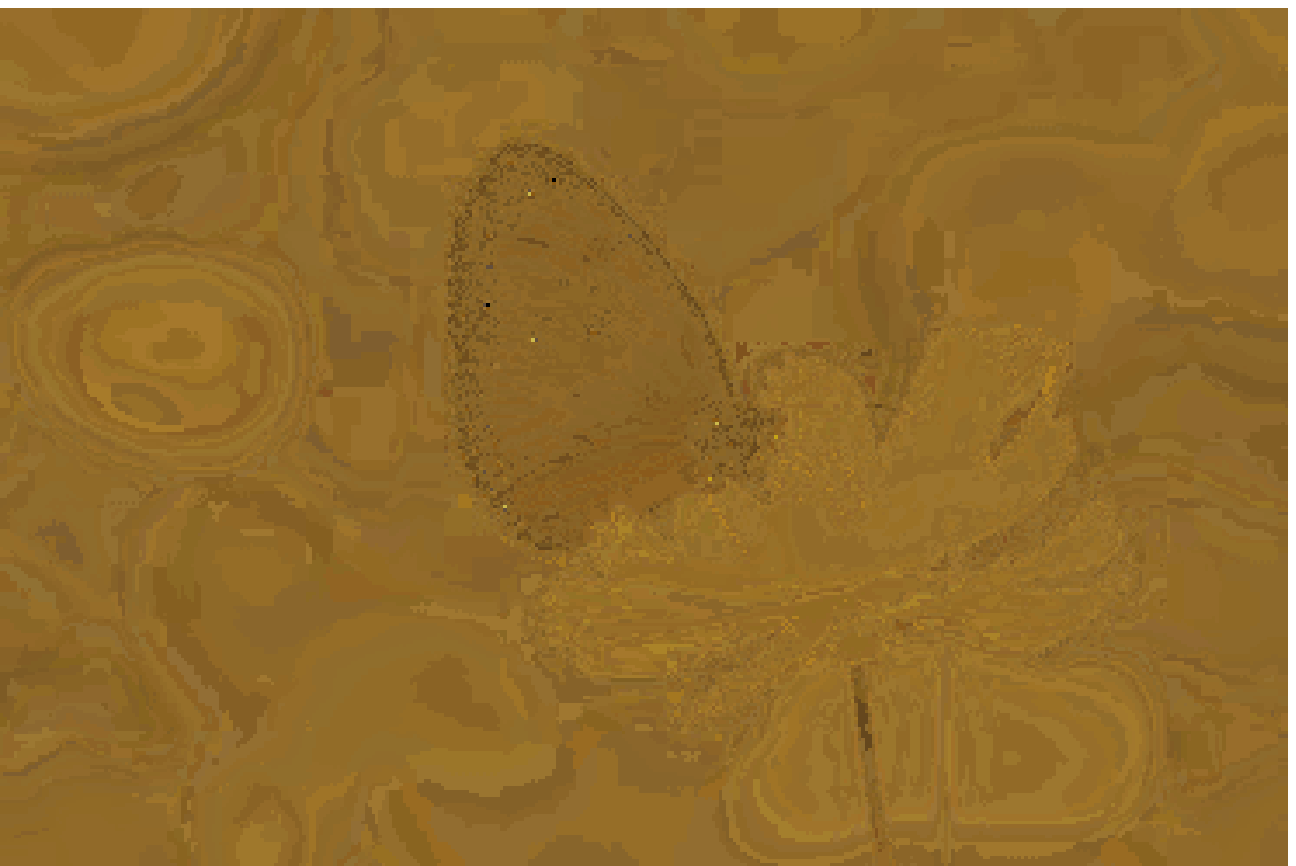}\\
		
		{\footnotesize  $k = 200$}		
	\end{center} 
	\end{minipage}	
	\begin{minipage}[t]{0.24\textwidth}
	\begin{center}
		\includegraphics[width=\textwidth]{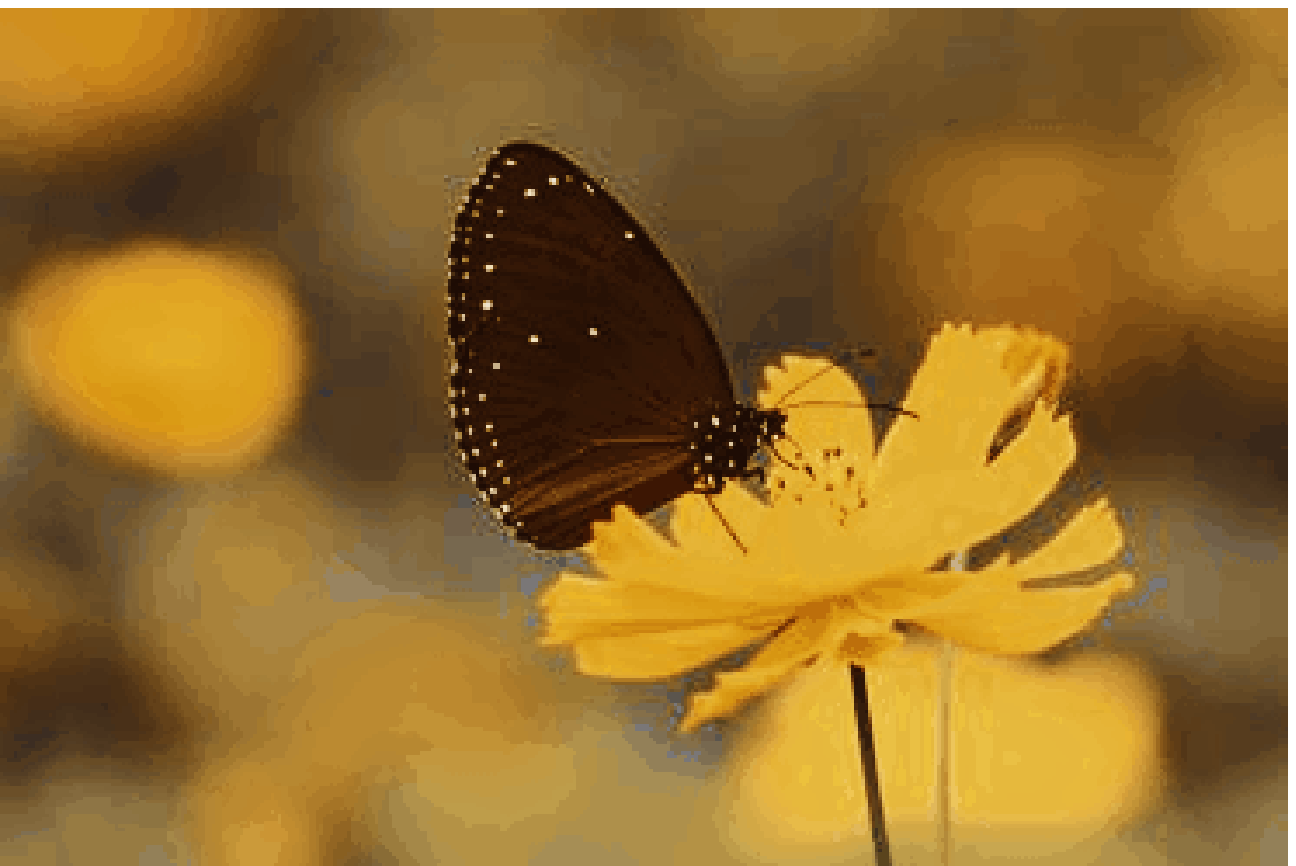}\\
		
		{\footnotesize  $k = 10^4$}			
	\end{center} 
	\end{minipage}	
	\begin{minipage}[t]{0.24\textwidth}
	\begin{center}
		\includegraphics[width=\textwidth]{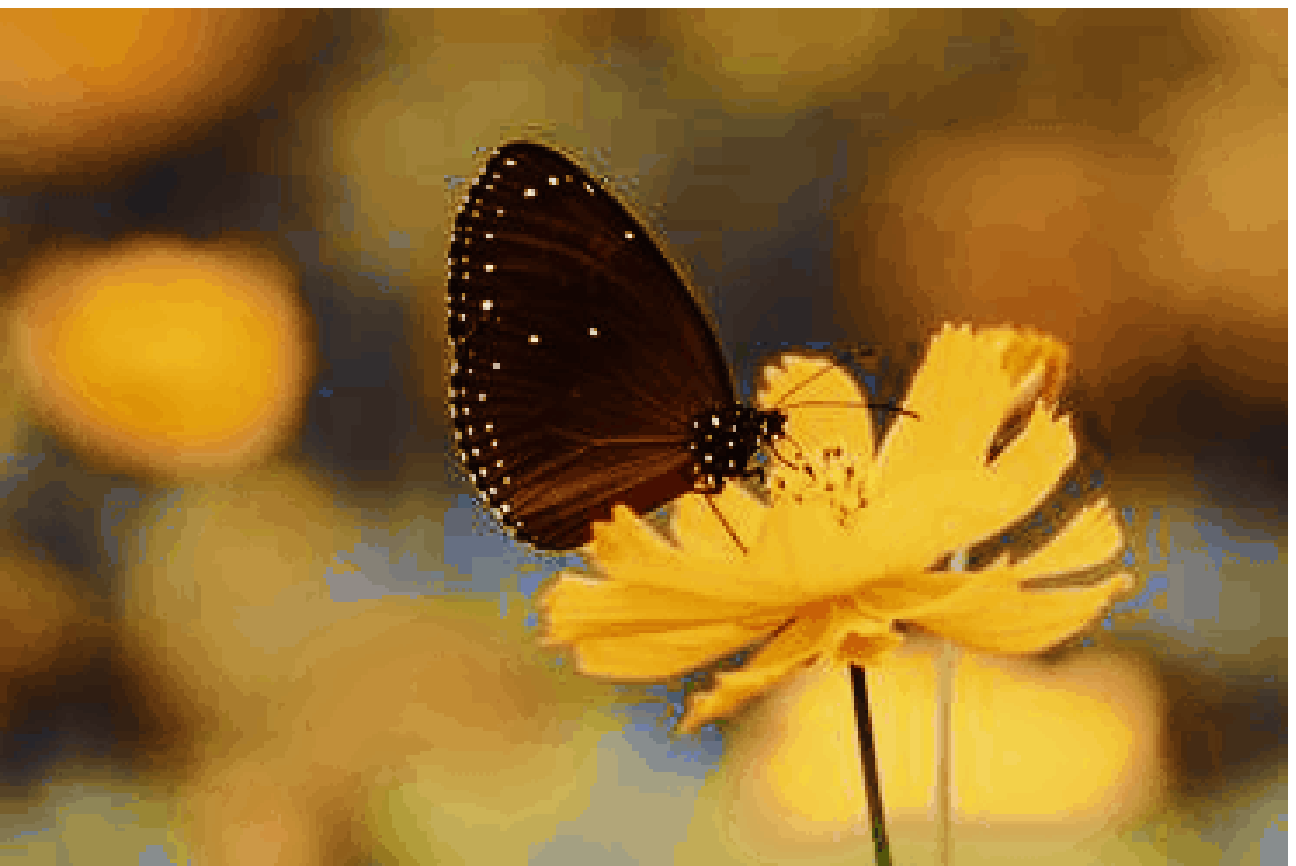}\\
		
		{\footnotesize  $k = 10^5$}			
	\end{center} 
	\end{minipage}
	\vspace*{0.2cm}	
	
	{\small  (a) transition of color-transferred images ($\lambda=10^{-9}$)}	
	\vspace*{0.5cm}
	
	\begin{minipage}[t]{0.33\textwidth}
	\begin{center}
		\includegraphics[width=\textwidth]{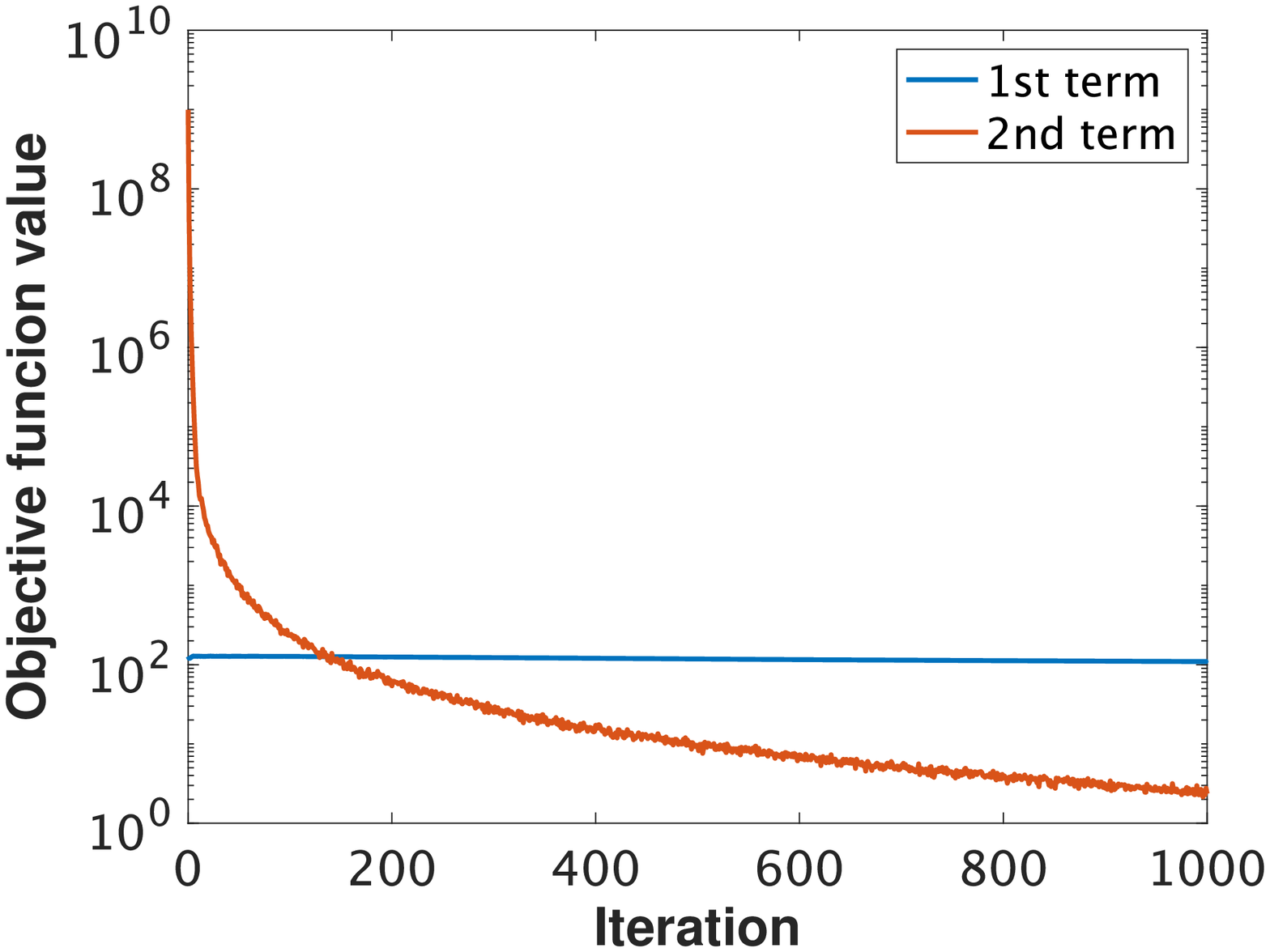}\\
	\end{center} 
	\end{minipage}
	\begin{minipage}[t]{0.33\textwidth}
	\begin{center}
		\includegraphics[width=\textwidth]{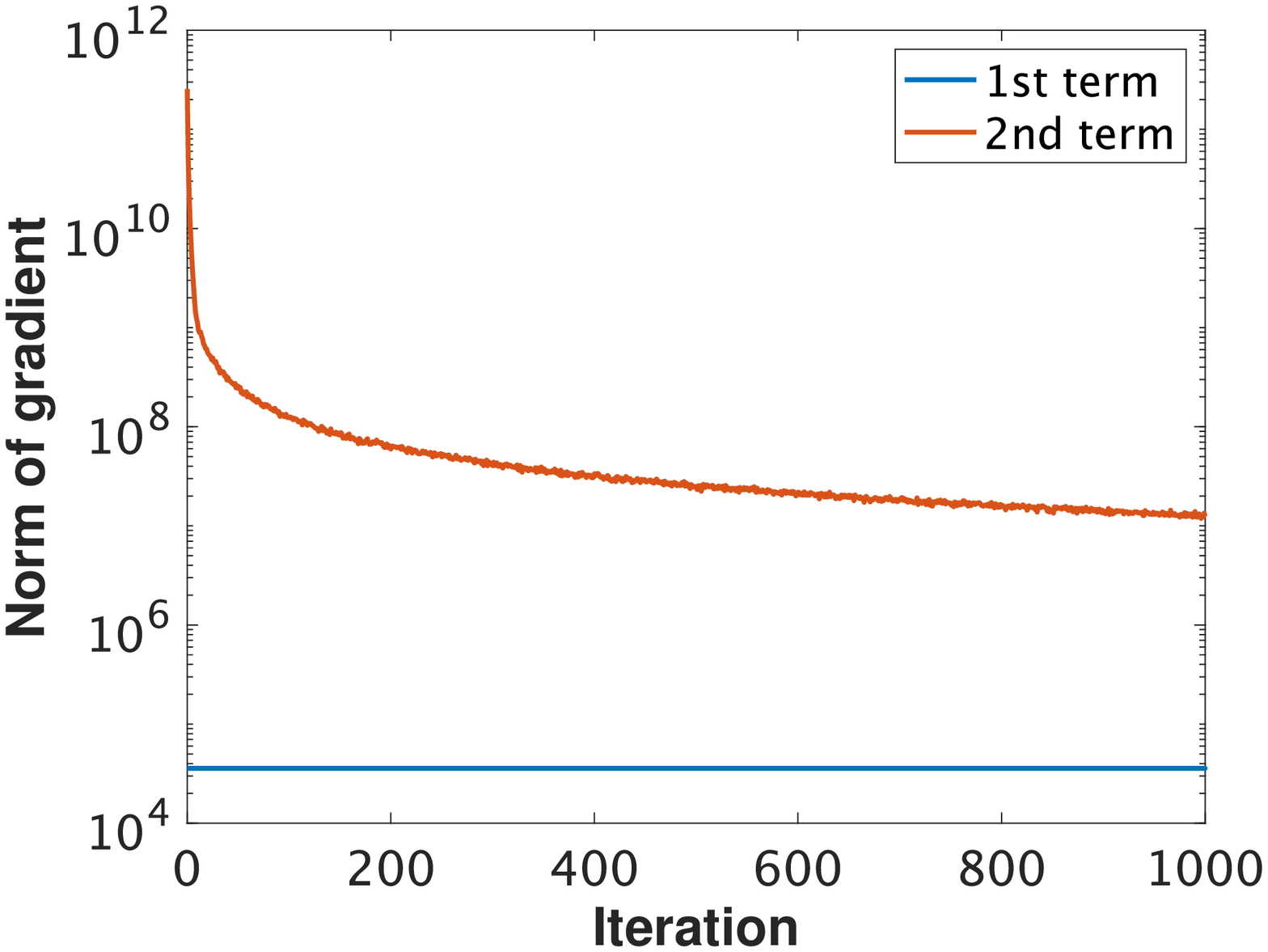}\\
	\end{center} 
	\end{minipage}	

	\begin{minipage}[t]{0.33\textwidth}
	\begin{center}
		{\small  (b) objective value ($\lambda=10^{-9}$)}	
	\end{center} 
	\end{minipage}	
	\begin{minipage}[t]{0.33\textwidth}
	\begin{center}
		{\small  (c) norm of gradient ($\lambda=10^{-9}$)}	
	\end{center} 
	\end{minipage}	
	\vspace*{1.5cm}		

	\begin{minipage}[t]{0.24\textwidth}
	\begin{center}
		\includegraphics[width=\textwidth]{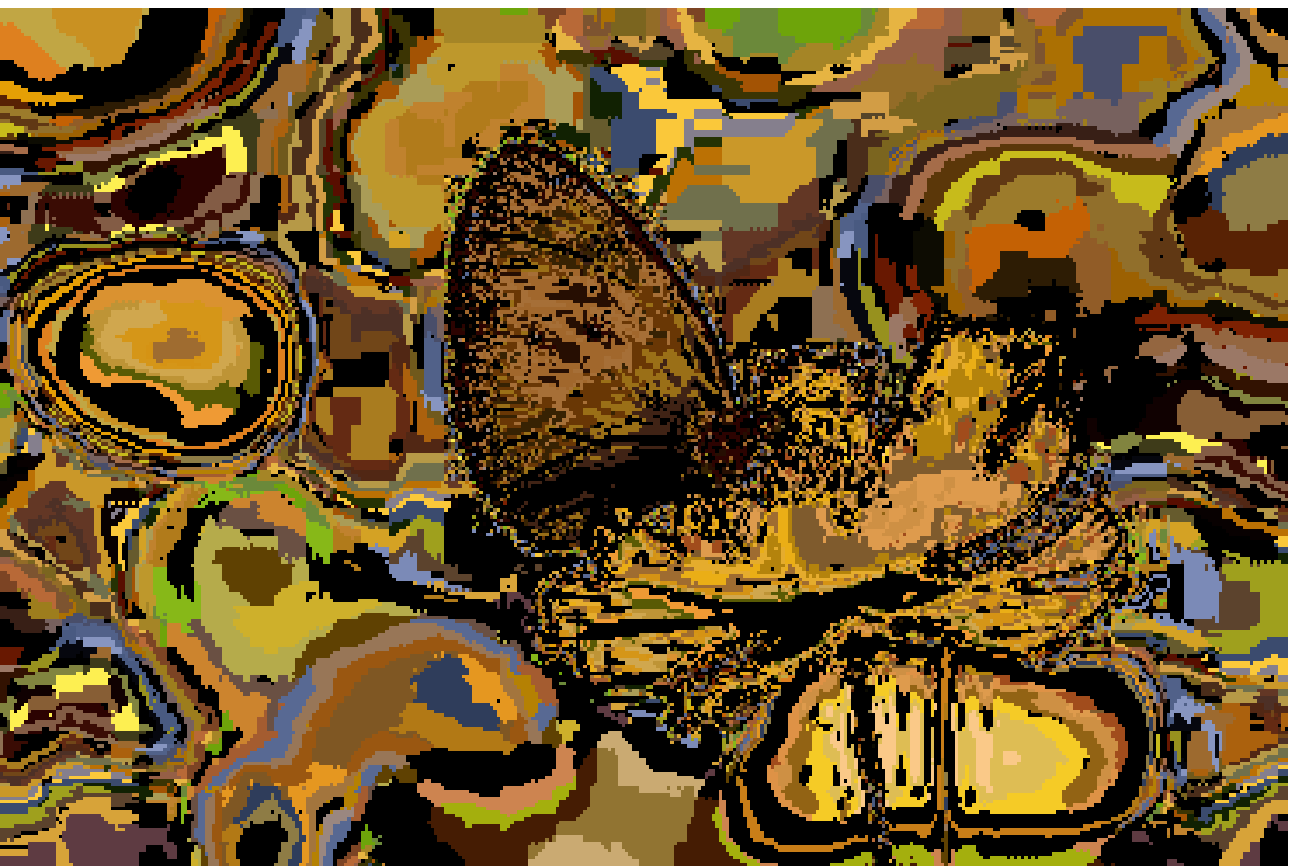}\\
		
		{\footnotesize  $k = 1$}			
	\end{center} 
	\end{minipage}
	\begin{minipage}[t]{0.24\textwidth}
	\begin{center}
		\includegraphics[width=\textwidth]{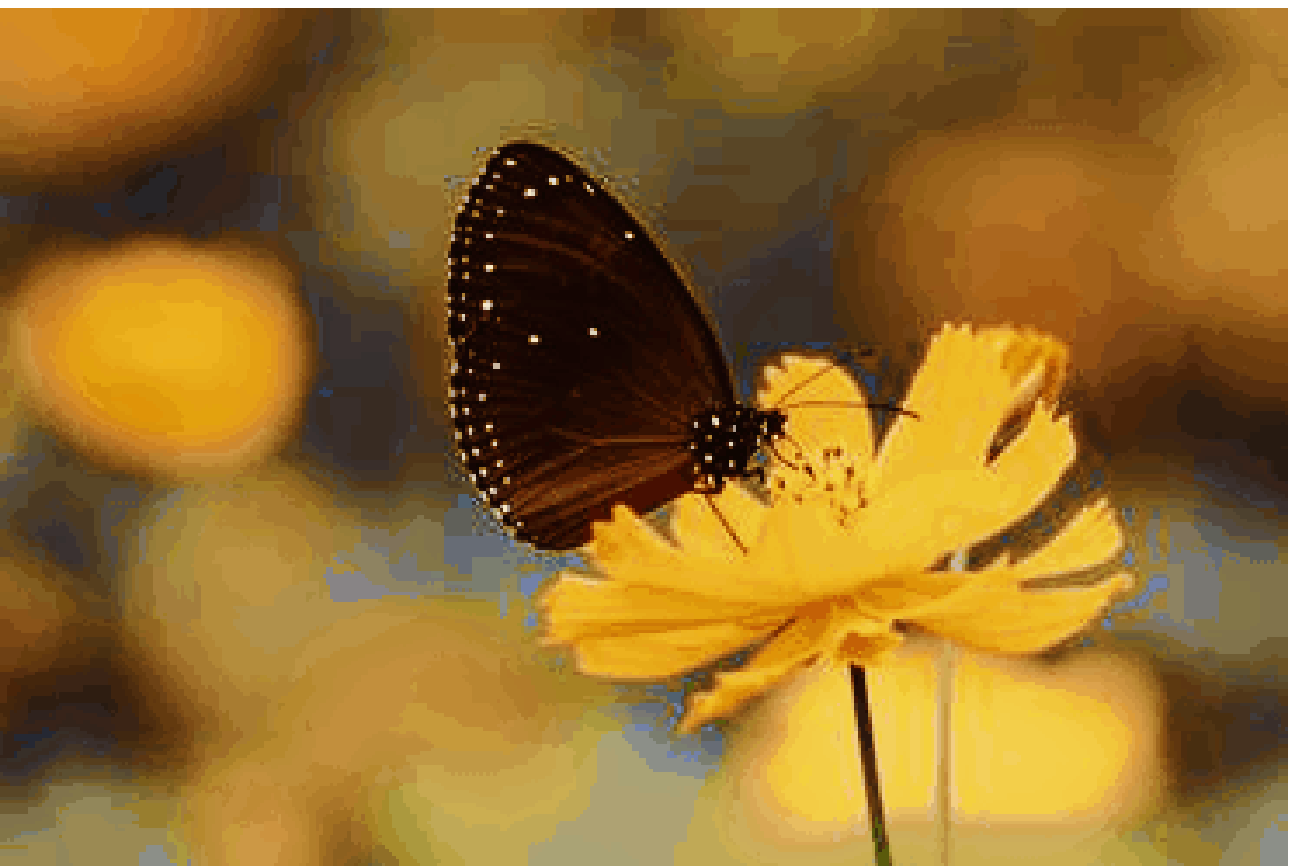}\\
		
		{\footnotesize  $k = 10$}			
	\end{center} 
	\end{minipage}	
	\begin{minipage}[t]{0.24\textwidth}
	\begin{center}
		\includegraphics[width=\textwidth]{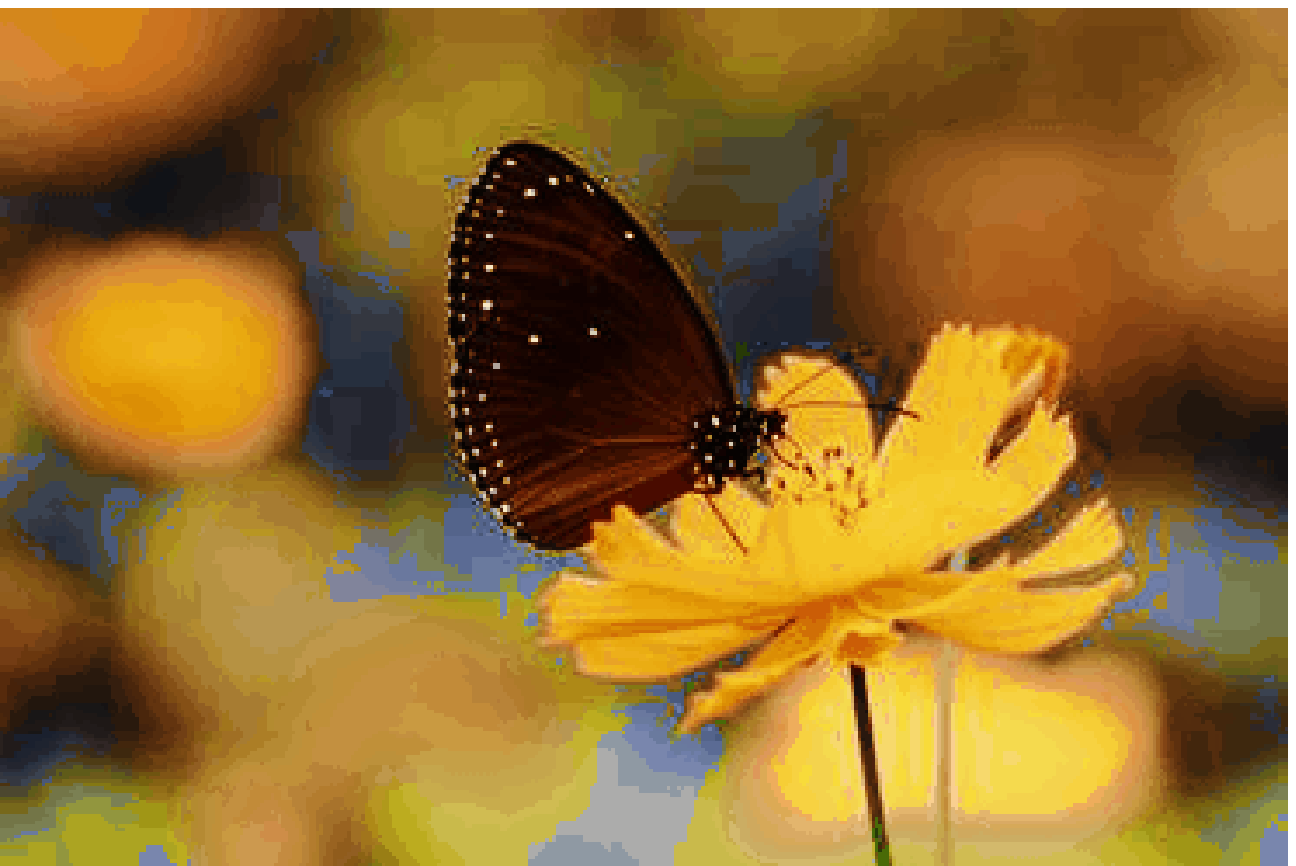}\\
		
		{\footnotesize  $k = 10^3$}			
	\end{center} 
	\end{minipage}	
	\begin{minipage}[t]{0.24\textwidth}
	\begin{center}
		\includegraphics[width=\textwidth]{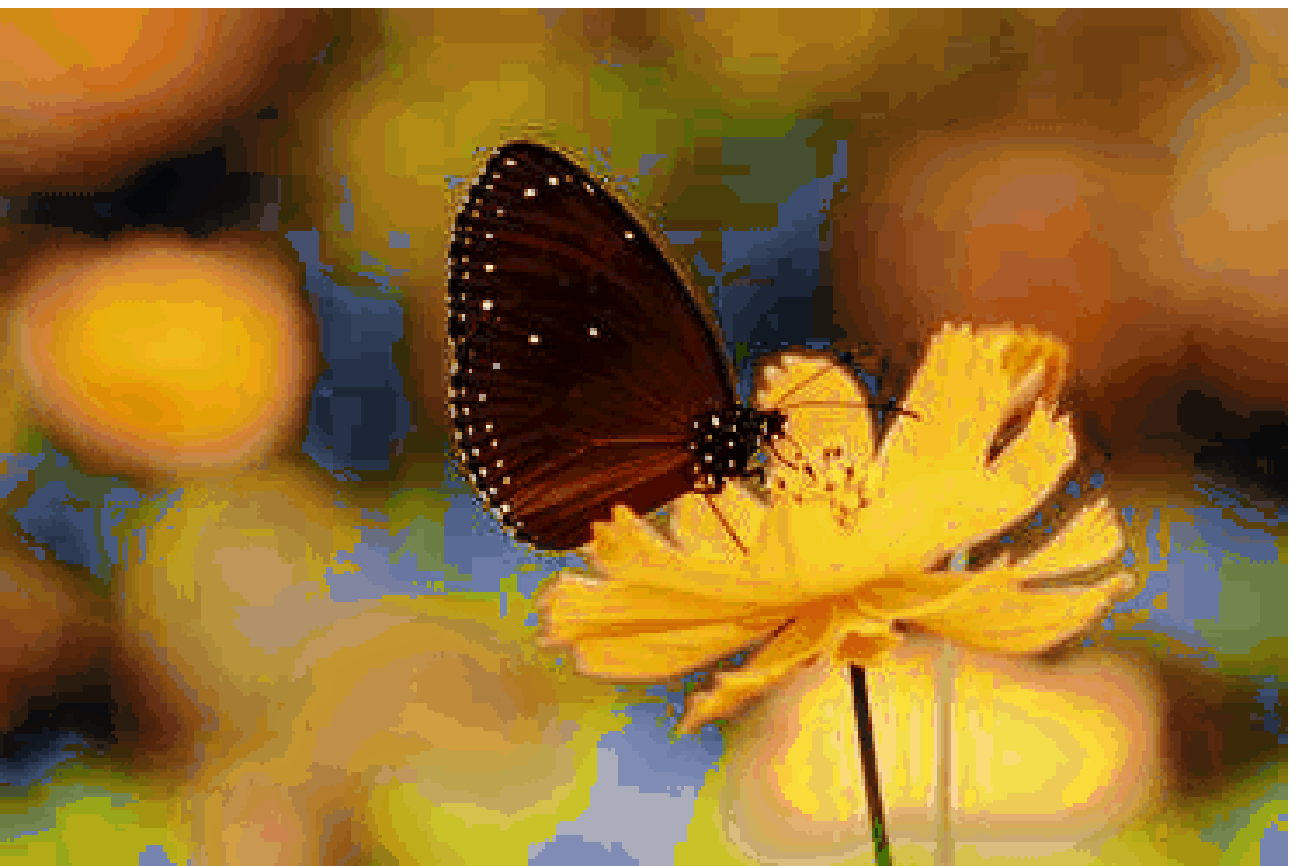}\\
		
		{\footnotesize  $k = 10^4$}			
	\end{center} 
	\end{minipage}	
	\vspace*{0.2cm}

	{\small  (d) transition of color-transferred images ($\lambda=10^{-6}$)}
	\vspace*{0.2cm}	

	\begin{minipage}[t]{0.33\textwidth}
	\begin{center}
		\includegraphics[width=\textwidth]{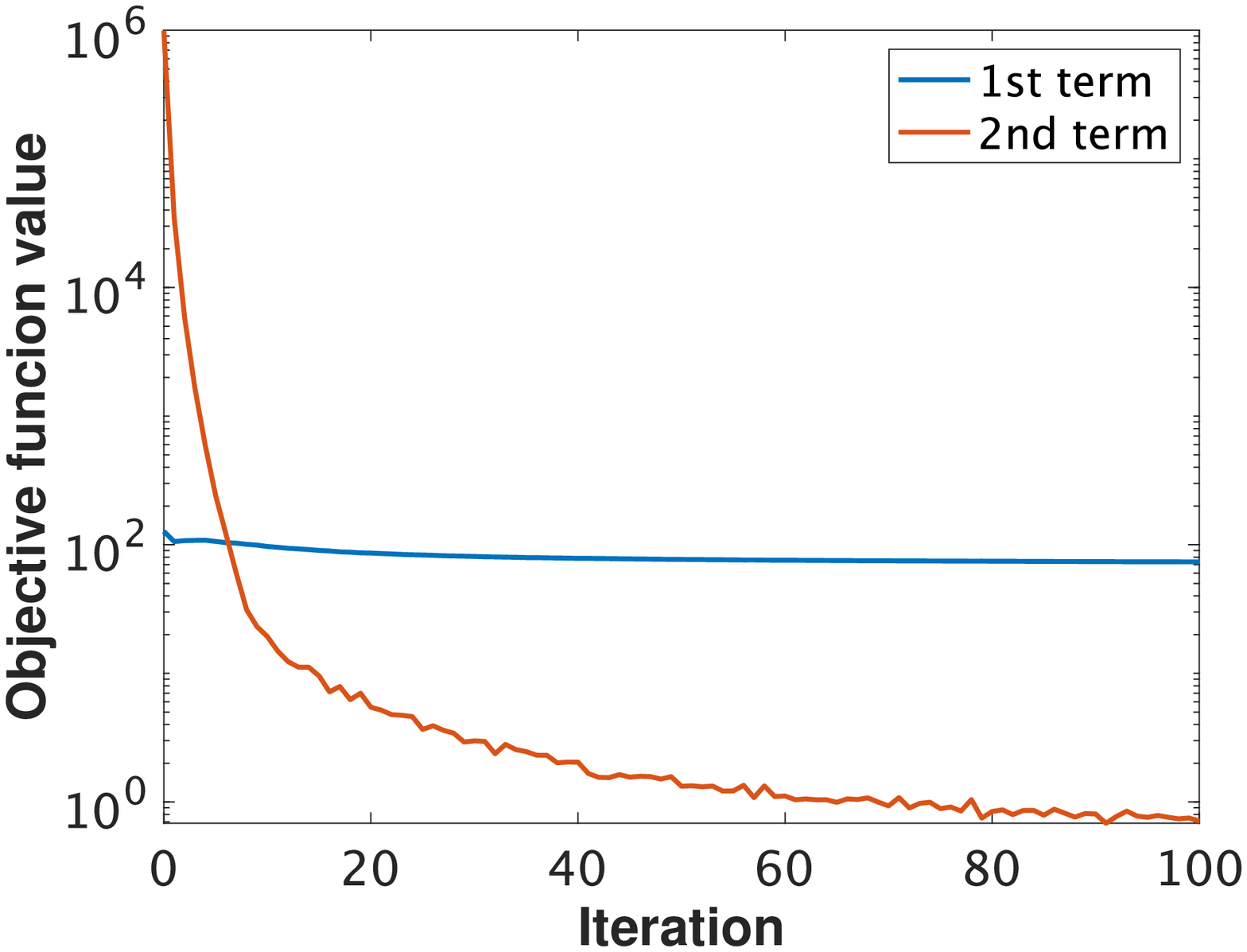}\\
	\end{center} 
	\end{minipage}	
	\begin{minipage}[t]{0.33\textwidth}
	\begin{center}
		\includegraphics[width=\textwidth]{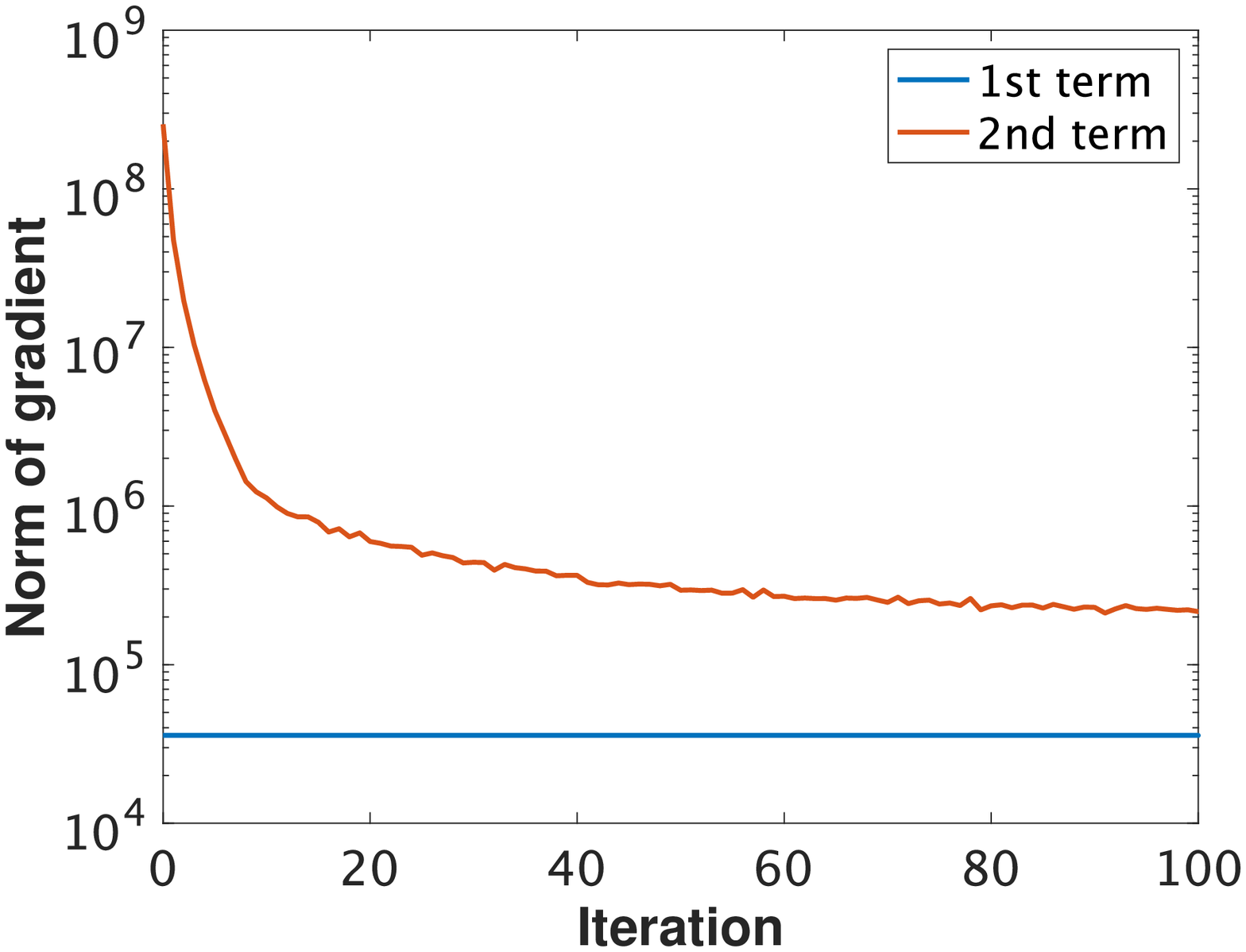}\\
	\end{center} 
	\end{minipage}	

	\begin{minipage}[t]{0.33\textwidth}
	\begin{center}
		{\small  (e) function value ($\lambda=10^{-6}$)}	
	\end{center} 
	\end{minipage}	
	\begin{minipage}[t]{0.33\textwidth}
	\begin{center}
		{\small  (g) norm of gradient ($\lambda=10^{-6}$)}	
	\end{center} 
	\end{minipage}	

\caption{Transition of color-transferred images and objective value and norm of gradient ($m=n=256$).}
\label{fig:CTImages_N256}
\end{center}		
\end{figure}

We evaluated two sets, i.e., $m=n=32$ and $m=n=256$. The baseline BCFW with the decay stepsize rule, BCFW-U-DEC, is used. The results of the former and the latter are shown in Figures \ref{fig:CTImages_N32} and \ref{fig:CTImages_N256}, respectively. 

\vspace*{0.2cm}
\noindent
{\bf Small size $m=n=32$ in Figure \ref{fig:CTImages_N32}.} As for the case $\lambda = 10^{-9}$ in Figure \ref{fig:CTImages_N32}(a), the same situation happens at $k=3300$ as the previous experiment using synthetic images. This single color is close to the $\vec{b}$-weighed averaged color of the centroids of the reference image. Next, at $k=10^5$, the single color is scattered and a naturally color-transferred images are produced. However, when the optimization process proceeds further, it can be seen that the image at $k=10^6$ contains artificial gray or blue pixels in the background. Figure \ref{fig:CTImages_N32}(b) illustrates the heat-map of the obtained transport matrices at the $k$-th iteration, $\mat{T}^{(k)}$. As the result of $\mat{T}^{(3000)}$ at $k=3000$, the matrix contains similar values along each column, in other words, the vertical blocks with similar colors can be recognized. However, this vertical structure disappears as the iteration proceeds. On the other hand, when $\lambda = 10^{-6}$, the convergence is extremely faster than that of $\lambda = 10^{-9}$. In addition, we cannot see such vertical structures and the single-colored image through the entire of iterations. 

\vspace*{0.2cm}
\noindent
{\bf Large size $m=n=256$ in Figure \ref{fig:CTImages_N256}.} Although the heat-map of the normalized transferred matrices as in Figure \ref{fig:CTImages_N32}(b) cannot be illustrated due to too small value of each pixels,  we find the same situation in the color-transferred images as the case with $m=n=32$. Additionally, the finally obtained image at $k=200$ when $(\lambda, k)=(10^{-6}, 10^4)$ looks quite similar to that of LP, which is shown in Figure \ref{fig:CTRealworldImages}(c). Therefore, it is reasonably understandable that the weight by $\lambda = 10^{-6}$ has similar impacts on the constraint of $\mat{T}\vec{1}_n = \vec{a}$ in the non-relaxed original OT problem.

\subsubsection{Discussion}
From these observations, when the relaxation parameter $\lambda$ is quite large, BCFW for the semi-relaxed OT problem converges faster. However, it pays a high price on dreadful artifacts on the color-transferred images. On the other hand, when using relatively small $\lambda$, BCFW can produce a natural image and avoids heavy artifacts because BCFW for the semi-relaxed OT problem does not necessarily transfer all the probability mass of the source image to the reference while keeping its solution closer to the original one. Nevertheless, it should be emphasized that it admits a slow convergence rate, and, more importantly, we will get similar images as LP if we further proceed its optimization process. Thus, it is perforable to terminate before its complete convergence.

\section{Conclusion}
We have proposed in this paper faster block-coordinate Frank-Wolfe algorithms for a convex semi-relaxed optimal transport problem. The main contributions are that we proved their upper-bounds of the worst convergence iterations, and the equivalence between the linearization duality gap and the Lagrangian duality gap. Three variants have been also proposed. Numerical evaluations demonstrated that the proposed algorithms outperform state-of-the-art algorithms across different settings.

\clearpage
\appendix

\noindent
{\bf\LARGE Supplementary}
\vspace*{0.2cm}

\renewcommand\thefigure{A.\arabic{figure}}  
\setcounter{figure}{0} 

\renewcommand\thetable{A.\arabic{table}}  
\setcounter{table}{0} 

\renewcommand{\theequation}{A.\arabic{equation}}
\setcounter{equation}{0}

\renewcommand\thealgorithm{A.\arabic{algorithm}}  
\setcounter{algorithm}{0} 

\renewcommand\thefigure{A.\arabic{figure}}  
\setcounter{figure}{0} 

\renewcommand\thetable{A.\arabic{table}}  
\setcounter{table}{0} 

\renewcommand{\theequation}{A.\arabic{equation}}
\setcounter{equation}{0}

\renewcommand\thealgorithm{A.\arabic{algorithm}}  
\setcounter{algorithm}{0} 

\vspace*{1.0cm}

This supplementary material presents the details of the proposed algorithms, the complete proof of the convergence analysis in the main material, additional theoretical results, and additional experiments. The structure is as follows:
\begin{itemize}
\item Section \ref{AppenSec:Algorithm}:

The complete algorithms of FW, BCAFW, BCPFW and BCFW-GA.

\item Sections \ref{AppenSec:ConvAnaFW}, \ref{AppenSec:ConvAnaBCFW}, \ref{AppenSec:DualityGapEquivalence} and  \ref{AppenSec:BCFWGAS}:

The complete proof of the convergence analysis in the main material, and additional theoretical results. 
\item Section \ref{AppenSec:AddNumEval}: 

Additional experiments for the main material. 

\end{itemize}

It should be first noted that, throughout this supplementary material, we dare to include a constant factor in the order of total complexity, for example, {$\mathcal{O}(\frac{8}{ \lambda\epsilon})$} as in {\bf Theorem \ref{thm:RelaxedConvergence}}, to discuss and compare obtained results in detail. 

\section{Algorithms}
\label{AppenSec:Algorithm}

This section summarizes the algorithms of FW, BCAFW, BCPFW and BCFW-GA in 
Algorithms \ref{AppenAlg:FW-SROT}, \ref{AppenAlg:BCAFW-SROT}, \ref{AppenAlg:BCPFW-SROT}, and \ref{AppenAlg:BCFW-GA-SROT}, respectively. 

\begin{algorithm}[htbp]
\caption{Frank-Wolfe (FW) algorithm for semi-relaxed OT}      
\label{AppenAlg:FW-SROT}
\begin{algorithmic}[1]       
\Require{$\mat{T}^{(0)} \in {b}_1\Delta_m \times \cdots \times {b}_n\Delta_m$}
\For {$k=0 \dots K$}
\For {$i=0 \dots n$}
\State Compute $\vec{s}_i = b_i \defargmin_{\scriptsize {\vec{e}_k \in \Delta_m}, k \in [m]}\ \langle \vec{e}_k,\nabla_if(\mat{T}^{(k)}) \rangle$
\EndFor
\If {$g(\mat{T}^{(k)}) \leq \epsilon$}
\State break
\EndIf
\State{Compute stepsize $\gamma$ as
\[
	\hspace*{-0.7cm}\gamma  = 
	\begin{cases}
	\gamma_{\rm LS},\quad \quad\quad\quad\quad\hfill {\rm (line-search\ in\ (\ref{AppenEq:SemiRelaxedFWStep}))}\\
	\displaystyle{\frac{2}{k+2}}, \hfill {\rm (decay\ rule)}
	\end{cases}	
\]
}
\State Update $\mat{T}^{(k+1)} = (1-\gamma)\mat{T}^{(k)} + \gamma \mat{S}$
\EndFor
\end{algorithmic}
\end{algorithm}

In the algorithm, the optimal stepsize $\gamma_{\rm LS}$ by the line-search is given as
\begin{equation}
\label{AppenEq:SemiRelaxedFWStep}
	\gamma_{\rm LS}  = \frac{{\lambda\langle \mat{T}^{(k)}-\mat{S},\mat{C}\rangle_F+\langle \mat{T}^{(k)}\vec{1}_n - \mat{S}\vec{1}_n,\mat{T}^{(k)}\vec{1}_n - \vec{a} \rangle}}{\displaystyle{ \|\mat{T}^{(k)}\vec{1}_n - \mat{S}\vec{1}_n\|^2}}.
\end{equation}

\begin{algorithm}[htbp]
\caption{Block-Coordinate Away-step Frank-Wolfe (BCAFW) for semi-relaxed OT}      
\label{AppenAlg:BCAFW-SROT}
\begin{algorithmic}[1]       
\Require{$\mat{T}^{(0)} = (\vec{t}^{(0)}_1,\vec{t}^{(0)}_2,\dots,\vec{t}^{(0)}_n) \in {b}_1\Delta_m \times {b}_2\Delta_m \times \cdots \times {b}_n\Delta_m$\,  $\mathcal{S}_i=\lbrace \vec{e}_1 \rbrace, \forall i \in [n]$. }

\For {$k=0 \dots K$}
\State Select index $i \in [n]$ randomly 
\State Compute $\vec{s}_i$ \Comment{Line 2 in {\bf Algorithm \ref{alg:BCFW-SROT}}}
\State Set $\vec{d}_{\rm FW} = \vec{s}_i - \vec{t}^{(k)}_i$

\State Compute $\vec{v}_i =b_i \defargmax_{\scriptsize {\vec{e}_j \in \mathcal{S}_i}, j \in [m] } \langle \vec{e},\nabla_i f(\mat{T}^{(k)}) \rangle$
\State Set $\vec{d}_{\rm Away} = \vec{t}^{(k)}_i - \vec{v}_i$

\If {$\langle -\nabla f(\mat{T}^{(k)}), \vec{d}_{\rm FW} \rangle \geq  \langle - \nabla f(\mat{T}^{(k)}), \vec{d}_{\rm Away} \rangle$}
	\State $\vec{d} = \vec{d}_{\rm FW}, \gamma_{\rm max} = 1$
\Else
	\State $\vec{d} = \vec{d}_{\rm Away}$, $\gamma_{\rm max} = \displaystyle{\frac{\alpha_{{\vec{v}_i}}}{1 - \alpha_{{\vec{v}_i}}}}$
\EndIf

\State{Compute stepsize $\gamma = \gamma_{\rm LS}$ using \vec{d}} \Comment{(\ref{Eq:SemiRelaxedBCAFWStep})}

\State Update $\vec{t}^{(k+1)}_j\quad\forall j \in [n]$ \Comment{Line 5 in {\bf Algorithm \ref{alg:BCFW-SROT}}}
\State Update $\mathcal{S}^{(k+1)}_i = \lbrace \vec{e} \in \Delta_m : \alpha^{(t+1)}_{{\vec{e}}} > 0 \rbrace$

\EndFor
\end{algorithmic}
\end{algorithm}

\begin{algorithm}[htbp]
\caption{Block-Coordinate Pairwise step Frank-Wolfe (BCPFW) for semi-relaxed OT}      
\label{AppenAlg:BCPFW-SROT}
\begin{algorithmic}[1]       
\Require{$\mat{T}^{(0)} = (\vec{t}^{(0)}_1,\vec{t}^{(0)}_2,\dots,\vec{t}^{(0)}_n) \in {b}_1\Delta_m \times {b}_2\Delta_m \times \cdots \times {b}_n\Delta_m$\,  $\mathcal{S}_i=\lbrace \vec{e}_1 \rbrace, \forall i \in [n]$. }

\For {$k=0 \dots K$}
\State Select index $i \in [n]$ randomly 
\State Compute $\vec{s}_i$ \Comment{Line 2 in {\bf Algorithm \ref{alg:BCFW-SROT}}}

\State Compute $\vec{v}_i$ \Comment{Line 5 in {\bf Algorithm \ref{AppenAlg:BCAFW-SROT}}}

\State Set $\vec{d} = \vec{d}_{\rm Pair} = \vec{s}_i - \vec{v}_i$ and $\gamma_{\rm max} = \alpha_{{\vec{v}_i}}$

\State Compute stepsize $\gamma = \gamma_{\rm LS}$ \Comment{Line 12 in {\bf Algorithm \ref{AppenAlg:BCAFW-SROT}}}

\State Update $\vec{t}^{(k+1)}_j\quad\forall j \in [n]$ \Comment{Line 5 in {\bf Algorithm \ref{alg:BCFW-SROT}}}
\State Update $\mathcal{S}^{(k+1)}_i$ \Comment{Line 14 in {\bf Algorithm \ref{AppenAlg:BCAFW-SROT}}}
\EndFor
\end{algorithmic}
\end{algorithm}

\begin{algorithm}[htbp]
\caption{Block-coordinate Frank-Wolfe with gap-adaptive sampling (BCFW-GA) for semi-relaxed OT}      
\label{AppenAlg:BCFW-GA-SROT}
\begin{algorithmic}[1]       
\Require{$\mat{T}^{(0)} = (\vec{t}^{(0)}_1,\vec{t}^{(0)}_2,\dots,\vec{t}^{(0)}_n) \in {b}_1\Delta_m \times {b}_2\Delta_m \times \cdots \times {b}_n\Delta_m$,  $g_i(\mat{T})=C,\ \forall i \in [n]$, where $C$ is a large constant. }

\For {$k=0 \dots K$}
\State Select index $i \in [n]$ randomly with probability $\propto g_i(\mat{T})$
\State Compute $\vec{s}_i$ \Comment{Line 3 in {\bf Algorithm \ref{alg:BCFW-SROT}}}
\State Compute stepsize $\gamma$  \Comment{Line 4 in {\bf Algorithm \ref{alg:BCFW-SROT}}}
\State Update $\vec{t}^{(k+1)}_j,\ \forall j \in [n]$ \Comment{Line 5 in {\bf Algorithm \ref{alg:BCFW-SROT}}}
\State {Update and store the duality gap for the $i$-th column $g_i(\mat{T})$ in (\ref{AppenEq:Gi}).
}
\If{$k$ satisfies the global update condition}
	\For {$i=1 \ldots n$}
	\State Compute $\vec{s}_i$  \Comment{Line 3 in {\bf Algorithm \ref{alg:BCFW-SROT}}}
	\State Update and store the duality gap for the $i$-th column $g_i(\mat{T})$ in (\ref{AppenEq:Gi}).
	\EndFor
\EndIf
\EndFor
\end{algorithmic}
\end{algorithm}

\section{Convergence analysis of the proposed FW algorithm for semi-relaxed OT problem}
\label{AppenSec:ConvAnaFW}

We first define the curvature constant $C_f$ for the semi-relaxed OT problem.
\begin{Def}[Curvature constant of FW algorithm for semi-relaxed OT problem]
\label{Def:CC-SROT}
\hspace*{1cm}

The curvature constant on the domain $\mathcal{M}$ of the semi-relaxed OT problem is defined as 
\begin{equation*}
	\label{eq:CurvatureFW}
	C_f := \!\!\!\!
	\sup_{\scriptsize\substack{\displaystyle{\mat{T},\mat{S} \in \mathcal{M}},\\
				\displaystyle{\gamma \in [0,1],}\\
				\displaystyle{\mat{Y} = \mat{T} + \gamma(\mat{S} - \mat{T})}
	}}\!\!\!\! \frac{2}{\gamma^2}(f(\mat{Y})-f(\mat{T})-\langle \mat{Y} -\mat{T},\nabla f(\mat{T}) \rangle),
\end{equation*}
where $\mathcal{M}$ represents ${b}_1\Delta_m \times {b}_2\Delta_m \times \dots \times {b}_n\Delta_m$. 
\end{Def}
We then drive the convergence analysis of the FW algorithm. 
\begin{Thm}
\label{thm:RelaxedConvergence}
Let $\mat{T}^*$ is the optimal solution of the semi-relaxed OT problem. Consider {Algorithm \ref{AppenAlg:FW-SROT}} with a decay stepsize rule $\gamma=\frac{2}{k+2}$. Then, we have $f(\mat{T}^{(k)}) - f(\mat{T}^{*}) \leq \frac{2C_f}{k+2}$ for $k \geq 1$, where $C_f$ the curvature constant with $C_f \leq \frac{4}{\lambda}$. Additionally, given a constant $\epsilon$, this means that {Algorithm \ref{AppenAlg:FW-SROT}} requires at most {$\mathcal{O}(\frac{8}{ \lambda\epsilon})$} for its convergence. 
\end{Thm}

To prove the theorem above, we first state the following theorem in the classical FW algorithm.
\begin{Thm}[Convergence analysis for a general FW algorithm \cite{lacostejulien,JaggiMartin2013}]
\label{thm:GeneralFrankWolfeConvergence}
\hspace*{1cm}

The curvature constant on the domain $\mathcal{M}$ is defined as 
\begin{equation*}
	\label{eq:curvature_constant_FW}
	C_f := \sup_{\scriptsize\substack{\displaystyle{\vec{x},\vec{s} \in \mathcal{M}},\\
				\displaystyle{\gamma \in [0,1],}\\
				\displaystyle{\vec{y} = \vec{x} + \gamma(\vec{s} - \vec{x})}
	}} \frac{2}{\gamma^2}(f(\vec{y})-f(\vec{x})-\langle \vec{y} -\vec{x},\nabla f(\vec{x}) \rangle),
\end{equation*}
where $\vec{y}$ is equal to $\vec{x}+\gamma(\vec{s}-\vec{x})$ for $\gamma \in [0,1]$. Then, for each $k \geq 1$, the iterate $\vec{x}^{(k)}$ of the Frank-Wolfe algorithm satisfies 
\begin{equation}
	\label{eq:PrimalConvergence}
	f(\vec{x}^{(k)}) - f(\vec{x}^{*}) \leq \frac{2C_f}{k+2}, 
\end{equation}
where $\vec{x}^*$ is the optimal solution. 
\end{Thm}
%

We also give the following lemma.
\begin{Lem} 
\label{lem:norm_maximum}
The upper-bound of squared $\ell_2$-norm $\|\vec{x}\|_2^2$ is  $1$ on a domain $\Delta_m$.
\end{Lem}
\begin{proof}
The square of  $\sum_{i=1}^{m}{x}_i$ satisfies 
\begin{equation*}
	\left(\sum_{i=1}^m{x}_i \right)^2 = \sum_{i=1}^m{x}_i^2 + 2\sum_{1\leq i < j \leq m}{x}_i {x}_j.
\end{equation*}
$\vec{x}$ satisfies $\sum_{i=1}^m{x}_i=1$ and $\sum_{i=1}^n{x}_i^2$ is equal to $\|\vec{x}\|_2^2$. Hence, squared $\ell_2$-norm is bounded as
\begin{eqnarray*}
	\|\vec{x}\|_2^2 \leq 1^2 -  2\sum_{1\leq i < j \leq m}{x}_i {x}_j \leq 1,
\end{eqnarray*}
where we note ${x}_i {x}_j \geq 0$. 

This complete the proof. 
\end{proof}


We now provide the complete proof of {\bf Theorem \ref{thm:RelaxedConvergence}}.

\begin{proof}
Since the objective of the semi-relaxed problem is twice differentiable, the Taylor expansion at $\gamma = 0$ of $f(\mat{T}+\gamma(\mat{S} - \mat{T}))$ is
\begin{equation}
\label{eq:TaylorExpansion}
	f(\mat{T} + \gamma(\mat{S}-\mat{T})) = f(\mat{T}) + \gamma(\mat{S}-\mat{T})^T\nabla f(\mat{T}) 
	+\frac{1}{2}\gamma^2(\mat{T}-\mat{S})^T\nabla^2f(\mat{T}')(\mat{S} - \mat{T}),
\end{equation}
where $\mat{T}'$ satisfies $\mat{T}' \in [\mat{T} ,\mat{T} + \gamma(\mat{S}-\mat{T})]$ \cite{AwayStepClarkson}.

Substituting (\ref{eq:TaylorExpansion}) to (\ref{eq:CurvatureFW}) gives the following formula transformation.
\begin{eqnarray*}
	C_f &=& \sup_{\scriptsize \substack{\displaystyle{\mat{T},\mat{S} \in \mathcal{M}},\\
				\displaystyle{\gamma \in [0,1],}\\
				\displaystyle{\mat{Y} = \mat{T} + \gamma(\mat{S} - \mat{T})}
	}}\frac{2}{\gamma^2}(f(\mat{Y})-f(\mat{T})-\langle \mat{Y} - \mat{T},\nabla f(\mat{T}) \rangle)\\
	    &\leq&\sup_{\scriptsize \substack{\displaystyle{\mat{T},\mat{S} \in \mathcal{M}},\\
				\displaystyle{\gamma \in [0,1],}\\
				\displaystyle{\mat{Y} = \mat{T} + \gamma(\mat{S} - \mat{T})}
	}}\frac{2}{\gamma^2}(\gamma(\mat{S}-\mat{T})^T\nabla f(\mat{T})
+ \frac{1}{2}\gamma^2(\mat{S}-\mat{T})^T\nabla^2f(\mat{T}')(\mat{S} - \mat{T})-\langle \gamma(\mat{S} - \mat{T}),\nabla f(\mat{T}) \rangle)\\
	  &\leq&\sup_{\scriptsize \substack{\displaystyle{\mat{T},\mat{S} \in \mathcal{M}},\\
				\displaystyle{\mat{T}' \in [\mat{T} ,\mat{T} + \gamma(\mat{S}-\mat{T})]}}}(\mat{S}-\mat{T})^T\nabla^2f(\mat{T}')(\mat{S} - \mat{T}).
\end{eqnarray*}
For arbitrary $\mat{X} \in \mathbb{R}^{m \times n}$, the Hessian matrix of $f(\mat{X})$ is
\begin{equation}
\label{eq:HessianSemiRelaxed}
	\nabla^2f(\mat{X}) = \frac{1}{\lambda}
	 \left(
		\begin{array}{ccc}
			\mat{E}_m & \cdots & \mat{E}_m\\
			\vdots & \vdots & \vdots \\
			\mat{E}_m  &\cdots & \mat{E}_m\\
		\end{array}
	\right)\quad ( \in \mathbb{R}^{mn \times mn}),
\end{equation}
where $\mat{E}_m$ is the identity matrix of size $m \times m$. Here, we denote a vectorization operation that stacks all column vectors of a matrix $\mat{X}=(\vec{x}_1, \vec{x}_2, \ldots, \vec{x}_n) \in \mathbb{R}^{m \times n}$, where $\vec{x}_i \in \mathbb{R}^m$, to make a column vector $(\vec{x}_1^T, \vec{x}_2^T, \ldots, \vec{x}_n^T)^T \in \mathbb{R}^{mn}$ as ${\rm Vec}(\mat{X})$. Then, the Hessian matrix in (\ref{eq:HessianSemiRelaxed}) satisfies 
\begin{eqnarray*}
	{\rm Vec}(\mat{X})^T \nabla^2f(\mat{X}) {\rm Vec}(\mat{X}) 
	= (\vec{x}_1^T, \vec{x}_2^T, \ldots, \vec{x}_n^T)
		\left(
		\begin{array}{ccc}
			\mat{E}_m & \cdots & \mat{E}_m\\
			\vdots & \vdots & \vdots \\
			\mat{E}_m  &\cdots & \mat{E}_m\\
		\end{array}
	\right) 
	\left(
		\begin{array}{c}
			\vec{x}_1\\
			\vec{x}_2\\
			\vdots \\
			\vec{x}_n\\
		\end{array}
	\right)
	=   \|\mat{X}\vec{1}_n\|^2 \geq 0.
\end{eqnarray*}
Thus, the Hessian matrix (\ref{eq:HessianSemiRelaxed}) is positive semi-definite. From this fact, the curvature constant $C_f$ is obtained as
\begin{eqnarray}
	\label{eq:CurvatureUpperBound}
	C_f&\leq&\sup_{\scriptsize {\mat{T},\mat{S} \in \mathcal{M}}}\frac{1}{\lambda}\|\mat{S}\vec{1}_n -\mat{T}\vec{1}_n \|^2
\end{eqnarray}
where $\mathcal{M} = {b}_1\Delta_m \times {b}_2\Delta_m \times \dots \times b_n \Delta_m$.
Consequently, the inequality (\ref{eq:CurvatureUpperBound}) is represented as
\begin{eqnarray*}
 C_f &\leq& \sup_{\scriptsize {\mat{T},\mat{S} \in \mathcal{M}}}\frac{1}{\lambda}\|\mat{S}\vec{1}_n -\mat{T}\vec{1}_n\|^2
 \leq   \sup_{\scriptsize{\mat{S} \in \mathcal{M}}}\frac{1}{\lambda}\|2\mat{S}\vec{1}_n\|^2
\leq\frac{4}{\lambda}\sup_{\scriptsize{\vec{s}_i \in {b}_i ,\forall i \in [n]}} \|\sum_{i=1}^n \vec{s}_i\|^2\\
&\leq & \frac{4}{\lambda}\sup_{\scriptsize{\vec{s}_i \in {b}_i ,\forall i \in [n]}} \left (\sum_{i=1}^n\|\vec{s}_i\|\right)^2
		 \leq  \frac{4}{\lambda} \left(\sum_{i=1}^nb_i\right)^2
		 \leq  \frac{4}{\lambda} 1^2 
		 \leq \frac{4}{\lambda},
\end{eqnarray*}
where the fifth inequality used {\bf Lemma \ref{lem:norm_maximum}}.

Finally, given a constant $\epsilon$, considering (\ref{eq:PrimalConvergence}) in {\bf Theorem \ref{thm:GeneralFrankWolfeConvergence}}, the FW algorithm for the semi-relaxed OT problem converges if it satisfies the following inequality.
\begin{eqnarray*}
	\epsilon \geq \frac{2C_f}{k+2}
	\quad \Longleftrightarrow \quad  \epsilon \geq \frac{2}{k+2}\cdot \frac{4}{\lambda}
	\quad \Longleftrightarrow \quad  k+2 \geq 2 \cdot \frac{4}{\epsilon \lambda}.
\end{eqnarray*}
Hence, the FW algorithm in {Algorithm \ref{AppenAlg:FW-SROT}} requires at most $\mathcal{O}(\frac{8}{\epsilon \lambda})$ iterations.

This completes the proof.
\end{proof}

\section{Proof of {\bf Theorem \ref{thm:relaxedBCFWconvergence}}}
\label{AppenSec:ConvAnaBCFW}

It is known that the convergence analysis of a general BCFW algorithm is presented as below.
\begin{Thm}[Convergence analysis of a general BCFW algorithm \cite{lacostejulien}]
\label{thm:BlockCoordinateFrankWolfeConvergence}
\hspace*{1cm}

For each $k \geq 1$, the iterates $\vec{x}^{k}$ of BCFW Algorithm satisfies
\begin{equation}
	\label{eq:ExpectedPrimalConvergence}
	\mathbb{E}[f(\vec{x}^{(k)})] - f(\vec{x}^{*}) \leq \frac{2n}{k+2n}(C_f^\otimes+h_0),
\end{equation}
where $\vec{x}^*$ is the optimal solution, and $h_0$ is $f(\vec{x}^{(0)})-f(\vec{x}^*)$. Here, 
$C_f^\otimes := \sum_{i=1}^nC^{(i)}_f$, where 
\begin{equation*}
C^{(i)}_f := 
	 \sup_{\substack{\scriptsize {\vec{x} \in \mathcal{M},\vec{s}_i} \in \mathcal{M}_i,\\
				{\gamma \in [0,1],}\\
				{\scriptsize \vec{y} = \vec{x} + \gamma(\vec{s}_{[i]} - \vec{x}_{[i]})}
	}}\frac{2}{\gamma^2}(f(\vec{y})-f(\vec{x})-\langle \vec{y}_i -\vec{x}_i,\nabla_i f(\vec{x})\rangle)
,
\end{equation*}
and $\vec{x}_{[i]}$ refers to the zero-padding of $\vec{x}_i$.
\end{Thm}

The following gives the complete proof of {\bf Theorem \ref{thm:relaxedBCFWconvergence}} in the main material.
\begin{proof}
We first bound the block curvature constant $C^{\otimes}_f$ in {\bf Definition \ref{def:CurvatureBCFW}}. 

When $f$ is twice differentiable, we can transform the block curvature constant as well as the curvature constant of FW (\ref{eq:CurvatureUpperBound}). Then, $C_f^{(i)}$ satisfies 
\begin{equation*}
\label{eq:BlockCurvatureMidwayBound}
C^{(i)}_f \leq \sup_{\scriptsize {\vec{t}_i,\vec{s}_i \in {b}_i \Delta_m}}\frac{1}{\lambda}\|\vec{s}_i - \vec{t}_i\|^2.
\end{equation*}
This can be rearranged by {\bf Lemma \ref{lem:norm_maximum}} as 
\begin{eqnarray*}
	C^{(i)}_f &\leq & \sup_{\scriptsize {\vec{t}_i,\vec{s}_i \in b_i \Delta_m}}\frac{1}{\lambda}\|\vec{s}_i - \vec{t}_i\|^2 
		 \ \leq \ \sup_{\scriptsize{\vec{s}_i \in b_i \Delta_m}}\frac{1}{\lambda}\|2\vec{s}_i \|^2 
		 \ \leq \  \frac{4}{\lambda}\sup_{\scriptsize\vec{s}_i \in b_i\Delta_m}\|\vec{s}\|^2
		\leq   \frac{4}{\lambda}{b}_i^2. 
\end{eqnarray*}
Consequently, the curvature constant $C_f^\otimes$ of the entire domain satisfies
\begin{eqnarray}
\label{Eq:CotimesUpperBound}
{C^{\otimes}_f = \sum_{i=1}^n C_f^{(i)} \leq \sum_{i=1}^n \frac{4{b}^2_i}{\lambda} \leq \frac{4}{\lambda},}
\end{eqnarray}
because ${b}_i$ is in the probability simplex and we used {\bf Lemma \ref{lem:norm_maximum} }.

We now prove the upper-bound of the linearization duality gap $g(\mat{T})$ at the initial point $k=0$ by considering the discussion above.
We define the initial matrix $\mat{T}^{(0)}$ as $\mat{T}^{(0)}=(b_1\vec{e}_1,\dots,b_i\vec{e}_1,\dots,b_n\vec{e}_1)$, where $\vec{e}_j \in \mathbb{R}^m$ is one of the extreme points and the $j$-th element is $1$ otherwise 0. $h_0$ satisfies $h_0 = f(\mat{T}^{(0)})-  f(\mat{T}^*)\leq g(\mat{T}^{(0)})$. Thus, we bound $g(\mat{T}^{(0)})$ as the following.
\begin{eqnarray}
\label{Eq:GtUpperBound}
	g(\mat{T}^{(0)}) 
	&=& \langle\mat{T}^{(0)}-\mat{S},\nabla f(\mat{T}^{(0)}) \rangle \notag\\
	&=&  \left(
		\begin{array}{c}
			\vec{t}_1^{(0)}-\vec{s}_1\\
			\vdots \\
			\vec{t}_i^{(0)}-\vec{s}_i\\
			\vdots \\
			\vec{t}_n^{(0)}-\vec{s}_n
		\end{array}
	\right)
	\cdot
	  \left(
		\begin{array}{c}
			\vec{c}_1\\
			\vdots \\
			\vec{c}_i\\
			\vdots \\
			\vec{c}_n
		\end{array}
	\right)
	+
	\frac{1}{\lambda}
	\left(
		\begin{array}{c}
			\vec{t}_1^{(0)}-\vec{s}_1\\
			\vdots \\
			\vec{t}_i^{(0)}-\vec{s}_i\\
			\vdots \\
			\vec{t}_n^{(0)}-\vec{s}_n
		\end{array}
	\right)
	\cdot
	\left(
		\begin{array}{c}
			\mat{T}^{(0)}\vec{1}_n - \vec{a} \\
			\vdots \\
			\mat{T}^{(0)}\vec{1}_n - \vec{a} \\
			\vdots \\
			\mat{T}^{(0)}\vec{1}_n - \vec{a} \\
		\end{array}
	\right) \notag\\
	&=&  \left(
		\begin{array}{c}
			{b}_1\vec{e}_1-\vec{s}_1\\
			\vdots \\
			{b}_i\vec{e}_1-\vec{s}_i\\
			\vdots \\
			{b}_n\vec{e}_1-\vec{s}_n
		\end{array}
	\right)
	\cdot
	  \left(
		\begin{array}{c}
			\vec{c}_1\\
			\vdots \\
			\vec{c}_i\\
			\vdots \\
			\vec{c}_n
		\end{array}
	\right)
	+
	\frac{1}{\lambda}
	\left(
		\begin{array}{c}
			{b}_1\vec{e}_1-\vec{s}_1\\
			\vdots \\
			{b}_i\vec{e}_1-\vec{s}_i\\
			\vdots \\
			{b}_n\vec{e}_1-\vec{s}_n
		\end{array}
	\right)
	\cdot
	\left(
		\begin{array}{c}
			\vec{e}_1 - \vec{a} \\
			\vdots \\
			\vec{e}_1 - \vec{a} \\
			\vdots \\
			\vec{e}_1 - \vec{a} \\
		\end{array}
	\right) \notag\\
	&\leq&
	\left(
		\begin{array}{c}
			{b}_1\vec{e}_1\\
			\vdots \\
			{b}_i\vec{e}_1\\
			\vdots \\
			{b}_n\vec{e}_1
		\end{array}
	\right)
	\cdot
	  \left(
		\begin{array}{c}
			\vec{c}_1\\
			\vdots \\
			\vec{c}_i\\
			\vdots \\
			\vec{c}_n
		\end{array}
	\right)
	+
	\frac{1}{\lambda}
	\left(
		\begin{array}{c}
			b_1\vec{e}_1-\vec{s}_1\\
			\vdots \\
			b_i\vec{e}_1-\vec{s}_i\\
			\vdots \\
			b_n\vec{e}_1-\vec{s}_n
		\end{array}
	\right)
	\cdot
	\left(
		\begin{array}{c}
			\vec{e}_1 - \vec{a} \\
			\vdots \\
			\vec{e}_1 - \vec{a} \\
			\vdots \\
			\vec{e}_1 - \vec{a} \\
		\end{array}
	\right) \notag\\
	&=& 
\sum_{i=1}^n{b}_i {C}_{1i,} + \frac{1}{\lambda}\sum_{i=1}^n \langle {b}_i \vec{e}_1,\vec{e}_1 - \vec{a} \rangle - \frac{1}{\lambda} \sum_{i=1}^n \langle \vec{s}_i,\vec{e}_1 - \vec{a} \rangle \notag\\
	&=& \sum_{i=1}^n{b}_i {C}_{1,i} + \frac{1}{\lambda}\sum_{i=1}^n{b}_i - \frac{1}{\lambda}\sum_{i=1}^n{b}_i \langle \vec{e}_1,\vec{a} \rangle -\sum_{i=1}^n\frac{1}{\lambda} \langle \vec{s}_i,\vec{e}_1 \rangle + \sum_{i=1}^n\frac{1}{\lambda} \langle \vec{s}_i,\vec{a} \rangle \notag\\
&\leq & \sum_{i=1}^n{b}_i {C}_{1,i} + \frac{1}{\lambda} + \frac{1}{\lambda}\sum_{i=1}^n \langle \vec{s}_i,\vec{a} \rangle \notag\\
&\leq& \sum_{i=1}^n{b}_i {C}_{1,i} + \frac{1}{\lambda} + \frac{1}{\lambda}.
\end{eqnarray}
$\vec{b}$ is in the probability simplex and its extreme points are $\vec{e}_1,\dots ,\vec{e}_n$. Therefore, $\sum_{i=1}^n{b}_i {C}_{1,i} \leq {C}_{1,1},\dots,{C}_{1,i},\dots,{C}_{1,n}) \leq  \|\mat{C}\|_\infty$. 

{If $\|\mat{C}\|_\infty \leq \frac{2}{\lambda}$, $C_f^\otimes + h_0 \leq \frac{8}{\lambda}$ holds.}
{Given an approximation precision constant $\epsilon$, considering (\ref{eq:ExpectedPrimalConvergence}) in {\bf Theorem \ref{thm:BlockCoordinateFrankWolfeConvergence}}, the BCFW algorithm for the semi-relaxed OT problem converges if it satisfies the following inequality.
\begin{eqnarray*}
\label{Eq:TotalCompAnaBCFW}
	\epsilon \geq \frac{2n}{k+2n} \cdot \frac{8}{\lambda} 
			\quad \Longleftrightarrow \quad  k \geq \frac{16n}{\lambda \epsilon} - 2n
			\quad \Longleftrightarrow \quad  k \geq \frac{16n}{\lambda \epsilon}.
\end{eqnarray*}
}
Hence, BCFW requires at most {$\mathcal{O}(\frac{16n}{\lambda \epsilon})$ iterations.} 

On the other hand, {If $\|\mat{C}\|_\infty > \frac{2}{\lambda}$, $h_0$ satisfies $h_0+C^\otimes_f \leq h_0 + \frac{4}{\lambda} \leq \|\mat{C}\|_\infty + \frac{2}{\lambda}+\frac{4}{\lambda}$.} Thereby, BCFW requires additional iterations $\mathcal{O}(\frac{2n\|\mat{C}\|_\infty}{\epsilon})$.

This derives the desired results. 
\end{proof}

\section{Proof of {\bf Theorem \ref{thm:RelaxedFenchelDualityGap}}}
\label{AppenSec:DualityGapEquivalence}

This section gives the complete proof of {\bf Theorem \ref{thm:RelaxedFenchelDualityGap}}.
\begin{proof}

The proof strategy follows that of \cite{AwayStepClarkson}.
We first consider the Lagrangian dual. The Lagrangian function $L(\mat{T},{\bf \Lambda},\vec{\mu})$ is defined as
\begin{equation*}
	\label{eq:Lagrangian}
	L(\mat{T},{\bf \Lambda},\vec{\mu}) = f(\mat{T})-\langle {\bf \Lambda},\mat{T} \rangle + \sum_{i=1}^n  \mu_i \left(\sum_{j=1}^mt_{j,i} - b_i\right),
\end{equation*}
where ${\bf \Lambda} =(\vec{\lambda}_1, \vec{\lambda}_2, \ldots, \vec{\lambda}_n)\geq \mat{0}$ and $\vec{\mu}$ are dual variables. Then, the Lagrangian dual function is defined as
\begin{eqnarray*}
	\label{eq:duality_Lagrangian}
	g({\bf \Lambda},\vec{\mu}) = \inf_{\scriptsize {\mat{T}}}\ L(\mat{T},{\bf \Lambda},\vec{\mu}).
\end{eqnarray*}
The Lagrangian function $L$ is also convex because of convexity of $f$. The gradient of Lagrangian at $\mat{T}$ is provided as
\begin{eqnarray*}
	\nonumber \nabla_{\scriptsize \mat{T}}L(\mat{T},{\bf \Lambda},\vec{\mu}) = \nabla f(\mat{T}) -(\vec{\lambda}_1, \vec{\lambda}_2, \ldots, \vec{\lambda}_n)^T+(\mu_1\vec{1}_m,\dots \mu_n\vec{1}_m)^T.
\end{eqnarray*}
Considering the case that  the gradient (\ref{eq:duality_Lagrangian}) is equal to \vec{0}, ${\bf \Lambda}$ satisfies $\vec{\lambda}_i = \nabla_if(\mat{T}) + \mu_i\vec{1}_m$. Substituting ${\bf \Lambda}$ into the Lagrangian dual function yeilds
\begin{eqnarray*}
	g(\mat{T},\vec{\mu}) =f(\mat{T}) - \sum_{i=1}^n\langle \nabla_if(\mat{T}),\vec{t}_i \rangle - \sum_{i=1}^n\mu_i{b}_i.
\end{eqnarray*}
Thus, the Lagrangian dual problem, denoted as $w(\mat{T})$, is defined as
\begin{eqnarray*}
	w(\mat{T}) &:=& \defmax_{\scriptsize {\bf \Lambda} \geq \mat{0}}\ g({\bf \Lambda},\vec{\mu}).
\end{eqnarray*}
This is written as
\begin{eqnarray*}
	\max_{{\scriptsize{{\bf T} \in \mathbb{R}^{m \times n}, {\mu} \in \mathbb{R}^n}}} && f(\mat{T}) - \sum_{i=1}^n\langle \vec{t}_i , \nabla_if(\mat{T}) \rangle - \sum_{i=1}^n\mu_i{b}_i\\
 {\rm subject\ to} &&-\mu_i \leq  \min_{j \in [m]} (\nabla_i f(\mat{T}))_j,\quad\quad \forall i \in [n].
\end{eqnarray*}
Denoting $-{\mu_i(\mat{T})}$ as $\min_{j \in [m]} (\nabla_i f(\mat{T}))_j $, the Lagrangian dual problem is derived as
\begin{equation*}
	\defmax_{{\scriptsize{\mat{T} \in \mathbb{R}^{m \times n}}}}\ f(\mat{T}) - \sum_{i=1}^n\langle \vec{t}_i , \nabla_if(\mat{T}) \rangle + \sum_{i=1}^{n}{b}_i\min_{j \in [m]} (\nabla_i f(\mat{T}))_j.
\end{equation*}
%
Now, denoting the Lagrangian gap as $g_L(\mat{T})$, the Lagrange duality gap is represented as
\begin{eqnarray*}	
g_{L}(\mat{T})&=&f(\mat{T})-w(\mat{T}) \\
	&=& \sum_{i=1}^n \langle \vec{t}_i,\nabla_i f(\mat{T})\rangle - \sum_{i=1}^n{b}_i\defmin_{j \in [m]} (\nabla_i f(\mat{T}))_j \\
&=&  \sum_{i=1}^n \langle \vec{t}_i,\nabla_i f(\mat{T})\rangle - \sum_{i=1}^n\defmin_{\vec{s}_i \in b_i\Delta_m}\langle \vec{s}_i ,\nabla_i f(\mat{T}) \rangle \\
&=& \sum_{i=1}^n\defmax_{\vec{s}_i \in b_i\Delta_m} \langle \vec{t}_i -\vec{s}_i , \nabla_if(\mat{T}) \rangle \\
&=& \defmax_{\scriptsize{\mat{S}' \in \mathcal{M}}} \langle \mat{T} - \mat{S}, \nabla f(\mat{T}) \rangle \\
&=& g(\mat{T}).
\end{eqnarray*}
As seen, we confirm that the Lagrange duality $g_{L}(\mat{T})$ is equal to the mineralization duality gap $g(\mat{T})$.

This completes the proof.
\end{proof}

\section{Proof of {\bf Theorem \ref{Thm:ComplexityBCFW-GA-SROT}} }
\label{AppenSec:BCFWGAS}

We give a convergence analysis {\bf Theorem \ref{Thm:ComplexityBCFW-GA-SROT}}  of BCFW-GA, which is a straightforward extension to the semi-relaxed OT problem from that of the structured SVM problem in \cite{Osokin_ICML_2016}). For this purpose, the following is given:

\begin{Def}[{\bf Definition 5} in \cite{Osokin_ICML_2016}]
\label{def:nonuni_measure}
The non-uniformity measure $\chi(\vec{x})$ of a vector $\vec{x} \in \mathbb{R}^{+}_n$ is defined as $\chi(\vec{x}) = \sqrt{1+n^2{\mathrm Var}[\vec{p}]}$, where $\vec{p} = \frac{\vec{x}}{\|\vec{x}\|_1}$ is the probability vector obtained by normalizing $\vec{x}$.
\end{Def}
In addition, the non-uniformity measure holds as follows.
\begin{Lem}({\bf Lemma 6} in \cite{Osokin_ICML_2016} and its remark)
\label{Lem:Lemma6inOsokin}
For $\vec{x} \in \mathbb{R}^n_+$, we have $\|\vec{x}\|_2 = \frac{\chi(\vec{x})}{\sqrt{n}}\|\vec{x}\|_1$. Therefore, the non-uniformity measure $\chi(\vec{x})$ satisfies $\chi(\vec{x})$ is in $[1,\sqrt{n}]$.
\end{Lem}

Then, the following theorem is given by extending Theorem 2 in \cite{Osokin_ICML_2016}.
\begin{Thm}(Convergence analysis of Algorithm \ref{AppenAlg:BCFW-GA-SROT} \cite{Osokin_ICML_2016})
\label{Thm:ConvAnaBCFW-GA-SROT}
Let $\mat{T}^*$ be the optimal solution of the semi-relaxed OT problem. Consider Algorithm \ref{AppenAlg:BCFW-GA-SROT}.
Let $C_f^{(:)}$ be $(C_f^{(1)},C_f^{(2)},\ldots,C_f^{(n)})^T$, where $C_f^{(i)}$ is 
defined in {Definition \ref{def:CurvatureBCFW}}. Also, denote $g_:(\mat{T}^{(k)})$ as $(g_1,\dots,g_n)^T$ as defined in (\ref{AppenEq:Gi}). Then, for each $k \geq 1$, the iterates of $\mat{T}^{(k)}$ of Algorithm \ref{AppenAlg:BCFW-GA-SROT} satisfies
\begin{equation}
\label{eq:GapSamplingInequality}
\mathbb{E}[f(\mat{T}^{(k)})] - f(\mat{T}^*) \leq \frac{2n}{k+2n}(C^\otimes_f\chi^\otimes + h_0),
\end{equation}
where the constant $\chi^\otimes$ is an upper bound on $\mathbb{E}[\displaystyle{\frac{\chi(C_f^{(:)})}{\chi(g_:(\mat{T}^{(k)}))^3}}]$.  Here, the expectation is taken with respect to the random selection of the blocks at the $k$-th iteration of the algorithm. 
\end{Thm}

\begin{proof}
The proof is identical to that of Theorem 2 in \cite{Osokin_ICML_2016}. Thus, we omit it. 
\end{proof}

We are ready to prove {\bf Theorem \ref{Thm:ComplexityBCFW-GA-SROT}} of Algorithm \ref{AppenAlg:BCFW-GA-SROT} using {\bf Theorem \ref{Thm:ConvAnaBCFW-GA-SROT}} and {\bf Lemma \ref{Lem:Lemma6inOsokin}}.

\begin{proof}
The proof mostly follows that of {\bf Theorem \ref{thm:relaxedBCFWconvergence}}. 

We first recall $C^\otimes_f \leq \frac{4}{\lambda}$ in (\ref{Eq:CotimesUpperBound}), 	and $h_0 \leq g(\mat{T}^{(0)}) \leq \| \mat{C}\|_{\infty} + \frac{2}{\lambda}$ in (\ref{Eq:GtUpperBound}). 
Furthermore, we also note the range of $\chi(\vec{x})$ for $\vec{x} \in \mathbb{R}^n_+$ as in {\bf Lemma \ref{Lem:Lemma6inOsokin}}, i.e., $1 \leq \chi(\vec{x}) \leq \sqrt{n}$. $\chi(\vec{x})=1$ 
corresponds to the uniform case whereas $\chi(\vec{x})\approx \sqrt{n}$ happens when the distribution is non-uniform. 
Then, considering $\chi^\otimes$ is an upper bound on ${\displaystyle \mathbb{E}[\frac{\chi(C_f^{(:)})}{\chi(g_:(\mat{T}^{(k)}))^3}]}$, we have the best case below when $\chi(C_f^{(:)})=1$ and $\chi(g_:(\mat{T}^{(k)})) \approx \sqrt{n}$ for the non-uniform case as
\[
	\chi^\otimes \leq \frac{1}{(\sqrt{n})^3}.
\]
On the other hand, we have the worst case below when $\chi(C_f^{(:)})=\sqrt{n}$ and $\chi(g_:(\mat{T}^{(k)})) = 1$ for the uniform case 
\[
\chi^\otimes \leq \sqrt{n}.
\]
Therefore, if $\|\mat{C}\|_\infty \leq \frac{2}{\lambda}$, we have 
\[
	C^\otimes_f\chi^\otimes + h_0 = \frac{4}{\lambda} \frac{1}{(\sqrt{n})^3} + \left(\frac{2}{\lambda} + \frac{2}{\lambda}\right) =  \frac{4}{\lambda}\left(\frac{1}{n\sqrt{n}} + 1\right),
\]
and
\[
	C^\otimes_f\chi^\otimes + h_0 = \frac{4}{\lambda} \sqrt{n} + \left(\frac{2}{\lambda} + \frac{2}{\lambda}\right) =  \frac{4}{\lambda}(\sqrt{n} + 1).
\]	
Consequently, from (\ref{eq:GapSamplingInequality}), {Algorithm \ref{AppenAlg:BCFW-GA-SROT}} satisfies in the best case
\begin{eqnarray}
\label{Eq:TotalCompBCFG-GA-best}
\epsilon \geq \frac{2n}{k+2n} \cdot \frac{4}{\lambda} \left(\frac{1}{n\sqrt{n}}+1 \right)
			 \Longleftrightarrow   k \geq \frac{8n}{\lambda \epsilon}+\frac{8}{\epsilon \lambda \sqrt{n}} - 2n
			 \overset{\rm approx.}{\Longleftrightarrow}   k \geq \frac{8n}{\lambda \epsilon}+ \underbrace{\frac{8}{\epsilon \lambda \sqrt{n}}}_{(*)}
			 \overset{\rm approx.}{\Longleftrightarrow}  k \geq \frac{8n}{\lambda \epsilon},
\end{eqnarray}	
and also has in the worst case 
\begin{eqnarray*}
\label{Eq:TotalCompBCFG-GA-worst}
\epsilon \geq \frac{2n}{k+2n} \cdot \frac{4}{\lambda} (\sqrt{n}+1)
			 \Longleftrightarrow   k \geq \frac{8n}{\lambda \epsilon}+\frac{8n\sqrt{n}}{\epsilon \lambda } - 2n
			 \overset{\rm approx.}{\Longleftrightarrow}   k \geq \frac{8n}{\lambda \epsilon}+\frac{8n\sqrt{n}}{\lambda \epsilon }
			 \overset{\rm approx.}{\Longleftrightarrow} k \geq \frac{8n\sqrt{n}}{\lambda \epsilon }.
\end{eqnarray*}
On the other hand, if  $\|\mat{C}\|_\infty > \frac{2}{\lambda}$, $h_0$ is bounded on $\|\mat{C}\|_\infty + \frac{2}{\lambda}$. Therefore, we have 
\[
	C^\otimes_f\chi^\otimes + h_0 = \frac{4}{\lambda} \frac{1}{(\sqrt{n})^3} + \left(\|\mat{C}\|_\infty + \frac{2}{\lambda}\right) ,
\]
and
\[
	C^\otimes_f\chi^\otimes + h_0 = \frac{4}{\lambda} \sqrt{n} + \left(\|\mat{C}\|_\infty + \frac{2}{\lambda}\right).
\]
Similarly, from (\ref{Eq:TotalCompBCFG-GA-best}), {Algorithm \ref{AppenAlg:BCFW-GA-SROT}} also satisfies in the best case
\begin{eqnarray*}
	k \geq \frac{4n}{\epsilon\lambda} + \frac{2n\|\mat{C}\|_\infty}{\epsilon}
\end{eqnarray*}
and also has in the worst case
\begin{eqnarray*}
	k \geq \frac{8n\sqrt{n}}{\epsilon \lambda} + \frac{2n\|\mat{C}\|_\infty}{\epsilon}.
\end{eqnarray*}

This completes the proof. 
\end{proof}

\clearpage
\section{Additional numerical evaluations}
\label{AppenSec:AddNumEval}

This section additionally presents numerical results. We used two publicly available images ``Grafiti" by Jon Ander and ``Rainbow Bridge National Monument Utah" by Bernard Sprangg. All the configurations of the experiments are the same as those in Section \ref{Sec:NumericalEvalations}.

\begin{figure}[htbp]
\begin{center}
	\begin{minipage}[t]{0.32\textwidth}
	\begin{center}
		\includegraphics[width=\textwidth]{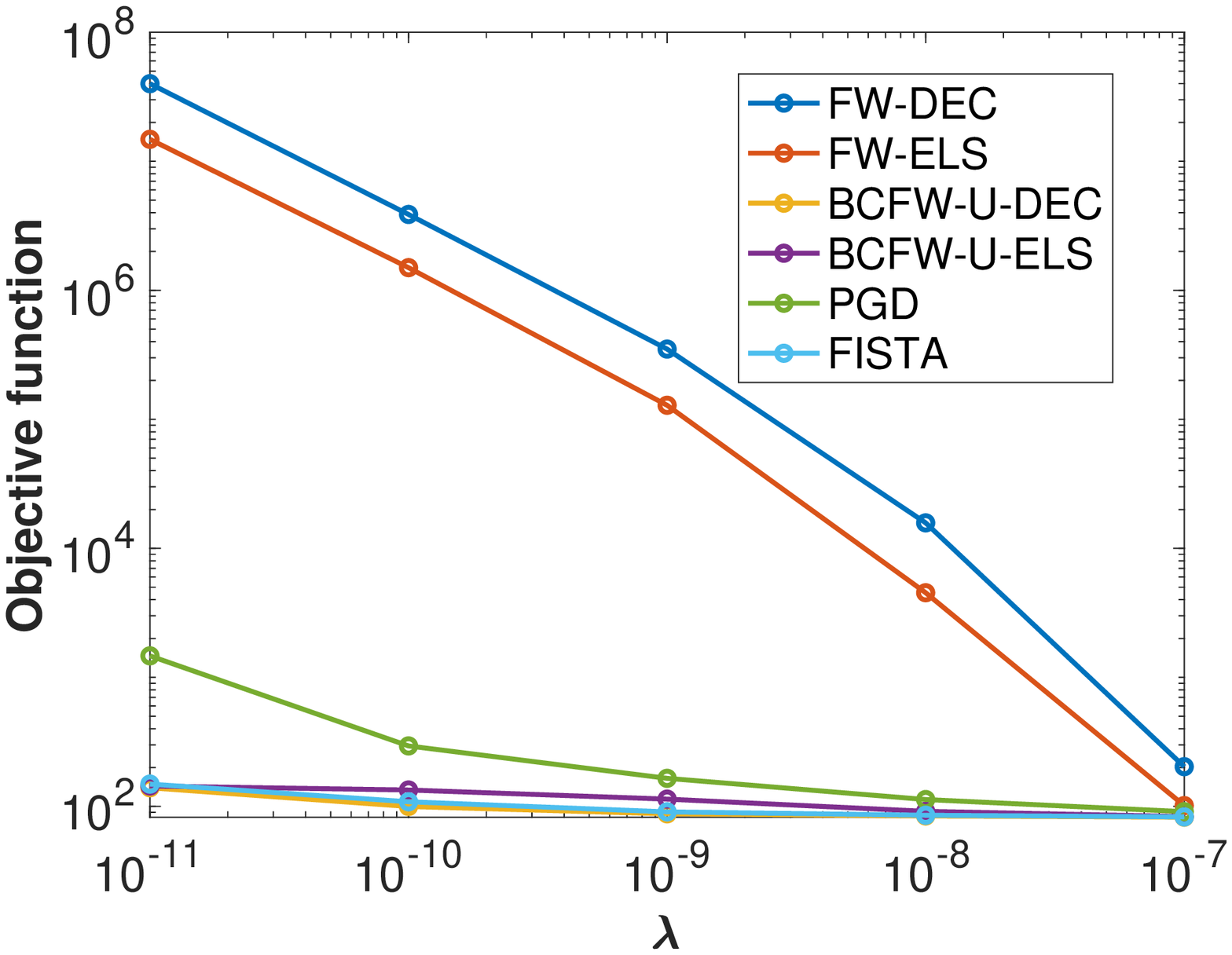}\\
		
		{\footnotesize  (a) objective value : $f(\mat{T})$}
		
	\end{center} 
	\end{minipage}
	\hspace*{-0.2cm}		
	\begin{minipage}[t]{0.32\textwidth}
	\begin{center}
		\includegraphics[width=\textwidth]{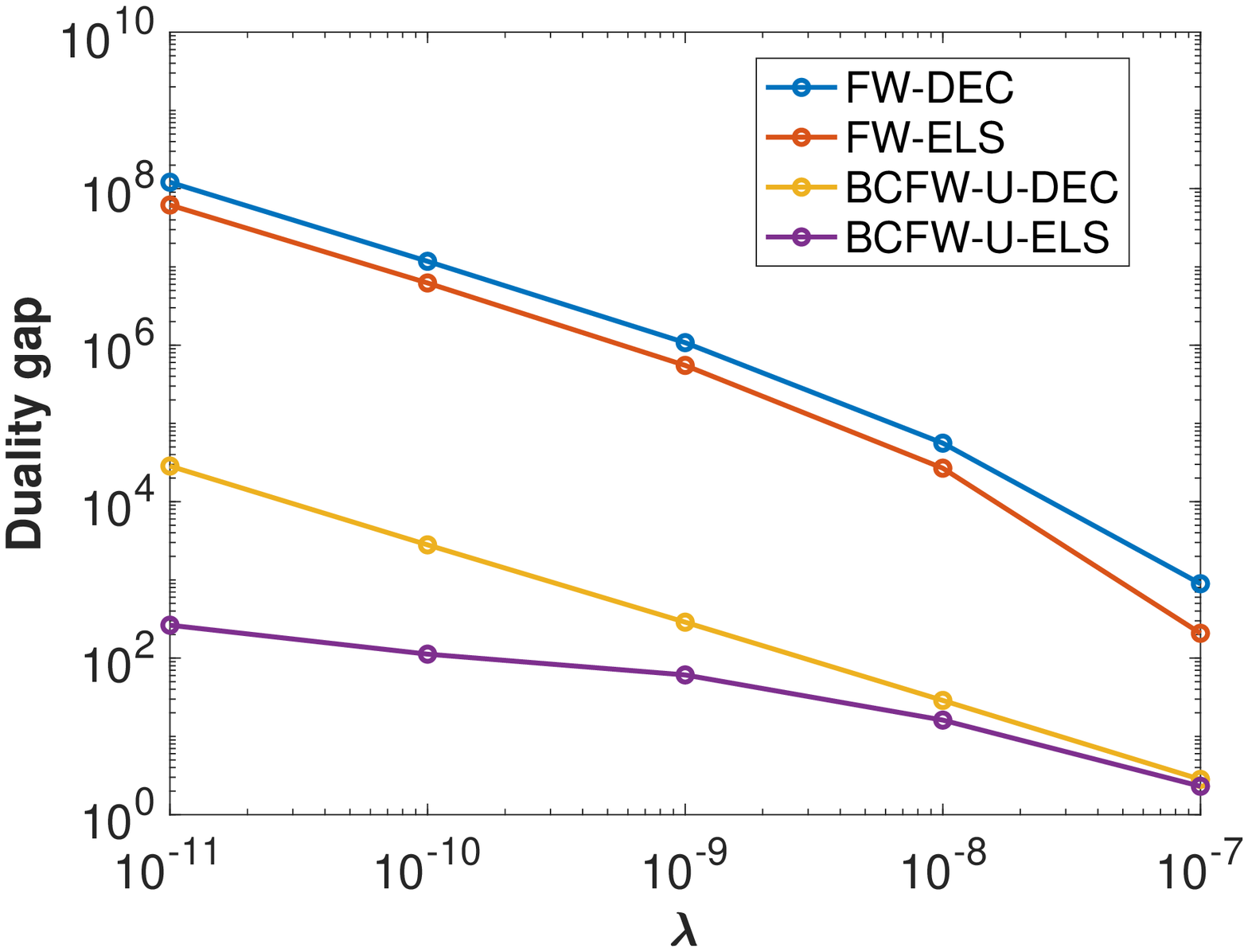}\\
		
		{\footnotesize  (b) duality gap : $g(\mat{T})$}
		
	\end{center} 
	\end{minipage}
	\hspace*{-0.2cm}
	\begin{minipage}[t]{0.32\textwidth}
	\begin{center}
		\includegraphics[width=\textwidth]{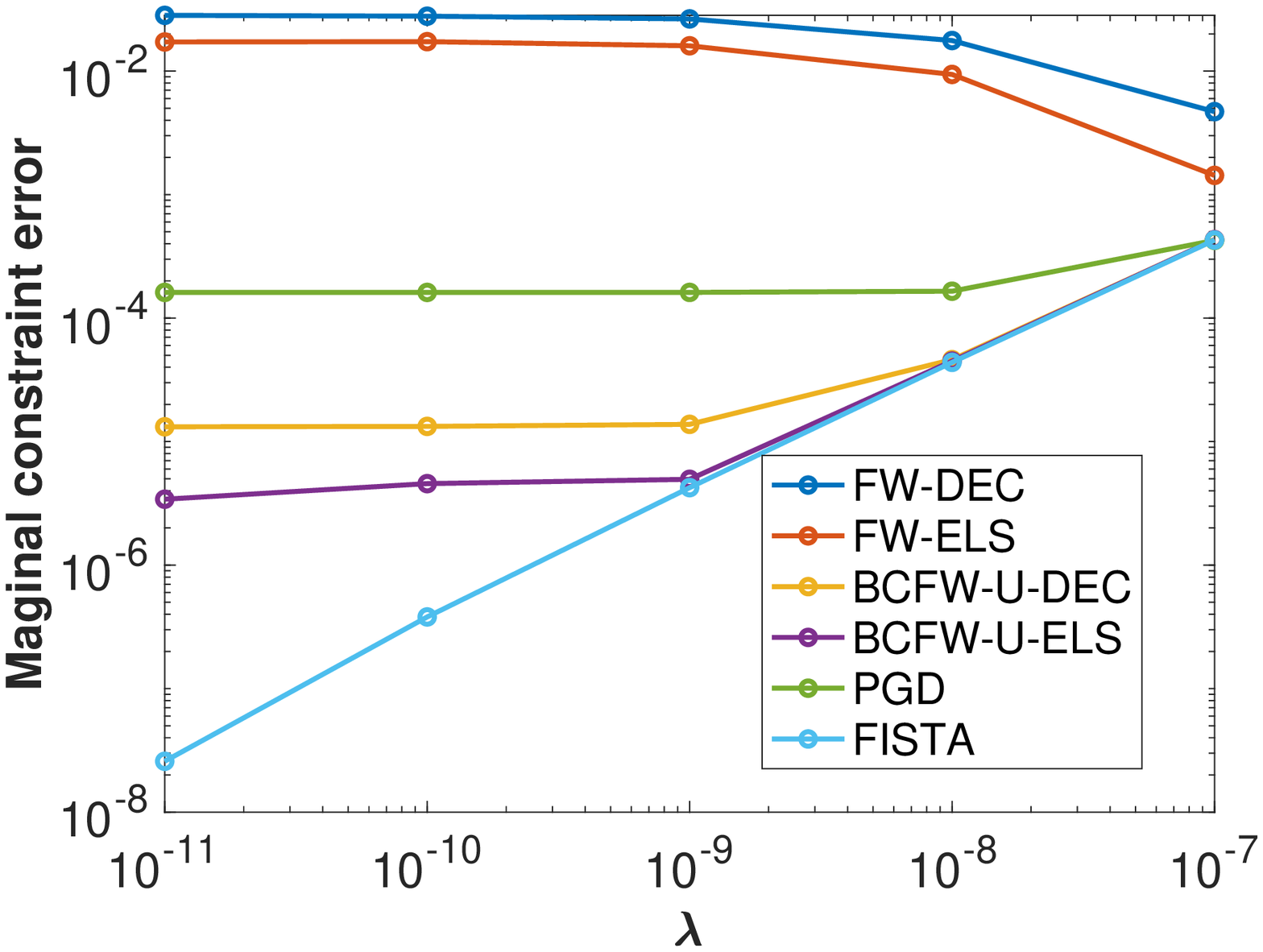}\\
		
		{\footnotesize  (c) marginal constraint error : $e_c$}
		
	\end{center} 
	\end{minipage}
	\vspace*{0.3cm}
	
	\begin{minipage}[t]{0.32\textwidth}
	\begin{center}
		\includegraphics[width=\textwidth]{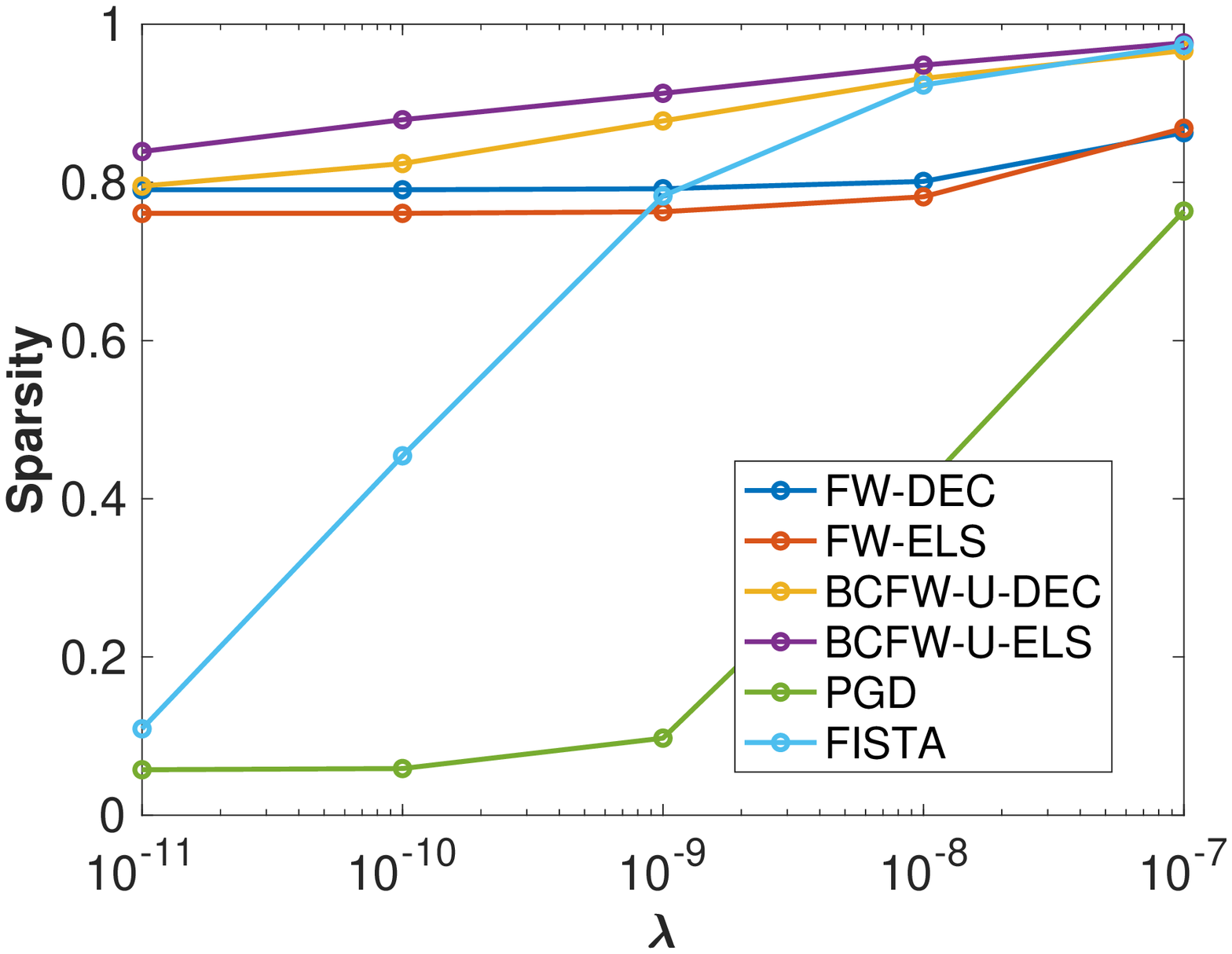}\\
		
		{\footnotesize  (d) sparsity}
		
	\end{center} 
	\end{minipage}
	\hspace*{-0.2cm}		
	\begin{minipage}[t]{0.32\textwidth}
	\begin{center}
		\includegraphics[width=\textwidth]{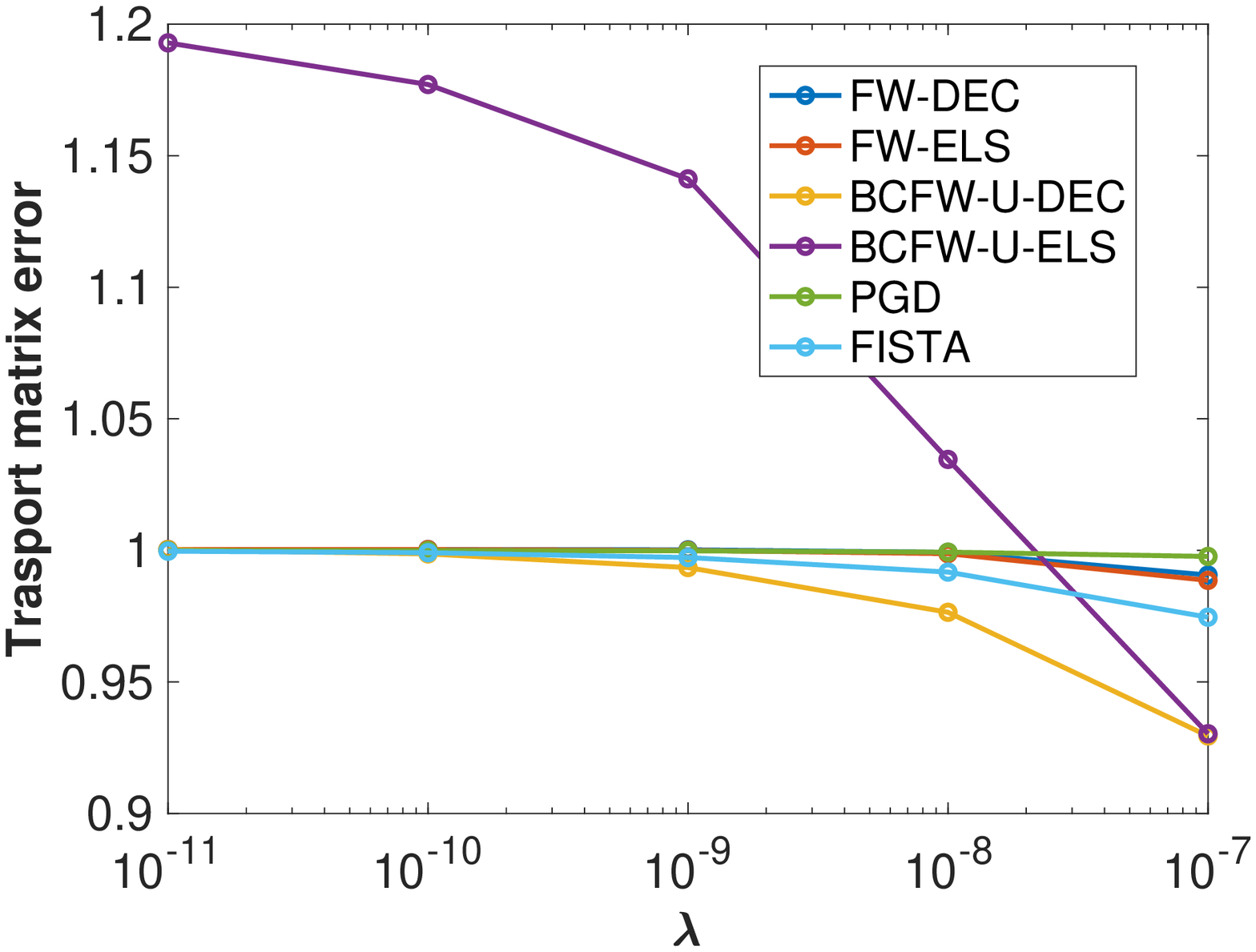}\\
		
		{\footnotesize  (e) matrix error : $e_M$}
		
	\end{center} 
	\end{minipage}	
	\hspace*{-0.2cm}			
	\begin{minipage}[t]{0.32\textwidth}
	\begin{center}
		\includegraphics[width=\textwidth]{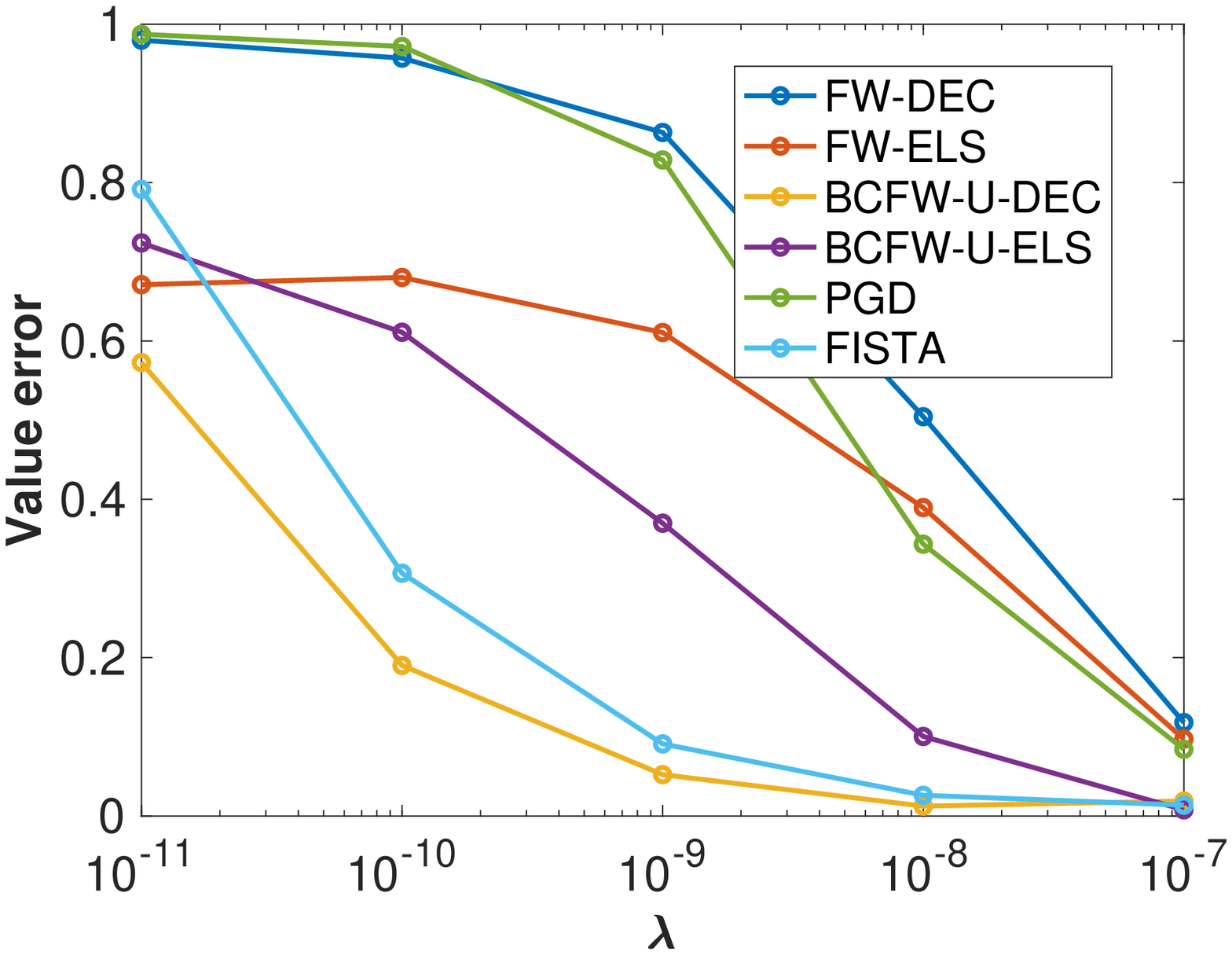}\\
		
		{\footnotesize  (f) value error : $e_v$}
		
	\end{center} 
	\end{minipage}
	\vspace*{0.3cm}	

	\begin{minipage}[t]{0.32\textwidth}
	\begin{center}
		\includegraphics[width=\textwidth]{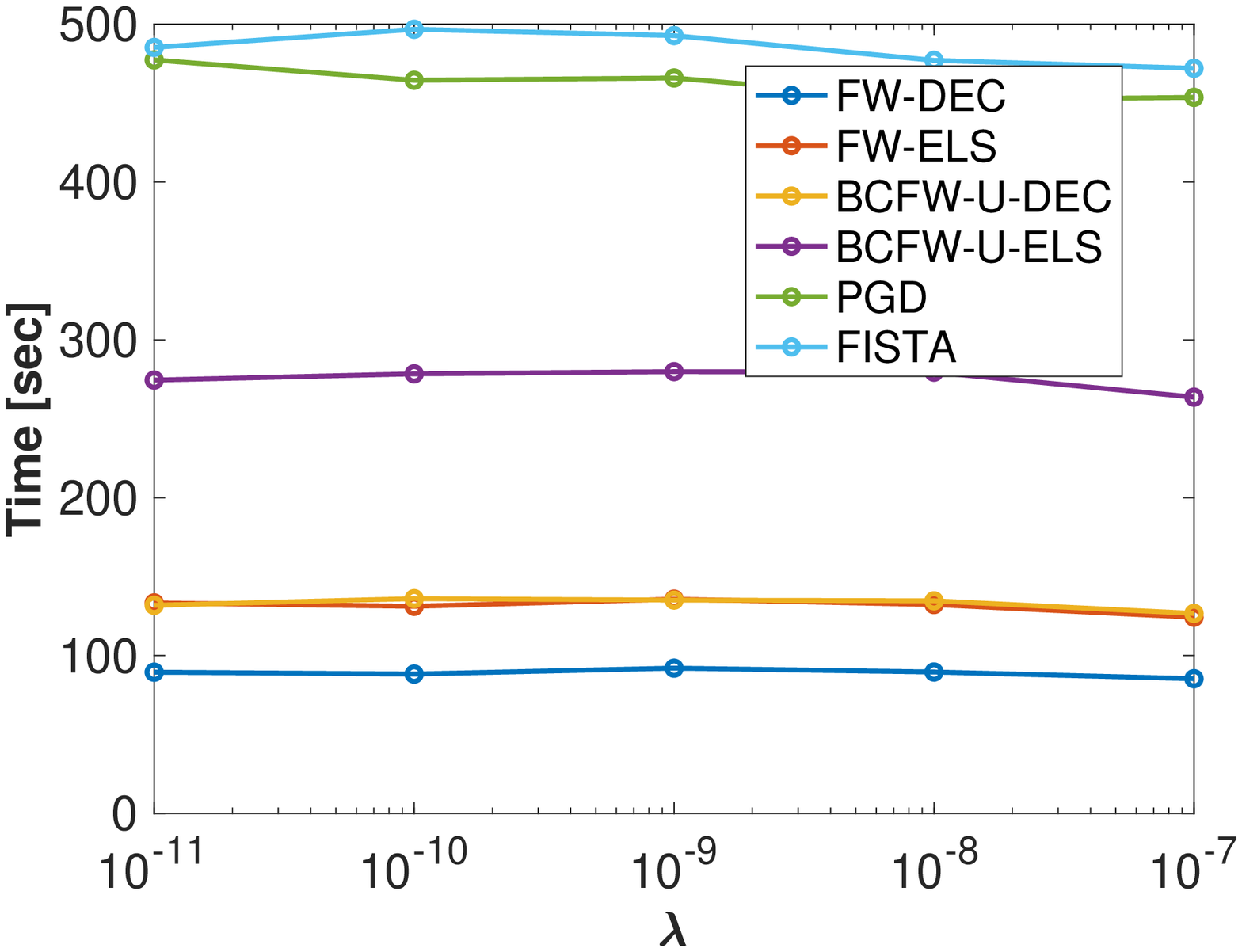}\\
		
		{\footnotesize  (g) computational time}
		
	\end{center} 
	\end{minipage}
	\hspace*{-0.2cm}	
	\begin{minipage}[t]{0.32\textwidth}
	\begin{center}
	\end{center} 
	\end{minipage}
	\hspace*{-0.2cm}	
	\begin{minipage}[t]{0.32\textwidth}
	\begin{center}
	\end{center} 
	\end{minipage}	
\caption{Evaluations on different relaxation parameters $\lambda$ (corresponding to Figure \ref{fig:PerformanceLambda}).}
\label{Appenfig:PerformanceLambda}
\end{center}		
\end{figure}

\begin{figure}[t]
\begin{center}
	\begin{minipage}[t]{0.32\textwidth}
	\begin{center}
		\includegraphics[width=\textwidth]{results/conv_comp_bcfw_pgd/conv_autumn_comunion_N4096_lambda1e-07_cost.eps}\\
		
		{\footnotesize  (a) objective value : $f(\mat{T})$ }
		
	\end{center} 
	\end{minipage}
	\hspace*{-0.2cm}	
	\begin{minipage}[t]{0.32\textwidth}
	\begin{center}
		\includegraphics[width=\textwidth]{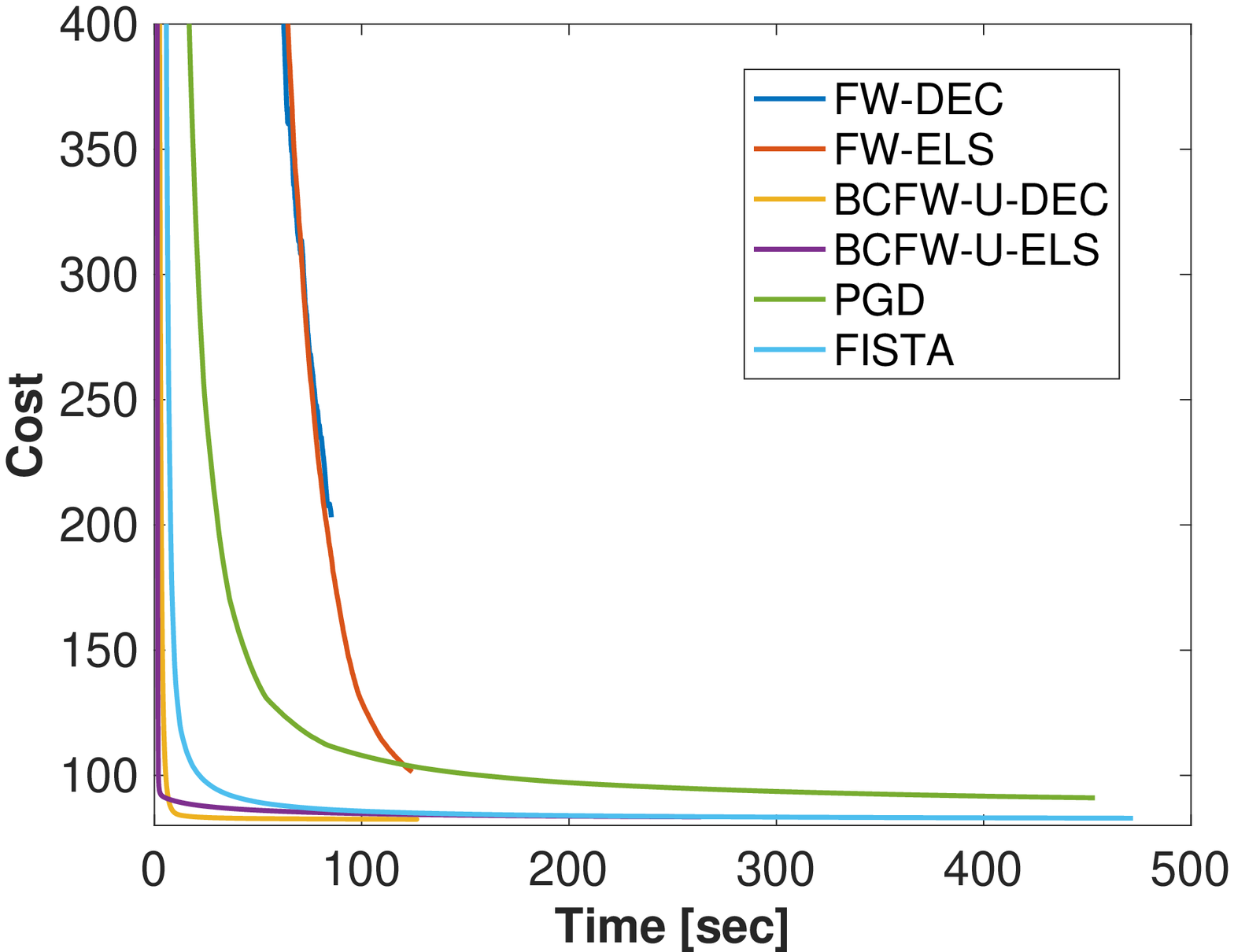}\\
		
		{\footnotesize  (b) objective value (time): $f(\mat{T})$ }
		
	\end{center} 
	\end{minipage}
	\hspace*{-0.2cm}	
	\begin{minipage}[t]{0.32\textwidth}
	\begin{center}
		\includegraphics[width=\textwidth]{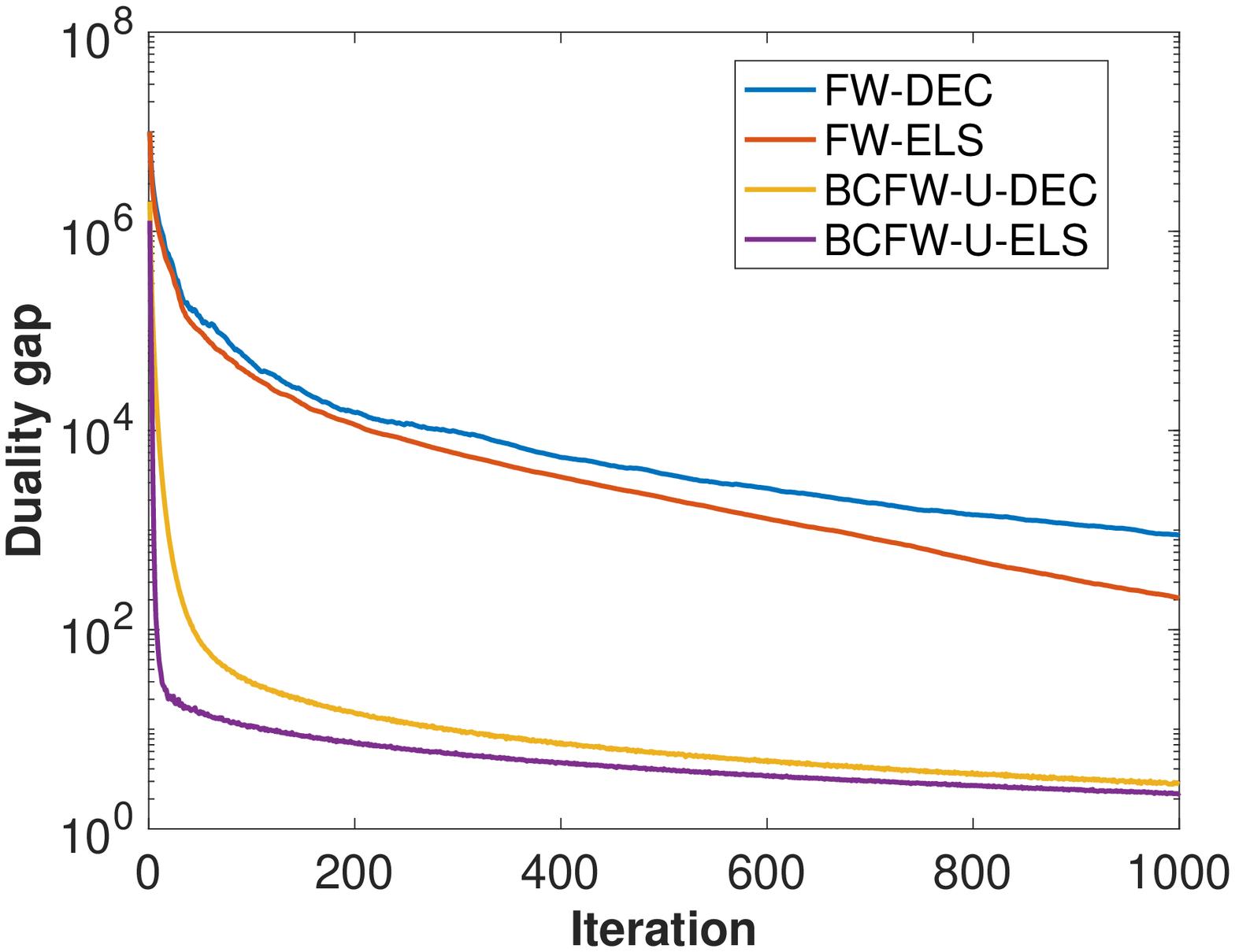}\\
		
		{\footnotesize  (c) duality gap : $g(\mat{T})$}
		
	\end{center} 
	\end{minipage}
	\vspace*{0.2cm}

	\begin{minipage}[t]{0.32\textwidth}
	\begin{center}
		\includegraphics[width=\textwidth]{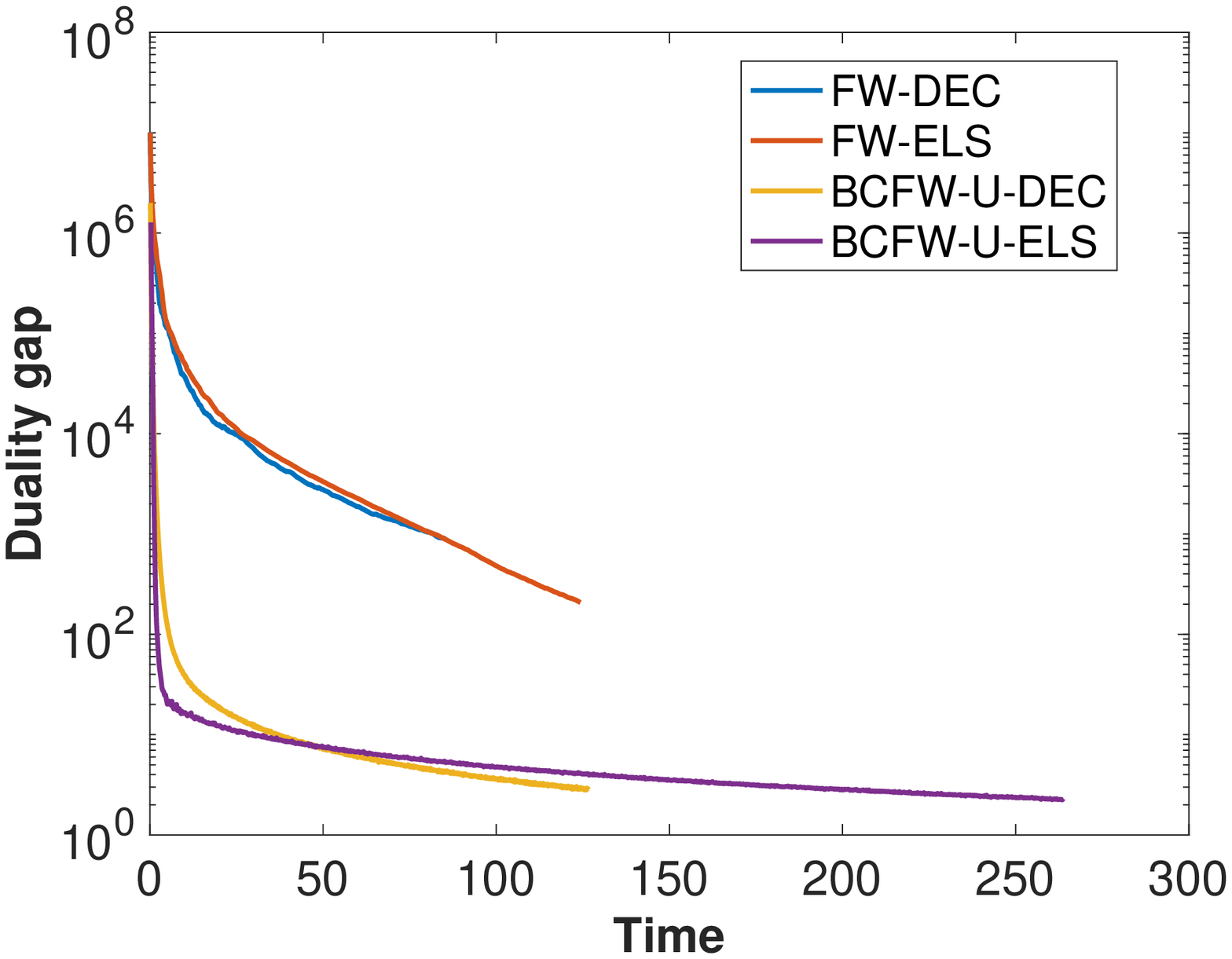}\\
		
		{\footnotesize  (d) duality gap (time) : $g(\mat{T})$}
		
	\end{center} 
	\end{minipage}
	\hspace*{-0.2cm}	
	\begin{minipage}[t]{0.32\textwidth}
	\begin{center}
		\includegraphics[width=\textwidth]{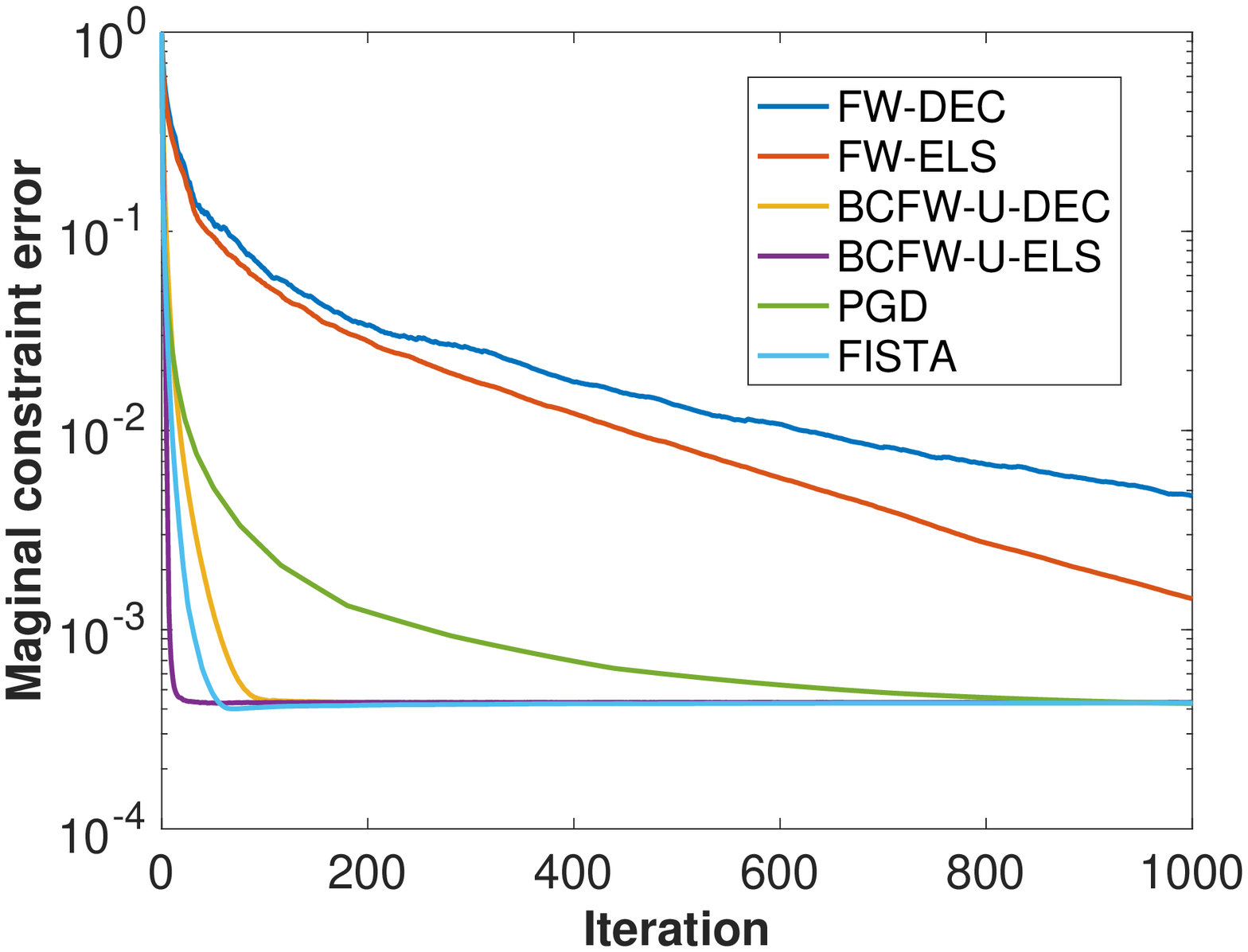}\\
		
		{\footnotesize  (e) marginal constraint error : $e_{c}$}
		
	\end{center} 
	\end{minipage}
	\hspace*{-0.2cm}
	\begin{minipage}[t]{0.32\textwidth}
	\begin{center}
		\includegraphics[width=\textwidth]{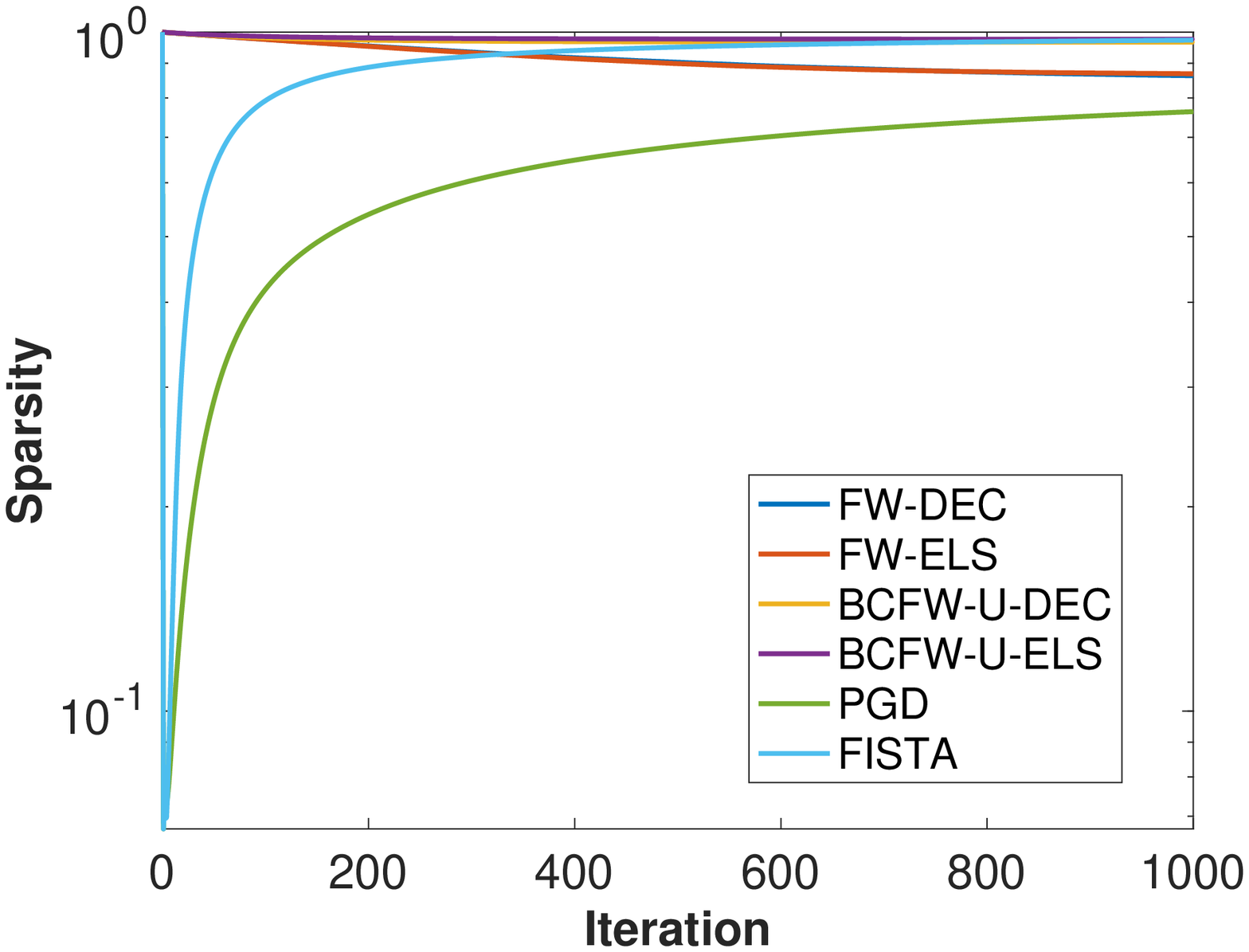}\\
		
		{\footnotesize  (f) sparsity}
		
	\end{center} 
	\end{minipage}
	\vspace*{0.2cm}
		
	\begin{minipage}[t]{0.32\textwidth}
	\begin{center}
		\includegraphics[width=\textwidth]{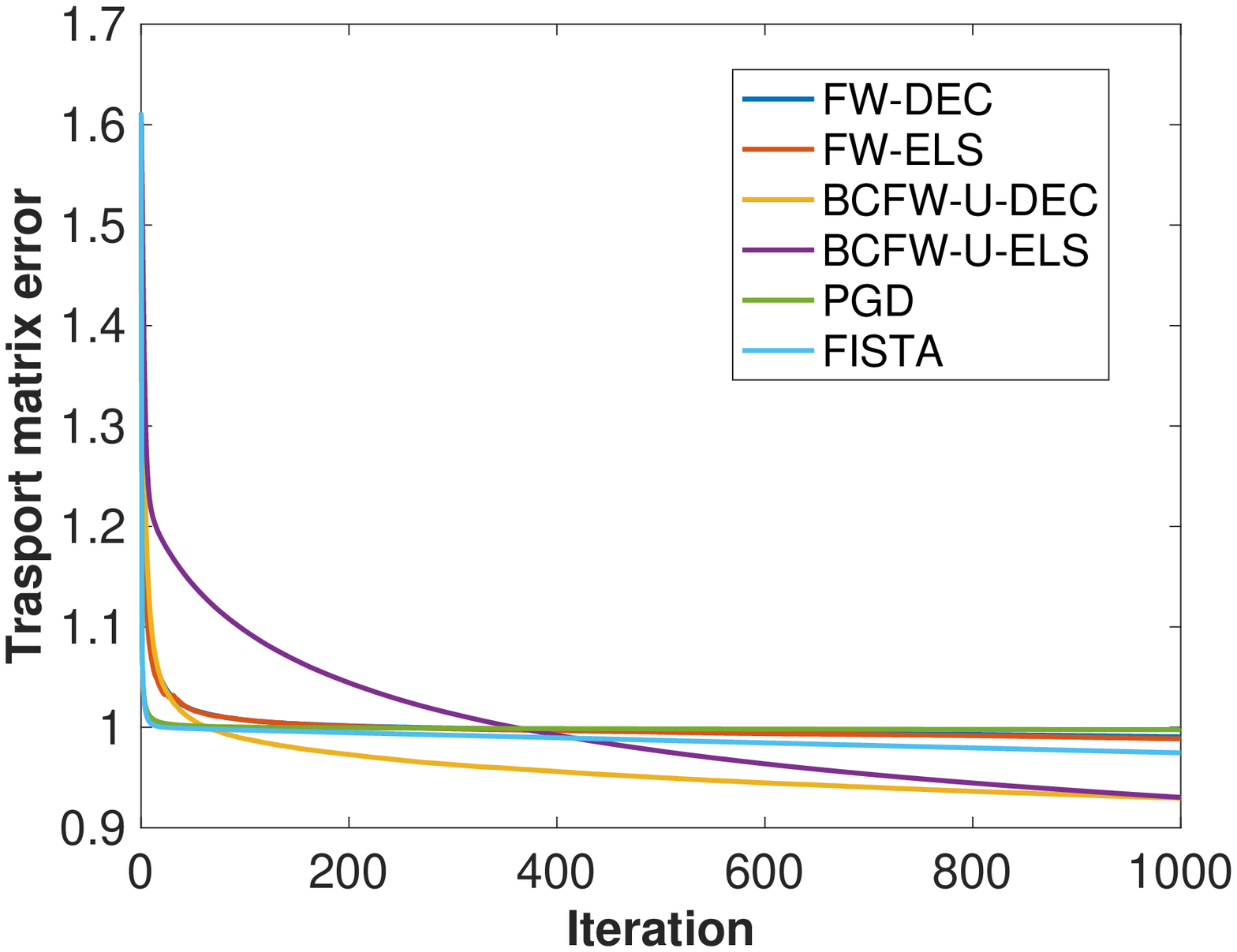}\\
		
		{\footnotesize  (g) matrix error : $e_{m}$ }
		
	\end{center} 
	\end{minipage}
	\hspace*{-0.2cm}
	\begin{minipage}[t]{0.32\textwidth}
	\begin{center}
		\includegraphics[width=\textwidth]{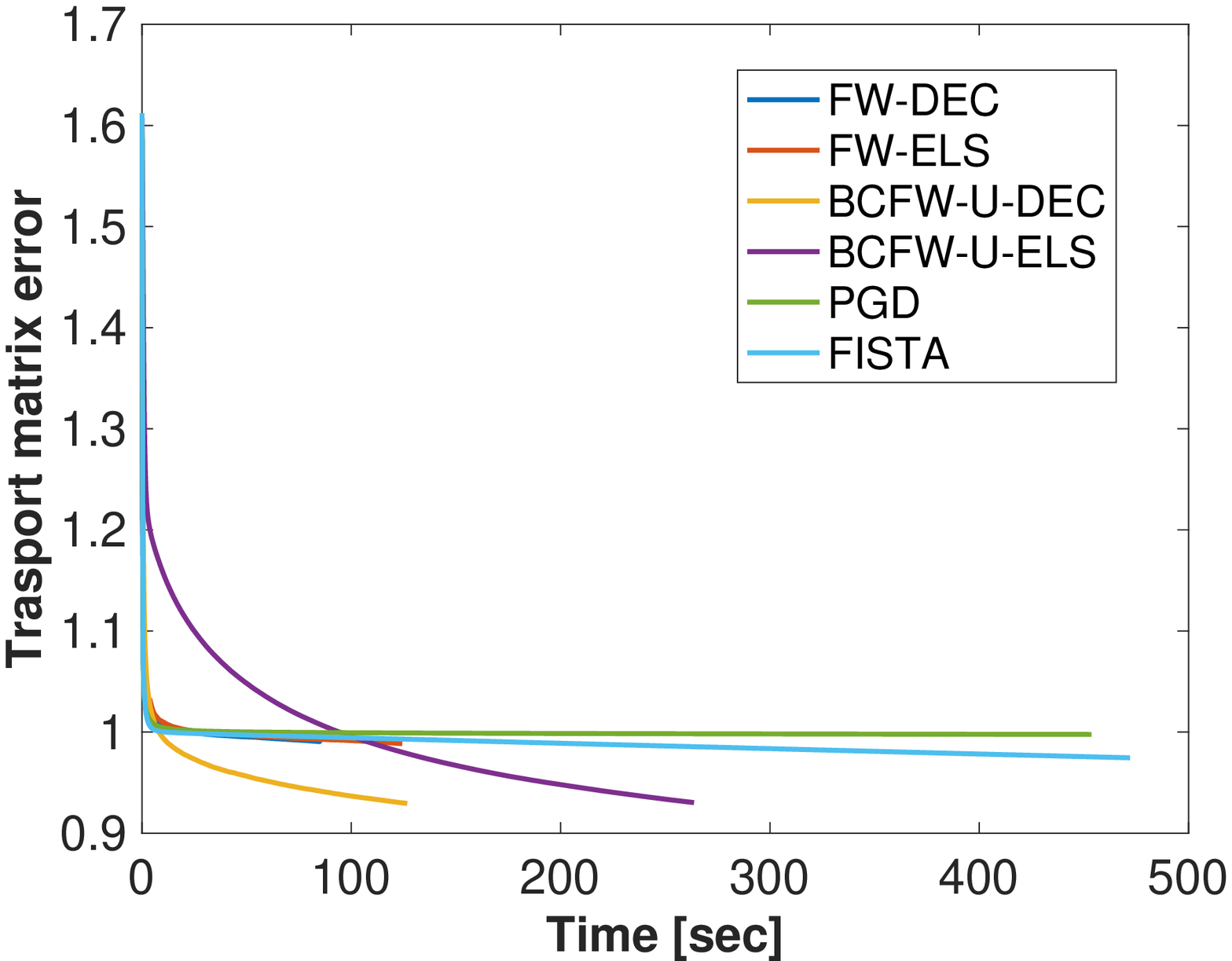}\\
		
		{\footnotesize  (h) matrix error (time) : $e_{m}$ }
		
	\end{center} 
	\end{minipage}
	\hspace*{-0.2cm}	
	\begin{minipage}[t]{0.32\textwidth}
	\begin{center}
		\includegraphics[width=\textwidth]{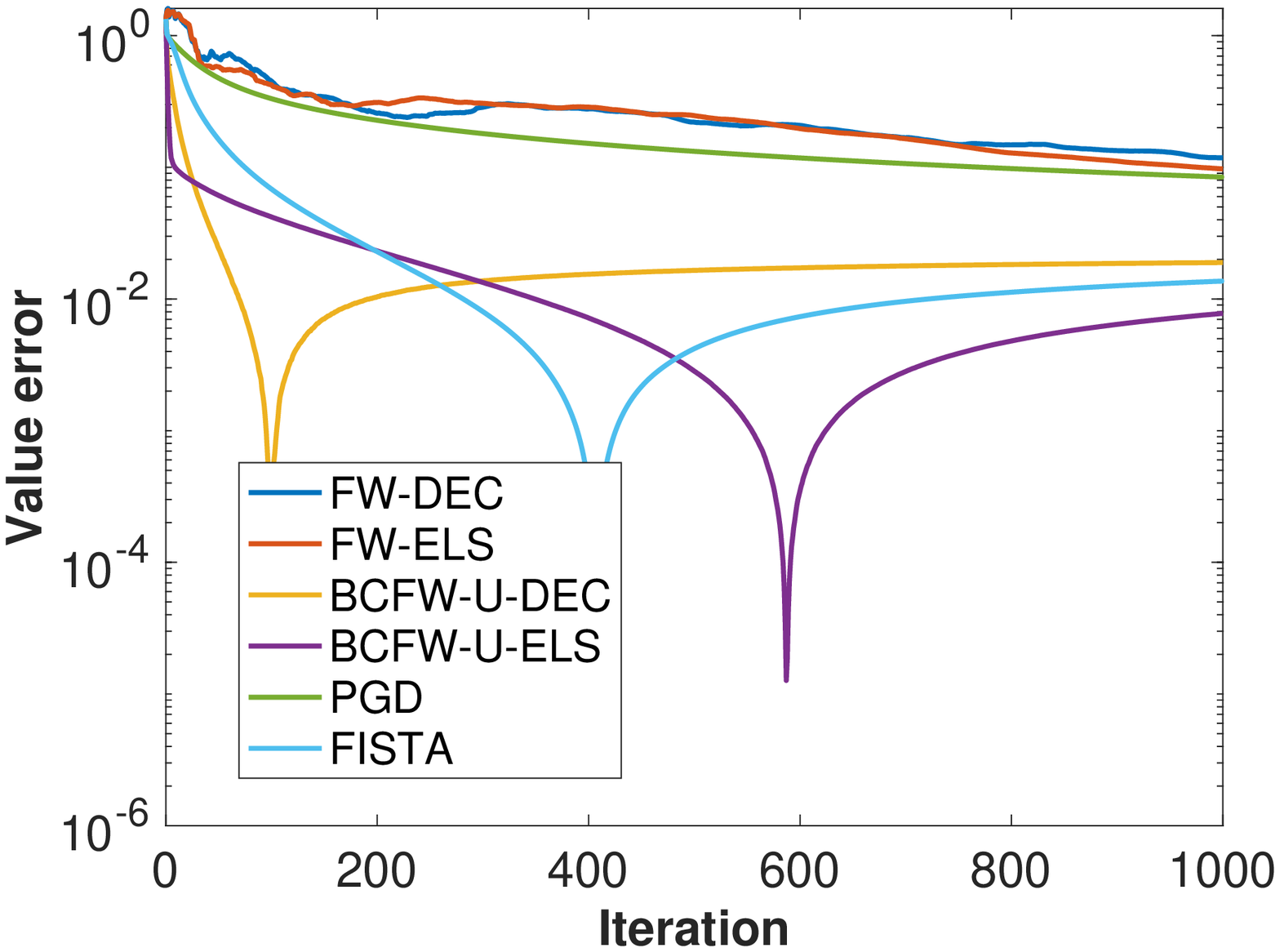}\\
		
		{\footnotesize  (i) value error : $e_v$}
		
	\end{center} 
	\end{minipage}		
\caption{Evaluations on convergence ($\lambda = 10^{-7}$) (corresponding to Figure \ref{fig:ConvergencePerformance}).}
\label{Appenfig:ConvergencePerformance}
\end{center}
\end{figure}

\begin{figure}[t]
\begin{center}
	\begin{minipage}[t]{0.32\textwidth}
	\begin{center}
		\includegraphics[width=\textwidth]{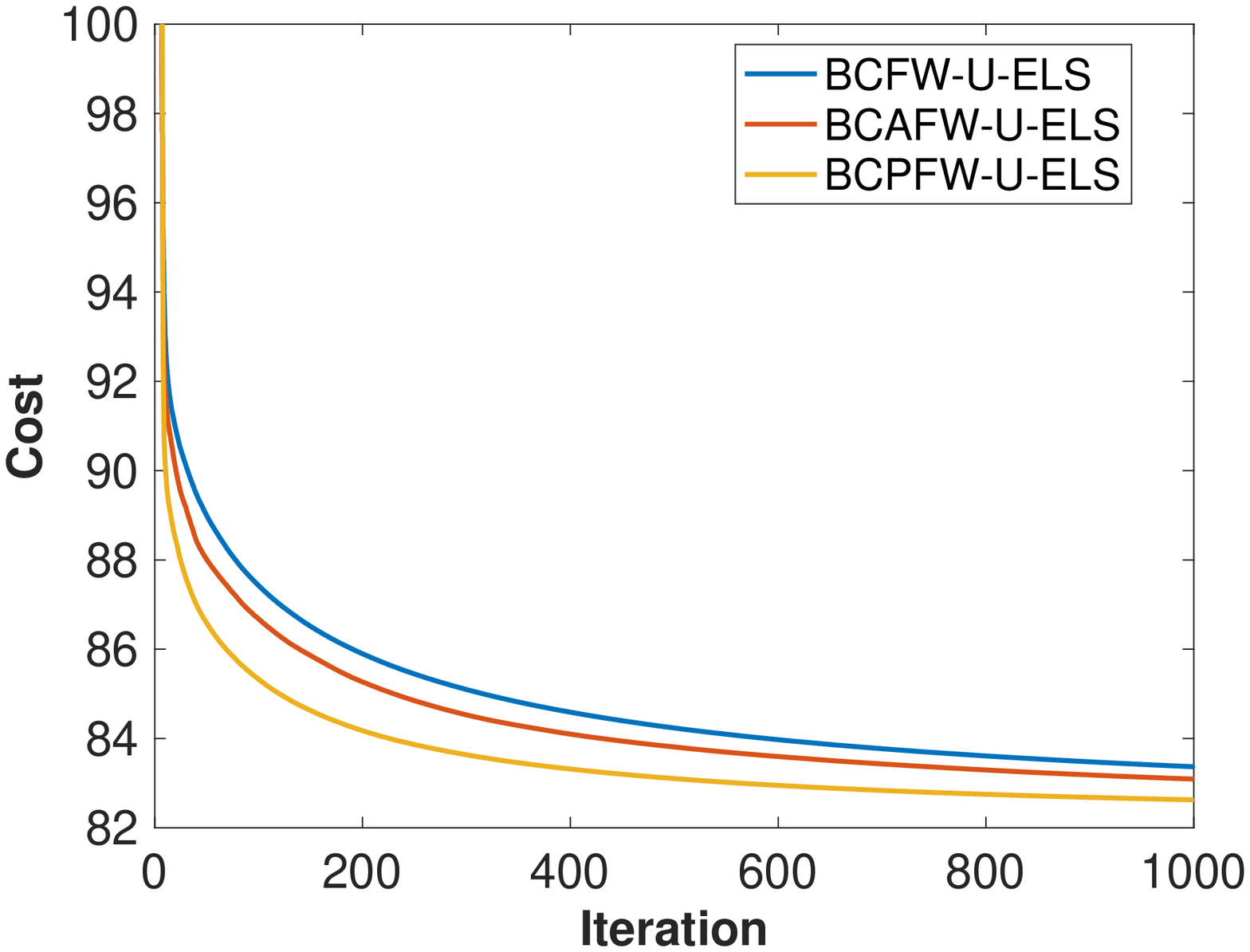}\\
		
		{\footnotesize  (a) objective value : $f(\mat{T})$ }
		
	\end{center} 
	\end{minipage}
	\hspace*{-0.2cm}	
	\begin{minipage}[t]{0.32\textwidth}
	\begin{center}
		\includegraphics[width=\textwidth]{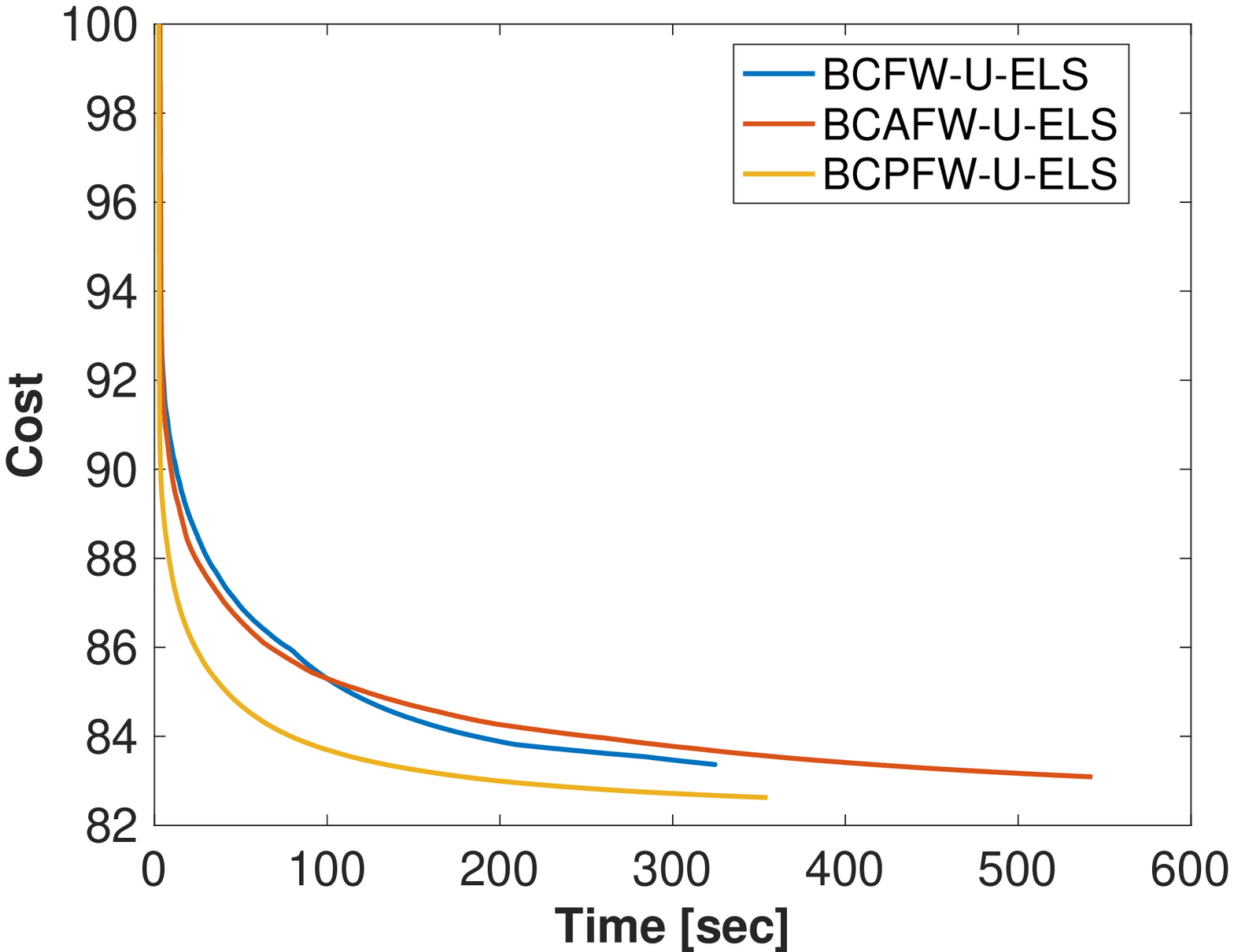}\\
		
		{\footnotesize  (b) objective value (time): $f(\mat{T})$ }
		
	\end{center} 
	\end{minipage}
	\hspace*{-0.2cm}	
	\begin{minipage}[t]{0.32\textwidth}
	\begin{center}
		\includegraphics[width=\textwidth]{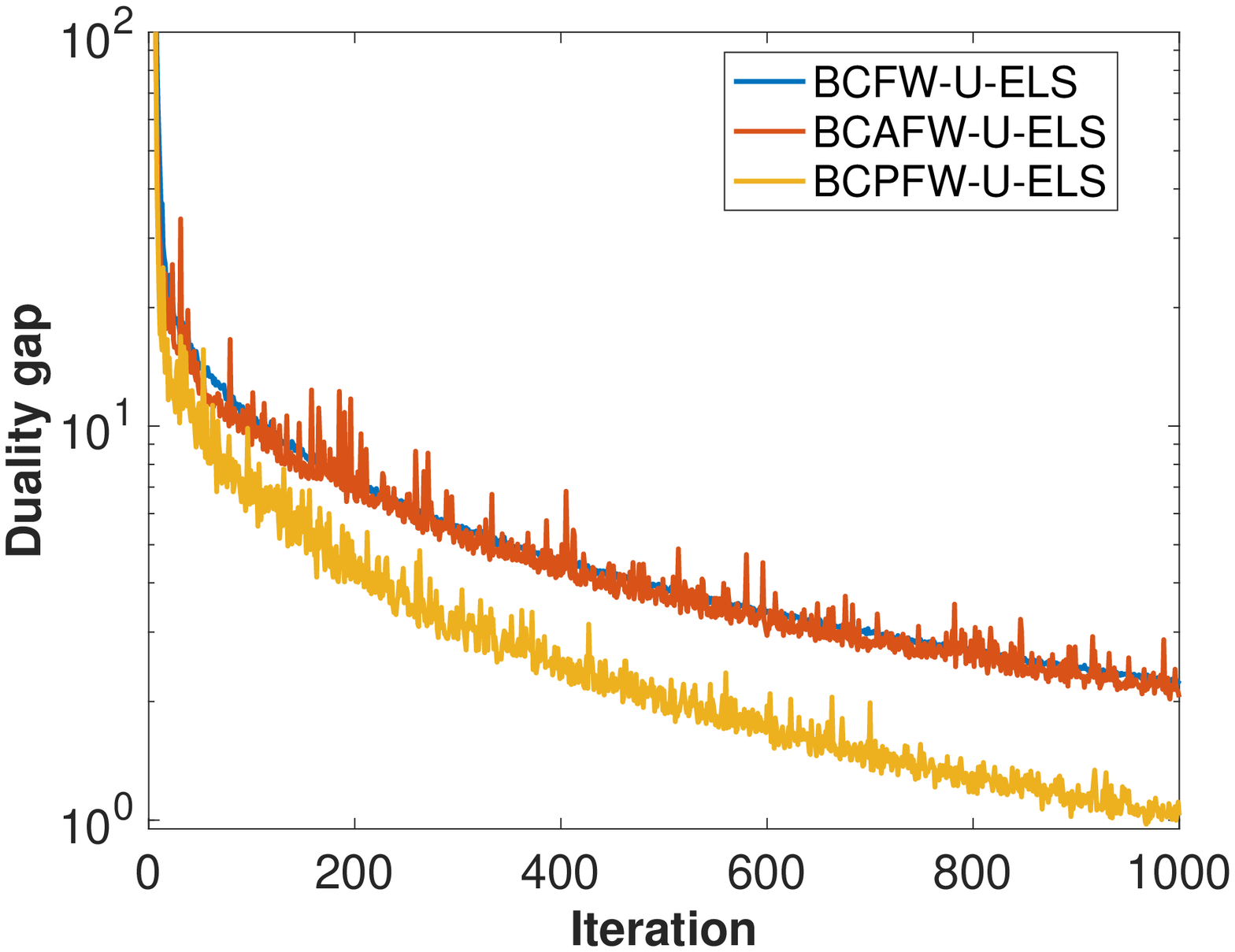}\\
		
		{\footnotesize  (c) duality gap : $g(\mat{T})$}
		
	\end{center} 
	\end{minipage}
	\vspace*{0.2cm}

	\begin{minipage}[t]{0.32\textwidth}
	\begin{center}
		\includegraphics[width=\textwidth]{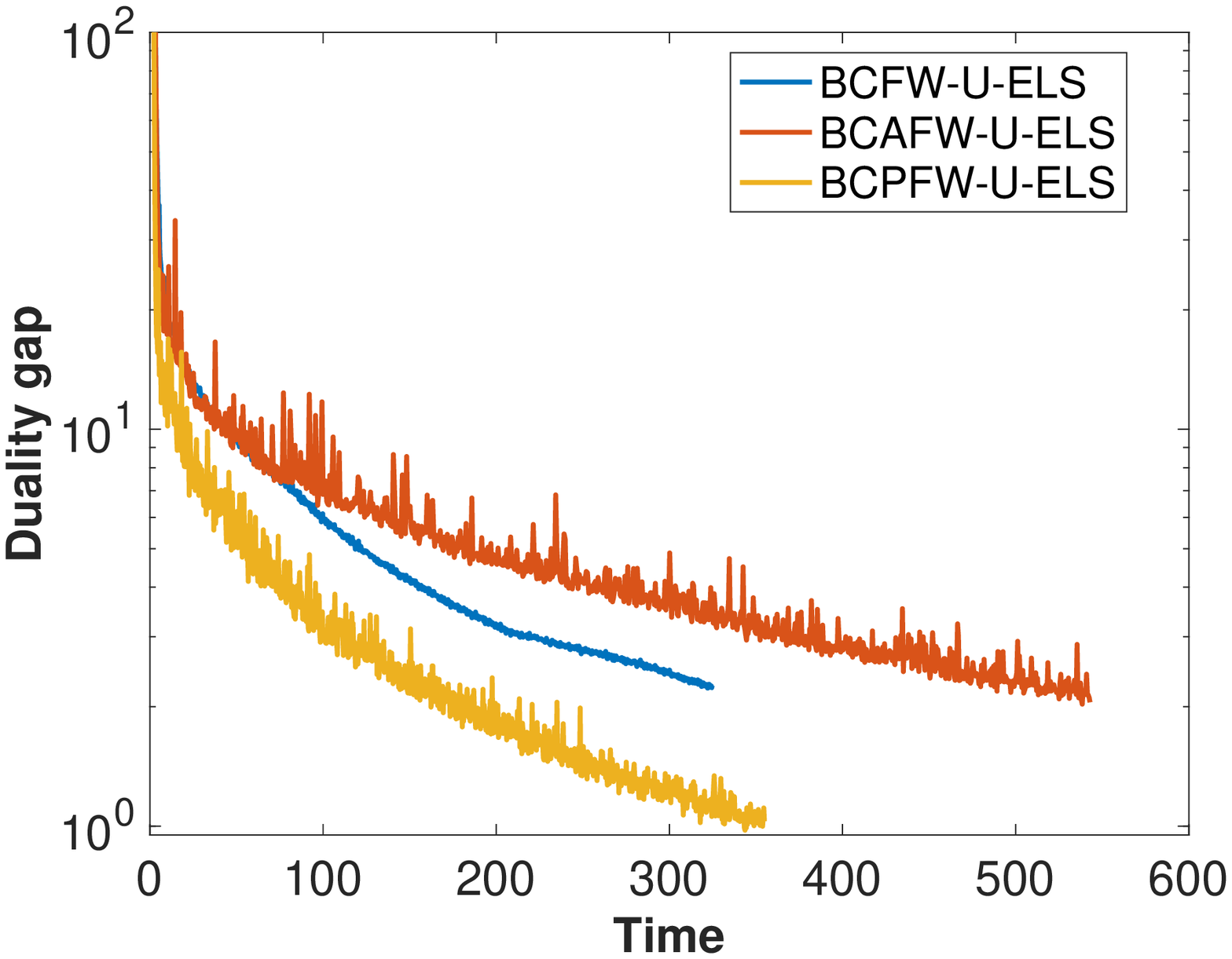}\\
		
		{\footnotesize  (d) duality gap (time) : $g(\mat{T})$}
		
	\end{center} 
	\end{minipage}
	\hspace*{-0.2cm}	
	\begin{minipage}[t]{0.32\textwidth}
	\begin{center}
		\includegraphics[width=\textwidth]{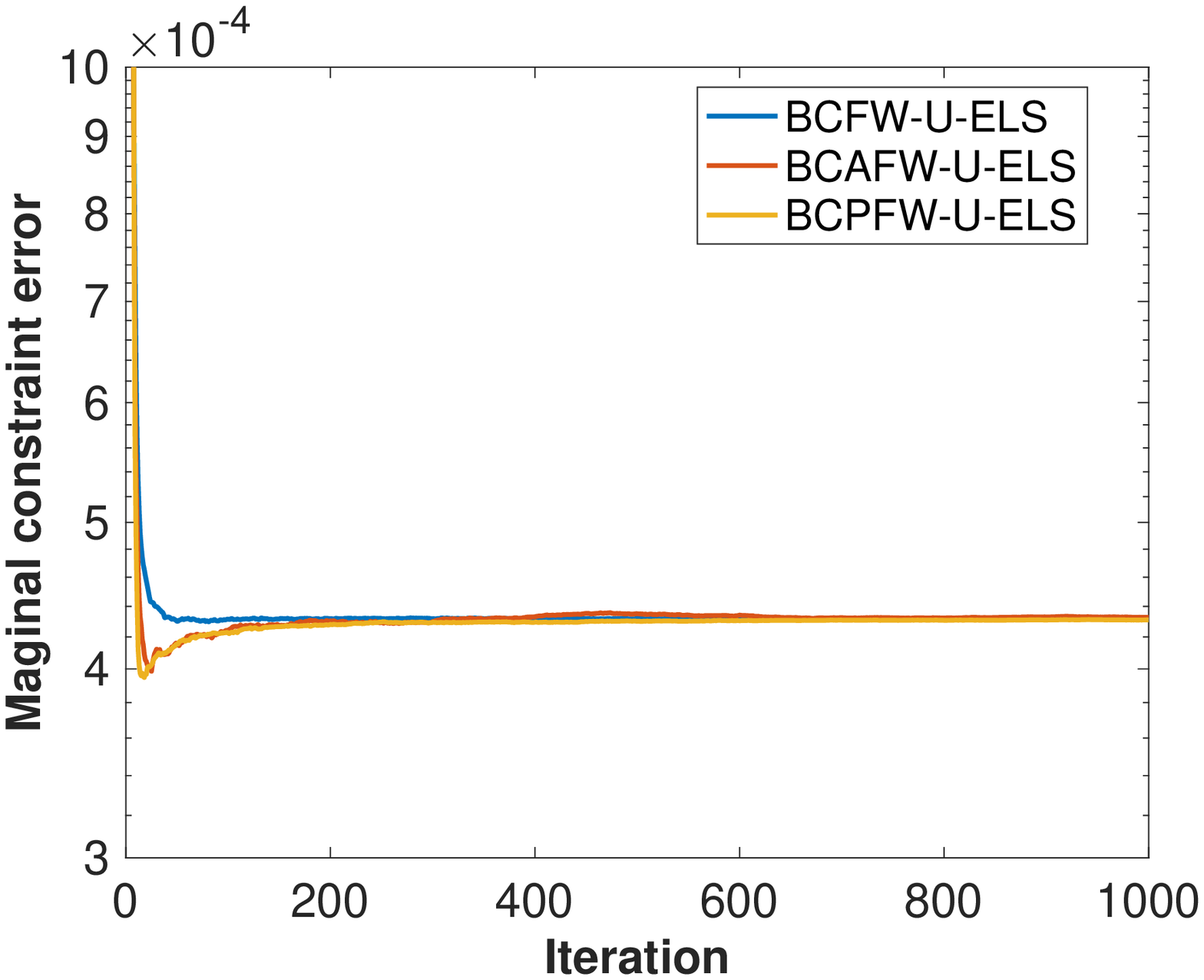}\\
		
		{\footnotesize  (e) marginal constraint error : $e_{c}$}
		
	\end{center} 
	\end{minipage}
	\hspace*{-0.2cm}
	\begin{minipage}[t]{0.32\textwidth}
	\begin{center}
		\includegraphics[width=\textwidth]{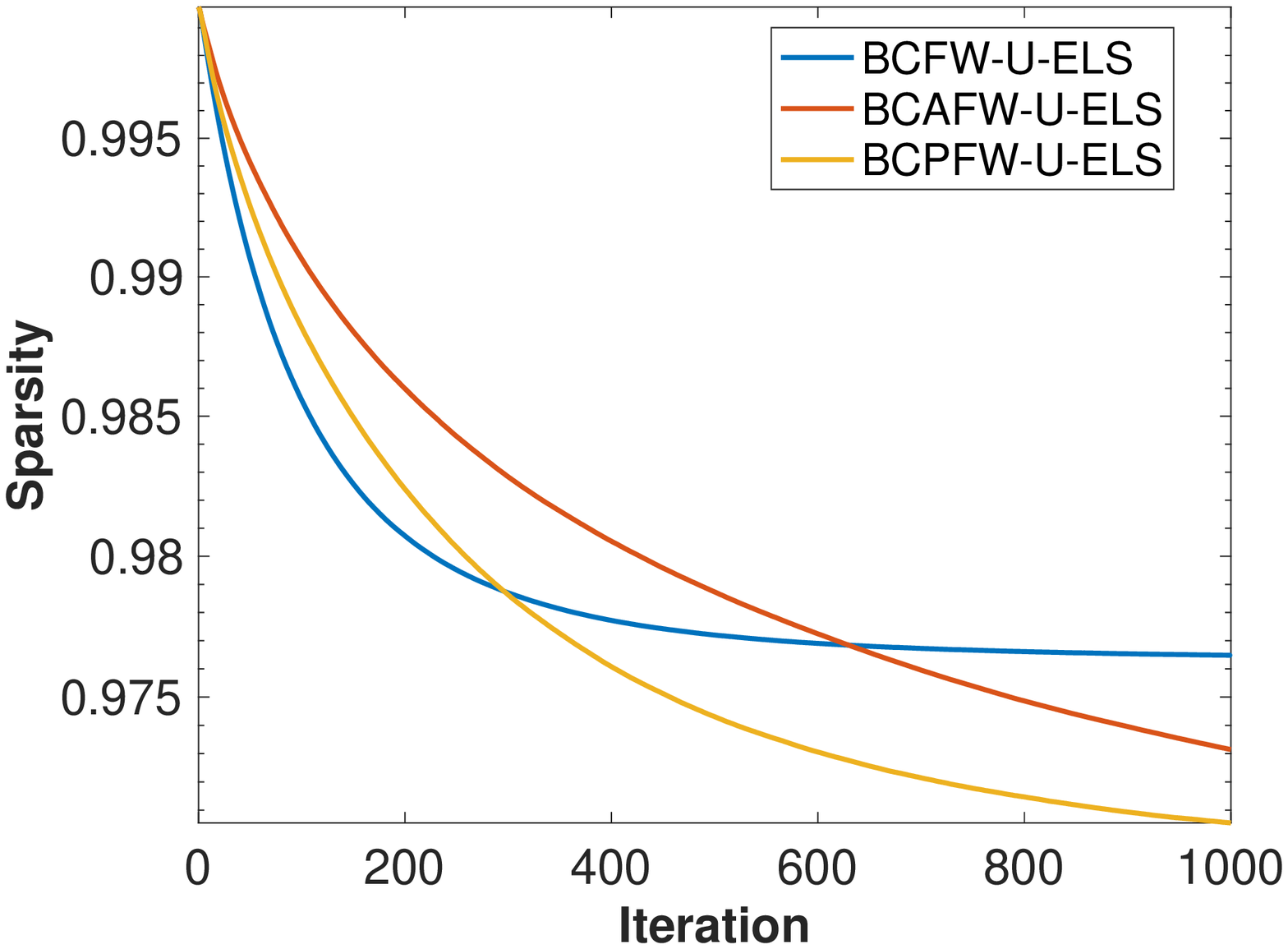}\\
		
		{\footnotesize  (f) sparsity}
		
	\end{center} 
	\end{minipage}	
	\vspace*{0.2cm}

	\begin{minipage}[t]{0.32\textwidth}
	\begin{center}
		\includegraphics[width=\textwidth]{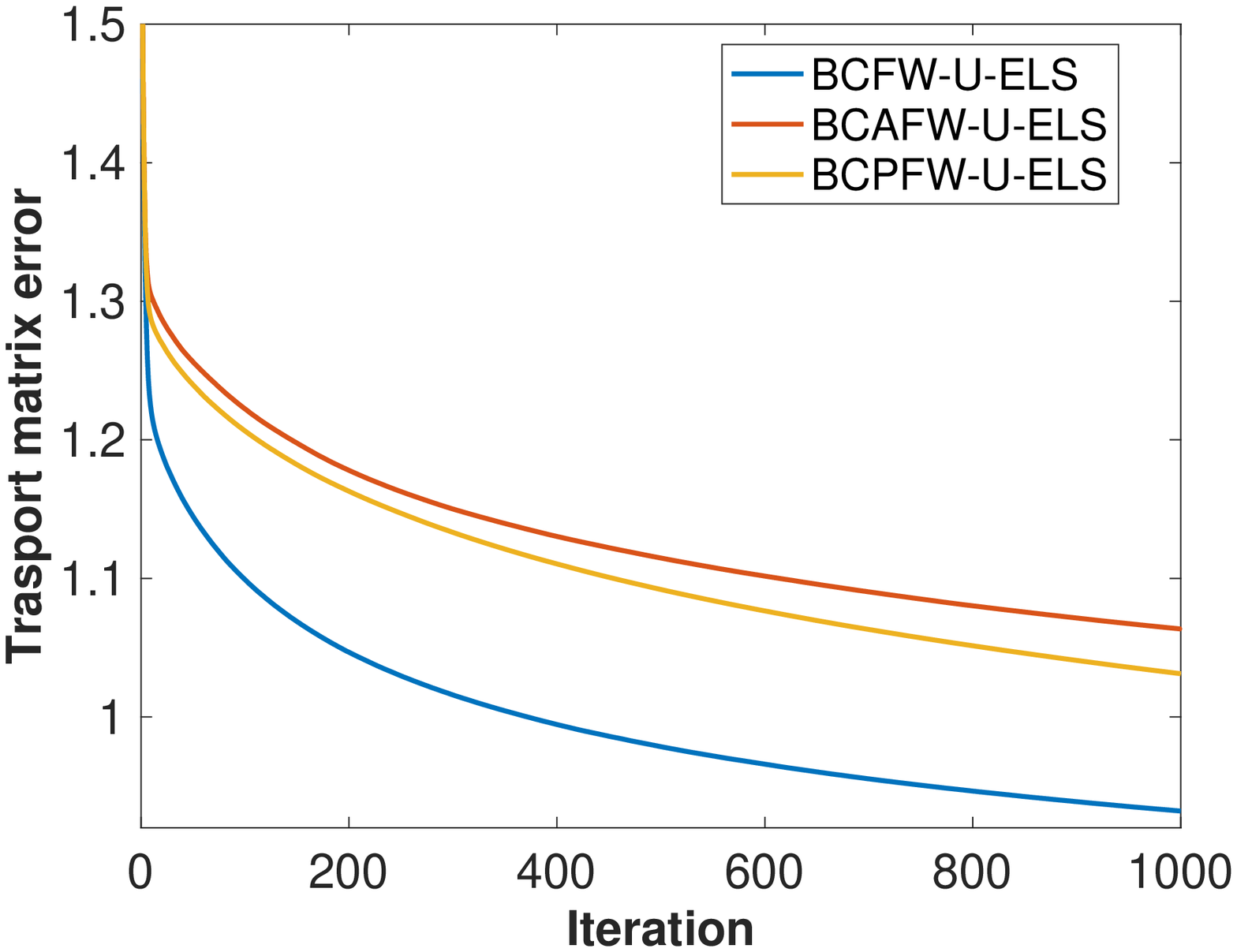}\\
		
		{\footnotesize  (g) matrix error : $e_{m}$ }
		
	\end{center} 
	\end{minipage}
	\hspace*{-0.2cm}
	\begin{minipage}[t]{0.32\textwidth}
	\begin{center}
		\includegraphics[width=\textwidth]{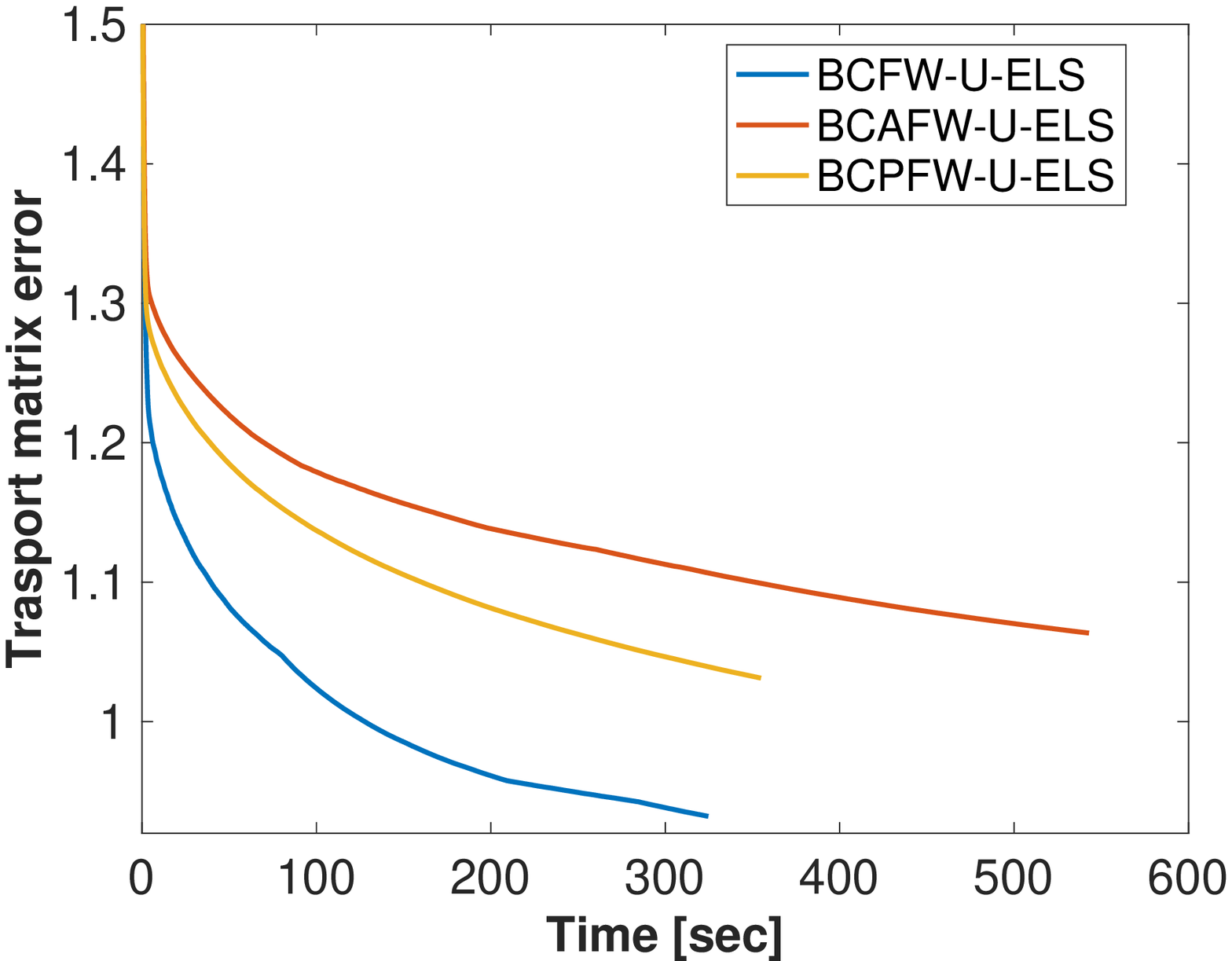}\\
		
		{\footnotesize  (h) matrix error (time) : $e_{m}$ }
		
	\end{center} 
	\end{minipage}		
	\hspace*{-0.2cm}			
	\begin{minipage}[t]{0.32\textwidth}
	\begin{center}
		\includegraphics[width=\textwidth]{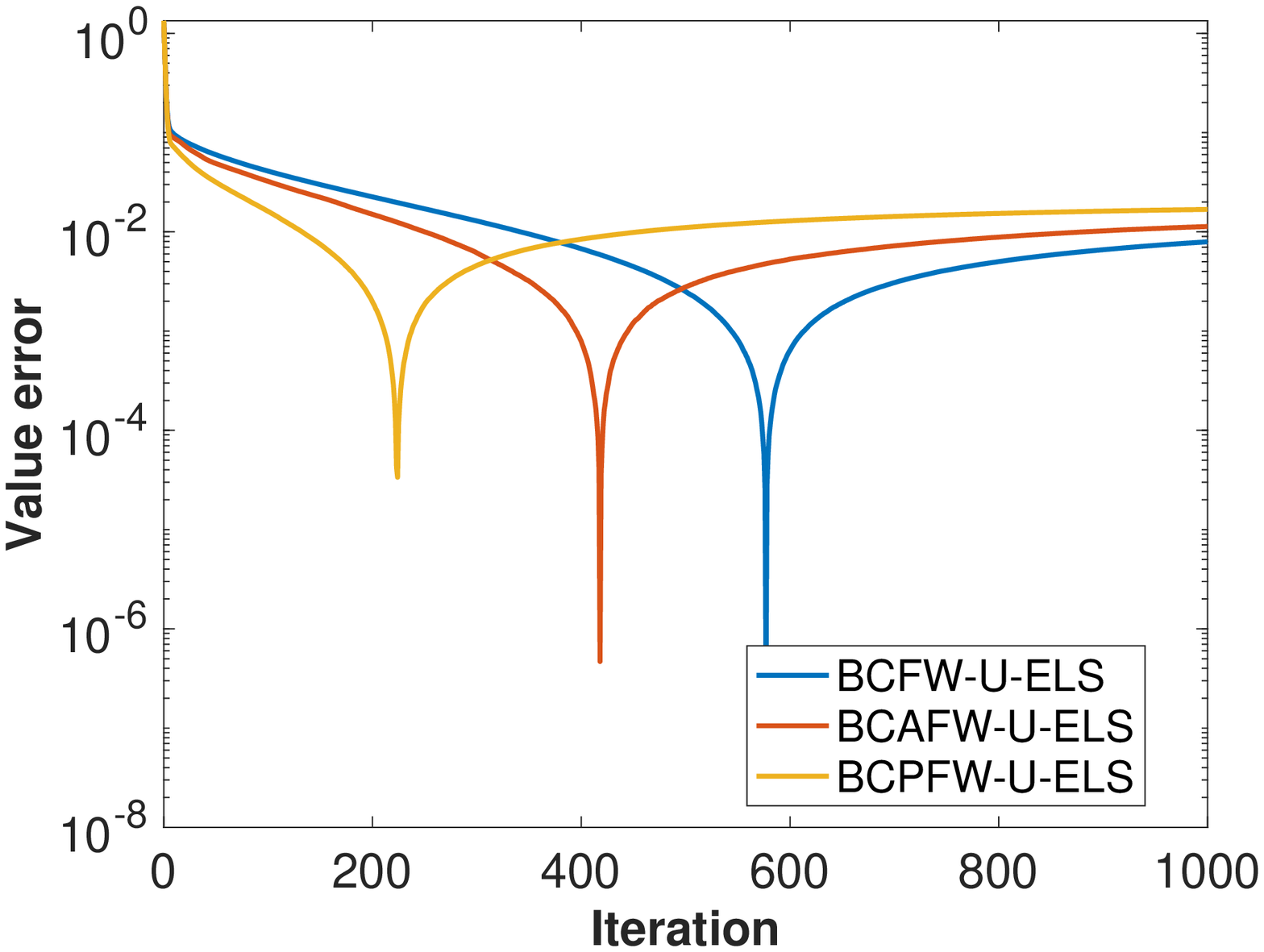}\\
		
		{\footnotesize  (i) value error : $e_v$}
		
	\end{center} 
	\end{minipage}

\caption{Evaluations on convergence of away-steps and pairwise-steps algorithms (corresponding to Figure \ref{fig:FastVariantConvergencePerformance}).}
\label{Appenfig:FastVariantConvergencePerformance}
\end{center}
\end{figure}

\begin{figure}[htbp]
\begin{center}
	\begin{minipage}[t]{0.24\textwidth}
	\begin{center}
		
		\includegraphics[width=\textwidth]{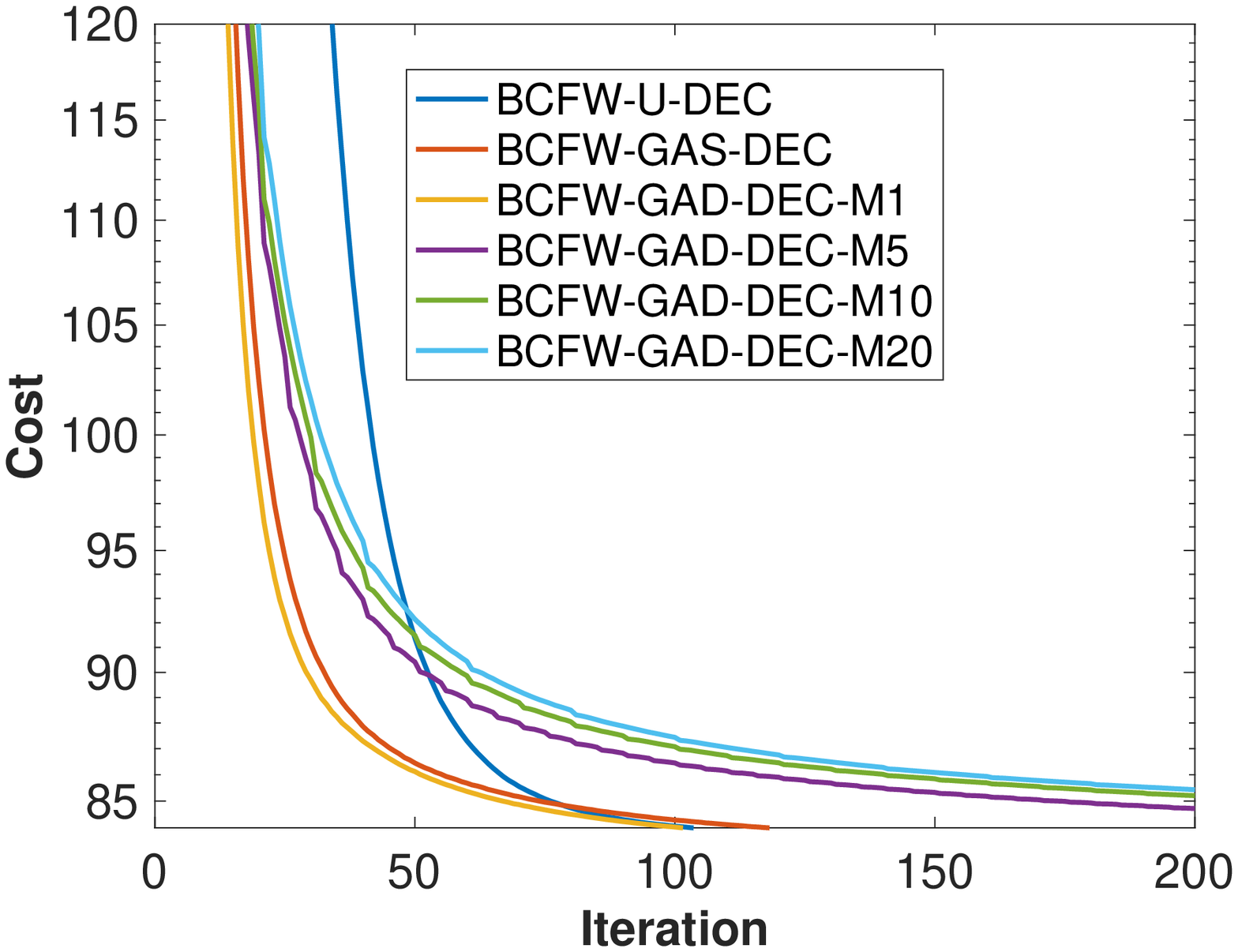}\\
		
		{\footnotesize (i) objective value : $f(\mat{T})$ }
		
	\end{center} 
	\end{minipage}
	\begin{minipage}[t]{0.236\textwidth}
	\begin{center}
		
		\includegraphics[width=\textwidth]{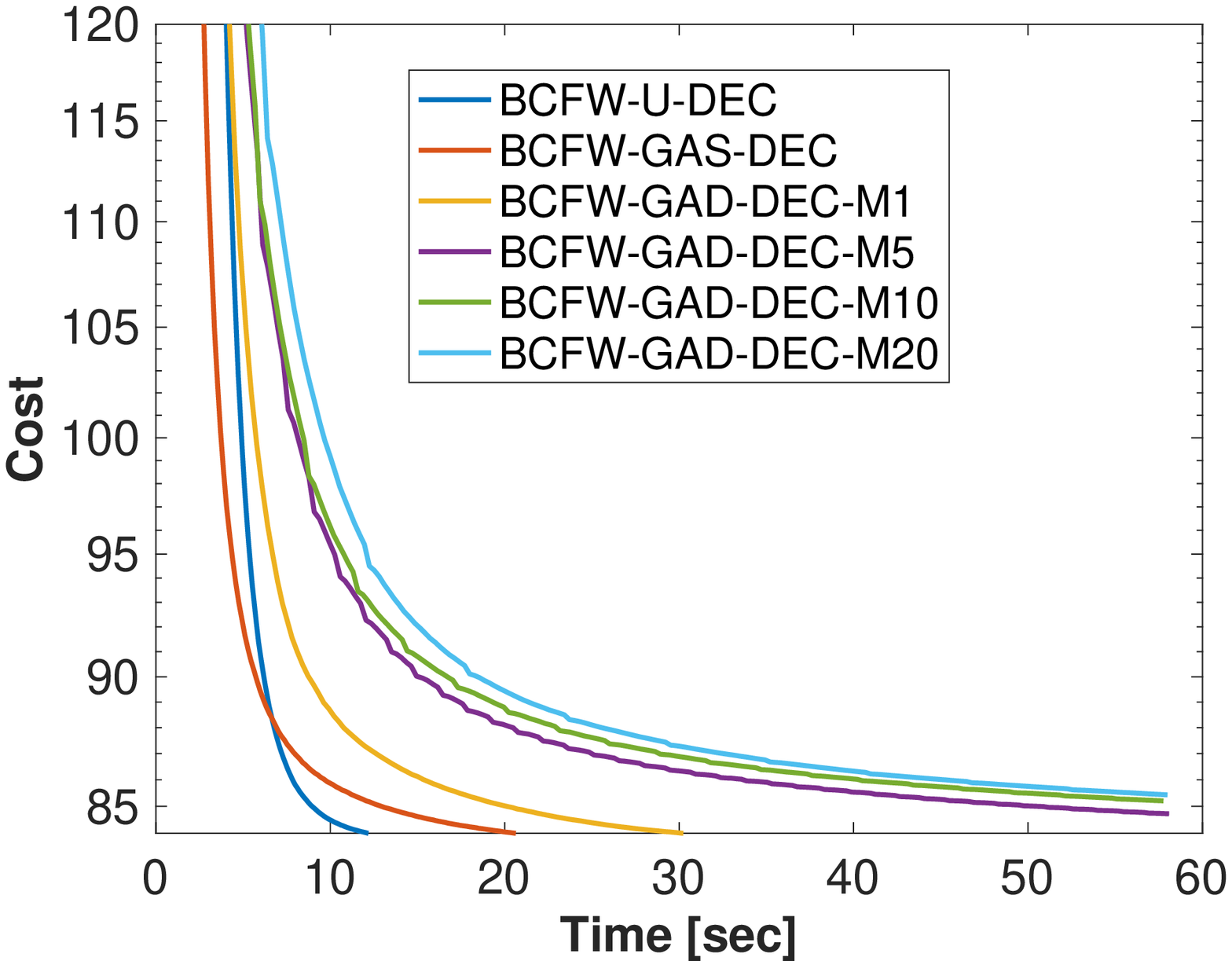}\\
		
		{\footnotesize  (ii) objective value (time)}
		
	\end{center} 
	\end{minipage}	
	\begin{minipage}[t]{0.24\textwidth}
	\begin{center}
		\includegraphics[width=\textwidth]{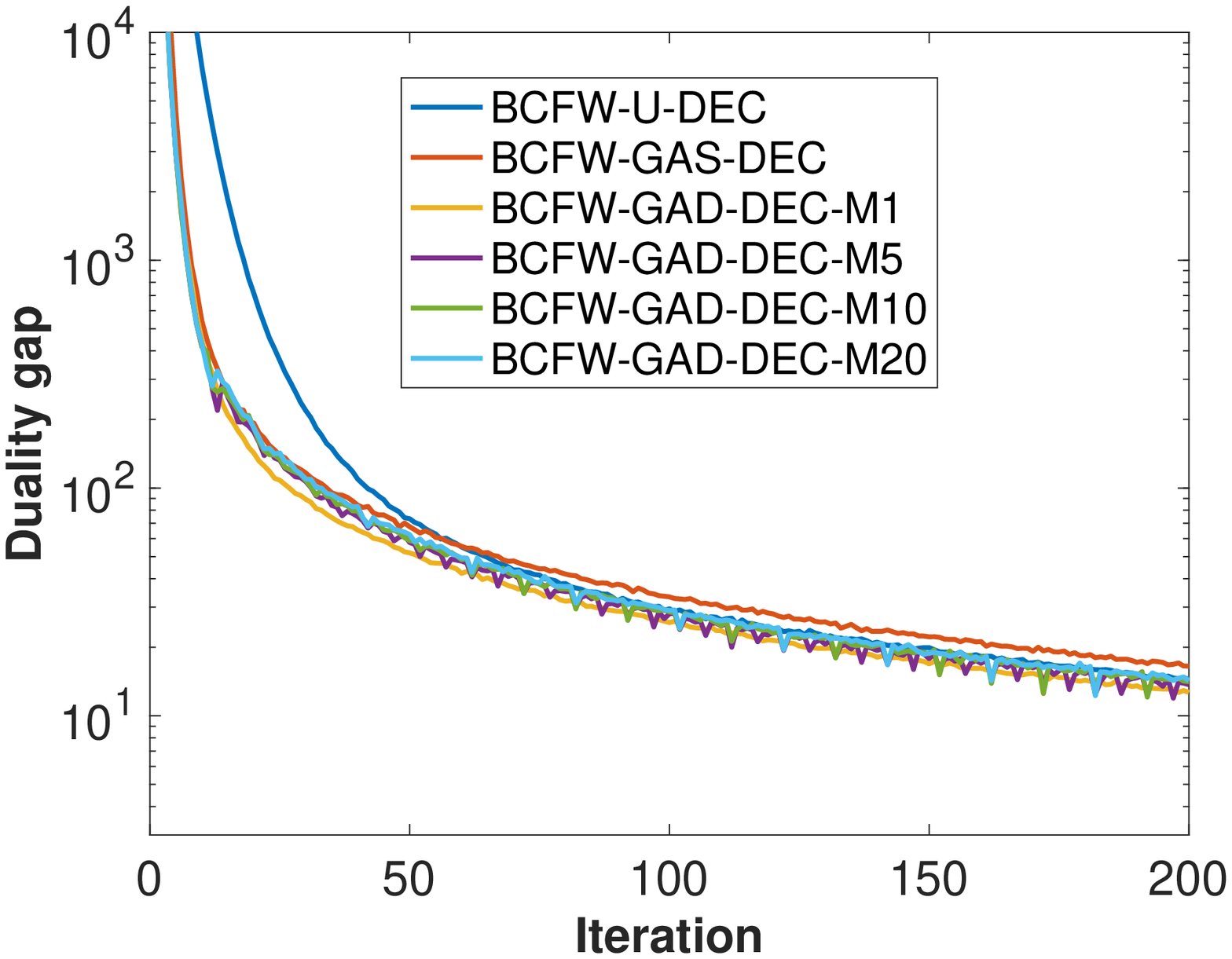}\\
		
		{\footnotesize  (iii) duality gap: $g(\mat{T})$}
		
	\end{center} 
	\end{minipage}
	\hspace*{0.1cm}
	\begin{minipage}[t]{0.23\textwidth}
	\begin{center}
		\includegraphics[width=\textwidth]{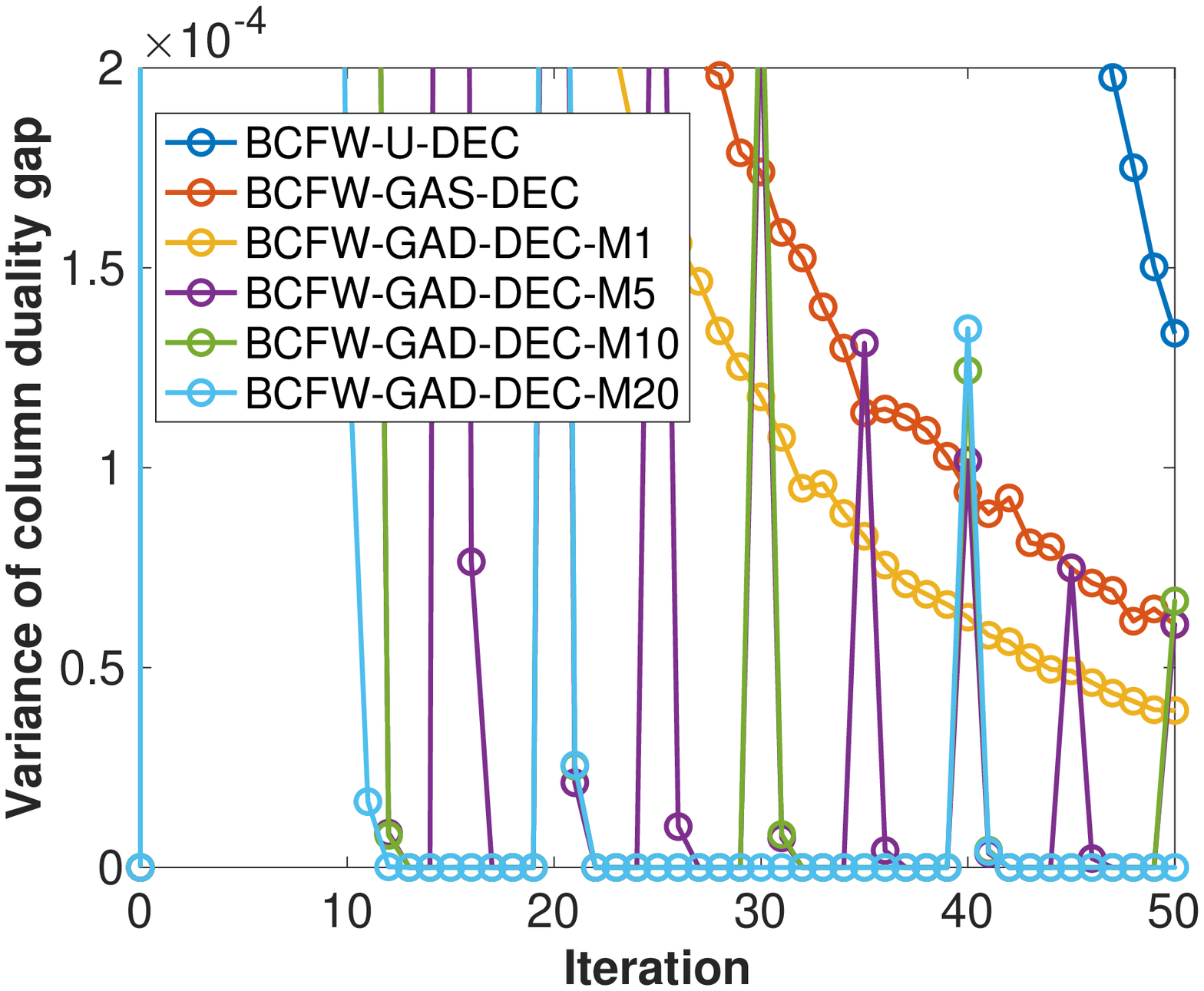}\\
		
		{\footnotesize  (iv) variance of $g_i(\mat{T})$}
		
	\end{center} 
	\end{minipage}
	\vspace*{0.2cm}	
	
	{\small (a) BCFW-U-DEC and BCFW-GA-DEC}
	\vspace*{0.6cm}
	
	\begin{minipage}[t]{0.24\textwidth}
	\begin{center}
		
		\includegraphics[width=\textwidth]{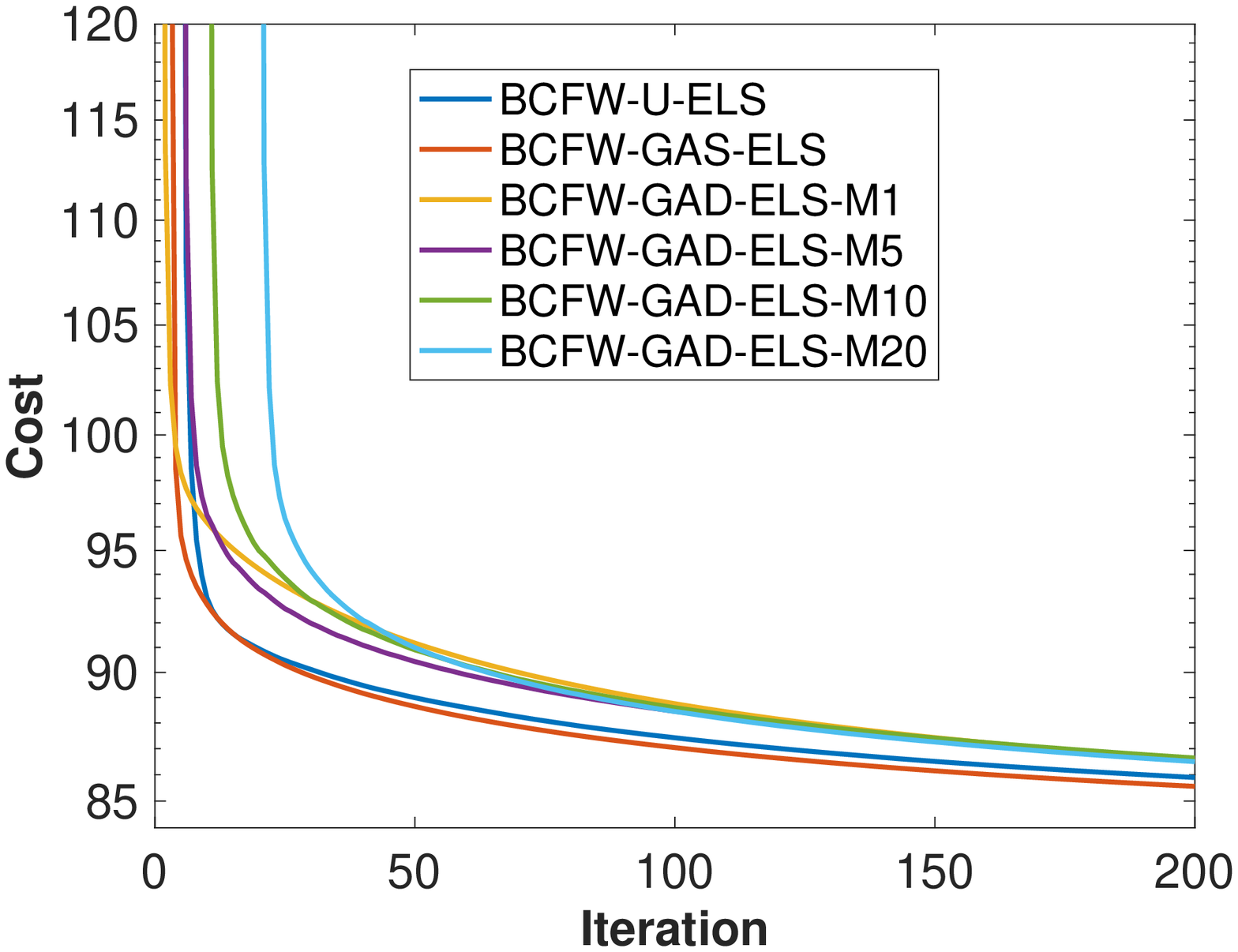}\\
		
		{\footnotesize  (i) objective value : $f(\mat{T})$ }
		
	\end{center} 
	\end{minipage}
	\begin{minipage}[t]{0.24\textwidth}
	\begin{center}
		
		\includegraphics[width=\textwidth]{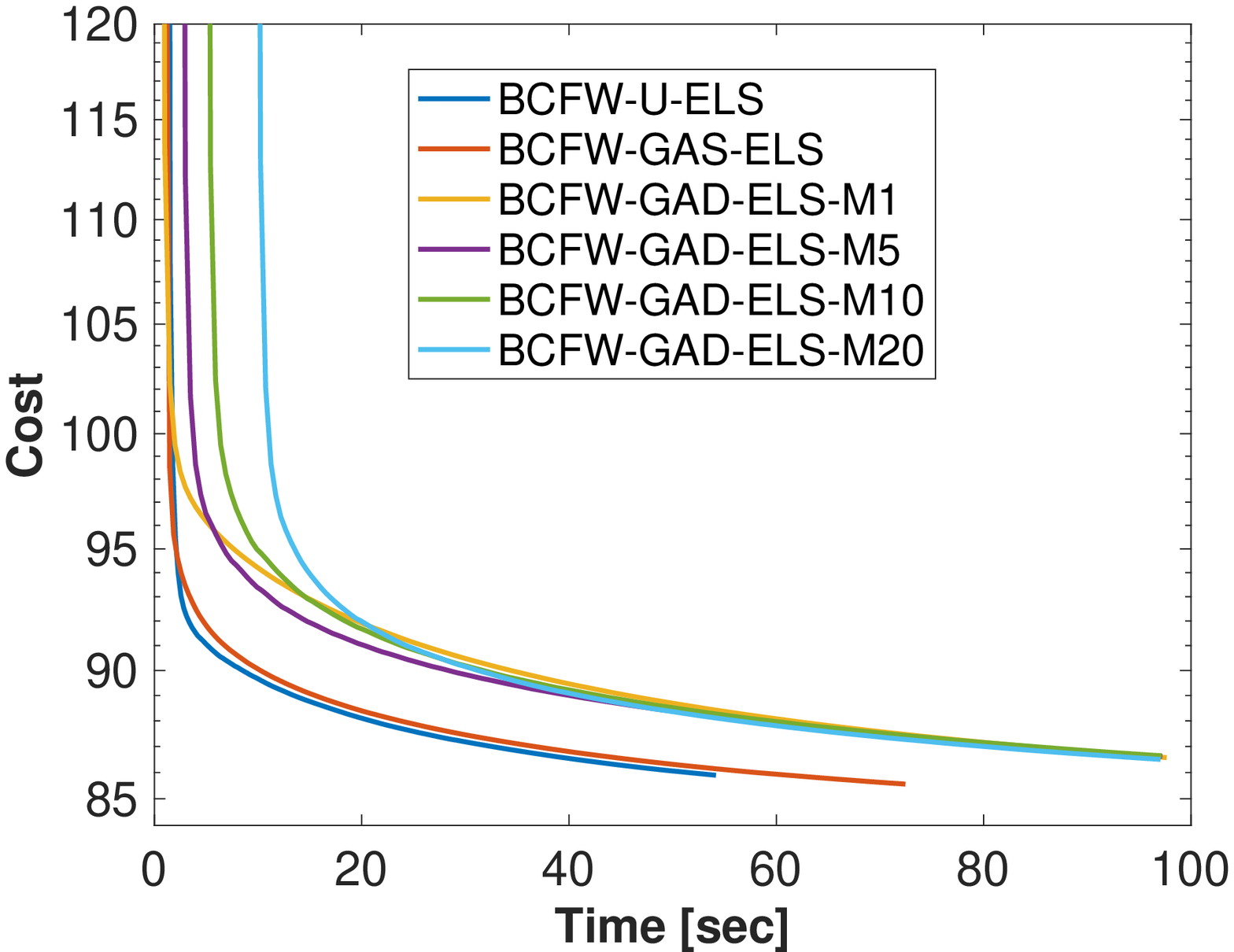}\\
		
		{\footnotesize  (ii) objective value (time)}
		
	\end{center} 
	\end{minipage}	
	\begin{minipage}[t]{0.24\textwidth}
	\begin{center}
		\includegraphics[width=\textwidth]{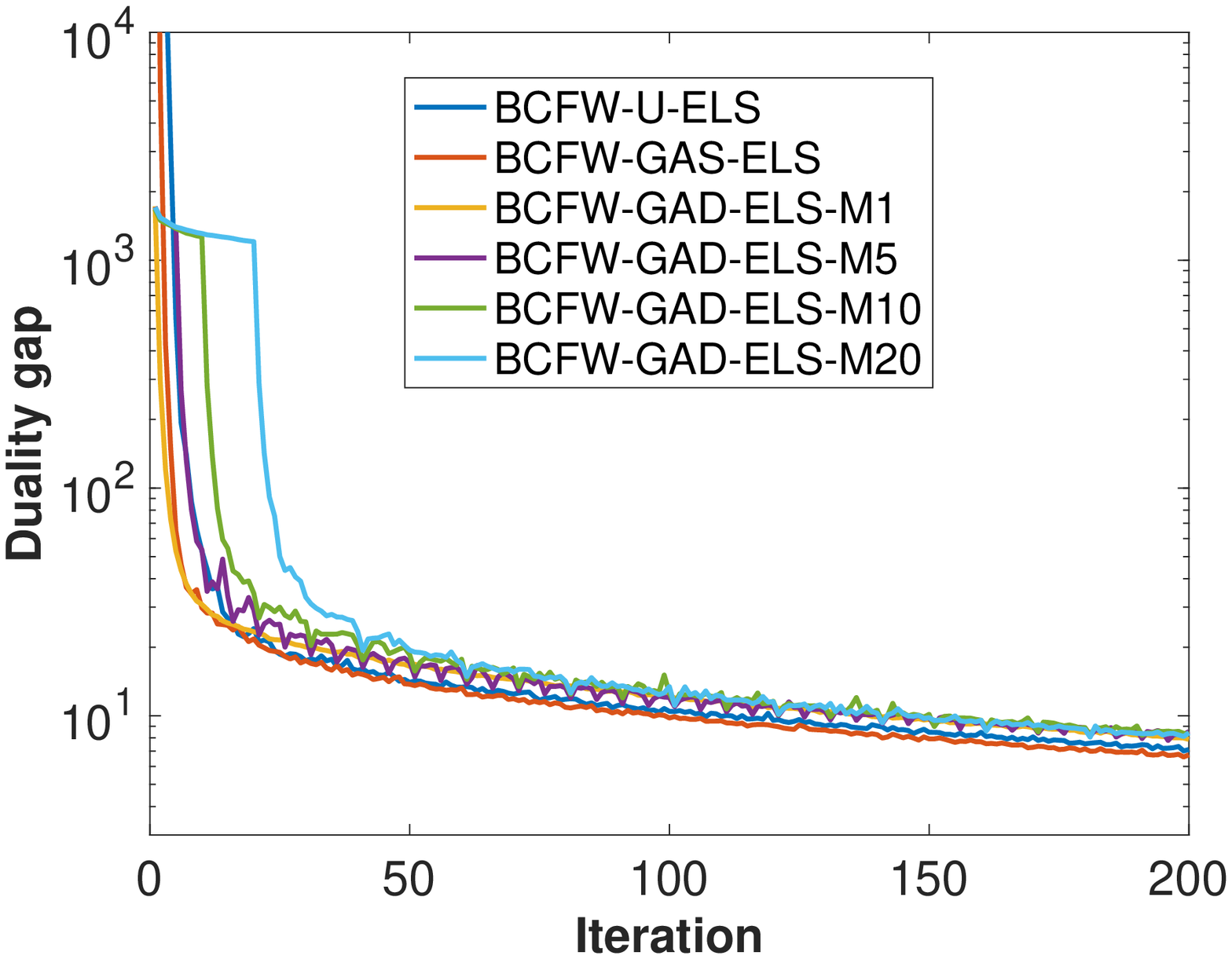}\\
		
		{\footnotesize  (iii) duality gap: $g(\mat{T})$}
		
	\end{center} 
	\end{minipage}
	\hspace*{0.1cm}
	\begin{minipage}[t]{0.23\textwidth}
	\begin{center}
		\includegraphics[width=\textwidth]{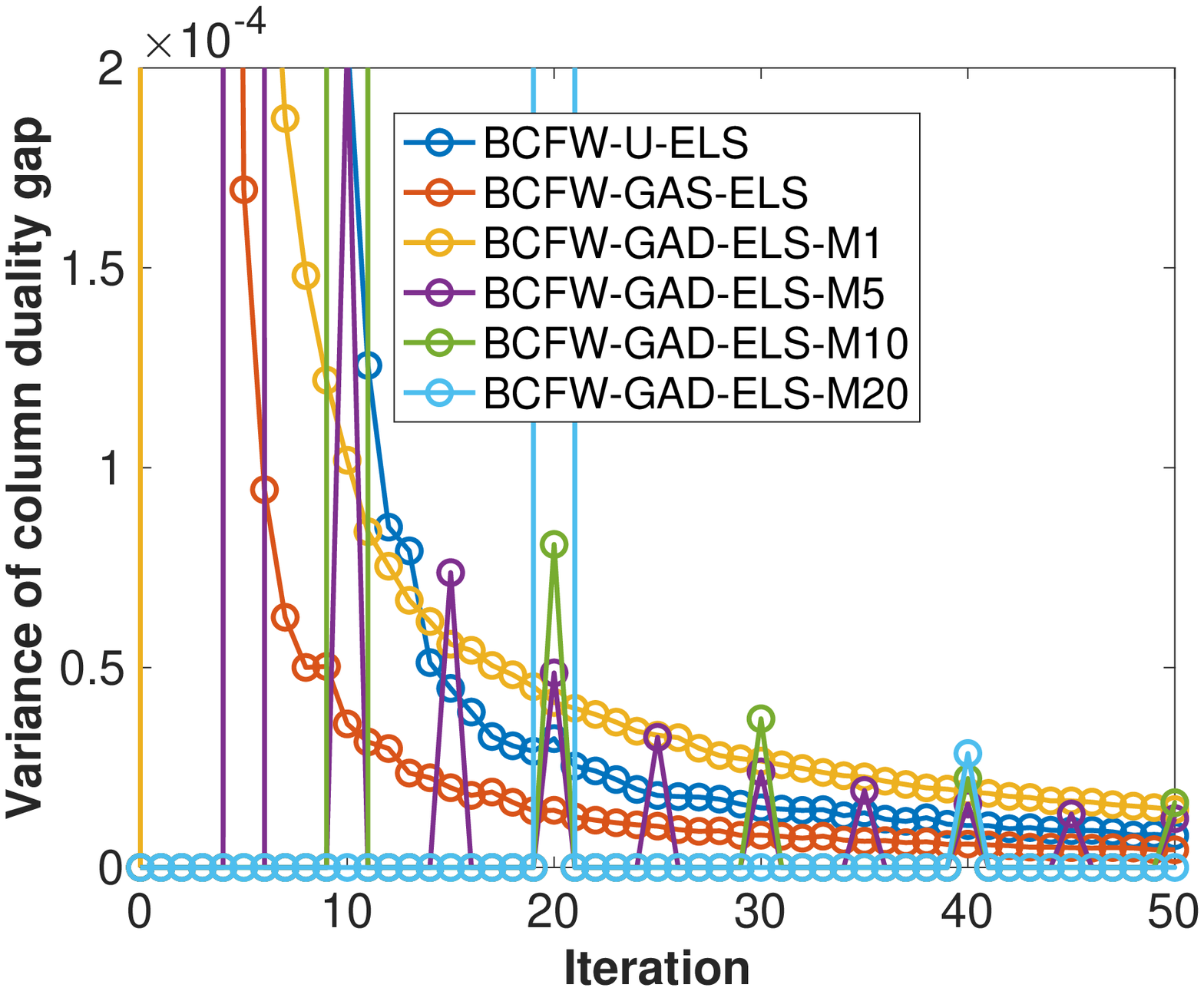}\\
		
		{\footnotesize  (iv) variances of $g_i(\mat{T})$}
		
	\end{center} 
	\end{minipage}	
	\vspace*{0.2cm}	
	
	{\small (b) BCFW-U-ELS and BCFW-GA-ELS}
	\vspace*{0.6cm}
	
	\begin{minipage}[t]{0.24\textwidth}
	\begin{center}
		
		\includegraphics[width=\textwidth]{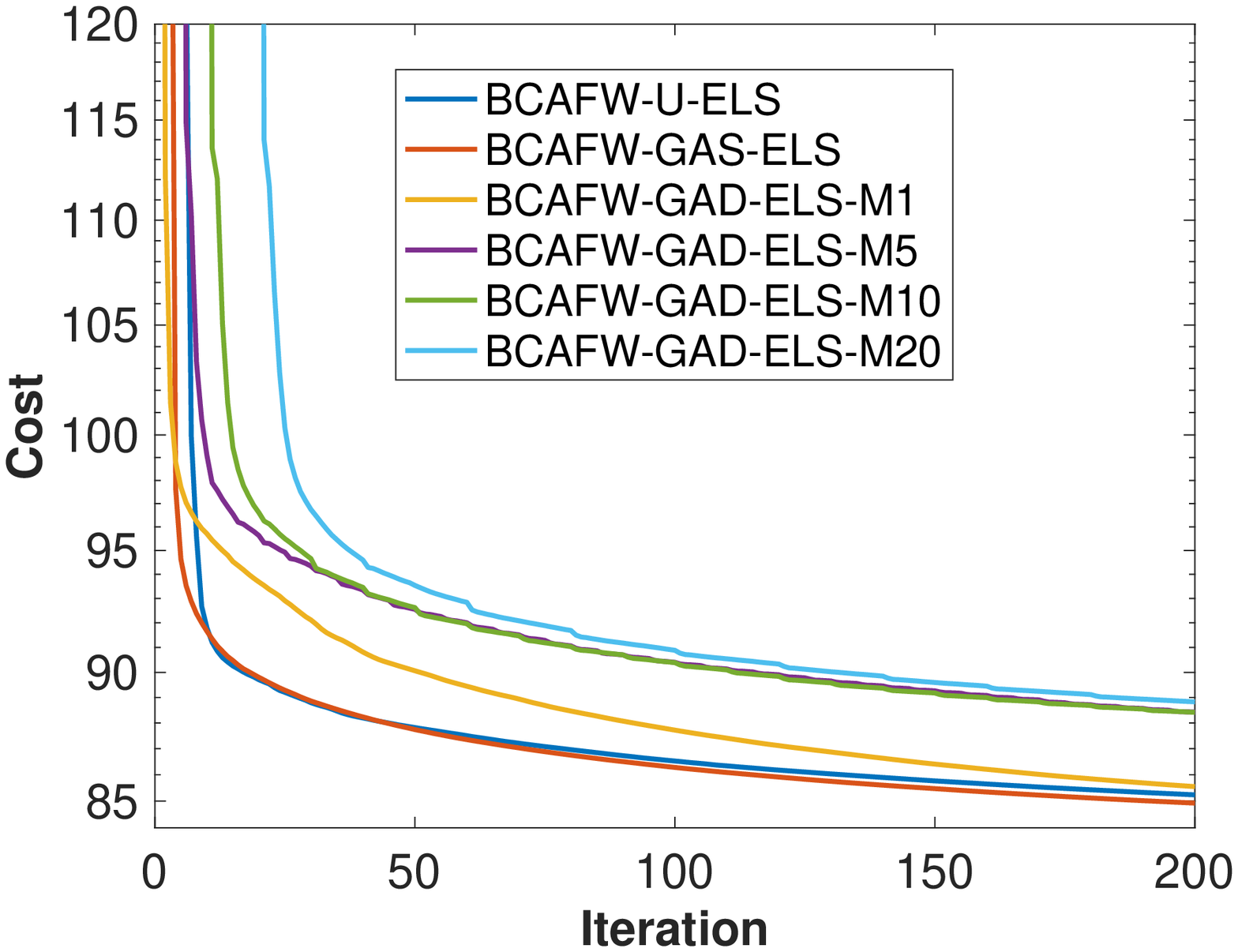}\\
		
		{\footnotesize  (i) objective value : $f(\mat{T})$ }
		
	\end{center} 
	\end{minipage}
	\begin{minipage}[t]{0.24\textwidth}
	\begin{center}
		
		\includegraphics[width=\textwidth]{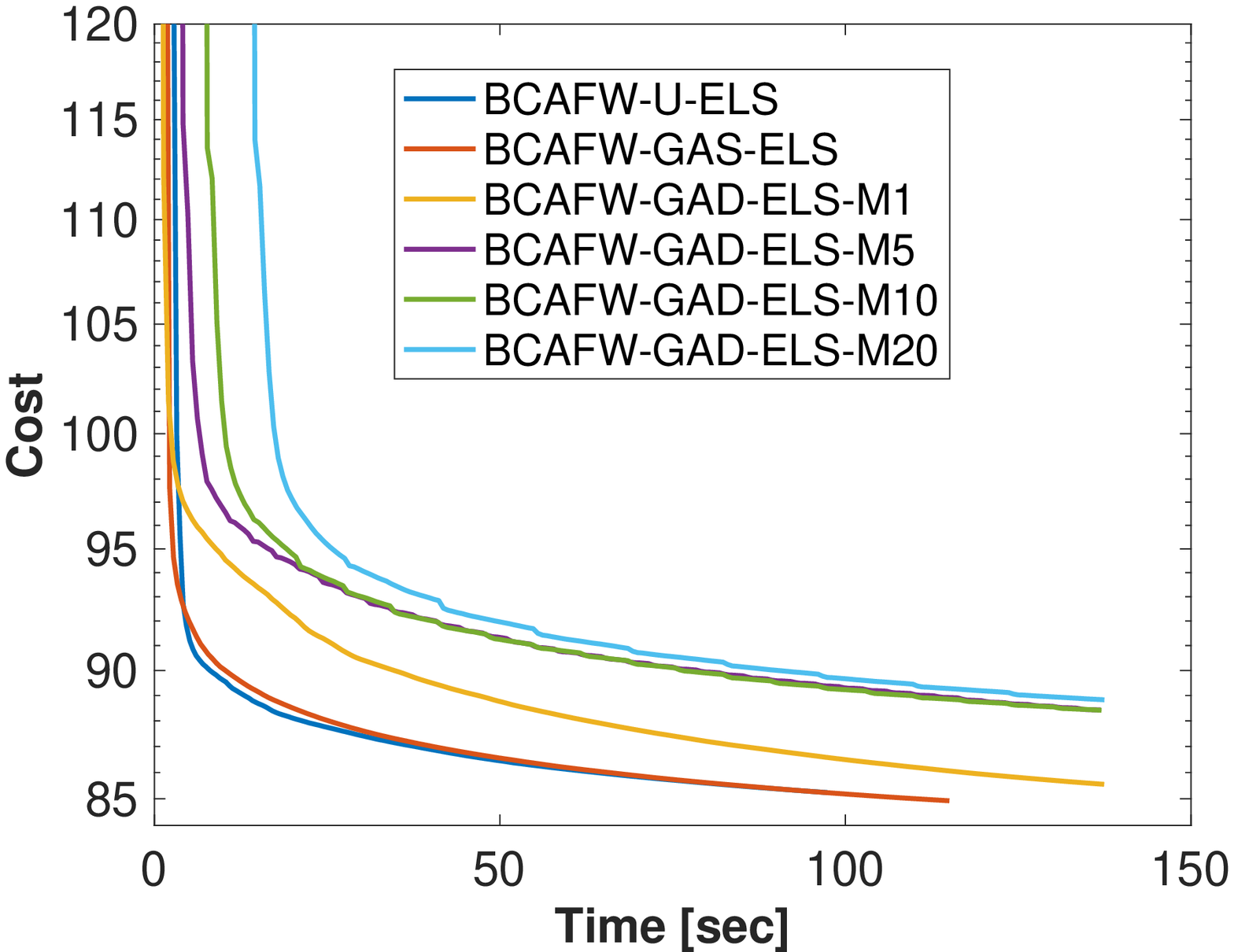}\\
		
		{\footnotesize  (ii) objective value (time)}
		
	\end{center} 
	\end{minipage}	
	\begin{minipage}[t]{0.24\textwidth}
	\begin{center}
		\includegraphics[width=\textwidth]{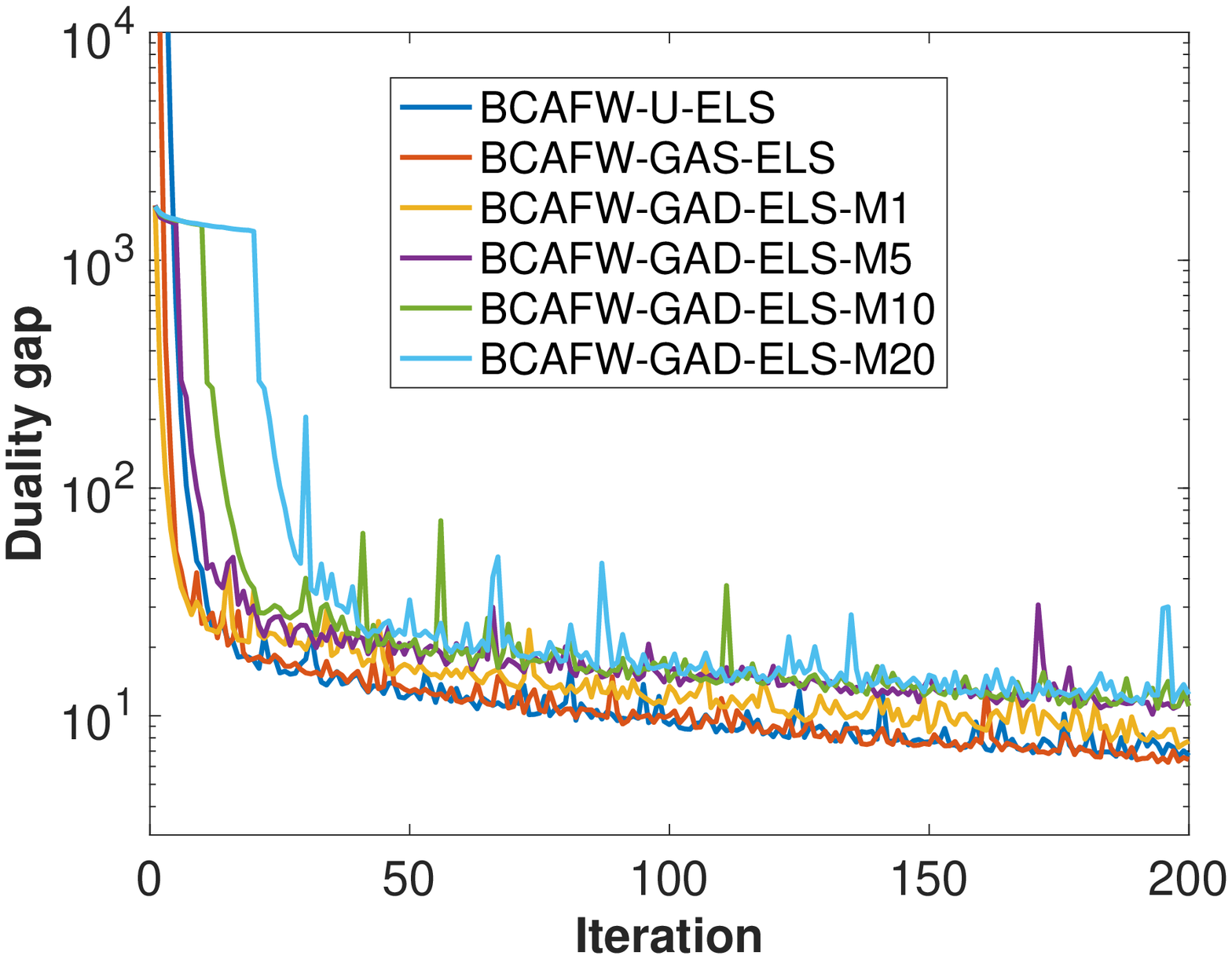}\\
		
		{\footnotesize  (iii) duality gap: $g(\mat{T})$}
		
	\end{center} 
	\end{minipage}
	\hspace*{0.1cm}
	\begin{minipage}[t]{0.23\textwidth}
	\begin{center}
		\includegraphics[width=\textwidth]{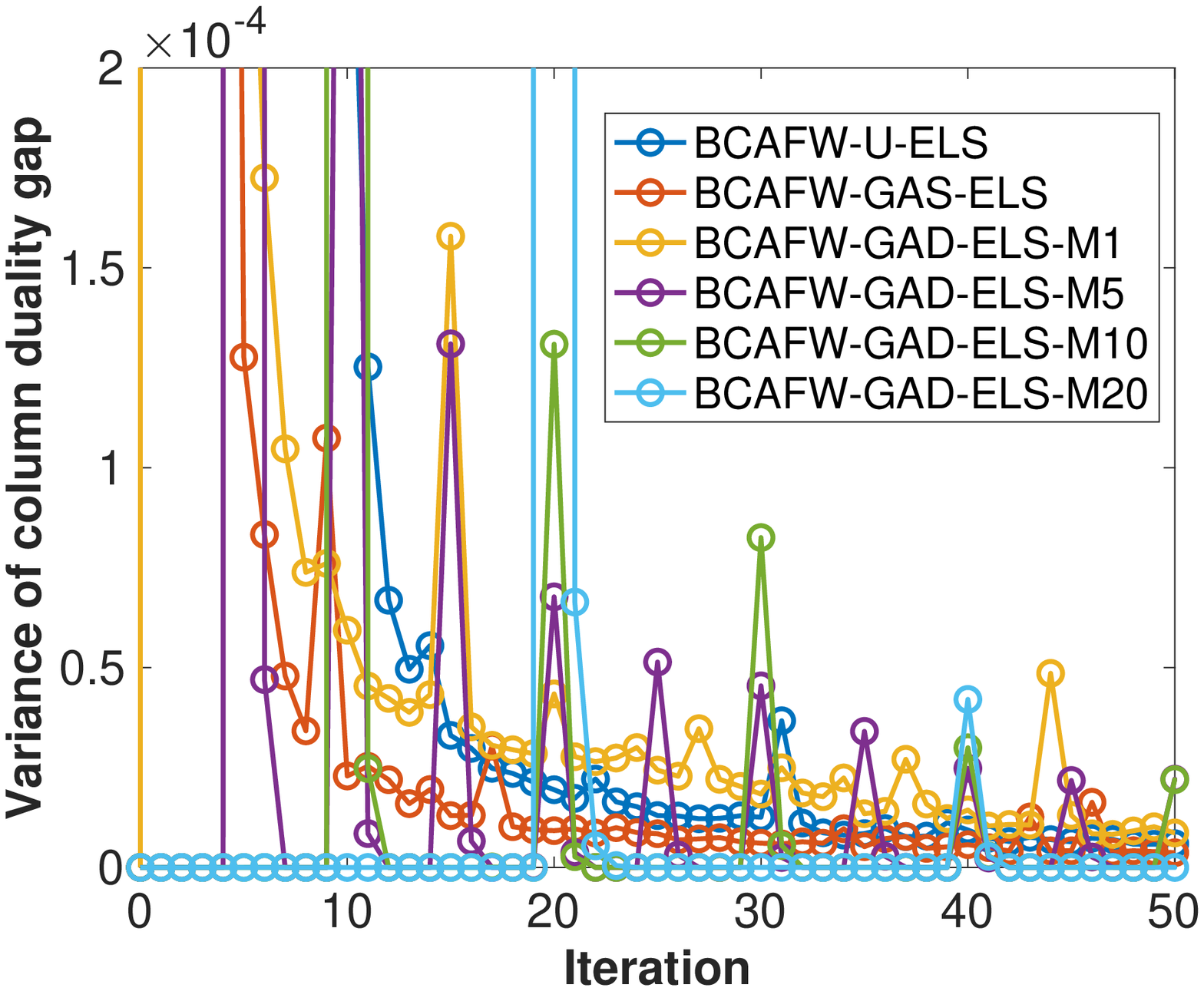}\\
		
		{\footnotesize  (iv) variances of $g_i(\mat{T})$}
		
	\end{center} 
	\end{minipage}
	\vspace*{0.2cm}		
	
	{\small (c) BCAFW-U-ELS and BCAFW-GA-ELS}
	\vspace*{0.6cm}		
	
	\begin{minipage}[t]{0.24\textwidth}
	\begin{center}
		
		\includegraphics[width=\textwidth]{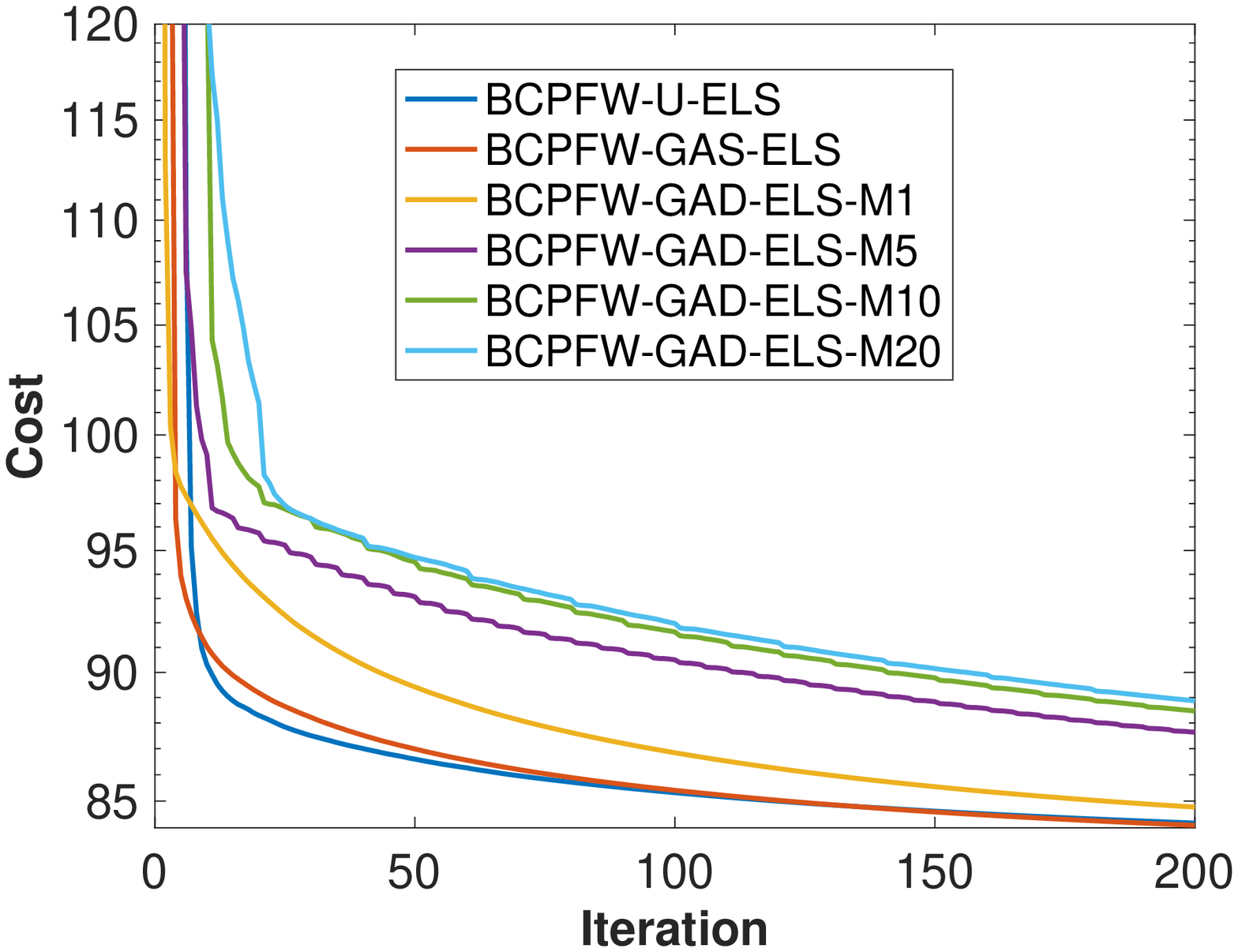}\\
		
		{\footnotesize  (i) objective value : $f(\mat{T})$ }
		
	\end{center} 
	\end{minipage}
	\begin{minipage}[t]{0.24\textwidth}
	\begin{center}
		
		\includegraphics[width=\textwidth]{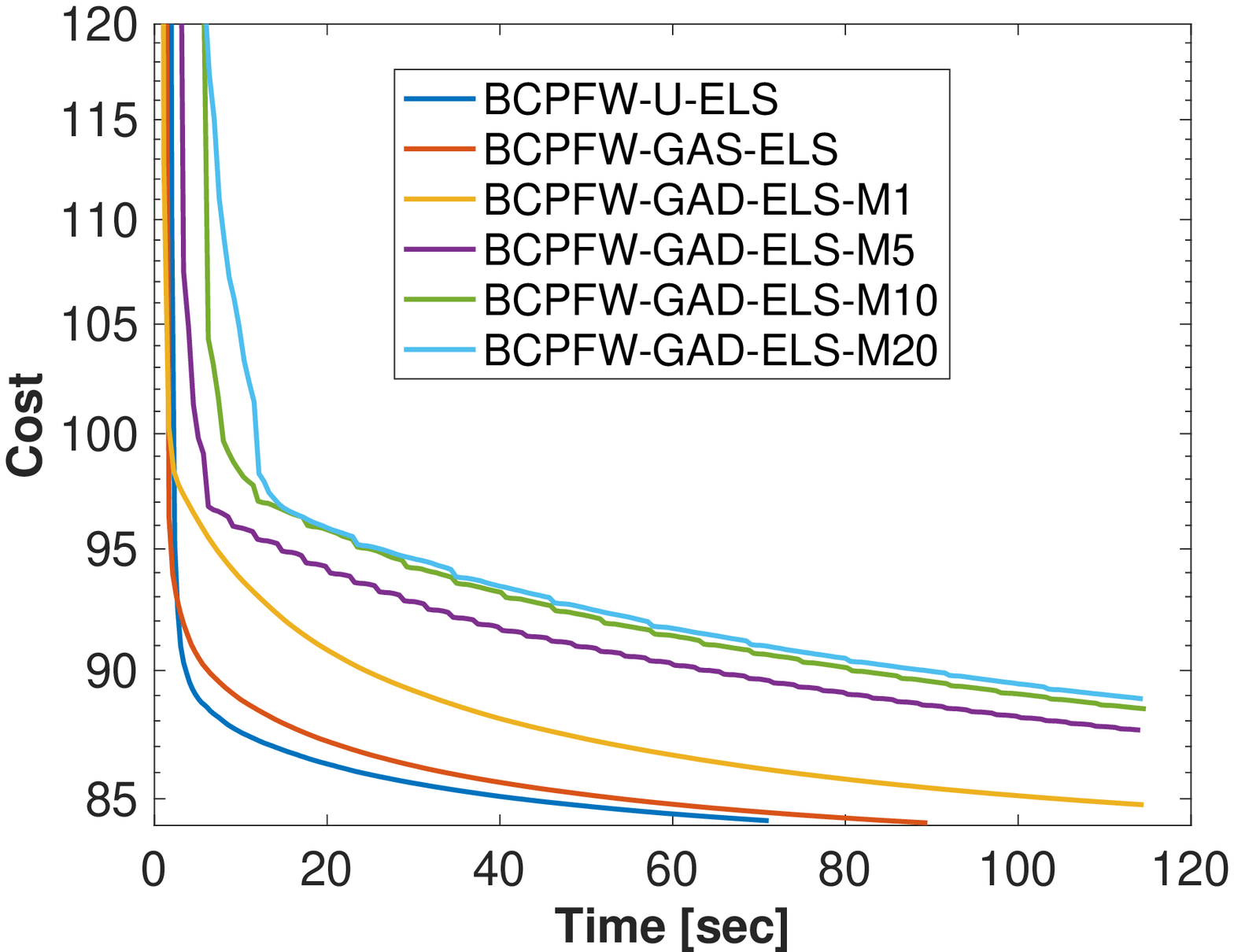}\\
		
		{\footnotesize  (ii) objective value (time)}
		
	\end{center} 
	\end{minipage}	
	\begin{minipage}[t]{0.24\textwidth}
	\begin{center}
		\includegraphics[width=\textwidth]{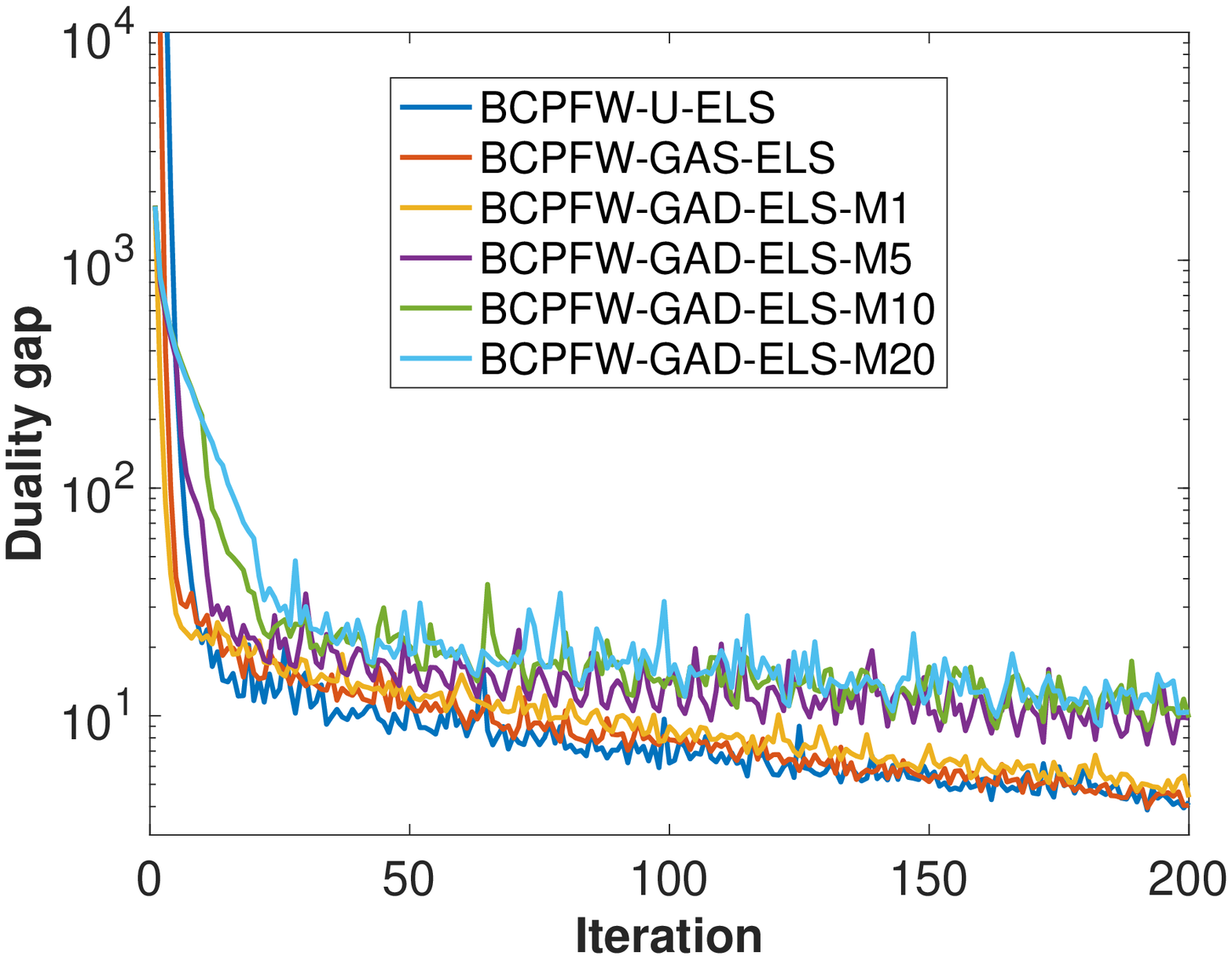}\\
		
		{\footnotesize  (iii) duality gap: $g(\mat{T})$}
		
	\end{center} 
	\end{minipage}
	\hspace*{0.1cm}
	\begin{minipage}[t]{0.23\textwidth}
	\begin{center}
		\includegraphics[width=\textwidth]{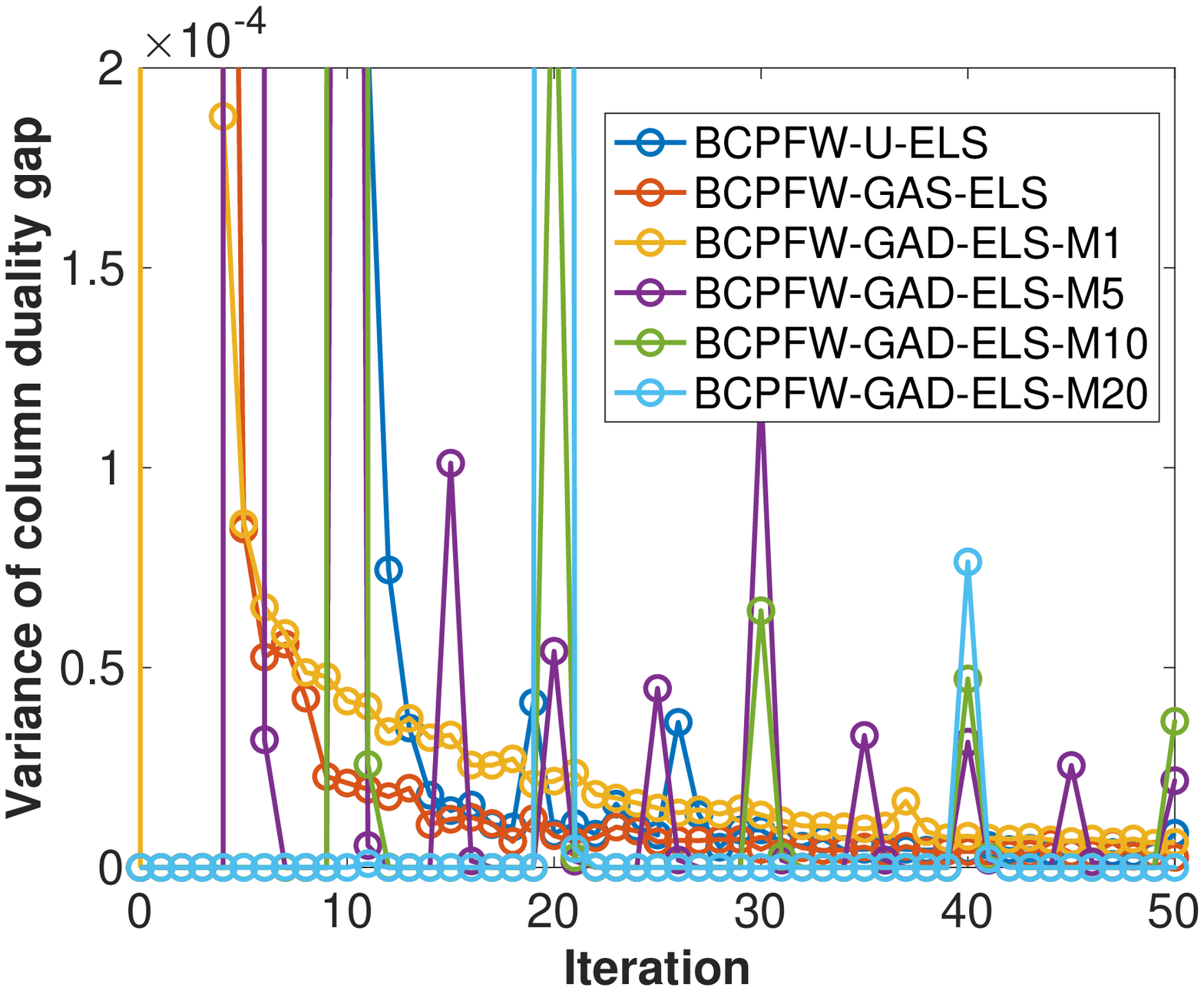}\\
		
		{\footnotesize  (iv) variances of $g_i(\mat{T})$}
		
	\end{center} 
	\end{minipage}
	\vspace*{0.2cm}		
	
	{\small (d) BCPFW-U-ELS and BCPFW-GA-ELS}
	
\caption{Evaluations on duality-adaptive sampling (corresponding to Figure \ref{fig:CompAdaptiveSampling}).}
\label{Appenfig:CompAdaptiveSampling}
\end{center}
\end{figure}

\clearpage	
\normalem
\bibliographystyle{unsrt}
\bibliography{OptimalTransport,VariantFrankWolfe,others,Supplementary}

\end{document}